\documentclass[preprint,3p,times,review]{elsarticle}

\usepackage{amssymb}
\usepackage{latexsym}
\usepackage{amsmath}
\usepackage{amsfonts,amssymb} 
\usepackage{afterpage}
\usepackage{multirow}
\usepackage{makecell}
\usepackage{threeparttable}
\usepackage{indentfirst}
\usepackage{booktabs}
\usepackage{pifont}
\usepackage{subfigure}
\usepackage{overpic}
\usepackage{appendix}

\PassOptionsToPackage{hyphens}{url}
\usepackage[table,xcdraw]{xcolor}
\usepackage{graphicx}
\usepackage{hyperref}
\usepackage[noend]{algpseudocode}
\usepackage[ruled,linesnumbered]{algorithm2e}

 \makeatletter

\newcommand{\topcaption}{%
    \setlength{\abovecaptionskip}{0pt}%
    \setlength{\belowcaptionskip}{10pt}%
	\caption}   

\graphicspath{{extras/all}{all}}
    
\journal{Pattern Recognition}

\begin{document}

\begin{frontmatter}
  \title{Robust Implementation of Foreground Extraction and Vessel Segmentation for X-ray Coronary Angiography Image Sequence}%

  \author[1,2]{Zeyu Fu}
  \author[1,2]{Zhuang Fu\corref{cor1}\fnref{fn1}}
  \cortext[cor1]{Corresponding author at: School of Mechanical Engineering, Shanghai Jiao Tong University, 800 Dongchuan RD. Minhang District, Shanghai 200240, China. Email: \url{zhfu@sjtu.edu.cn} (Zhuang Fu)}
  \fntext[fn1]{Zeyu Fu and Zhuang Fu contributed equally to this work.}
  \author[1,2]{Chenzhuo Lu}
  \author[1,2]{Jun Yan}
  \author[3,4,5,6]{Jian Fei}
  \author[7]{Hui Han}

  \address[1]{School of Mechanical Engineering, Shanghai Jiao Tong University, Shanghai 200240, China}
  \address[2]{State Key Laboratory of Mechanical System and Vibration, Shanghai 200240, China}
  \address[3]{Department of General Surgery, Pancreatic Disease Center, Ruijin Hospital, Shanghai Jiao Tong University School of Medicine, Shanghai 200025, China.}
  \address[4]{Research Institute of Pancreatic Diseases, Shanghai Jiao Tong University School of Medicine, Shanghai 200025, China}
  \address[5]{State Key Laboratory of Oncogenes and Related Genes, Shanghai 200240, China}
  \address[6]{Institute of Translational Medicine, Shanghai Jiao Tong University, Shanghai 200240, China.}
  \address[7]{Department of cardiovascular medicine, Ruijin Hospital, Shanghai Jiao Tong University School of Medicine, Shanghai 200025, China}

  \begin{abstract}
    The extraction of contrast-filled vessels from X-ray coronary angiography (XCA) image sequence has important clinical significance for intuitively diagnosis and therapy. In this study, the XCA image sequence is regarded as a 3D tensor input, the vessel layer is regarded as a sparse tensor, and the background layer is regarded as a low-rank tensor. Using tensor nuclear norm (TNN) minimization, a novel method for vessel layer extraction based on tensor robust principal component analysis (TRPCA) is proposed. Furthermore, considering the irregular movement of vessels and the low-frequency dynamic disturbance of surrounding irrelevant tissues, the total variation (TV) regularized spatial-temporal constraint is introduced to smooth the foreground layer. Subsequently, for vessel layer images with uneven contrast distribution, a two-stage region growing (TSRG) method is utilized for vessel enhancement and segmentation. A global threshold method is used as the preprocessing to obtain main branches, and the Radon-Like features (RLF) filter is used to enhance and connect broken minor segments, the final binary vessel mask is constructed by combining the two intermediate results. The visibility of TV-TRPCA algorithm for foreground extraction is evaluated on clinical XCA image sequences and third-party dataset, which can effectively improve the performance of commonly used vessel segmentation algorithms. Based on TV-TRPCA, the accuracy of TSRG algorithm for vessel segmentation is further evaluated. Both qualitative and quantitative results validate the superiority of the proposed method over existing state-of-the-art approaches.
  \end{abstract}

  \begin{keyword}
    X-ray coronary angiography, Tensor RPCA, TV regularization, Two-stage region growing, Foreground extraction,  Vessel segmentation
  \end{keyword}
\end{frontmatter}


\section{Introduction} \label{Introduction}

Coronary artery disease (CAD) is one of the most deadly diseases globally\cite{world2021world}. Compared with drug therapy and surgical procedures such as coronary artery bypass grafting (CABG), minimally invasive interventional surgery represented by percutaneous coronary intervention (PCI) has become the main clinical means for the treatment of coronary artery disease with advantages of high precision, small wound and quick recovery. X-ray coronary angiography (XCA) is a method of developing vessels under X-rays by injecting contrast agents. It can clearly display the anatomical structure of coronary artery tree, which has important clinical significance for pre-operative diagnosis and is also regarded as the gold standard\cite{kiricsli2013standardized}. XCA also plays a key role in multimodal registration, stenosis assessment, interventional path planning, biplanar 3D reconstruction, stent implantation, and balloon angioplasty\cite{jin2017extracting,qin2019accurate}.

\subsection{Background and Motivation}
Although XCA can provide a fluoroscopic model of coronary arteries, it still has many aspects to be improved. (1) Peripheral irrelevant tissues such as ribs, spine, diaphragm and lung will also be developed by X-ray, resulting in poor visibility of coronary vessels. (2) Images of different organs overlap after projection and affect the observation of lesions. (3) Different blood flow rates cause uneven distribution of contrast agents, and high-frequency Gaussian noise affects the detail performance of contrast images. See \hyperref[fig1]{Fig.1.(a1)}.

Therefore, it is a prerequisite for accurate diagnosis and treatment to isolate background structure and noise signal from raw XCA image sequence. The main purpose of this study is to extract the high visibility contrast-filled vessel layers and then perform accurate segmentation to restore the topologically correct tubular structures.

\subsection{Related Works and Methods}
Methods of vessel extraction can be divided into three categories: vessel segmentation for a single image, digital subtraction angiography (DSA) for two different images, and vessel layer separation for continuous angiography image sequence.

\begin{figure*}[tbp]
  \centering
  \subfigure[]{
    \begin{overpic}[width=0.19\linewidth]{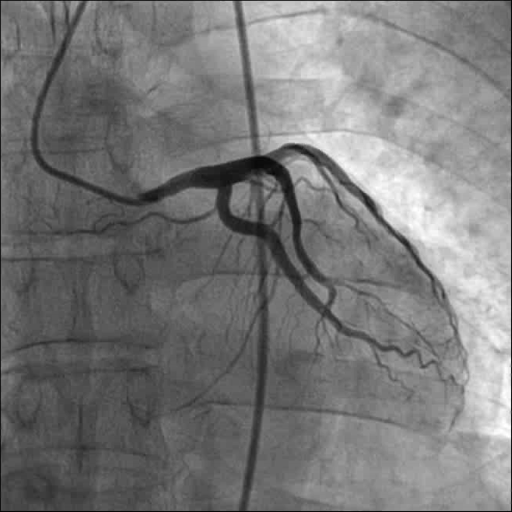}
      \put(3,5){\color{white}{(a1)}}
    \end{overpic}
  }
  \subfigure[]{
    \hspace{-0.3cm}
    \begin{overpic}[width=0.19\linewidth]{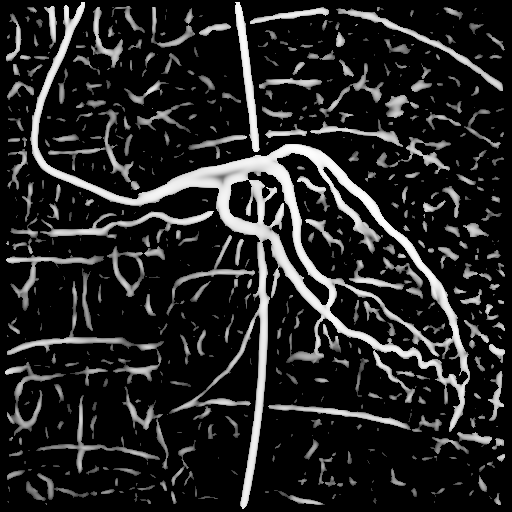}
      \put(3,5){\color{white}{(b1)}}
    \end{overpic}
  }
  \subfigure[]{
    \hspace{-0.3cm}
    \begin{overpic}[width=0.19\linewidth]{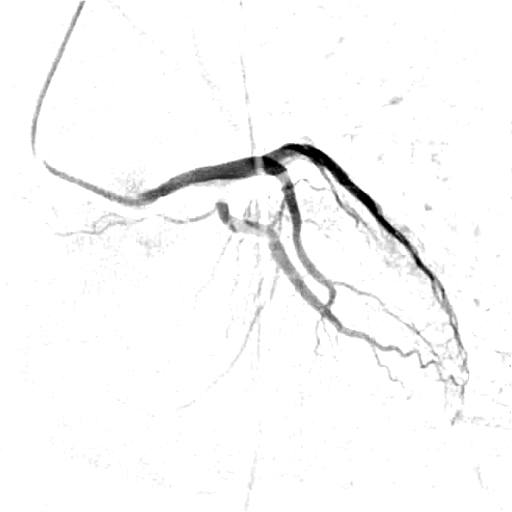}
      \put(3,5){\color{black}{(c1)}}
    \end{overpic}
  }
  \subfigure[]{
    \hspace{-0.3cm}
    \begin{overpic}[width=0.19\linewidth]{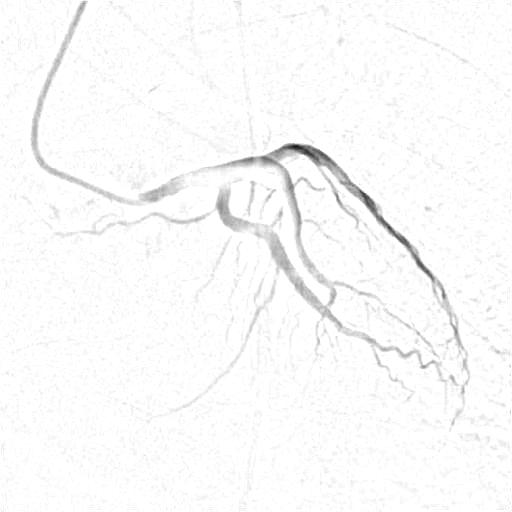}
      \put(3,5){\color{black}{(d1)}}
    \end{overpic}
  }
  \subfigure[]{
    \hspace{-0.3cm}
    \begin{overpic}[width=0.19\linewidth]{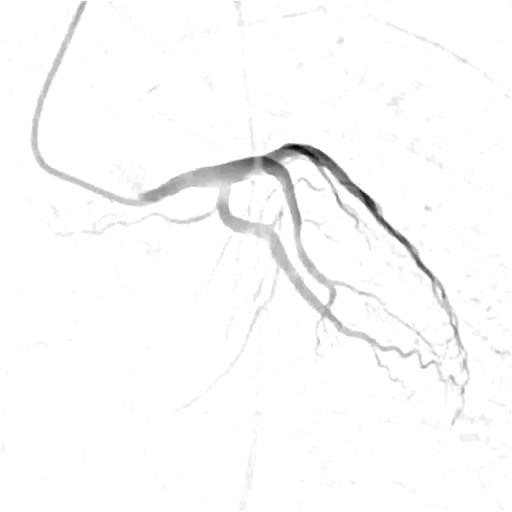}
      \put(3,5){\color{black}{(e1)}}
    \end{overpic}
  }

  \vspace{-0.95cm}
  \subfigure[]{
    \begin{overpic}[width=0.19\linewidth]{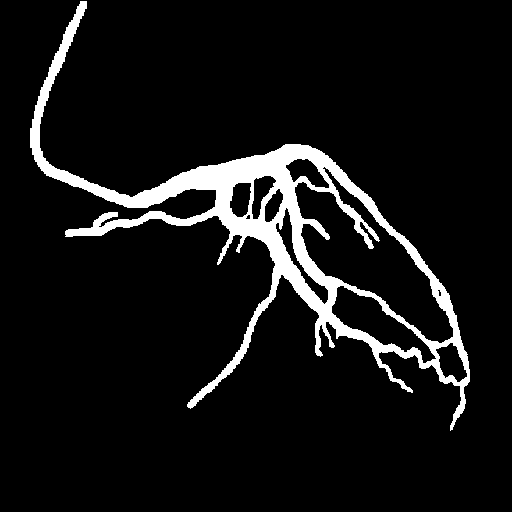}
      \put(3,5){\color{white}{(a2)}}
    \end{overpic}
  }
  \subfigure[]{
    \hspace{-0.3cm}
    \begin{overpic}[width=0.19\linewidth]{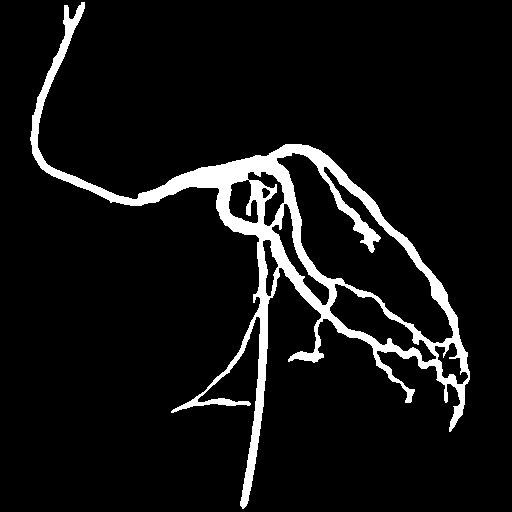}
      \put(3,5){\color{white}{(b2)}}
    \end{overpic}
  }
  \subfigure[]{
    \hspace{-0.3cm}
    \begin{overpic}[width=0.19\linewidth]{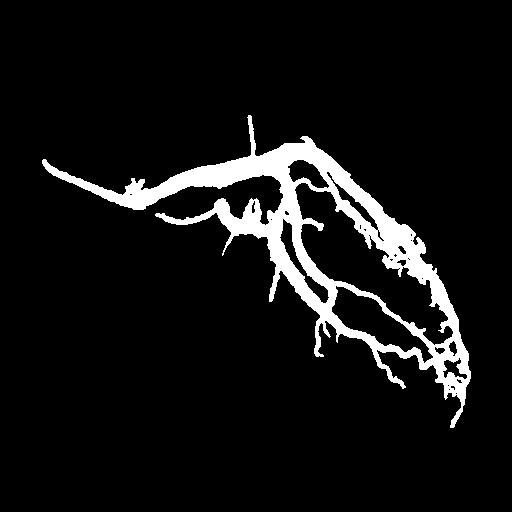}
      \put(3,5){\color{white}{(c2)}}
    \end{overpic}
  }
  \subfigure[]{
    \hspace{-0.3cm}
    \begin{overpic}[width=0.19\linewidth]{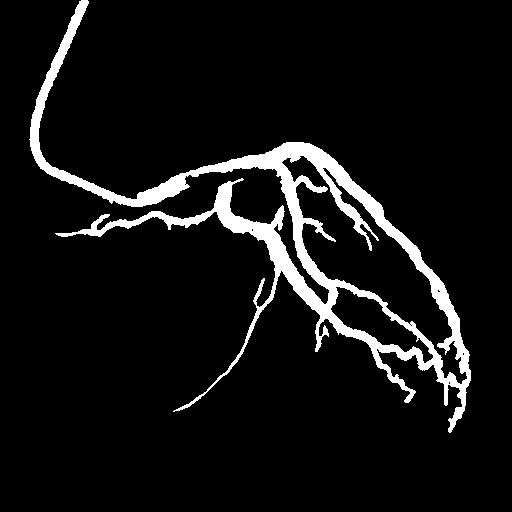}
      \put(3,5){\color{white}{(d2)}}
    \end{overpic}
  }
  \subfigure[]{
    \hspace{-0.3cm}
    \begin{overpic}[width=0.19\linewidth]{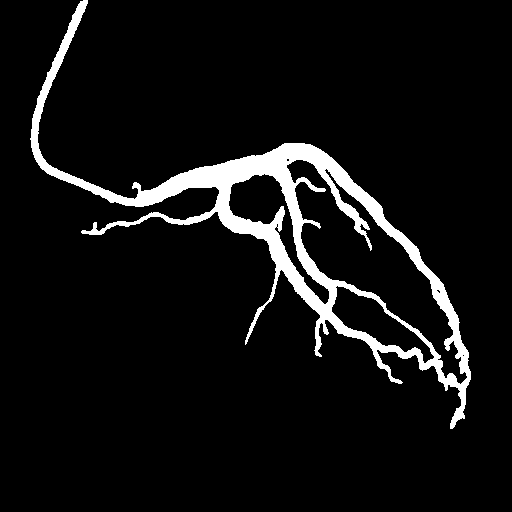}
      \put(3,5){\color{white}{(e2)}}
    \end{overpic}
  }

  \label{fig1}
  \vspace{-0.8cm}
  \caption{Methods of vessel extraction. (a1)-(a2) are the raw XCA image and its ground truth vessel mask selected by the surgeon. (b1)-(b2) are the multi-scale Frangi filtering vessel enhancement and its segmentation result by region growing. (c1)-(c2) are the vessel layer extracted by DSA and its segmentation result. (d1)-(d2) are the vessel layer extracted by TNN-TRPCA and its segmentation result. (e1)-(e2) are the vessel layer extracted by proposed TV-TRPCA and its segmentation result. The vessel segmentation of (c1)-(e1) utilizes the proposed TSRG method.}
\end{figure*}

Vessel segmentation is the most classical method, which is widely utilized in biomedical image processing. In practical application, in order to obtain better segmentation effect, it is often used in combination with other methods. Segmentation methods based on vessel enhancement, deformable models, tracking approaches and deep learning are the commonly used\cite{moccia2018blood}. Considering the characteristics of tubular structure, most vessel enhancement methods realize multi-scale filtering enhancement by constructing different vessel model response functions for convolution processing based on Hessian matrix\cite{frangi1998multiscale}. Different from Hessian matrix-based filtering, the method of Radon-like features (RLF) filter\cite{kumar2010radon} has been proposed in recent years. It provides non-isotropic sampling of the neighborhood by edge sensing along different directions, which can highlight minor vessel segments. Syeda-Mahmood adopted RLF filter for vessel enhancement in coronary image segmentation and achieved remarkable results\cite{syeda2012finding}. However, a variety of noise signals are also enhanced at the same time. Generally, threshold methods like OTSU\cite{otsu1979threshold} are used for postprocessing to obtain a high signal-to-noise ratio (SNR) vessel mask. \hyperref[fig1]{Fig.1.(b)} show the vessel enhancement by multi-scale Frangi filtering and the corresponding vessel segmentation result. The advantage of vessel enhancement is that no prior model is required, but it amplifies artifacts and irrelevant tissues. Moreover, thick blood vessels are prone to voids, so appropriate parameters must be selected for vessels of different gray scales and diameters, resulting in poor portability. The method based on deformable models extract vessels by selecting the appropriate initial vascular model. It is more suitable for medical images with complex background by combining prior information and utilizing the grayscale, gradient, texture and other characteristics of images. Typically, region-based deformable vessel segmentation algorithm selects seed points inside vessels as the starting point of region growing and expands continuously by traversing the surrounding voxel points. Kerkeni\cite{kerkeni2016coronary} implemented vessel segmentation in XCA images using multi-scale region growing algorithm. Most model-based methods are semi-automatic and sensitive to the selection of initial models, vessel enhancement is usually used as preprocessing to improve segmentation accuracy. Tracking approaches are divided into model-based and minimal-cost-path-based. It starts from the seed point and grows iteratively along a certain location and direction according to preset constraints to obtain a tubular structure (model-based) or a centerline tree (minimal-cost-path-based). This method is more frequently used in the extraction of vessel centerline. The method based on deep learning has the widest application scope in vessel segmentation. It does not need to extract features from raw images and can judge the contour of vessels adaptively. At present, high accuracy machine learning methods such as basic networks CNN\cite{nasr2018segmentation}, FCN\cite{wan2021automatic}, U-Net\cite{ronneberger2015u} and advanced networks VGN\cite{shin2019deep}, SVS-Net\cite{hao2020sequential} are supervised, the template calibration and network training are time-consuming and labor-intensive. Recently, weakly supervised method that do not require accurate calibration have been proposed\cite{zhang2020weakly}, but this methods need to take the vessel layer obtained from foreground extraction as the label, thus the segmentation accuracy will be affected by the layer separation algorithm.

Digital subtraction angiography (DSA)\cite{meaney1980digital} eliminates bone and soft tissue on the angiography images by subtraction, and the difference between the anterior and posterior mask is regarded as the vessel area. Due to the breathing and movement of human body, direct subtraction will produce motion artifacts. In order to obtain the spatial-temporal corresponding initial mask image, Bentoutou\cite{bentoutou2002invariant} improved the image registration method, Song\cite{song2019patch} adopted the patch-based adaptive background subtraction method, both of which achieved certain results. \hyperref[fig1]{Fig.1.(c)} show  the vessel layer extracted by DSA and its segmentation result. The spatial-temporal consistency of the initial mask of DSA could be seriously disturbed by body movements and contrast agent distribution changes, high-frequency noise and dynamic disturbance will also affect its calculation accuracy.

Layer separation is a video segmentation algorithm commonly used in machine vision. It regards the scene in the video as the superposition of multiple element layers such as foreground, background and noise, then extracts different layers separately. In XCA image sequence, different tissues have different motion patterns. Coronary artery motion is the most intense and irregular, lung respiration is slow and uniform, ribs and diaphragm move periodically with lung respiration, and spine remains basically static. Therefore, vessel layers can be extracted by layer separation. Different from the filtering method which simply enhances the vessel pixel mathematically, the layer separation algorithm can obtain the image sequence that truly reflects the continuous motion information of vessels. Dimension reduction is the mainly used method for layer separation. Tang\cite{tang2012application} used independent component analysis (ICA) to separate coronary artery and background in DSA, which has the advantage of being more continuous than traditional subtraction, but the registration accuracy of the initial mask is limited by the dynamic disturbance. Xia\cite{xia2019vessel} proposed a matrix decomposition model (MDM) with gradient sparsity to capture and segment blood vessels through container framework. The theory decomposed non-vessel background into static background structure, background structure away from blood vessels and background structure close to vessels, achieving certain effects on XCA image sequences with complex backgrounds. However, this method will delete a large number of broken vessel segments similar to the background structure and cannot completely display the coronary tree.

Robust principal component analysis (RPCA)\cite{candes2011robust} is a typical dimension reduction method with global low-rank property. The original observation matrix $\pmb{O}\in \mathbb{R}^{MN\times T}$  is separated into two parts: the low-rank matrix $\pmb{L}\in \mathbb{R}^{MN\times T}$ and the sparse matrix $\pmb{S}\in \mathbb{R}^{MN\times T}$, which represents to the static background and the moving foreground, respectively. The iterative decomposition operation is carried out by minimizing the cost function. A typical RPCA solution via principal component pursuit can be calculated according to \hyperref[eq1]{Eq.1}

\begin{equation}
  \label{eq1}
  \min_{\pmb{L},\pmb{S}}\ \Vert\pmb{L}\Vert_{\ast} + \lambda\Vert\pmb{S}\Vert_1\ \ \ s.t.\ \pmb{O}=\pmb{L}+\pmb{S}
\end{equation}
where $\Vert\pmb{L}\Vert_{\ast}=\sum\sigma(\pmb{L})$ is the nuclear norm of matrix $\pmb{L}$, which is the sum of all singular values, $\Vert\pmb{S}\Vert_1=\sum_{i,j}|\pmb{S}_{i,j}|$ is the $\ell_1$ norm of $\pmb{S}$, which is the sum of absolute values of all elements, $\lambda$ is a positive scalar equilibrium factor for balancing foreground and background. In recent years, RPCA and its derivative algorithms have been widely utilized in XCA image layer separation. Ma\cite{ma2017automatic} decomposed XCA images into three layers by using morphological filtering superimposed RPCA method, which significantly improved the visibility of blood vessels. On this basis, an automatic online layer separation method was designed to guide PCI surgery. Zhang\cite{zhang2018vesselness} proposed a vesselness-constrained RPCA approach (VC-RPCA) that incorporated the vessel-like appearance into the layer separation framework to achieve precise vessel enhancement, and an adaptive regularization strategy is used to capture vessels without significant movement. Qin\cite{qin2019accurate} performed global logarithmic transformation on XCA images, used IALM-RPCA to separate layers in logarithmic space, and restored the background layer by tensor completion. This method replaced addition operation with multiplication operation in the original image space, which is more consistent with the physical model of XCA. Song\cite{song2019spatio} integrated the motion consistency of quasi-static structure and the smoothness of blood vessels into the layer separation of XCA image sequence, and proposed an online RPCA method adaptively calculated spatial constraints using the proportion of the vascular region in the previous frame, which can retain the local blood vessel information after the contrast agent dissipates.

The RPCA in matrix form above has one disadvantage: it can only process two-dimensional data. However, the actual data is usually high-dimensional in nature, so to use RPCA, the original data must be matricized first, but matricization will destroy its inherent spatial structure, resulting in information loss and performance degradation. In addition, the results obtained after video matricization become huge, which occupies a large memory space and increases the computational complexity. To solve this problem, convex optimization method tensor robust principal component analysis (TRPCA) was introduced for low-rank tensor decomposition (LRTD). Since tensor rank\cite{kolda2009tensor,lu2019tensor} can be defined as CP-rank based on CANDECOMP/PARAFAC (CP) decomposition, Tucker-rank based on higher-order principal component analysis (HOPCA), Tubal-rank, Multi-rank and Average-rank based on higher-order singular value decomposition (HOSVD), etc., various TRPCA methods have been extended. In order to reduce the loss of high-dimensional information, Goldfarb\cite{goldfarb2014robust} proposed high-order RPCA (HoRPCA) model. Xie\cite{xie2017kronecker} proposed Kronecke-Basis-Representation based RPCA (KBR-RPCA) model, which uses tensor sparse detection to replace the background of nuclear norm modeling, and assigns appropriate weights to the rank of each dimension. Liu\cite{liu2018improved} proposed a new tensor nuclear norm for background constraint based on low-rank components of core tensors, which improved the accuracy of foreground background separation. Lu\cite{lu2019tensor,lu2016tensor} proposed an improved tensor nuclear norm TRPCA (TNN-TRPCA) model for background modeling to strengthen the low-rank property of background. Gao\cite{gao2020enhanced} applied the weighted Schatten $p$-norm minimization (WSNM)\cite{xie2016weighted} model for LRTD and proposed an enhanced TRPCA (ETRPCA) algorithm, which further improved the stability of background modeling in complex fields. Most of the above methods adopt $\ell_1$ norm constraint for foreground, but $\ell_1$ norm treats each pixel independently without considering the spatial continuity and temporal persistence of foreground target, which makes the algorithm insensitive to the discontinuous moving foreground and dynamic disturbance. RPCA and TRPCA algorithms with $\ell_1$ norm regularization constraint are difficult to avoid blurred edges and internal voids in moving object detection, as shown in \hyperref[fig1]{Fig.1.(d)} .

Total variation (TV) regularization can compensate for the deficiency of $\ell_1$ norm, and improve the spatial-temporal continuity of foreground layer and the spatial-temporal correlation of background layer in LRTD. Chan\cite{chan2011augmented} proposed a total variation norm regularization method based on $\ell_1$  norm and $\ell_{2,1}$ norm for video deblurring, denoising and disparity refinement, where $\ell_1$ norm and $\ell_{2,1}$ norm correspond to isotropic and anisotropic inputs, respectively. This method has been widely promoted in image processing\cite{cao2016total}, it has a strong inhibitory effect on discontinuous changes, effectively reduces the interference of dynamic disturbance to foreground extraction, and makes layer separation more accurate. Jin\cite{jin2017extracting} introduced total variation regularization into the foreground extraction of XCA video and proposed the motion coherent regularization RPCA (MCR-RPCA), which improved the motion continuity of vessel layer without the benefit of prior knowledge. However, this method still needs to matricize the video, which destroys the original structure and cannot naturally retain the high-dimensional spatial-temporal information obtained by continuous angiography.

\subsection{Main Contributions} \label{Contibutions}
Before presenting our method, it is necessary to discuss the characteristics of XCA image sequences. Different from general moving object detection, the moving foreground objects (blood vessels) in XCA video will gradually extend various branches over time, and they move irregularly with the heartbeat rather than the common translatory motion, so the vessel layer tensor is not completely sparse. The dynamic disturbance moves periodically with respiration and has both low-rank and sparse properties, which need to be separated as the component of background layer. At the same time, in XCA video, the proximal vessel branches are visually thick and have small motion amplitude, meanwhile the distal vessel branches are visually thin and have large motion amplitude. The vessel layer extracted by layer separation will have uneven gray distribution and local fracture, which will affect the accuracy of subsequent quantitative analysis.

To solve the above problems, we propose a total variation regularized tensor robust principal component analysis (TV-TRPCA) method to extract the contrast-filled vessel mask. The model regards XCA sequence as the superposition of vessel layer, background layer and noise layer, uses the tensor nuclear norm (TNN) minimization to constrain the background and  $TV/\ell_1$ norm regularization to constrain the foreground, which strengthen the low-rank property of the background and the spatial-temporal continuity of the foreground. Moreover, a two-stage region growing (TRSG) method is used to enhance and segment the vessels extracted by layer separation. To the best of our knowledge, we are the first to apply TRPCA to vessel layer separation. Comparing with the existing matrix decomposition-based methods, the recovery of high-dimensional spatial-temporal information is more accurate. The result of proposed foreground extraction and vessel segmentation method is shown in \hyperref[fig1]{Fig.1.(e)}.

The main contributions of this paper are as follows:

(1) We propose a total variation regularized tensor robust principal component analysis (TV-TRPCA) method, which effectively improves the visibility of XCA image sequence foreground extraction and provides a high contrast template for existing vessel segmentation algorithms.

(2) We adopt a two-stage regional growing (TSRG) method to segment the extracted contrast-filled vessel layer. Specifically, the main branches are obtained based on global threshold segmentation, and the minor segments are enhanced and connected based on Radon-Like features (RLF) filtering. The two intermediate results are combined to get the topologically accurate binary coronary artery mask.

The rest of this paper is organized as follows. In \hyperref[Methods]{Section 2}, the proposed TV-TRPCA model and TSRG algorithm are introduced in detail. The experimental results, including a thorough comparison and detailed analysis of the method performance, are presented in \hyperref[Experiment]{Section 3}. The conclusion is given in \hyperref[Conclusion]{Section 4}.

\section{Methods} \label{Methods}
In this section, we will introduce the theory of tensor robust principal component analysis, total variation, and some vessel enhancement and segmentation methods.

\subsection{Foreground Extraction}

\subsubsection{Basic Definitions and Notations of Tensors}
In this paper, tensors are denoted by Euler script letters, e.g., $\mathcal{X}$. Matrices are denoted by boldface capital letters, e.g., $\pmb{X}$, vectors are denoted by boldface lowercase letters, e.g., $\pmb{x}$, and scalars are denoted by lowercase letters, e.g., $x$. For a 3D tensor in the field of real numbers denoted as $\mathcal{X}\in\mathbb{R}^{M\times N\times T}$, in which $M$, $N$, $T$ are the side lengths of each dimension, some used notations are summarized in \hyperref[tab1]{Tab.1}. In which $\mathcal{X}_{mnt}\in\mathbb{R}$ is a scalar, $\mathcal{X}^{(t)}\in\mathbb{R}^{M\times N}$ is a matrix, $\textbf{Vec}(\mathcal{X})\in\mathbb{R}^{MNT}$ is a vector, $\widetilde{\mathcal{X}}\in\mathbb{C}^{M\times N\times T}$ is the Discrete Fourier Transformation(DFT) on $\mathcal{X}$ along the 3-rd dimension. $\left\langle \pmb{X}, \pmb{Y}\right\rangle=Tr(\pmb{X}^\ast\pmb{Y})$ represents the inner product of two matrices. $bcirc(\mathcal{X})\in\mathbb{R}^{MT\times NT}$ and $bstack(\mathcal{X})\in\mathbb{R}^{MT\times N}$ are two tensor operators as defined below.

\begin{table}[!t]
  \renewcommand{\arraystretch}{1.5}
  \centering
  \topcaption{Notations of 3-D Tensors\cite{lu2019tensor,lu2016tensor,cao2016total,kilmer2011factorization}}
  \label{tab1}
  \begin{tabular}{@{}ll@{}}
    \toprule \toprule
    \textbf{Notations}                                                                                                                     & \textbf{Explanations}                                     \\ \midrule
    $\mathcal{X}, \pmb{X}, \pmb{x}, x$                                                                                                     & tensor, matrix, vector, scalar                            \\
    $\mathcal{X}(m,n,t)$ or $\mathcal{X}_{mnt}$                                                                                            & the $(m,n,t)^{th}$ entry of $\mathcal{X}$                 \\
    $\mathcal{X}(m,:,:), \mathcal{X}(:,n,:), \mathcal{X}(:,:,t)$                                                                           & \makecell[l]{the $m^{th}$, $n^{th}$, $t^{th}$ horizontal, \\ lateral and frontal slice of $\mathcal{X}$}  \\
    $\mathcal{X}^{(t)}$                                                                                                                    & the $t^{th}$ frontal slice of $\mathcal{X}$               \\
    $\Vert\mathcal{X}\Vert_1 = \sum_{m,n,t} \mathcal|{X}_{mnt}|$                                                                           & $\ell_1$ norm  of $\mathcal{X}$                           \\
    $\Vert\mathcal{X}\Vert_F = \sqrt{\sum_{m,n,t} |\mathcal{X}_{mnt}|^2}$                                                                  & Frobenius norm of $\mathcal{X}$                           \\
    $\left\langle \mathcal{X}, \mathcal{Y}\right\rangle = \sum_{t = 1}^{T} \left\langle \mathcal{X}^{(t)}, \mathcal{Y}^{(t)}\right\rangle$ & inner product of $\mathcal{X}$ and $\mathcal{Y}$          \\
    $\mathcal{X}\ast \mathcal{Y} = fold\left(bcirc(\mathcal{X})\cdot bstack(\mathcal{Y})\right)$                                           & tensor product of $\mathcal{X}$ and $\mathcal{Y}$         \\
    $\textbf{Vec}(\mathcal{X})$ or $f(\mathcal{X})$                                                                                        & vectorization of $\mathcal{X}$                            \\
    $\widetilde{\mathcal{X}}=\pmb{fft}\left(\mathcal{X},[\ ],3\right)$                                                                     & \makecell[l]{DFT on $\mathcal{X}$ along the               \\3-rd dimension}              \\
    $\pmb{rank}_a(\mathcal{X})=\frac{1}{T}\pmb{rank}(bcirc(\mathcal{X}))$                                                                  & tensor average rank                                       \\
    $\Vert\mathcal{X}\Vert_\circledast = \frac{1}{T} \sum_{t=1}^{T} \Vert \widetilde{\mathcal{X}}^{(t)}\Vert_\ast$                         & tensor nuclear norm of $\mathcal{X}$                      \\
    \bottomrule \bottomrule
  \end{tabular}
\end{table}

$bcirc(\mathcal{X})$ is the block circulant matrix formed by circular combination of the frontal slice of tensor $\mathcal{X}$:
\[ bcirc(\mathcal{X})=\begin{bmatrix}
    \mathcal{X}^{(1)} & \mathcal{X}^{(T)}   & \cdots & \mathcal{X}^{(2)} \\
    \mathcal{X}^{(2)} & \mathcal{X}^{(1)}   & \cdots & \mathcal{X}^{(3)} \\
    \vdots            & \vdots              & \ddots & \vdots            \\
    \mathcal{X}^{(T)} & \mathcal{X}^{(T-1)} & \cdots & \mathcal{X}^{(1)} \\
  \end{bmatrix}
\]
Compared with matricizations along certain dimension, the block circulant matricization can preserve more spatial relationship among entries\cite{lu2016tensor}.

$bstack(\mathcal{X})$ is the block stacking matrix formed by stacking the frontal slice of tensor $\mathcal{X}$ and $fold(\cdot)$ is the inverse operator:
\[ bstack(\mathcal{X})=\begin{bmatrix}
    \mathcal{X}^{(1)} \\
    \mathcal{X}^{(2)} \\
    \vdots            \\
    \mathcal{X}^{(T)} \\
  \end{bmatrix},\ fold(bstack(\mathcal{X}))=\mathcal{X}
\]

The conjugate transpose of complex tensor $\mathcal{X}\in\mathbb{C}^{M\times N\times T}$ is the tensor $\mathcal{X}^\ast\in\mathbb{C}^{N\times M\times T}$ obtained by conjugate transposing each of the frontal slices and then reversing the order of transposed frontal slices 2 through $T$\cite{lu2019tensor}.

The identity tensor $\mathcal{I}\in\mathbb{R}^{M\times M\times T}$ is a tensor whose first frontal slice is a $M\times M$ identity matrix, and other frontal slices are all zeros\cite{kilmer2011factorization}.

A complex tensor $\mathcal{Q}\in\mathbb{C}^{M\times M\times T}$ is unitary if it satisfies $\mathcal{Q}^\ast\ast\mathcal{Q}=\mathcal{Q}\ast\mathcal{Q}^\ast=\mathcal{I}$. The unitary tensor expends orthogonal tensor\cite{kilmer2011factorization} in the complex field.

The f-diagonal tensor is a tensor with each frontal slice being diagonal matrix\cite{kilmer2011factorization}.

\subsubsection{T-SVD}
For $\mathcal{X}\in\mathbb{C}^{M\times N\times T}$, it can be factorized as
\begin{equation}
  \mathcal{X}=\mathcal{U}\ast\mathcal{S}\ast\mathcal{V}^\ast
  \label{eq2}
\end{equation}
where $\mathcal{U},\mathcal{V}\in\mathbb{C}^{M\times M\times T}$ are unitary tensors, $\mathcal{S}\in\mathbb{R}^{M\times N\times T}$ is an f-diagonal positive real tensor. \hyperref[eq2]{Eq.2} is called the tensor singular value decomposition (t-SVD)\cite{kilmer2011factorization}. In order to compute t-SVD, it is supposed to analyze the properties of tensor product (t-product) first.

Based on circular combination and stacking operations, the tensor product can be defined as follows:
\begin{equation}
  \mathcal{Z}=\mathcal{X}\ast\mathcal{Y}=fold(bcirc(\mathcal{X})\cdot bstack(\mathcal{Y}))
  \label{eq3}
\end{equation}
where $\mathcal{Y}\in\mathbb{R}^{N\times L\times T}$ and $\mathcal{Z}\in\mathbb{R}^{M\times L\times T}$. The t-product has a similar form to matrix products except that cyclic convolution replaces the product operation of elements, it acts on tensors in the original domain which corresponds to the matrix product of the frontal slices in the Fourier domain\cite{kilmer2011factorization}:
\begin{equation}
  \widetilde{\mathcal{Z}}^{(t)}=\widetilde{\mathcal{X}}^{(t)}\ast\widetilde{\mathcal{Y}}^{(t)},\ t=1,2,...,T.
  \label{eq4}
\end{equation}
Based on this property, t-SVD can be computed in Fourier domain efficiently. Specifically, each frontal slice of $\widetilde{\mathcal{U}}$, $\widetilde{\mathcal{S}}$ and $\widetilde{\mathcal{V}}$ can be solved by matrix singular value decomposition:
\begin{equation}
  \left[\widetilde{\mathcal{U}}^{(t)},\widetilde{\mathcal{S}}^{(t)},\widetilde{\mathcal{V}}^{(t)}\right]=\textbf{SVD}\left(\widetilde{\mathcal{X}}^{(t)}\right),\ t=1,2,...,T.
  \label{eq5}
\end{equation}

Finally, $\mathcal{U}$, $\mathcal{S}$, $\mathcal{V}$ can be computed by performing inverse DFT on $\widetilde{\mathcal{U}}$, $\widetilde{\mathcal{S}}$, $\widetilde{\mathcal{V}}$ obtained in \hyperref[eq5]{Eq.5} along the 3-rd dimension:
\begin{equation}
  \mathcal{J}=\pmb{ifft}\left(\widetilde{\mathcal{J}},[\ ],3\right),\ \mathcal{J}=\mathcal{U},\ \mathcal{S},\ \mathcal{V}
  \label{eq6}
\end{equation}

\subsubsection{Tensor Nuclear Norm}
In robust principal component analysis (RPCA), low-rank constraint and sparse constraint are applied to the background and foreground. But the computation of matrix rank and $\ell_0$ norm is NP-hard in mathematics. In practice, these two terms are generally convex relaxed by nuclear norm and $\ell_1$ norm, and the object is transformed into a convex optimization problem. This approach can be extended to tensor robust principal component analysis (TRPCA).

In this paper, we adopted tensor average rank and its tight convex surrogate tensor nuclear norm (TNN) for low-rank tensor recovery\cite{lu2019tensor,lu2016tensor}. For $\mathcal{X}\in\mathbb{R}^{M\times N\times T}$, the tensor average rank can be defined as
\begin{equation}
  \pmb{rank}_a(\mathcal{X})=\frac{1}{T}\pmb{rank}(bcirc(\mathcal{X}))
  \label{eq7}
\end{equation}

According to \cite{lu2019tensor}, a low tubal rank tensor always has low average rank, and the low average rank assumption is weaker than the low Tucker-rank and low CP-rank assumptions. Thus, the low-rank property of background can be strengthened by using tensor average rank.

The tensor nuclear norm of $\mathcal{X}\in\mathbb{R}^{M\times N\times T}$ is defined as the average of the nuclear norm of all the frontal slices of $\widetilde{\mathcal{X}}$ in Fourier domain, it is also closely related to the nuclear norm of the block circulant matrix $bcirc(\mathcal{X})$ in the original domain\cite{lu2016tensor}, i.e.,
\begin{equation}
  \Vert\mathcal{X}\Vert_\circledast  =  \frac{1}{T} \sum_{t=1}^{T} \Vert \widetilde{\mathcal{X}}^{(t)}\Vert_\ast =\frac{1}{T}\Vert bcirc(\mathcal{X})\Vert_\ast
  \label{eq8}
\end{equation}

\subsubsection{Total Variation}
Total variation (TV) was first proposed to solve image denoising problems. Recently, TV model has been widely used in many computer vision (CV) problems\cite{chan2011augmented}.

In XCA image sequence, the trajectory of vessel foreground is usually smooth in time and space domain, while the significant change of dynamic disturbance is periodic and discontinuous. In mathematics, total variation can smooth the signal and restrain the discontinuous change\cite{cao2016total}. Therefore, TV regularization can effectively suppress the dynamic disturbance. Pixel gradient regards the XCA images as a discrete function, which can be used as the approximation to variation. Suppose the vessel foreground tensor is $\mathcal{V}\in\mathbb{R}^{M\times N\times T}$, we can represent its horizontal, vertical, and temporal variances as
\begin{equation}
  \left\{\begin{aligned}
    \nabla_m \mathcal{V}(m, n, t) & = \mathcal{V}(m+1, n, t) - \mathcal{V}(m, n, t) \\
    \nabla_n \mathcal{V}(m, n, t) & = \mathcal{V}(m, n+1, t) - \mathcal{V}(m, n, t) \\
    \nabla_t \mathcal{V}(m, n, t) & = \mathcal{V}(m, n, t+1) - \mathcal{V}(m, n, t)
  \end{aligned}\
  \right.
  \label{eq9}
\end{equation}
\[m=1,2,...,M;\ n=1,2,...,N;\ t=1,2,...,T\]
To avoid data overflow, define $\mathcal{V}(M+1, n, t)=\mathcal{V}(1, n, t)$, $\mathcal{V}(m, N+1, t)=\mathcal{V}(m, 1, t)$, $\mathcal{V}(m, n, T+1)=\mathcal{V}(m, n, 1)$. The total variation of $\mathcal{V}$, denoted as $\nabla \mathcal{V}\in \mathbb{R}^{M\times N\times 3T}$, is defined as the weighted superposition of variance in each dimension:
\begin{equation}
  \nabla \mathcal{V} = fold\left(\left[\sigma_m\pmb{V}_m^T,\sigma_n\pmb{V}_n^T,\sigma_t\pmb{V}_t^T\right]^T\right)=fold\left(\pmb{V}\right)
  \label{eq10}
\end{equation}
where $\sigma_m,\sigma_n,\sigma_t$ are the scalar weights of each dimension, $\pmb{V}_m,\pmb{V}_n,\pmb{V}_t\in \mathbb{R}^{MT\times N}$ are the block stacking matrices of $\nabla_m \mathcal{V},\nabla_n \mathcal{V},\nabla_t \mathcal{V}\in \mathbb{R}^{M\times N\times T}$, i.e., \[\pmb{V}_i=bstack(\nabla_i \mathcal{V}),\ i=m,n,t\]

Since the $\ell_{2,1}$ norm based sparsity constraint forcing exact zero columns or
rows, it is not suitable for anisotropic inputs such as XCA image sequence\cite{jin2017extracting}. Therefore, we adopt $TV/\ell_1$ norm to constrain the foreground total variation, based on \hyperref[eq10]{Eq.10}, it can be computed as
\begin{equation}
  \Vert \mathcal{V}\Vert_{TV1}  = \sigma_m\Vert \nabla_m \mathcal{V}\Vert _1+\sigma_n\Vert \nabla_n \mathcal{V}\Vert _1+\sigma_t\Vert \nabla_t \mathcal{V}\Vert _1 \\
  = \Vert \nabla \mathcal{V}\Vert _1
  \label{eq11}
\end{equation}

\subsubsection{TV-TRPCA}
On the basis of the previous definitions, in this section we will derive the total variation regularized tensor robust principal component analysis (TV-TRPCA) in detail.

Firstly, the raw grayscale image sequence is normalized into a 3D tensor $\mathcal{O}\in\mathbb{R}^{M\times N\times T}$, $M,N$ are image height and width, $T$ is the number of frames. In ideal cases, $\mathcal{O}$ is the superposition of low-rank static background tensor $\mathcal{B}\in\mathbb{R}^{M\times N\times T}$, Gaussian noise tensor $\mathcal{G}\in\mathbb{R}^{M\times N\times T}$ and residual completely sparse vessel foreground (Laplacian noise) tensor $\mathcal{V}\in\mathbb{R}^{M\times N\times T}$\cite{zhao2014robust}. In TRPCA with convex relaxation, optimal $\mathcal{B},\mathcal{V},\mathcal{G}$ can be calculated through minimizing the following objective function:
\begin{equation}
  \min_{\mathcal{B},\mathcal{V},\mathcal{G}}\ \Vert\mathcal{B}\Vert_\circledast + \lambda_1\Vert\mathcal{V}\Vert_1 + \frac{\lambda_2}{2}\Vert\mathcal{G}\Vert^2_F\ \ \ s.t.\ \mathcal{O}=\mathcal{B}+\mathcal{V}+\mathcal{G}
  \label{eq12}
\end{equation}
where $\lambda_1,\lambda_2$ are positive scalar equilibrium factors.

\begin{algorithm}[!t]
  \SetAlgoVlined 
  \DontPrintSemicolon
  \SetCommentSty{mycommfont} 
  \caption{TV-TRPCA optimization method}
  \label{al1}
  \KwIn{1.Normalized observation tensor $\mathcal{O}$.\\\ \ \ \ \ \ \ \ \ \ \ \ 2.Equilibrium factors $\lambda_1,\lambda_2,\lambda_3$.\\\ \ \ \ \ \ \ \ \ \ \ \ 3.Penalty scalars $\mu,\nu,\mu_{max},\nu_{max}$, ratio $\rho$.\\ \ \ \ \ \ \ \ \ \ \ \ \ 4.Total variation weights $\sigma_m,\sigma_n,\sigma_t$.\\\ \ \ \ \ \ \ \ \ \ \ \ 5.Stopping criteria $\epsilon$.\\\ \ \ \ \ \ \ \ \ \ \ \ 6.Maximal iteration $imax$.}
  \KwOut{1.Background layer $\mathcal{B}$.\\\ \ \ \ \ \ \ \ \ \ \ \ \ \ \ 2.Foreground layer $\mathcal{V}$.\\\ \ \ \ \ \ \ \ \ \ \ \ \ \ \ 3.Gaussian noise layer $\mathcal{G}$.\\\textbf{Initialization:}\\$\mathcal{B}_0=\mathcal{V}_0=\mathcal{G}_0=\mathcal{H}_0=\mathcal{X}_0=\mathcal{Y}_0=\mathbb{O}^{M\times N\times T}$. $\mathcal{T}_0=\mathcal{Z}_0=\mathbb{O}^{M\times N\times 3T}$. $k=0$.}
  \While{not converged \textbf{and} $k<imax$ }
  {
    Update $\mathcal{B}_{k+1}$ via \hyperref[eq19]{Eq.19};\\
    Update $\mathcal{V}_{k+1}$ via \hyperref[eq20]{Eq.20};\\
    Update $\mathcal{G}_{k+1}$ via \hyperref[eq21]{Eq.21};\\
    Update $\mathcal{H}_{k+1}$ via \hyperref[eq29]{Eq.29};\\
    Update $\mathcal{T}_{k+1}$ via \hyperref[eq31]{Eq.31};\\
    Update $\mathcal{X}_{k+1},\mathcal{Y}_{k+1},\mathcal{Z}_{k+1},\mu_{k+1},\nu_{k+1}$ via \hyperref[eq32]{Eq.32};\\
    $k=k+1$.\\
    Check the convergence condition:\\
    $\Vert\mathcal{O}-\mathcal{B}_{k+1}-\mathcal{V}_{k+1}-\mathcal{G}_{k+1}\Vert_F^2\leqslant \epsilon\Vert\mathcal{O}\Vert_F^2$
  }
  \textbf{end while}\\
  \KwRet{$\mathcal{B},\mathcal{V},\mathcal{G}$}
\end{algorithm}

The above decomposition works well assuming that the background is completely stationary. However, according to the \hyperref[Contibutions]{previous analysis}, there exists non-negligible low-frequency dynamic disturbance in XCA video, which makes the accuracy of \hyperref[eq12]{Eq.12} decline sharply. Therefore, the TV regularization constraint on the vessel foreground layer $\mathcal{V}$ is required to achieve precise separation. \hyperref[eq12]{Eq.12} can be updated into a more accurate form:
\begin{equation}
  \begin{aligned}
    \min_{\mathcal{B},\mathcal{V},\mathcal{G},\mathcal{H}}\  & \Vert\mathcal{B}\Vert_\circledast + \lambda_1\Vert\mathcal{V}\Vert_1 + \frac{\lambda_2}{2}\Vert\mathcal{G}\Vert^2_F + \lambda_3\Vert\mathcal{H}\Vert_{TV1} \\
                                                             & s.t.\ \mathcal{O}=\mathcal{B}+\mathcal{V}+\mathcal{G},\ \mathcal{H}=\mathcal{V}
  \end{aligned}
  \label{eq13}
\end{equation}
where $\lambda_1,\lambda_2,\lambda_3$ are positive scalar equilibrium factors. Considering $\Vert\mathcal{H}\Vert_{TV1}=\Vert\nabla \mathcal{H}\Vert_1$, let $\mathcal{T}\in\mathbb{R}^{M\times N\times 3T}=\nabla \mathcal{H}$, then \hyperref[eq13]{Eq.13} can be converted to
\begin{equation}
  \begin{aligned}
    \min_{\mathcal{B},\mathcal{V},\mathcal{G},\mathcal{H},\mathcal{T}}\  & \Vert\mathcal{B}\Vert_\circledast + \lambda_1\Vert\mathcal{V}\Vert_1 + \frac{\lambda_2}{2}\Vert\mathcal{G}\Vert^2_F + \lambda_3\Vert\mathcal{T}\Vert_1 \\
    s.t.\ \mathcal{O}                                                    & =\mathcal{B}+\mathcal{V}+\mathcal{G},\ \mathcal{H}=\mathcal{V},\ \mathcal{T}=\nabla \mathcal{H}
  \end{aligned}
  \label{eq14}
\end{equation}
\hyperref[eq14]{Eq.14} is the final form of TV-TRPCA optimization problem.

\subsubsection{Optimization Strategy}
For the convex optimization problem of matrix recovery, the common solving methods include accelerated proximal gradient (APG), dual approach (DA), iterative thresholding (IT), augmented Lagrange multipliers (ALM) and alternating direction method of multipliers (ADMM) algorithms\cite{lin2010augmented}.

Since ADMM strategy has low solving complexity and can guarantee the convergence in most cases, we adopt ADMM based on ALM to solve the optimization problem of \hyperref[eq14]{Eq.14}.

The augmented Lagrange function of \hyperref[eq14]{Eq.14} is given by:
\begin{equation}
  \begin{aligned}
    \pmb{\mathcal{L}} (\mathcal{B},\mathcal{V},\mathcal{G},\mathcal{H},\mathcal{T};\mu,\nu)\  & = \Vert\mathcal{B}\Vert_\circledast + \lambda_1\Vert\mathcal{V}\Vert_1 + \frac{\lambda_2}{2}\Vert\mathcal{G}\Vert^2_F + \lambda_3\Vert\mathcal{T}\Vert_1                                                                                                \\
                                                                                              & + \frac{\mu}{2}\Vert\mathcal{O}-\mathcal{B}-\mathcal{V}-\mathcal{G}\Vert_F^2 + \left\langle \mathcal{X},\mathcal{O}-\mathcal{B}-\mathcal{V}-\mathcal{G}\right\rangle                                                                                    \\
                                                                                              & + \frac{\mu}{2}\Vert\mathcal{H}-\mathcal{V}\Vert_F^2 + \left\langle \mathcal{Y},\mathcal{H}-\mathcal{V}\right\rangle + \frac{\nu}{2}\Vert\mathcal{T}-\nabla \mathcal{H}\Vert_F^2 + \left\langle \mathcal{Z},\mathcal{T}-\nabla \mathcal{H}\right\rangle
  \end{aligned}
  \label{eq15}
\end{equation}
where $\mathcal{X},\mathcal{Y}\in\mathbb{R}^{M\times N\times T}$ and $\mathcal{Z}\in\mathbb{R}^{M\times N\times 3T}$ are the Lagrange multiplier tensors, $\mu$ and $\nu$ are positive penalty scalars.

Because of the huge number of variables in the augmented Lagrange function, it is difficult to optimize them simultaneously. With the ADMM strategy, we optimize one variable at a time and constrain the other variables to remain constant, alternating until the convergence condition is satisfied. At the end of each iteration, the values of $\mu$ and $\nu$ will increase, making objective function more sensitive to the change of variables. Specifically, we can decompose the minimization problem of \hyperref[eq15]{Eq.15} into five sub-problems:

\ding{172} $\mathcal{B}$ sub-problem:

In the $(k+1)^{th}$ iteration, for the $\mathcal{B}$ sub-problem optimization, the objective function is:
\begin{equation}
  \begin{aligned}
    \mathcal{B}_{k+1} = & \mathop{\arg\min}\limits_\mathcal{B}\ \pmb{\mathcal{L}}\left(\mathcal{B},\mathcal{V}_k,\mathcal{G}_k,\mathcal{H}_k,\mathcal{T}_k;\mu_k,\nu_k\right)                                                                                                                              \\
    =                   & \mathop{\arg\min}\limits_\mathcal{B}\ \Vert\mathcal{B}\Vert_\circledast + \frac{\mu_k}{2}\Vert\mathcal{O}-\mathcal{B}-\mathcal{V}_k-\mathcal{G}_k\Vert^2_F                          +\left\langle \mathcal{X}_k,\mathcal{O}-\mathcal{B}-\mathcal{V}_k-\mathcal{G}_k\right\rangle \\
    =                   & \mathop{\arg\min}\limits_\mathcal{B}\ \Vert\mathcal{B}\Vert_\circledast + \frac{\mu_k}{2}\Vert\mathcal{O}-\mathcal{B}-\mathcal{V}_k-\mathcal{G}_k+\frac{\mathcal{X}_k}{2}\Vert^2_F                                                                                               \\
    =                   & \mathop{\arg\min}\limits_\mathcal{B}\ \Vert\mathcal{B}\Vert_\circledast + \frac{\mu_k}{2}\Vert\mathcal{A}-\mathcal{B}\Vert^2_F
    \label{eq16}
  \end{aligned}
\end{equation}
where $\mathcal{A}=\mathcal{O}-\mathcal{V}_k-\mathcal{G}_k+\mathcal{X}_k/2$. In \cite{lu2019tensor}, TNN was proved to have a closed-form solution as the proximal operator of the matrix nuclear norm. In this case, suppose $\mathcal{A}=\mathcal{U}\ast\mathcal{S}\ast\mathcal{V}^\ast$ is the t-SVD of $\mathcal{A}\in\mathbb{R}^{M\times N\times T}$, the tensor Singular Value Thresholding (t-SVT) operator is defined as
\begin{equation}
  \pmb{\mathcal{D}}_{\mu_k}(\mathcal{A})=\mathcal{U}\ast\mathcal{S}_{\mu_k^{-1}}\ast\mathcal{V}^\ast
  \label{eq17}
\end{equation}
where
\begin{equation}
  \mathcal{S}_{\mu_k^{-1}}=\pmb{ifft}\left(max\left(\widetilde{\mathcal{S}}-\mu_k^{-1},0\right),[\ ],3\right)
  \label{eq18}
\end{equation}
The t-SVT operator $\pmb{\mathcal{D}}_{\mu_k}(\mathcal{A})$ simply applies a soft-thresholding rule to the singular values $\widetilde{\mathcal{S}}$ of the frontal slices of $\widetilde{\mathcal{A}}$, effectively shrinking these towards zero\cite{lu2019tensor}. Finally, the $\mathcal{B}$ sub-problem optimization can be computed by
\begin{equation}
  \mathcal{B}_{k+1}=\pmb{\mathcal{D}}_{\mu_k}\left(\mathcal{O}-\mathcal{V}_k-\mathcal{G}_k+\frac{\mathcal{X}_k}{2}\right)
  \label{eq19}
\end{equation}

\ding{173} $\mathcal{V}$ sub-problem:

In the $(k+1)^{th}$ iteration, for the $\mathcal{V}$ sub-problem optimization, the objective function is:
\begin{equation}
  \begin{aligned}
    \mathcal{V}_{k+1} = & \mathop{\arg\min}\limits_\mathcal{V}\ \pmb{\mathcal{L}}\left(\mathcal{B}_{k+1},\mathcal{V},\mathcal{G}_k,\mathcal{H}_k,\mathcal{T}_k;\mu_k,\nu_k\right)                                                                                                         \\
    =                   & \mathop{\arg\min}\limits_\mathcal{V}\ \lambda_1\Vert\mathcal{V}\Vert_1 + \frac{\mu_k}{2}\Vert\mathcal{O}-\mathcal{B}_{k+1}-\mathcal{V}-\mathcal{G}_k\Vert^2_F + \left\langle \mathcal{X}_k,\mathcal{O}-\mathcal{B}_{k+1}-\mathcal{V}-\mathcal{G}_k\right\rangle \\
                        & + \frac{\mu_k}{2}\Vert\mathcal{H}_k-\mathcal{V}\Vert^2_F + \left\langle \mathcal{Y}_k,\mathcal{H}_k-\mathcal{V}\right\rangle                                                                                                                                    \\
    =                   & \mathop{\arg\min}\limits_\mathcal{V}\ \lambda_1\Vert\mathcal{V}\Vert_1 + \mu_k\Vert\mathcal{Q}-\mathcal{V}\Vert^2_F                                                                                                                                             \\
    =                   & \pmb{\mathcal{S}}_{\frac{\lambda_1}{2\mu_k}}(\mathcal{Q})
    \label{eq20}
  \end{aligned}
\end{equation}
where $\mathcal{Q}=(\mathcal{O}-\mathcal{B}_{k+1}-\mathcal{G}_k+\mathcal{H}_k)/2+(\mathcal{X}_k+\mathcal{Y}_k)/2\mu_k$. $\pmb{\mathcal{S}}_a(b)=sgn(b)max(|b|-a,0)$ is the shrinkage operator of a scalar and \hyperref[eq20]{Eq.20} is the extension to tensor $\mathcal{Q}$ element-wisely.

\ding{174} $\mathcal{G}$ sub-problem:

In the $(k+1)^{th}$ iteration, for the $\mathcal{G}$ sub-problem optimization, the objective function is:
\begin{equation}
  \begin{aligned}
    \mathcal{G}_{k+1} = & \mathop{\arg\min}\limits_\mathcal{G}\ \pmb{\mathcal{L}}\left(\mathcal{B}_{k+1},\mathcal{V}_{k+1},\mathcal{G},\mathcal{H}_k,\mathcal{T}_k;\mu_k,\nu_k\right)                                                                                                                         \\
    =                   & \mathop{\arg\min}\limits_\mathcal{G}\ \frac{\lambda_2}{2}\Vert\mathcal{G}\Vert^2_F + \frac{\mu_k}{2}\Vert\mathcal{O}-\mathcal{B}_{k+1}-\mathcal{V}_{k+1}-\mathcal{G}\Vert^2_F + \left\langle \mathcal{X}_k,\mathcal{O}-\mathcal{B}_{k+1}-\mathcal{V}_{k+1}-\mathcal{G}\right\rangle \\
    =                   & \frac{\mu_k}{\lambda_2+\mu_k}\left(\mathcal{O}-\mathcal{B}_{k+1}-\mathcal{V}_{k+1}+\frac{\mathcal{X}_k}{\mu_k}\right)
    \label{eq21}
  \end{aligned}
\end{equation}

\ding{175} $\mathcal{H}$ sub-problem:
In the $(k+1)^{th}$ iteration, for the $\mathcal{H}$ sub-problem optimization, the objective function is:
\begin{equation}
  \begin{aligned}
    \mathcal{H}_{k+1} = & \mathop{\arg\min}\limits_\mathcal{H}\ \pmb{\mathcal{L}}\left(\mathcal{B}_{k+1},\mathcal{V}_{k+1},\mathcal{G}_{k+1},\mathcal{H},\mathcal{T}_k;\mu_k,\nu_k\right)                                                                                                                                                     \\
    =                   & \mathop{\arg\min}\limits_\mathcal{H}\ \frac{\mu_k}{2}\Vert\mathcal{H}-\mathcal{V}_{k+1}\Vert^2_F + \left\langle \mathcal{Y}_k,\mathcal{H}-\mathcal{V}_{k+1}\right\rangle + \frac{\nu_k}{2}\Vert\mathcal{T}_k-\nabla \mathcal{H}\Vert^2_F + \left\langle \mathcal{Z}_k,\mathcal{T}_k-\nabla \mathcal{H}\right\rangle
    \label{eq22}
  \end{aligned}
\end{equation}

Since there are tensors of different sizes in \hyperref[eq22]{Eq.22}, for convenience, we use tensor vectorization to simplify the computation. Define operator $\textbf{Vec}(\mathcal{X})=f(\mathcal{X})\in\mathbb{R}^{MNT}$ transforms the 3-D tensor $\mathcal{X}\in\mathbb{R}^{M\times N\times T}$ to a 1-D vector and $f^{-1}(\mathcal{X})$ is the inverse operator, the variances of $\mathcal{X}$ in each dimension can be vectorized into the product of a block circulant matrix and the tensor vectorization $f(\mathcal{X})$, i.e.,
\begin{equation}
  f(\nabla_i \mathcal{H}) = \pmb{D}_if(\mathcal{H}),\ i=m,n,t
  \label{eq23}
\end{equation}
where $\pmb{D}_m,\pmb{D}_n,\pmb{D}_t\in\mathbb{R}^{MNT\times MNT}$ are block circulant matrices. Let $\pmb{D}=\left[\sigma_m\pmb{D}_m^T,\sigma_n\pmb{D}_n^T,\sigma_t\pmb{D}_t^T\right]^T\in\mathbb{R}^{3MNT\times MNT}$, according to \hyperref[eq10]{Eq.10}, we have
\begin{equation}
  f(\nabla \mathcal{H})= \pmb{D}f(\mathcal{H})
  \label{eq24}
\end{equation}

Now we can rewrite \hyperref[eq22]{Eq.22} as follows (the derivation use the property that the transpose of a scalar is itself):
\begin{equation}
  \begin{aligned}
    \mathcal{H}_{k+1} = & \mathop{\arg\min}\limits_\mathcal{H}\ \frac{\mu_k}{2}\Vert\mathcal{H}-\mathcal{V}_{k+1}\Vert^2_F + \left\langle \mathcal{Y}_k,\mathcal{H}-\mathcal{V}_{k+1}\right\rangle + \frac{\nu_k}{2}\Vert f(\mathcal{T}_k)-\pmb{D}f(\mathcal{H})\Vert^2_F + \left\langle f(\mathcal{Z}_k),f(\mathcal{T}_k)-\pmb{D}f(\mathcal{H})\right\rangle \\
    =                   & \mathop{\arg\min}\limits_\mathcal{H}\ \frac{1}{2}\left(\mu_kf^T(\mathcal{H})f(\mathcal{H})+\nu_kf^T(\mathcal{H})\pmb{D}^T\pmb{D}f(\mathcal{H})\right) - \mu_kf^T(\mathcal{H})f\left(\mathcal{V}_{k+1}-\frac{\mathcal{Y}_k}{\mu_k}\right) - \nu_kf^T(\mathcal{H})\pmb{D}^Tf\left(\mathcal{T}_k+\frac{\mathcal{Z}_k}{\nu_k}\right)    \\
    =                   & \mathop{\arg\min}\limits_\mathcal{H}\ \pmb{\phi}(f(\mathcal{H}))
    \label{eq25}
  \end{aligned}
\end{equation}
Notice that
\begin{equation}
  \pmb{D}^T\pmb{D}=\sigma_m^2\pmb{D}_m^T\pmb{D}_m+\sigma_n^2\pmb{D}_n^T\pmb{D}_n+\sigma_t^2\pmb{D}_t^T\pmb{D}_t
  \label{eq26}
\end{equation}
is positive definite, so \hyperref[eq25]{Eq.25} can be solved by the normal equation of convex function $\pmb{\phi}(f(\mathcal{H}))$.

Let $\partial\pmb{\phi}(f(\mathcal{H}))/\partial f(\mathcal{H})=0$, we can get
\begin{equation}
  \left(\mu_k\pmb{I}+\nu_k\pmb{D}^T\pmb{D}\right)f(\mathcal{H}_{k+1})= f(\mathcal{K})
  \label{eq27}
\end{equation}
where
\begin{equation}
  \mathcal{K}=\mu_k\left(\mathcal{V}_{k+1}-\dfrac{\mathcal{Y}_k}{\mu_k}\right)+\nu_kf^{-1}\left(\pmb{D}^Tf\left(\mathcal{T}_k+\dfrac{\mathcal{Z}_k}{\nu_k}\right)\right)
  \label{eq28}
\end{equation}

$\mathcal{H}_{k+1}$ can be directly solved by computing the Moore-Penrose pseudoinverse of matrix $\mu_k\pmb{I}+\nu_k\pmb{D}^T\pmb{D}\in\mathbb{R}^{MNT\times MNT}$, but this will occupy huge memory space, resulting in low computational efficiency. Consider that the block circulant matrix can be diagonalized by the DFT matrix\cite{golub2013matrix}, hence \hyperref[eq27]{Eq.27} has the following solution (See \hyperref[APPENDIX]{Appendix A} for the details)
\begin{equation}
  \mathcal{H}_{k+1}= f^{-1}\left(\pmb{ifft}\left(\dfrac{\pmb{fft}(f(\mathcal{K}))}{diagv\left(\mu_k\pmb{I}+\nu_k\pmb{K}\right)}\right)\right)
  \label{eq29}
\end{equation}
where
\begin{equation}
  \pmb{K}=\sigma_m^2|\pmb{fft}(\pmb{D}_m)|^2+\sigma_n^2|\pmb{fft}(\pmb{D}_n)|^2+\sigma_t^2|\pmb{fft}(\pmb{D}_t)|^2
  \label{eq30}
\end{equation}
where $diagv(\cdot)$ forms a column vector from the elements of the diagonal of a matrix, $|\cdot|^2$ is the element-wise square, and the division is also performed element-wisely. The denominator in \hyperref[eq29]{Eq.29} can be precalculated outside the iteration, reducing the computational complexity.

\ding{176} $\mathcal{T}$ sub-problem:

In the $(k+1)^{th}$ iteration, for the $\mathcal{T}$ sub-problem optimization, the objective function is:
\begin{equation}
  \begin{aligned}
    \mathcal{T}_{k+1} = & \mathop{\arg\min}\limits_\mathcal{T}\ \pmb{\mathcal{L}}\left(\mathcal{B}_{k+1},\mathcal{V}_{k+1},\mathcal{G}_{k+1},\mathcal{H}_{k+1},\mathcal{T};\mu_k,\nu_k\right)                                                       \\
    =                   & \mathop{\arg\min}\limits_\mathcal{T}\ \lambda_3\Vert\mathcal{T}\Vert_1 + \frac{\nu_k}{2}\Vert\mathcal{T}-\nabla \mathcal{H}_{k+1}\Vert^2_F + \left\langle \mathcal{Z}_k,\mathcal{T}-\nabla \mathcal{H}_{k+1}\right\rangle \\
    =                   & \mathop{\arg\min}\limits_\mathcal{T}\ \lambda_3\Vert\mathcal{T}\Vert_1 + \frac{\nu_k}{2}\Vert\mathcal{T}-\nabla \mathcal{H}_{k+1}+\frac{\mathcal{Z}_k}{\nu_k}\Vert^2_F                                                    \\
    =                   & \pmb{\mathcal{S}}_{\frac{\lambda_3}{\nu_k}}(\mathcal{W})
    \label{eq31}
  \end{aligned}
\end{equation}
where $\mathcal{W}=\nabla \mathcal{H}_{k+1}-\mathcal{Z}_k/\nu_k$. The solution of $\mathcal{T}$ sub-problem is similar to $\mathcal{V}$ sub-problem.

After solving all the five sub-problems, the Lagrange multipliers and penalty
scalars are updated by
\begin{equation}
  \left\{\begin{aligned}
    \mathcal{X}_{k+1} & = \mathcal{X}_k + \mu_k(\mathcal{O}-\mathcal{B}_{k+1}-\mathcal{V}_{k+1}-\mathcal{G}_{k+1}) \\
    \mathcal{Y}_{k+1} & = \mathcal{Y}_k + \mu_k(\mathcal{H}_{k+1}-\mathcal{V}_{k+1})                               \\
    \mathcal{Z}_{k+1} & = \mathcal{Z}_k + \nu_k(\mathcal{T}_{k+1}-\nabla \mathcal{H}_{k+1})                        \\
    \mu_{k+1}         & =\min(\rho\mu_k,\mu_{max})                                                                 \\
    \nu_{k+1}         & =\min(\rho\nu_k,\nu_{max})
  \end{aligned}\
  \right.
  \label{eq32}
\end{equation}
where $\rho$ is the increase ratio of penalty scalars $\mu$ and $\nu$.

The entire algorithm of optimizing \hyperref[eq14]{Eq.14} has been summarized in \hyperref[al1]{Algorithm 1}. The iteration terminates when the convergence condition is satisfied, or the maximal iteration is reached.

\begin{figure}[!t]
  \centering
  \hspace{1.7cm}
  \subfigure[]{
    \begin{overpic}[width=0.25\linewidth]{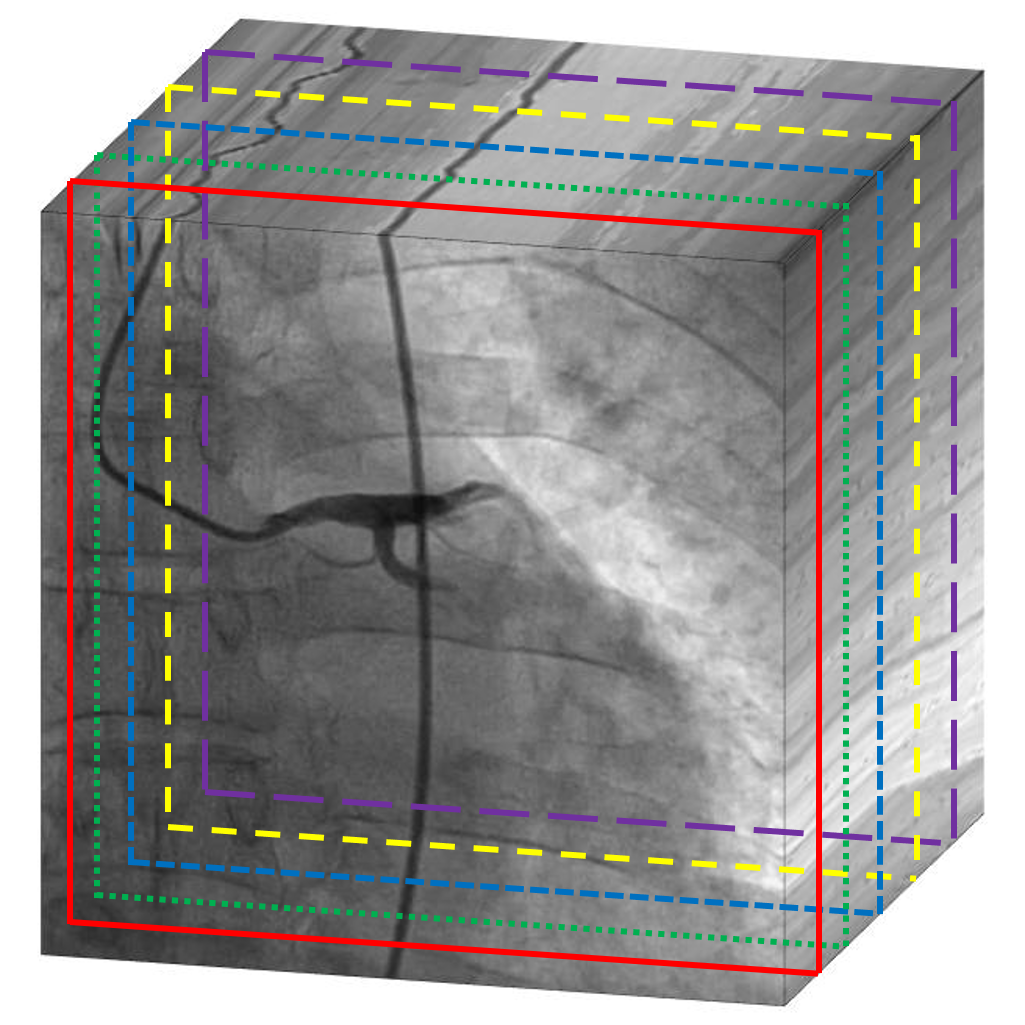}
      \put(0,90){\color{black}{(a)}}
    \end{overpic}
  }
  \hspace{2cm}
  \subfigure[]{
    \begin{overpic}[width=0.16\linewidth]{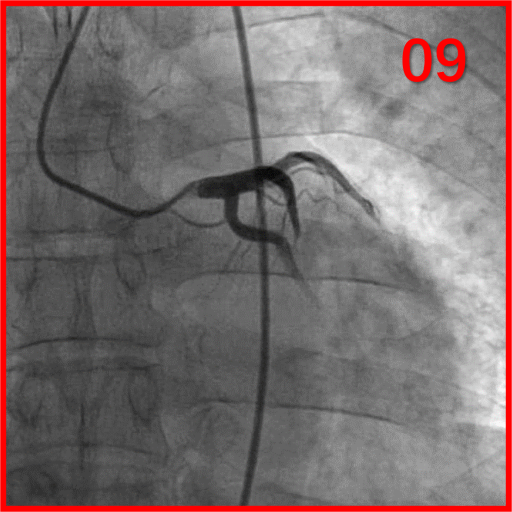}
      \put(3,5){\color{white}{(b1)}}
    \end{overpic}
  }
  \subfigure[]{
    \hspace{-0.3cm}
    \begin{overpic}[width=0.16\linewidth]{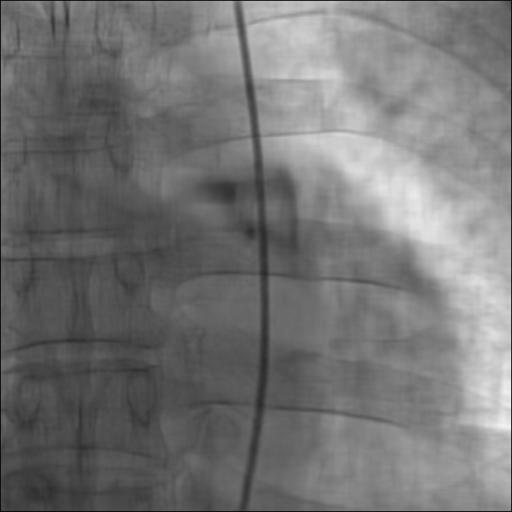}
      \put(3,5){\color{white}{(b2)}}
    \end{overpic}
  }
  \subfigure[]{
    \hspace{-0.3cm}
    \begin{overpic}[width=0.16\linewidth]{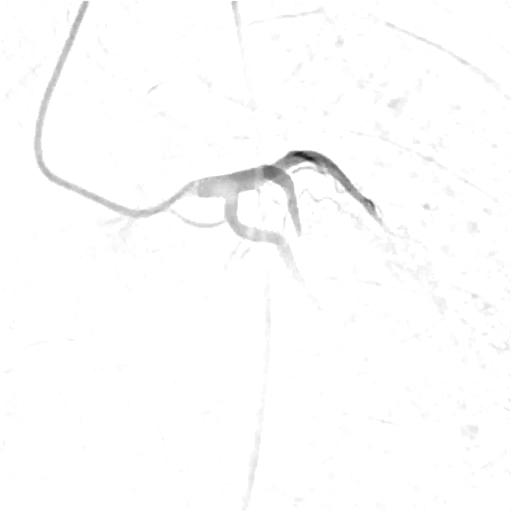}
      \put(3,5){\color{black}{(b3)}}
    \end{overpic}
  }

  \vspace{-0.95cm}
  \subfigure[]{
    \begin{overpic}[width=0.16\linewidth]{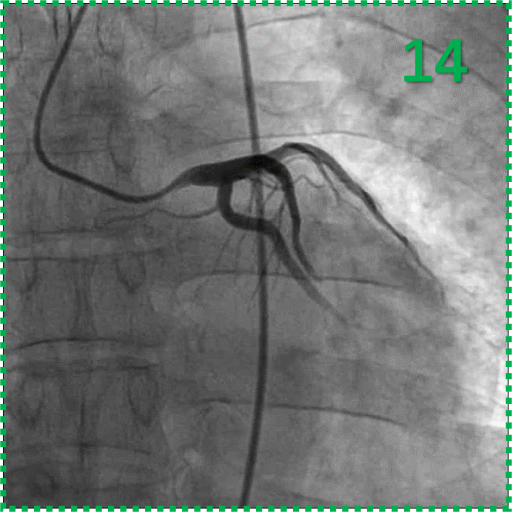}
      \put(3,5){\color{white}{(c1)}}
    \end{overpic}
  }
  \subfigure[]{
    \hspace{-0.3cm}
    \begin{overpic}[width=0.16\linewidth]{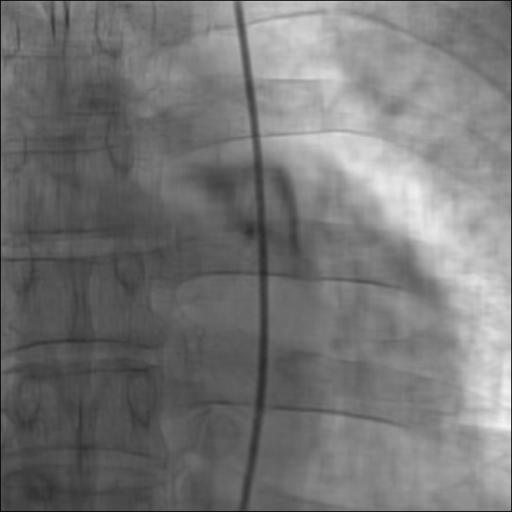}
      \put(3,5){\color{white}{(c2)}}
    \end{overpic}
  }
  \subfigure[]{
    \hspace{-0.3cm}
    \begin{overpic}[width=0.16\linewidth]{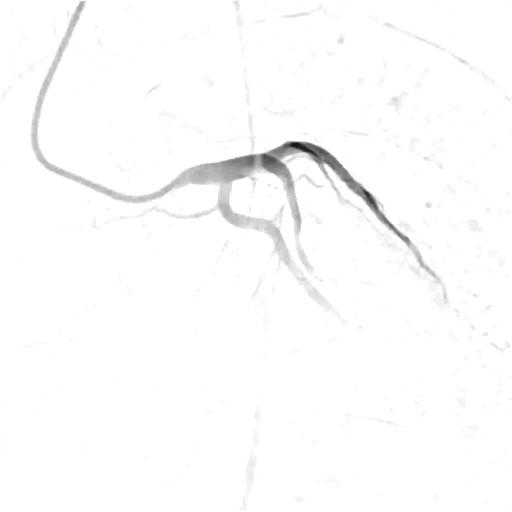}
      \put(3,5){\color{black}{(c3)}}
    \end{overpic}
  }
  \subfigure[]{
    \begin{overpic}[width=0.16\linewidth]{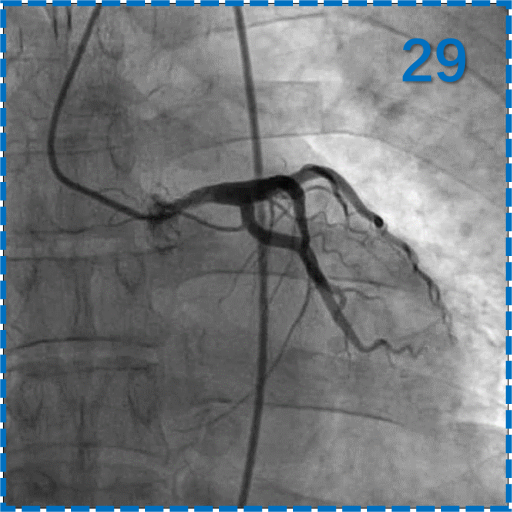}
      \put(3,5){\color{white}{(d1)}}
    \end{overpic}
  }
  \subfigure[]{
    \hspace{-0.3cm}
    \begin{overpic}[width=0.16\linewidth]{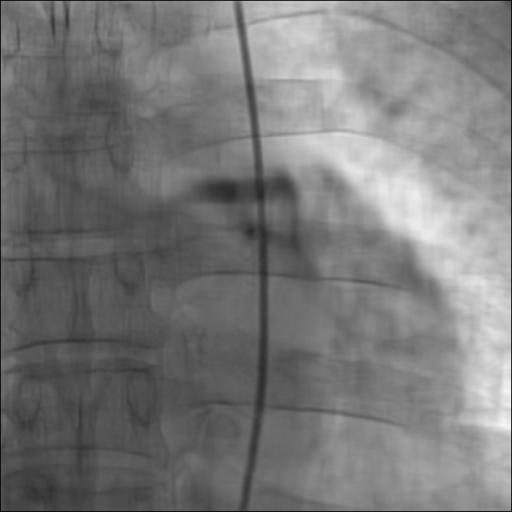}
      \put(3,5){\color{white}{(d2)}}
    \end{overpic}
  }
  \subfigure[]{
    \hspace{-0.3cm}
    \begin{overpic}[width=0.16\linewidth]{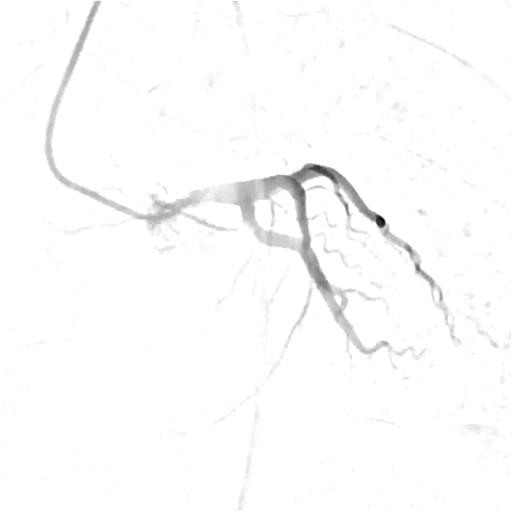}
      \put(3,5){\color{black}{(d3)}}
    \end{overpic}
  }

  \vspace{-0.95cm}
  \subfigure[]{
    \begin{overpic}[width=0.16\linewidth]{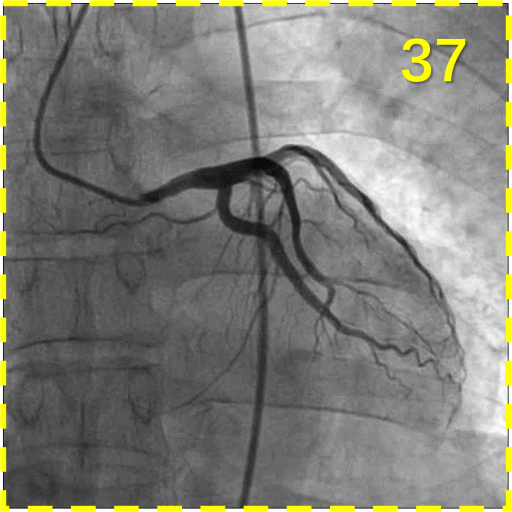}
      \put(3,5){\color{white}{(e1)}}
    \end{overpic}
  }
  \subfigure[]{
    \hspace{-0.3cm}
    \begin{overpic}[width=0.16\linewidth]{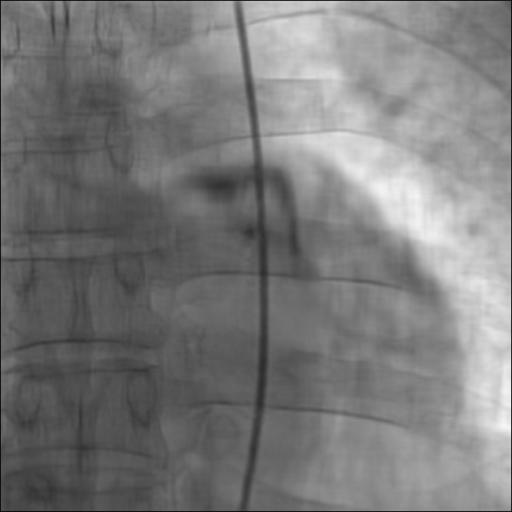}
      \put(3,5){\color{white}{(e2)}}
    \end{overpic}
  }
  \subfigure[]{
    \hspace{-0.3cm}
    \begin{overpic}[width=0.16\linewidth]{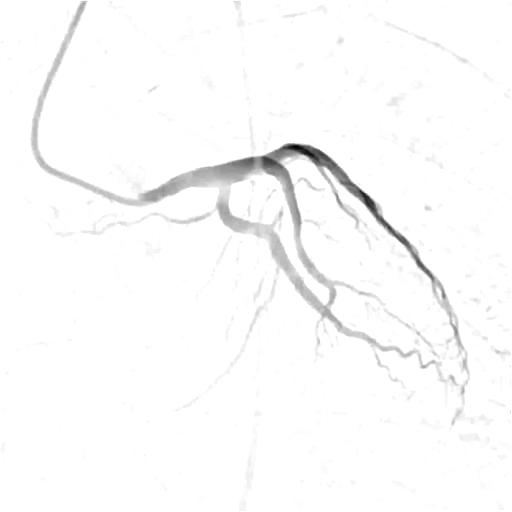}
      \put(3,5){\color{black}{(e3)}}
    \end{overpic}
  }
  \subfigure[]{
    \begin{overpic}[width=0.16\linewidth]{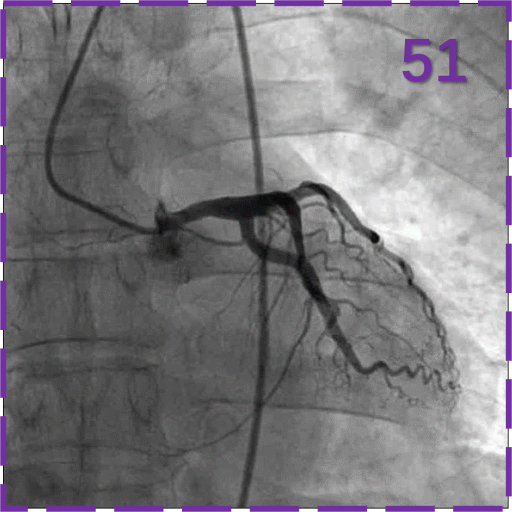}
      \put(3,5){\color{white}{(f1)}}
    \end{overpic}
  }
  \subfigure[]{
    \hspace{-0.3cm}
    \begin{overpic}[width=0.16\linewidth]{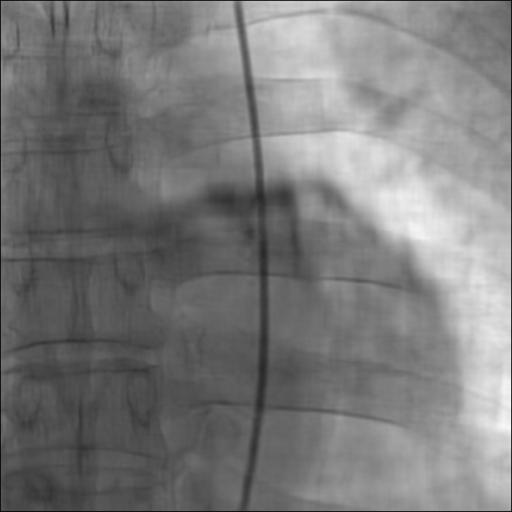}
      \put(3,5){\color{white}{(f2)}}
    \end{overpic}
  }
  \subfigure[]{
    \hspace{-0.3cm}
    \begin{overpic}[width=0.16\linewidth]{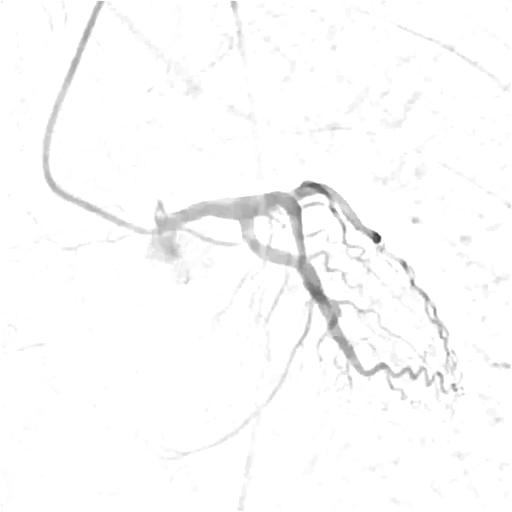}
      \put(3,5){\color{black}{(f3)}}
    \end{overpic}
  }

  \label{fig2}
  \vspace{-0.8cm}
  \caption{TV-TRPCA foreground extraction. (a) is the 3D-tensor formed by XCA image sequence and its $9^{th},14^{th},29^{th},37^{th},51^{th}$ frontal slices. (b1)-(f1) are the raw image, (b2)-(f2) are the background layers, (b3)-(f3) are the vessel foreground layers. In order to improve the visibility of vessel layers, gray scale normalization is adopted.}
\end{figure}

We use TV-TRPCA method to process an XCA image sequence and show the results of foreground extraction for different frames in \hyperref[fig2]{Fig.2}. In frame 9, the contrast agent had just been injected. In frame 14-37, the contrast agent gradually diffused completely. In frame 51, the contrast agent began to dissipate. In order to facilitate observation, we normalized the gray scale of the foreground layer. It can be found that the vessel layer is well separated during the whole process, and the periodic dynamic disturbance and high-frequency noise are well suppressed. The background also retains high low-rank property.

\subsection{Vessel Segmentation}
The vessel layer extracted by TV-TRPCA could have uneven grayscale distribution and local fracture, affecting the accuracy of subsequent quantitative analysis. The method of vessel segmentation can effectively solve these two problems.

\subsubsection{Region Growing}
Region growing is a model-based image segmentation method. Based on the similarity of gray values, region growing method combines pixels in digital images according to certain rules to form regions of interest (ROI)\cite{tremeau1997region}. Therefore, this method mainly needs to determine the seed point of growing, the region growing rule and the termination condition of stopping growing.

In this paper, seed point is automatically selected based on the maximum criterion of local gray intensity. To be specific, a disk with its diameter close to aorta is used to match the extracted vessel layer image. When all points in the disk have the densest gray scale, the center of the disk is the initial seed point of region growing, which can effectively avoid the interference caused by residual noise. According to the gray threshold, the points in its 4-neighborhood are classified and  the next seed point is selected until there is no pixel point that meets the growing rule.

\subsubsection{RLF filter}
The region growing method can extract continuous vessel branches extending from the aorta, but broken segments will be deleted. Consider the unique ridge structure of the minor segment, we adopt the Radon-like features (RLF) filter\cite{kumar2010radon} to enhance the extracted vessel layer image. It uses edge sensing in different directions to make non-isotropic sampling of the neighborhood, which can highlight and extend the tubular structures.

Before extracting radon features of coronary artery, the image was first convolved by an edge enhancing Gaussian second derivative (GSD) filter as follows\cite{kumar2010radon}
\begin{equation}
  \pmb{G}(x,y)=\max_{\sigma,\theta}\Delta\pmb{G}(\sigma,\theta)\otimes \pmb{I}(x,y)
  \label{eq33}
\end{equation}
where $\sigma$ and $\theta$ are the scale and direction of GSD filter, $\otimes$ is the convolution operator, $\pmb{I}$ is the original image.

The nodes of Radon-like features are defined by the edge graph of image $\pmb{G}(x,y)$ while the extraction function is defined as a mean function between two nodes, i.e.,\cite{kumar2010radon}
\begin{equation}
  \pmb{T}(\pmb{I},\pmb{l}(t))=\frac{\int_{t_i}^{t_{i+1}}\pmb{G}(\pmb{l}(t))\partial t}{\Vert\pmb{l}(t_{i+1})-\pmb{l}(t_i)\Vert_2},\ t\in\left[t_i,t_{i+1}\right]
  \label{eq34}
\end{equation}
where $\pmb{l}$ is the scanning line along which features are enhanced, $t$ is the nodes set where $\pmb{l}$ intersects the edge graph. Changing the slope and intercept of $\pmb{l}$, different images can be obtained by combine the extraction function value of each pixel. Averaging the gray values of these images gives the final RLF filtering result $\pmb{R}$.

\begin{figure*}[!t]
  \centering
  \subfigure[]{
    \includegraphics[width=0.99\linewidth]{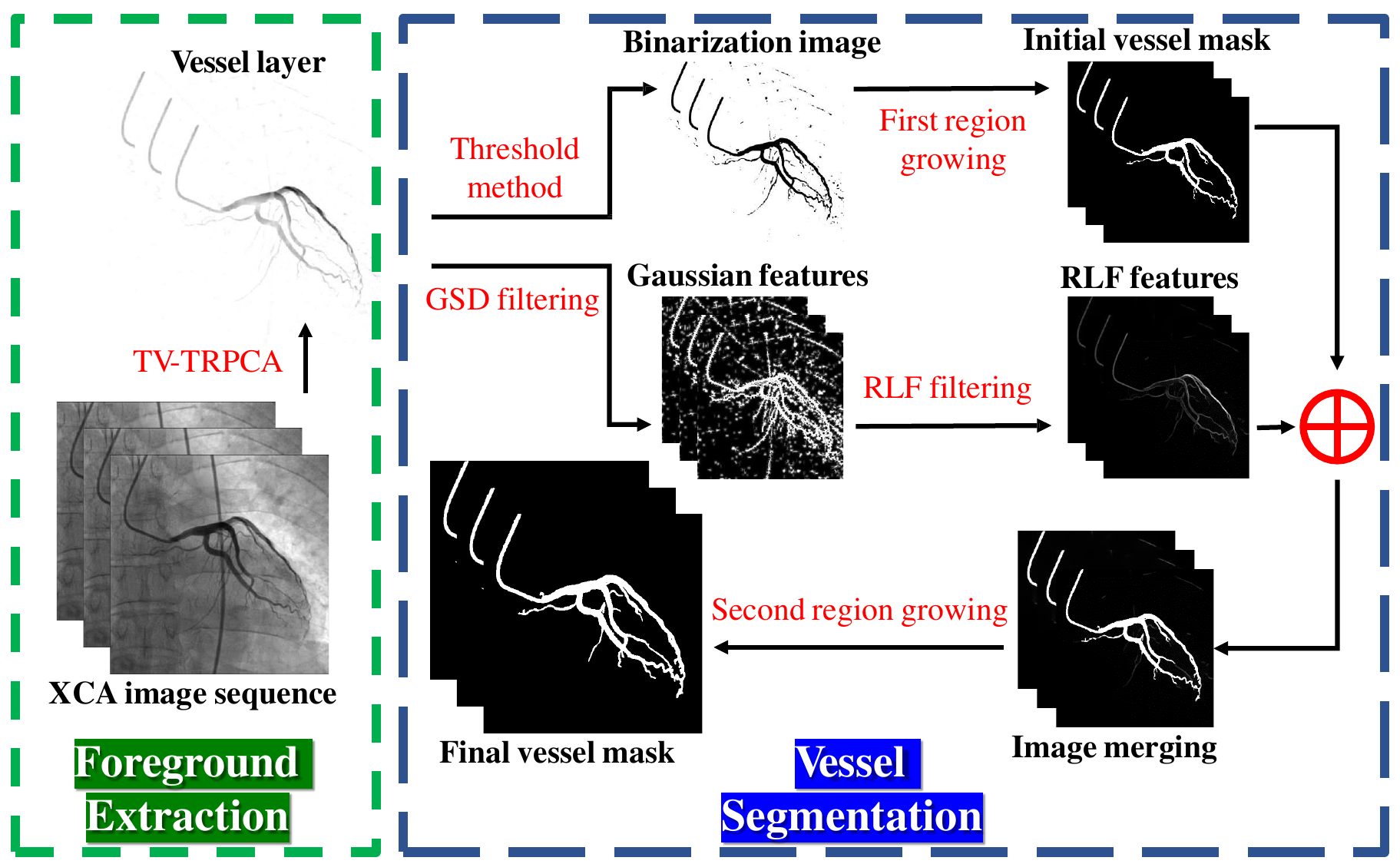}
  }

  \label{fig3}
  \vspace{-0.8cm}
  \caption{The whole process of proposed method, including TV-TRPCA foreground extraction and TSRG vessel segmentation.}
\end{figure*}

\subsubsection{Two-stage Region Growing}
Although RLF filtering enhance and connect the minor segments. As shown in \hyperref[fig3]{Fig.3}, for thick vessels such as the aorta, it only enhances the edge information, producing internal voids. So RLF filtering is not suitable for global vessel segmentation. Based on the previous analysis, a two-stage regional growing (TSRG) vessel segmentation method is presented in this section. The sections in the blue box in \hyperref[fig3]{Fig.3} gives a more intuitive representation of the details.

In the first stage, a global threshold method is used as the preprocessing to delete the residual low gray level noise. Since most of the noise has been removed after foreground extraction, we use 95\% of the maximum gray intensity as the threshold for binarization. Compared with other complex threshold segmentation algorithms like OTSU\cite{otsu1979threshold}, this procedure can preserve more vessel information in low-noise images. The vessel mask of main branches can be obtained by performing region growing on the binary image.

In the second stage, the RLF filtering is utilized on the extracted vessel layer to get an edge-like graph. Superimposing (summing their gray scale values) this edge-like graph on the vessel mask obtained in the first stage gives an image with more detailed features. The second region growing is performed on this image to obtain the final vessel mask.

\subsection{Overview of the Whole Process}
The whole process of our method is shown in \hyperref[fig3]{Fig.3}. We first use TV-TRPCA algorithm to extract the vessel foreground layer, then perform TSRG algorithm for vessel segmentation. Finally, a complete topologically correct binary coronary artery mask was obtained.

\section{Experimental Results} \label{Experiment}

\subsection{Raw XCA Data and Gold Standard}

We first took 7 clinical XCA image sequences as the raw experimental data for the evaluation of proposed TV-TRPCA and TSRG methods. All these sequences were obtained from Ruijin Hospital affiliated to Shanghai Jiao Tong University, including left coronary artery (LCA) and right coronary artery (RCA) angiograms in different positions such as AP, LAO/RAO and CRAN/CAUD. Each sequence is consisted of 60 continuous frames with resolution resized to 512$\times$512 pixels and 8 bits per pixel. Furthermore, to exam the stability and applicability of our methods, we did the same experiment with the dataset of TÜRKER TUNCER uploaded on \textit{Kaggle}\footnote{\url{https://www.kaggle.com/datasets/turkertuncer/coronary-angiography-print}}. This dataset contains 51 groups of XCA image sequences in four different positions, and each sequence is consisted of 10 continuous frames with resolution of 512$\times$512 pixels and 8 bits per pixel. In the experiment, we randomly selected 10 sequences from the dataset for calculation.

\begin{figure*}[tbp]
  \centering
  \subfigure[]{
    \begin{overpic}[width=0.089\linewidth]{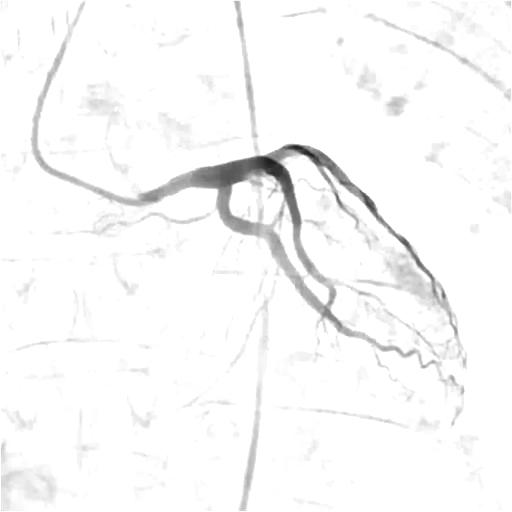}
      \begin{footnotesize}
        \put(3,5){\color{black}{(A1)}}
      \end{footnotesize}
      \begin{tiny}
        \put(52,83){\color{blue}{$\frac{\lambda_1}{\lambda_0}$=0.05}}
      \end{tiny}
    \end{overpic}
  }
  \subfigure[]{
    \hspace{-0.33cm}
    \begin{overpic}[width=0.089\linewidth]{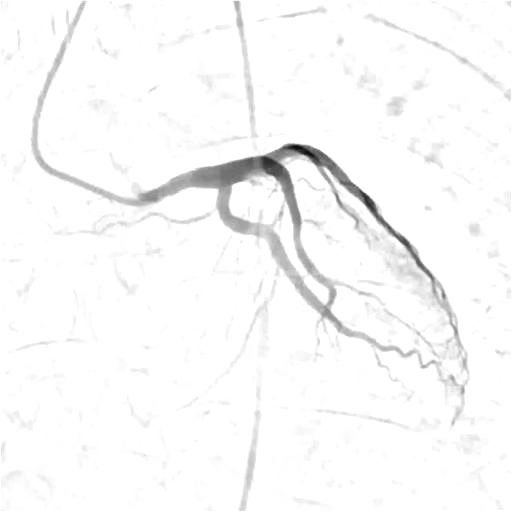}
      \begin{footnotesize}
        \put(3,5){\color{black}{(B1)}}
      \end{footnotesize}
      \begin{tiny}
        \put(52,83){\color{blue}{$\frac{\lambda_1}{\lambda_0}$=0.1}}
      \end{tiny}
    \end{overpic}
  }
  \subfigure[]{
    \hspace{-0.33cm}
    \begin{overpic}[width=0.089\linewidth]{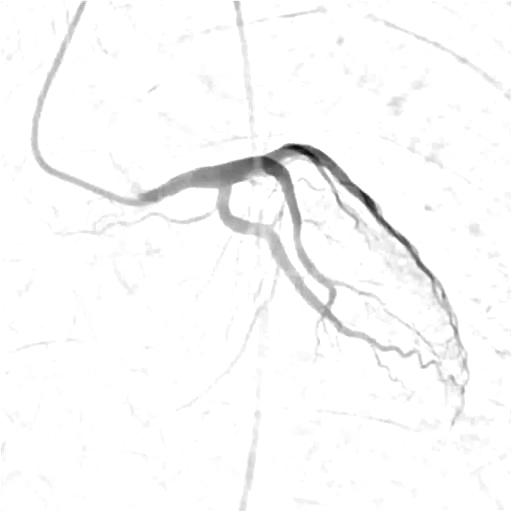}
      \begin{footnotesize}
        \put(3,5){\color{black}{(C1)}}
      \end{footnotesize}
      \begin{tiny}
        \put(52,83){\color{blue}{$\frac{\lambda_1}{\lambda_0}$=0.15}}
      \end{tiny}
    \end{overpic}
  }
  \subfigure[]{
    \hspace{-0.33cm}
    \begin{overpic}[width=0.089\linewidth]{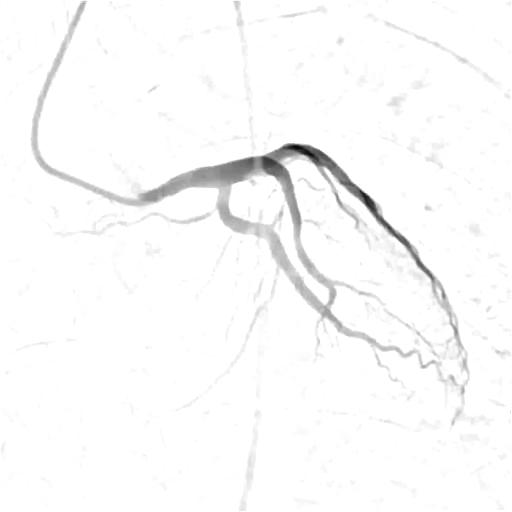}
      \begin{footnotesize}
        \put(3,5){\color{black}{(D1)}}
      \end{footnotesize}
      \begin{tiny}
        \put(52,83){\color{blue}{$\frac{\lambda_1}{\lambda_0}$=0.2}}
      \end{tiny}
    \end{overpic}
  }
  \subfigure[]{
    \hspace{-0.33cm}
    \begin{overpic}[width=0.089\linewidth]{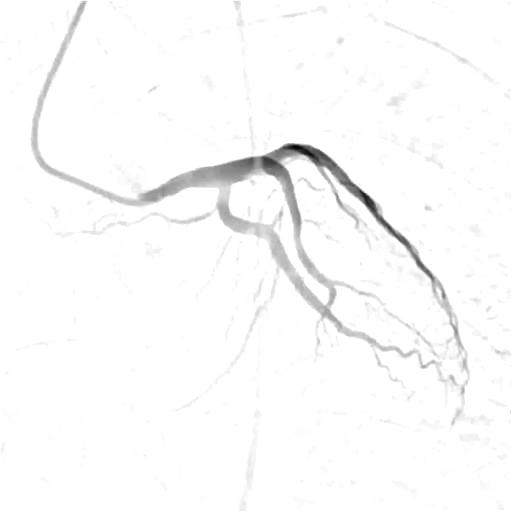}
      \begin{footnotesize}
        \put(3,5){\color{black}{(E1)}}
      \end{footnotesize}
      \begin{tiny}
        \put(52,83){\color{blue}{$\frac{\lambda_1}{\lambda_0}$=0.25}}
      \end{tiny}
    \end{overpic}
  }
  \subfigure[]{
    \hspace{-0.33cm}
    \begin{overpic}[width=0.089\linewidth]{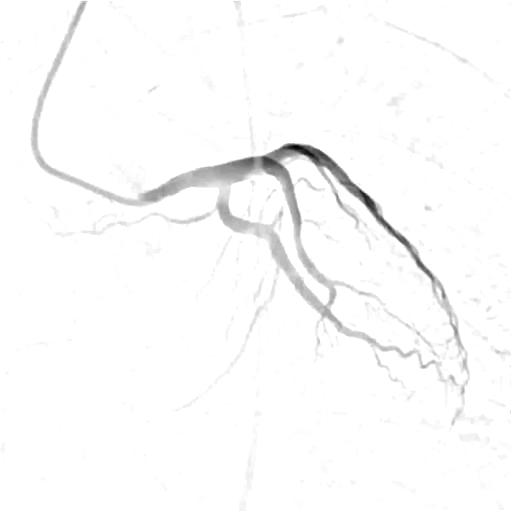}
      \begin{footnotesize}
        \put(3,5){\color{black}{(F1)}}
      \end{footnotesize}
      \begin{tiny}
        \put(52,83){\color{blue}{$\frac{\lambda_1}{\lambda_0}$=0.3}}
      \end{tiny}
    \end{overpic}
  }
  \subfigure[]{
    \hspace{-0.33cm}
    \begin{overpic}[width=0.089\linewidth]{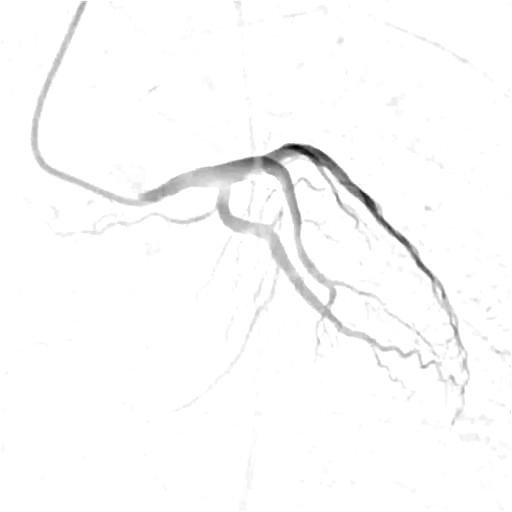}
      \begin{footnotesize}
        \put(3,5){\color{black}{(G1)}}
      \end{footnotesize}
      \begin{tiny}
        \put(52,83){\color{blue}{$\frac{\lambda_1}{\lambda_0}$=0.35}}
      \end{tiny}
    \end{overpic}
  }
  \subfigure[]{
    \hspace{-0.33cm}
    \begin{overpic}[width=0.089\linewidth]{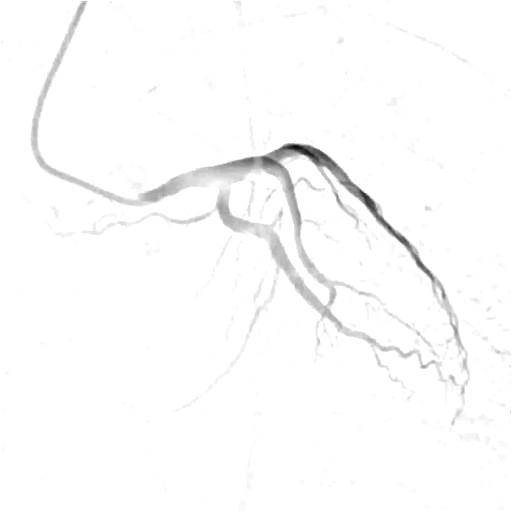}
      \begin{footnotesize}
        \put(3,5){\color{black}{(H1)}}
      \end{footnotesize}
      \begin{tiny}
        \put(52,83){\color{blue}{$\frac{\lambda_1}{\lambda_0}$=0.4}}
      \end{tiny}
    \end{overpic}
  }
  \subfigure[]{
    \hspace{-0.33cm}
    \begin{overpic}[width=0.089\linewidth]{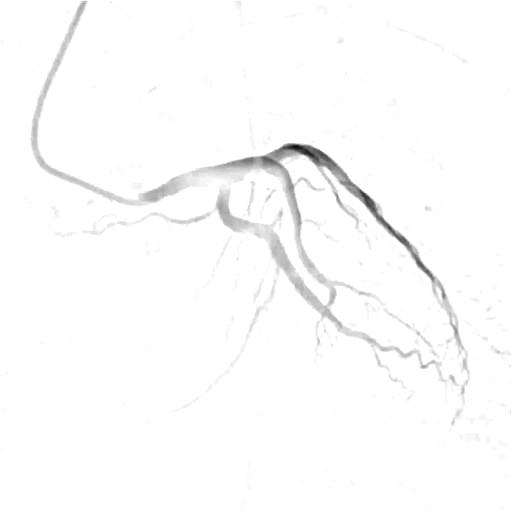}
      \begin{footnotesize}
        \put(3,5){\color{black}{(I1)}}
      \end{footnotesize}
      \begin{tiny}
        \put(52,83){\color{blue}{$\frac{\lambda_1}{\lambda_0}$=0.45}}
      \end{tiny}
    \end{overpic}
  }
  \subfigure[]{
    \hspace{-0.33cm}
    \begin{overpic}[width=0.089\linewidth]{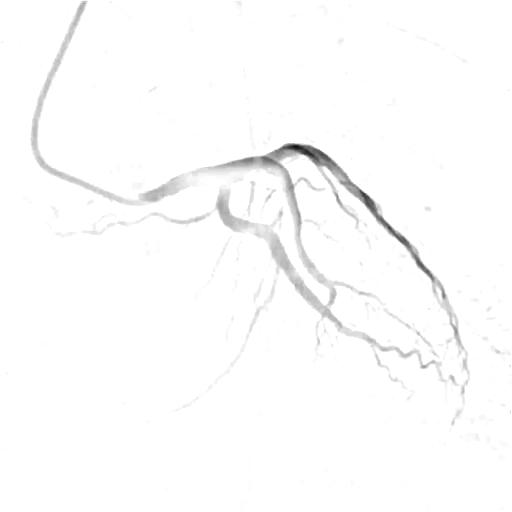}
      \begin{footnotesize}
        \put(3,5){\color{black}{(J1)}}
      \end{footnotesize}
      \begin{tiny}
        \put(52,83){\color{blue}{$\frac{\lambda_1}{\lambda_0}$=0.5}}
      \end{tiny}
    \end{overpic}
  }
  \subfigure[]{
    \hspace{-0.33cm}
    \begin{overpic}[width=0.089\linewidth]{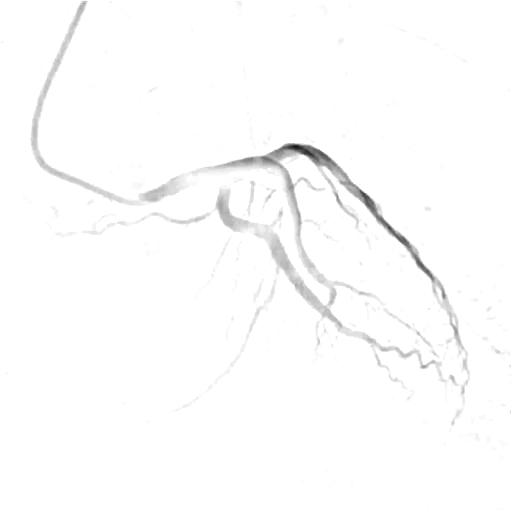}
      \begin{footnotesize}
        \put(3,5){\color{black}{(K1)}}
      \end{footnotesize}
      \begin{tiny}
        \put(52,83){\color{blue}{$\frac{\lambda_1}{\lambda_0}$=0.55}}
      \end{tiny}
    \end{overpic}
  }

  \vspace{-0.95cm}
  \subfigure[]{
    \begin{overpic}[width=0.089\linewidth]{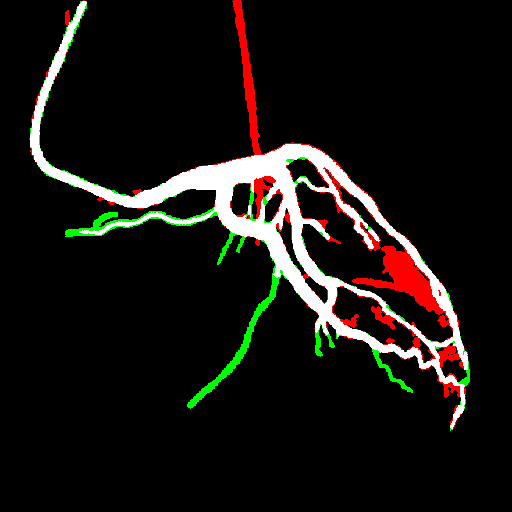}
      \begin{footnotesize}
        \put(3,5){\color{white}{(a1)}}
      \end{footnotesize}
      \begin{tiny}
        \put(52,83){\color{yellow}{$\frac{\lambda_1}{\lambda_0}$=0.05}}
      \end{tiny}
    \end{overpic}
  }
  \subfigure[]{
    \hspace{-0.33cm}
    \begin{overpic}[width=0.089\linewidth]{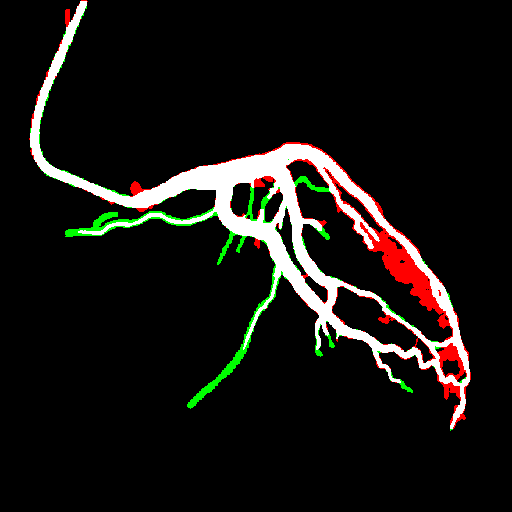}
      \begin{footnotesize}
        \put(3,5){\color{white}{(b1)}}
      \end{footnotesize}
      \begin{tiny}
        \put(52,83){\color{yellow}{$\frac{\lambda_1}{\lambda_0}$=0.1}}
      \end{tiny}
    \end{overpic}
  }
  \subfigure[]{
    \hspace{-0.33cm}
    \begin{overpic}[width=0.089\linewidth]{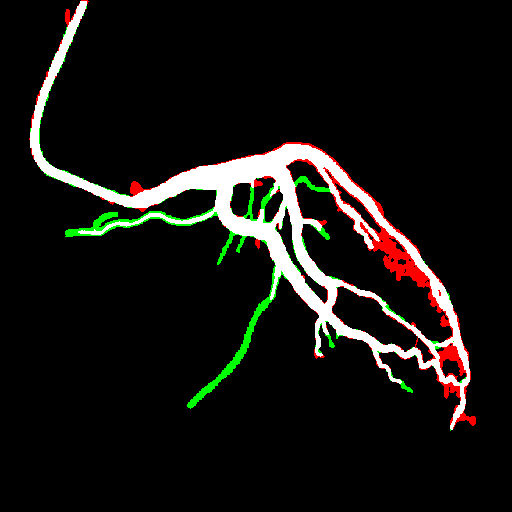}
      \begin{footnotesize}
        \put(3,5){\color{white}{(c1)}}
      \end{footnotesize}
      \begin{tiny}
        \put(52,83){\color{yellow}{$\frac{\lambda_1}{\lambda_0}$=0.15}}
      \end{tiny}
    \end{overpic}
  }
  \subfigure[]{
    \hspace{-0.33cm}
    \begin{overpic}[width=0.089\linewidth]{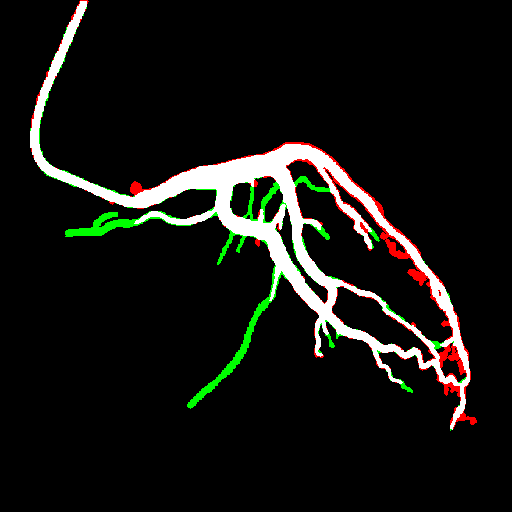}
      \begin{footnotesize}
        \put(3,5){\color{white}{(d1)}}
      \end{footnotesize}
      \begin{tiny}
        \put(52,83){\color{yellow}{$\frac{\lambda_1}{\lambda_0}$=0.2}}
      \end{tiny}
    \end{overpic}
  }
  \subfigure[]{
    \hspace{-0.33cm}
    \begin{overpic}[width=0.089\linewidth]{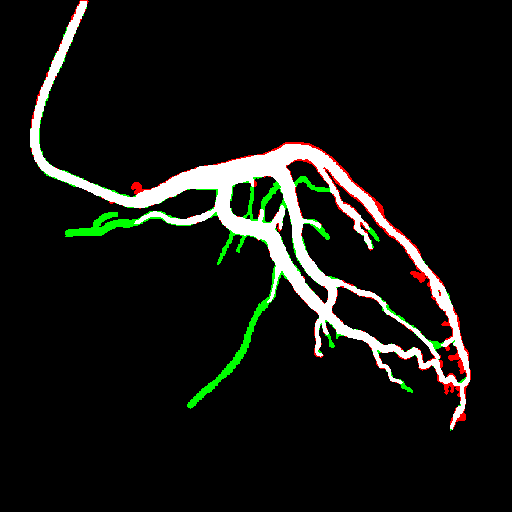}
      \begin{footnotesize}
        \put(3,5){\color{white}{(e1)}}
      \end{footnotesize}
      \begin{tiny}
        \put(52,83){\color{yellow}{$\frac{\lambda_1}{\lambda_0}$=0.25}}
      \end{tiny}
    \end{overpic}
  }
  \subfigure[]{
    \hspace{-0.33cm}
    \begin{overpic}[width=0.089\linewidth]{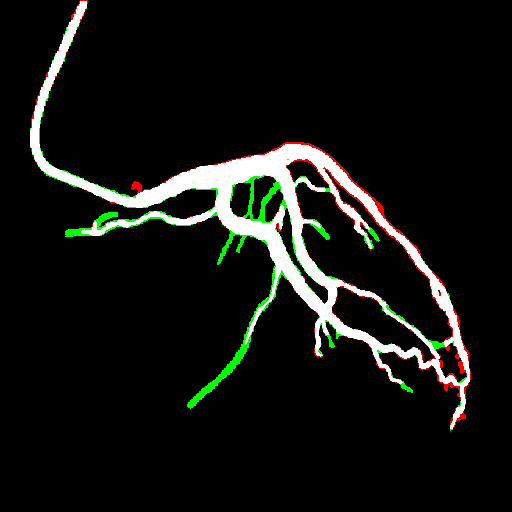}
      \begin{footnotesize}
        \put(3,5){\color{white}{(f1)}}
      \end{footnotesize}
      \begin{tiny}
        \put(52,83){\color{yellow}{$\frac{\lambda_1}{\lambda_0}$=0.3}}
      \end{tiny}
    \end{overpic}
  }
  \subfigure[]{
    \hspace{-0.33cm}
    \begin{overpic}[width=0.089\linewidth]{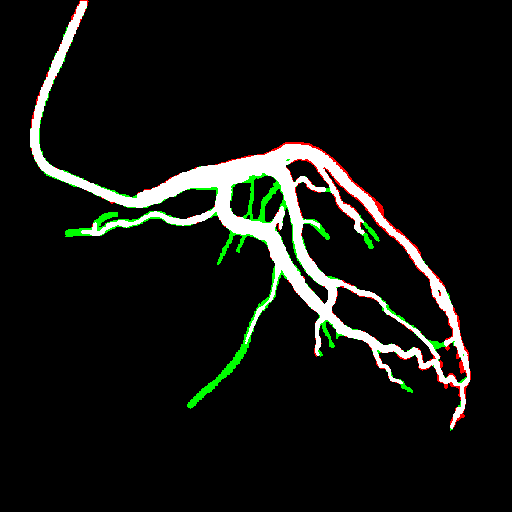}
      \begin{footnotesize}
        \put(3,5){\color{white}{(g1)}}
      \end{footnotesize}
      \begin{tiny}
        \put(52,83){\color{yellow}{$\frac{\lambda_1}{\lambda_0}$=0.35}}
      \end{tiny}
    \end{overpic}
  }
  \subfigure[]{
    \hspace{-0.33cm}
    \begin{overpic}[width=0.089\linewidth]{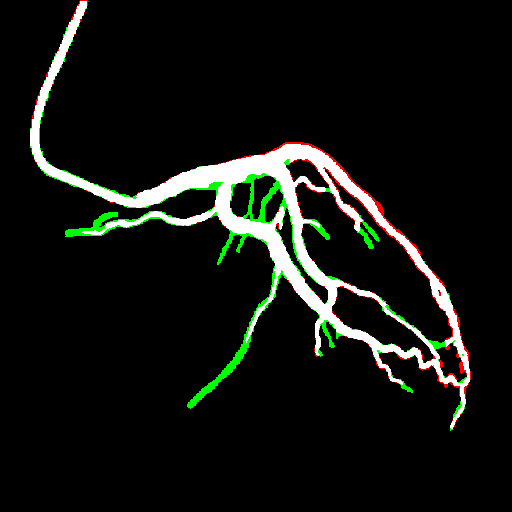}
      \begin{footnotesize}
        \put(3,5){\color{white}{(h1)}}
      \end{footnotesize}
      \begin{tiny}
        \put(52,83){\color{yellow}{$\frac{\lambda_1}{\lambda_0}$=0.4}}
      \end{tiny}
    \end{overpic}
  }
  \subfigure[]{
    \hspace{-0.33cm}
    \begin{overpic}[width=0.089\linewidth]{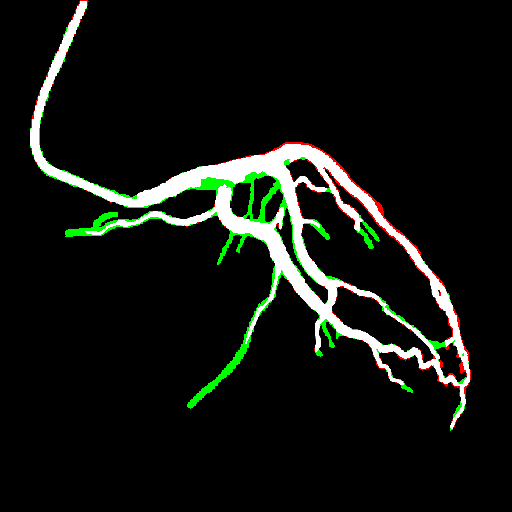}
      \begin{footnotesize}
        \put(3,5){\color{white}{(i1)}}
      \end{footnotesize}
      \begin{tiny}
        \put(52,83){\color{yellow}{$\frac{\lambda_1}{\lambda_0}$=0.45}}
      \end{tiny}
    \end{overpic}
  }
  \subfigure[]{
    \hspace{-0.33cm}
    \begin{overpic}[width=0.089\linewidth]{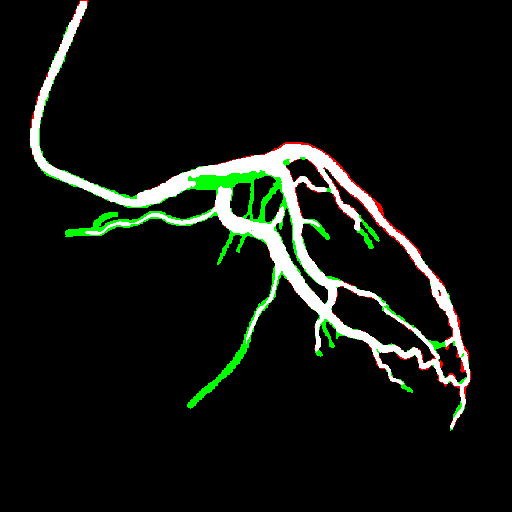}
      \begin{footnotesize}
        \put(3,5){\color{white}{(j1)}}
      \end{footnotesize}
      \begin{tiny}
        \put(52,83){\color{yellow}{$\frac{\lambda_1}{\lambda_0}$=0.5}}
      \end{tiny}
    \end{overpic}
  }
  \subfigure[]{
    \hspace{-0.33cm}
    \begin{overpic}[width=0.089\linewidth]{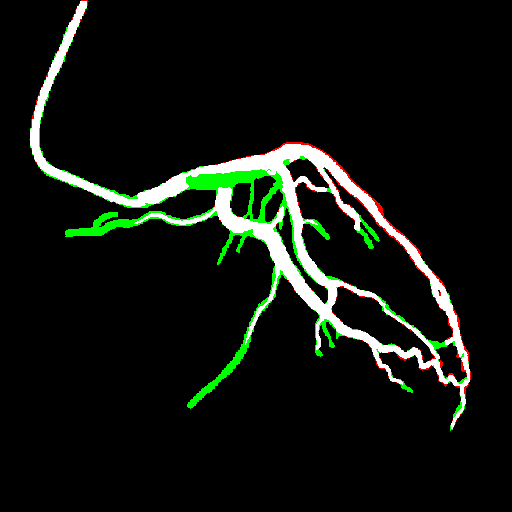}
      \begin{footnotesize}
        \put(3,5){\color{white}{(k1)}}
      \end{footnotesize}
      \begin{tiny}
        \put(52,83){\color{yellow}{$\frac{\lambda_1}{\lambda_0}$=0.55}}
      \end{tiny}
    \end{overpic}
  }

  \vspace{-0.5cm}
  \subfigure[]{
    \begin{overpic}[width=0.089\linewidth]{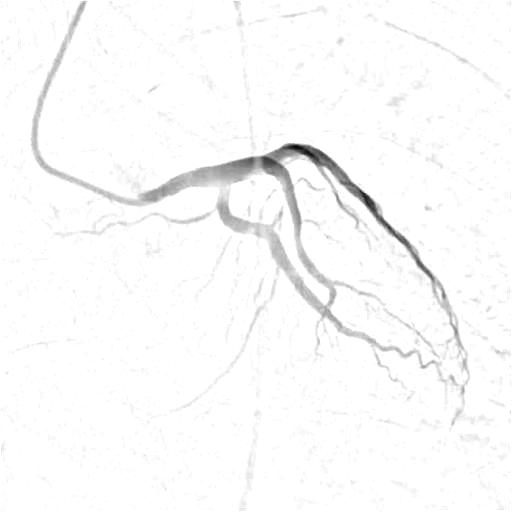}
      \begin{footnotesize}
        \put(3,5){\color{black}{(A2)}}
      \end{footnotesize}
      \begin{tiny}
        \put(58,83){\color{blue}{$\frac{\lambda_3}{\lambda_1}$=0.1}}
      \end{tiny}
    \end{overpic}
  }
  \subfigure[]{
    \hspace{-0.33cm}
    \begin{overpic}[width=0.089\linewidth]{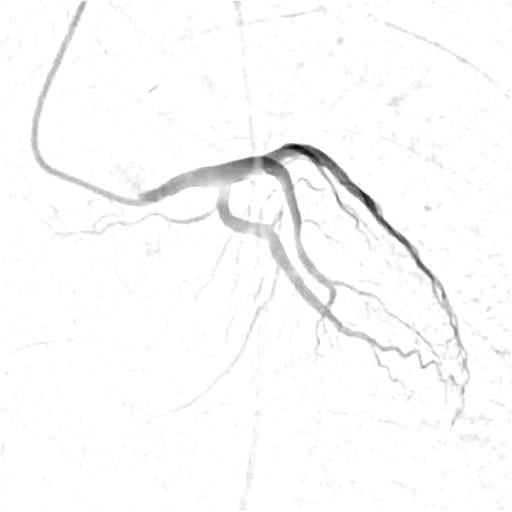}
      \begin{footnotesize}
        \put(3,5){\color{black}{(B2)}}
      \end{footnotesize}
      \begin{tiny}
        \put(58,83){\color{blue}{$\frac{\lambda_3}{\lambda_1}$=0.2}}
      \end{tiny}
    \end{overpic}
  }
  \subfigure[]{
    \hspace{-0.33cm}
    \begin{overpic}[width=0.089\linewidth]{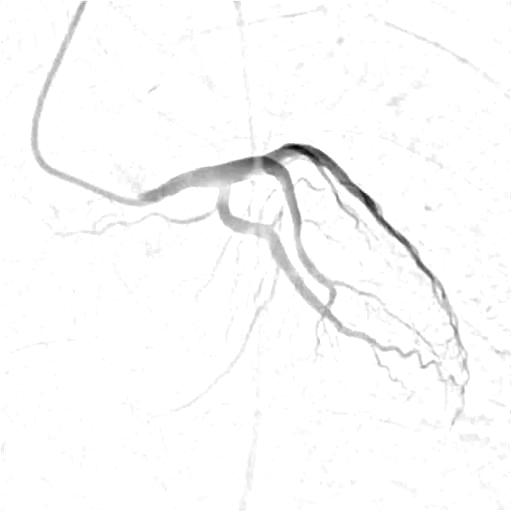}
      \begin{footnotesize}
        \put(3,5){\color{black}{(C2)}}
      \end{footnotesize}
      \begin{tiny}
        \put(58,83){\color{blue}{$\frac{\lambda_3}{\lambda_1}$=0.3}}
      \end{tiny}
    \end{overpic}
  }
  \subfigure[]{
    \hspace{-0.33cm}
    \begin{overpic}[width=0.089\linewidth]{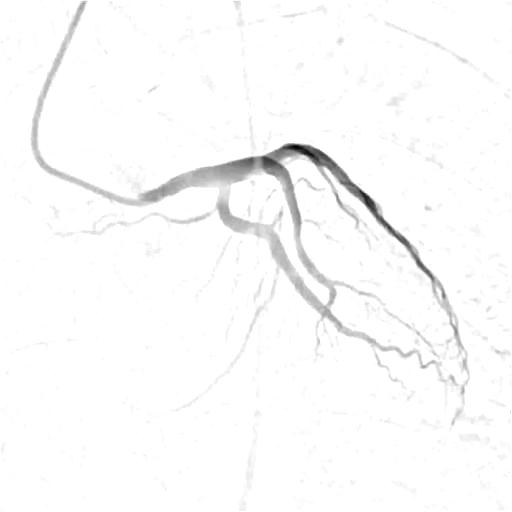}
      \begin{footnotesize}
        \put(3,5){\color{black}{(D2)}}
      \end{footnotesize}
      \begin{tiny}
        \put(58,83){\color{blue}{$\frac{\lambda_3}{\lambda_1}$=0.4}}
      \end{tiny}
    \end{overpic}
  }
  \subfigure[]{
    \hspace{-0.33cm}
    \begin{overpic}[width=0.089\linewidth]{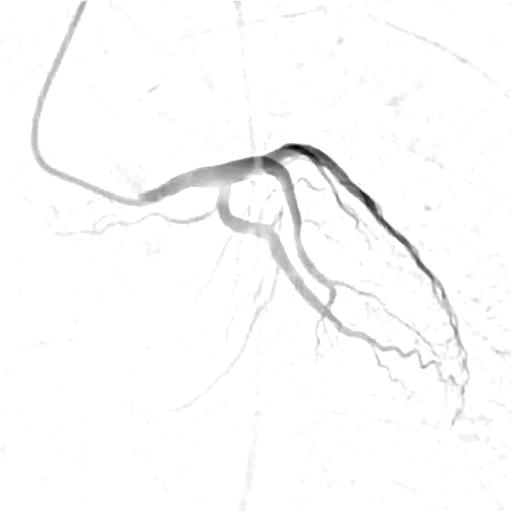}
      \begin{footnotesize}
        \put(3,5){\color{black}{(E2)}}
      \end{footnotesize}
      \begin{tiny}
        \put(58,83){\color{blue}{$\frac{\lambda_3}{\lambda_1}$=0.5}}
      \end{tiny}
    \end{overpic}
  }
  \subfigure[]{
    \hspace{-0.33cm}
    \begin{overpic}[width=0.089\linewidth]{0.3_0.6_0137f.png}
      \begin{footnotesize}
        \put(3,5){\color{black}{(F2)}}
      \end{footnotesize}
      \begin{tiny}
        \put(58,83){\color{blue}{$\frac{\lambda_3}{\lambda_1}$=0.6}}
      \end{tiny}
    \end{overpic}
  }
  \subfigure[]{
    \hspace{-0.33cm}
    \begin{overpic}[width=0.089\linewidth]{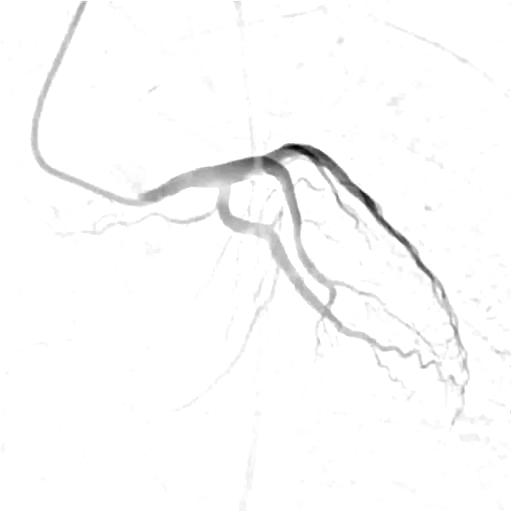}
      \begin{footnotesize}
        \put(3,5){\color{black}{(G2)}}
      \end{footnotesize}
      \begin{tiny}
        \put(58,83){\color{blue}{$\frac{\lambda_3}{\lambda_1}$=0.7}}
      \end{tiny}
    \end{overpic}
  }
  \subfigure[]{
    \hspace{-0.33cm}
    \begin{overpic}[width=0.089\linewidth]{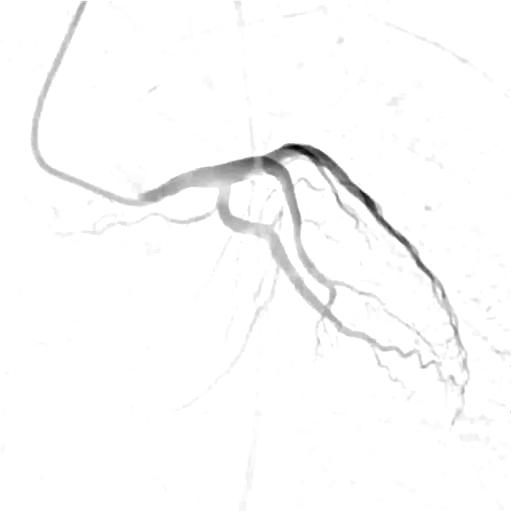}
      \begin{footnotesize}
        \put(3,5){\color{black}{(H2)}}
      \end{footnotesize}
      \begin{tiny}
        \put(58,83){\color{blue}{$\frac{\lambda_3}{\lambda_1}$=0.8}}
      \end{tiny}
    \end{overpic}
  }
  \subfigure[]{
    \hspace{-0.33cm}
    \begin{overpic}[width=0.089\linewidth]{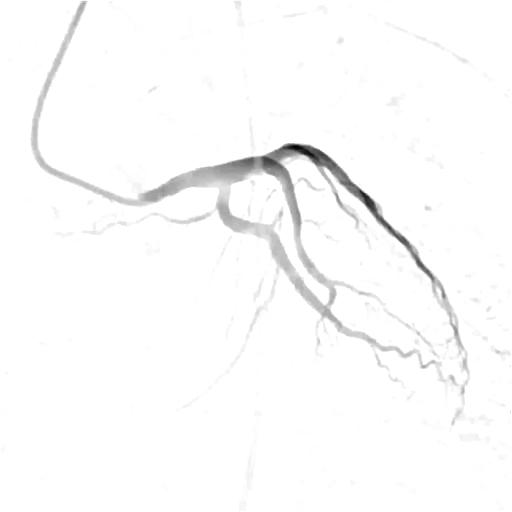}
      \begin{footnotesize}
        \put(3,5){\color{black}{(I2)}}
      \end{footnotesize}
      \begin{tiny}
        \put(58,83){\color{blue}{$\frac{\lambda_3}{\lambda_1}$=0.9}}
      \end{tiny}
    \end{overpic}
  }
  \subfigure[]{
    \hspace{-0.33cm}
    \begin{overpic}[width=0.089\linewidth]{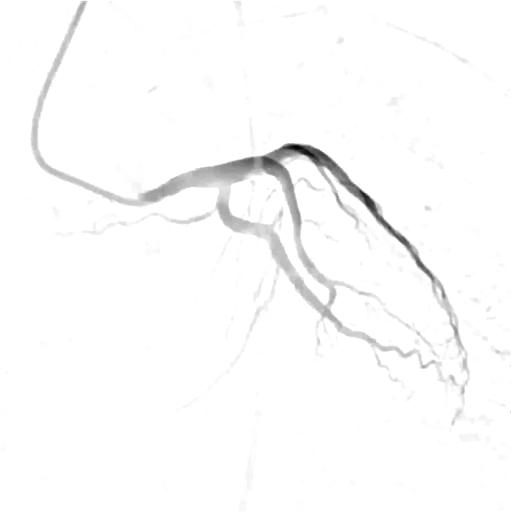}
      \begin{footnotesize}
        \put(3,5){\color{black}{(J2)}}
      \end{footnotesize}
      \begin{tiny}
        \put(58,83){\color{blue}{$\frac{\lambda_3}{\lambda_1}$=1}}
      \end{tiny}
    \end{overpic}
  }
  \subfigure[]{
    \hspace{-0.33cm}
    \begin{overpic}[width=0.089\linewidth]{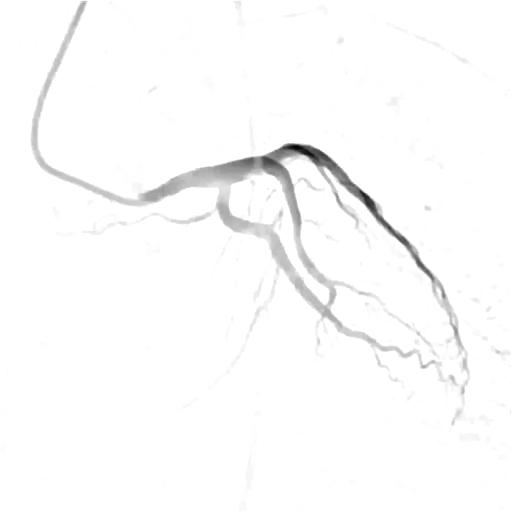}
      \begin{footnotesize}
        \put(3,5){\color{black}{(K2)}}
      \end{footnotesize}
      \begin{tiny}
        \put(58,83){\color{blue}{$\frac{\lambda_3}{\lambda_1}$=1.1}}
      \end{tiny}
    \end{overpic}
  }

  \vspace{-0.95cm}
  \subfigure[]{
    \begin{overpic}[width=0.089\linewidth]{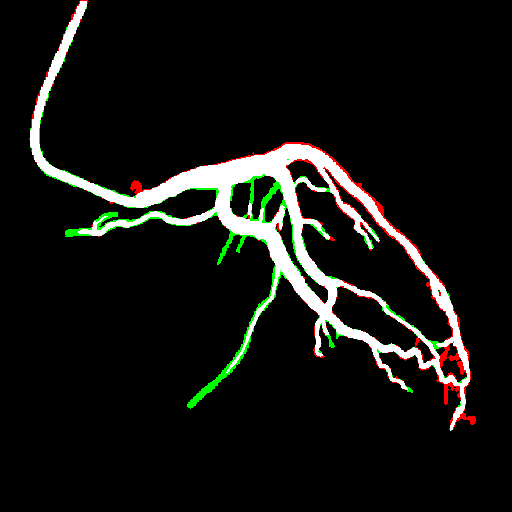}
      \begin{footnotesize}
        \put(3,5){\color{white}{(a2)}}
      \end{footnotesize}
      \begin{tiny}
        \put(58,83){\color{yellow}{$\frac{\lambda_3}{\lambda_1}$=0.1}}
      \end{tiny}
    \end{overpic}
  }
  \subfigure[]{
    \hspace{-0.33cm}
    \begin{overpic}[width=0.089\linewidth]{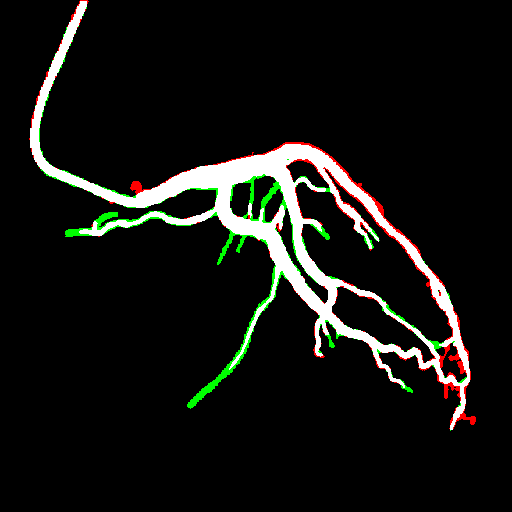}
      \begin{footnotesize}
        \put(3,5){\color{white}{(b2)}}
      \end{footnotesize}
      \begin{tiny}
        \put(58,83){\color{yellow}{$\frac{\lambda_3}{\lambda_1}$=0.2}}
      \end{tiny}
    \end{overpic}
  }
  \subfigure[]{
    \hspace{-0.33cm}
    \begin{overpic}[width=0.089\linewidth]{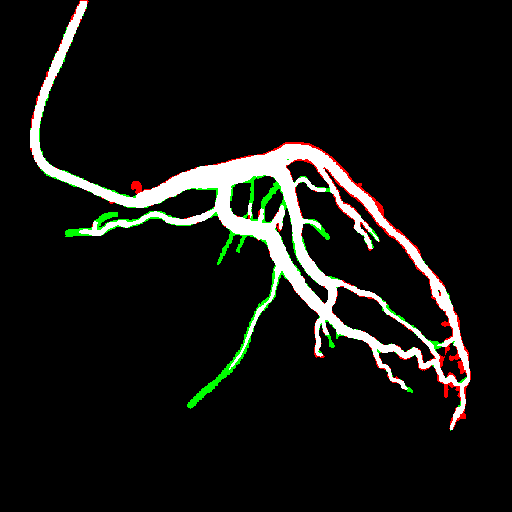}
      \begin{footnotesize}
        \put(3,5){\color{white}{(c2)}}
      \end{footnotesize}
      \begin{tiny}
        \put(58,83){\color{yellow}{$\frac{\lambda_3}{\lambda_1}$=0.3}}
      \end{tiny}
    \end{overpic}
  }
  \subfigure[]{
    \hspace{-0.33cm}
    \begin{overpic}[width=0.089\linewidth]{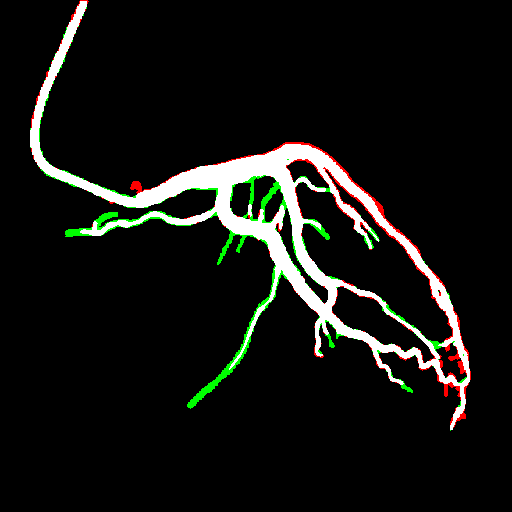}
      \begin{footnotesize}
        \put(3,5){\color{white}{(d2)}}
      \end{footnotesize}
      \begin{tiny}
        \put(58,83){\color{yellow}{$\frac{\lambda_3}{\lambda_1}$=0.4}}
      \end{tiny}
    \end{overpic}
  }
  \subfigure[]{
    \hspace{-0.33cm}
    \begin{overpic}[width=0.089\linewidth]{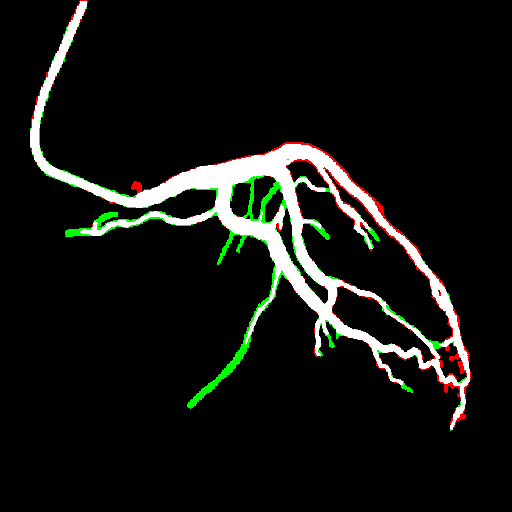}
      \begin{footnotesize}
        \put(3,5){\color{white}{(e2)}}
      \end{footnotesize}
      \begin{tiny}
        \put(58,83){\color{yellow}{$\frac{\lambda_3}{\lambda_1}$=0.5}}
      \end{tiny}
    \end{overpic}
  }
  \subfigure[]{
    \hspace{-0.33cm}
    \begin{overpic}[width=0.089\linewidth]{0.3_0.6_0137_gt2.png}
      \begin{footnotesize}
        \put(3,5){\color{white}{(f2)}}
      \end{footnotesize}
      \begin{tiny}
        \put(58,83){\color{yellow}{$\frac{\lambda_3}{\lambda_1}$=0.6}}
      \end{tiny}
    \end{overpic}
  }
  \subfigure[]{
    \hspace{-0.33cm}
    \begin{overpic}[width=0.089\linewidth]{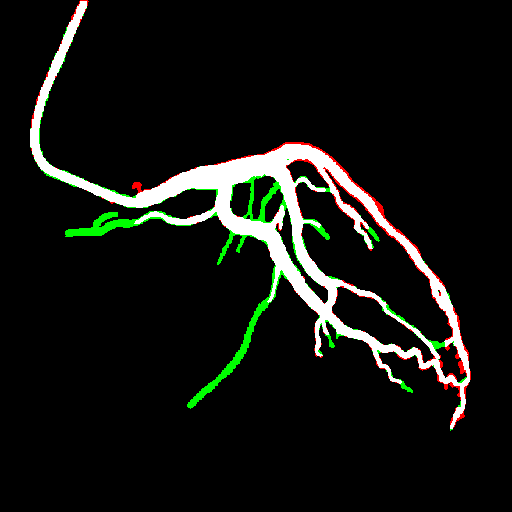}
      \begin{footnotesize}
        \put(3,5){\color{white}{(g2)}}
      \end{footnotesize}
      \begin{tiny}
        \put(58,83){\color{yellow}{$\frac{\lambda_3}{\lambda_1}$=0.7}}
      \end{tiny}
    \end{overpic}
  }
  \subfigure[]{
    \hspace{-0.33cm}
    \begin{overpic}[width=0.089\linewidth]{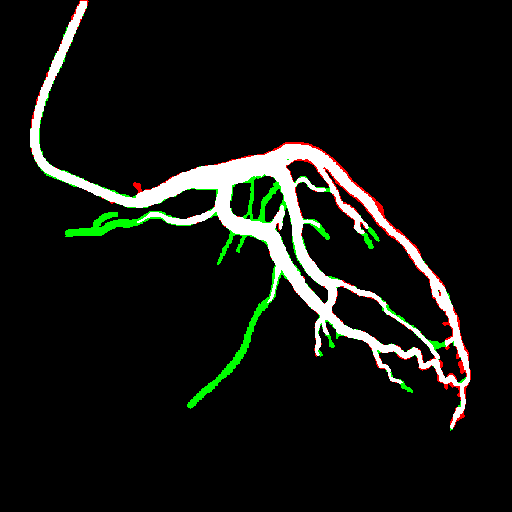}
      \begin{footnotesize}
        \put(3,5){\color{white}{(h2)}}
      \end{footnotesize}
      \begin{tiny}
        \put(58,83){\color{yellow}{$\frac{\lambda_3}{\lambda_1}$=0.8}}
      \end{tiny}
    \end{overpic}
  }
  \subfigure[]{
    \hspace{-0.33cm}
    \begin{overpic}[width=0.089\linewidth]{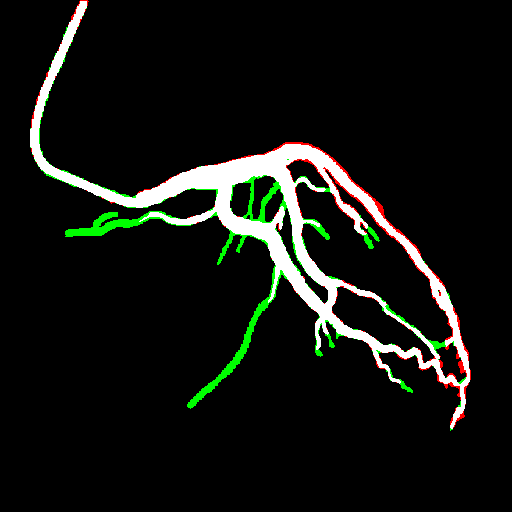}
      \begin{footnotesize}
        \put(3,5){\color{white}{(i2)}}
      \end{footnotesize}
      \begin{tiny}
        \put(58,83){\color{yellow}{$\frac{\lambda_3}{\lambda_1}$=0.9}}
      \end{tiny}
    \end{overpic}
  }
  \subfigure[]{
    \hspace{-0.33cm}
    \begin{overpic}[width=0.089\linewidth]{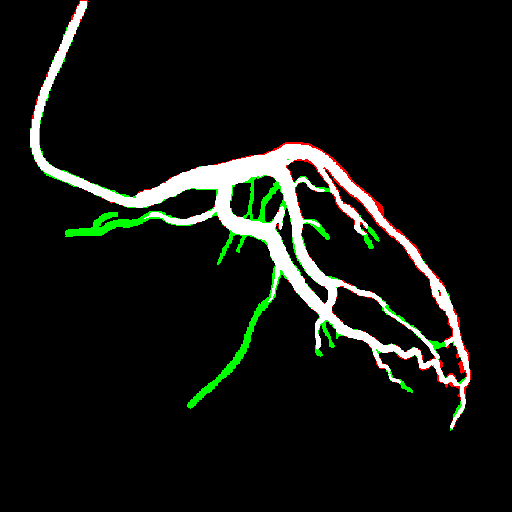}
      \begin{footnotesize}
        \put(3,5){\color{white}{(j2)}}
      \end{footnotesize}
      \begin{tiny}
        \put(58,83){\color{yellow}{$\frac{\lambda_3}{\lambda_1}$=1}}
      \end{tiny}
    \end{overpic}
  }
  \subfigure[]{
    \hspace{-0.33cm}
    \begin{overpic}[width=0.089\linewidth]{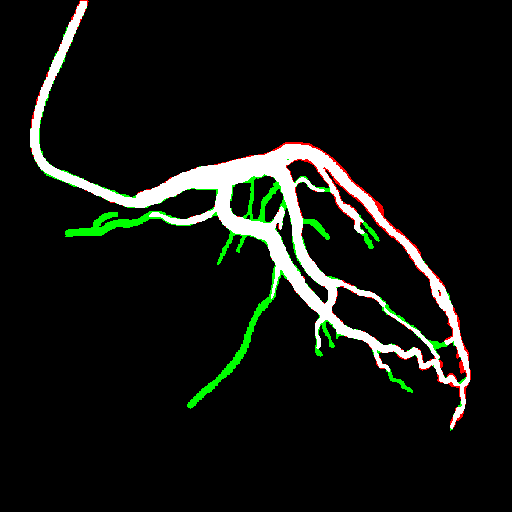}
      \begin{footnotesize}
        \put(3,5){\color{white}{(k2)}}
      \end{footnotesize}
      \begin{tiny}
        \put(58,83){\color{yellow}{$\frac{\lambda_3}{\lambda_1}$=1.1}}
      \end{tiny}
    \end{overpic}
  }

  \label{fig4}
  \vspace{-0.8cm}
  \caption{Parameter optimization experiment for TV-TRPCA. (A1-K1) and (a1-k1) are the foreground extraction and vessel segmentation results as $\lambda_1$ varies, in which $\lambda_0=1/\sqrt{MT}$. (A2-K2) and (a2)-(k2) are the foreground extraction and vessel segmentation results as $\lambda_3$ varies. All other parameters were kept constant during each test.}
\end{figure*}

\begin{figure*}[!t]
  \centering
  \subfigure[]{
    \begin{overpic}[width=0.49\linewidth]{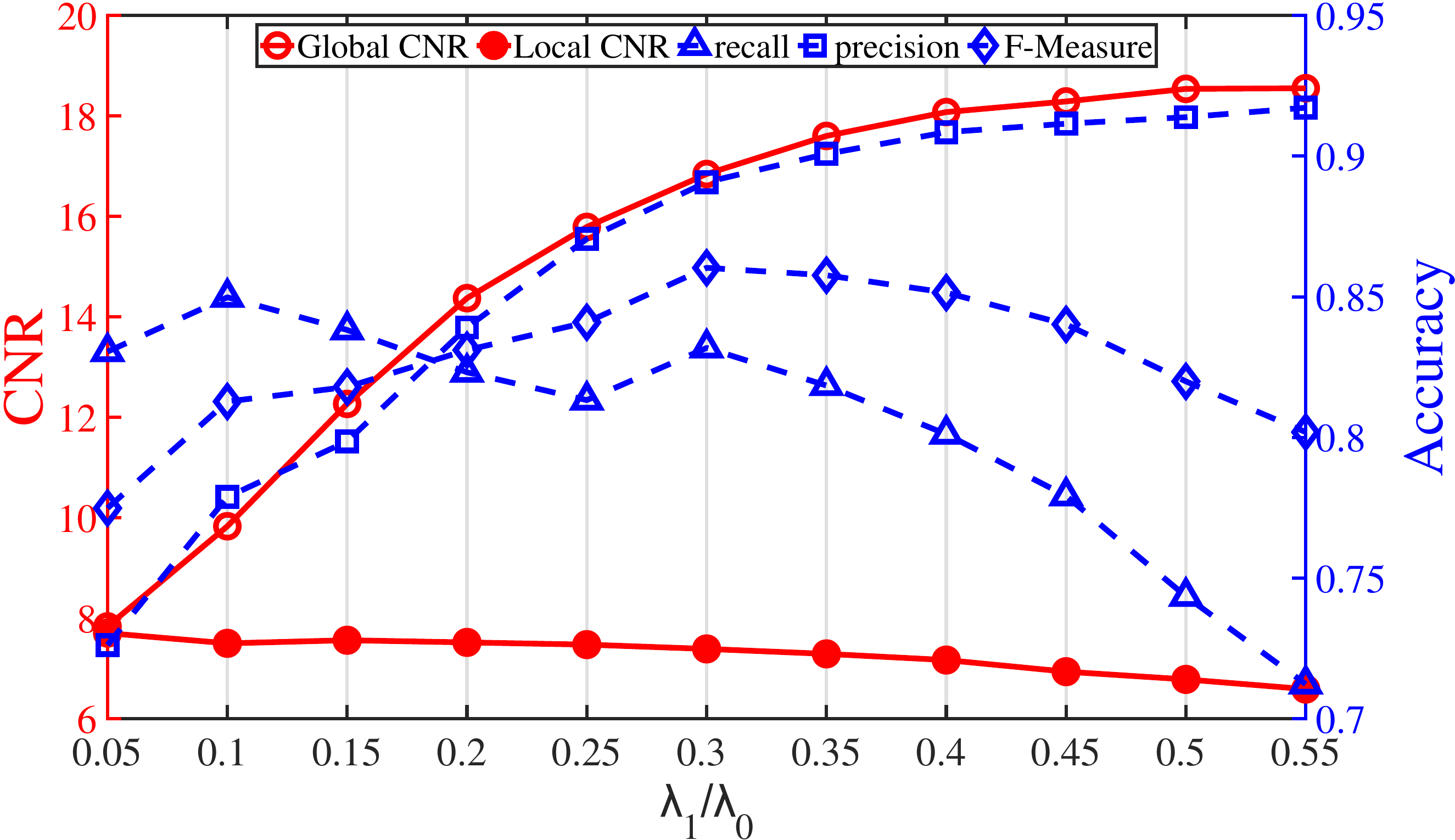}
      \put(9,52){\color{black}{(A)}}
    \end{overpic}
  }
  \subfigure[]{
    \hspace{-0.2cm}
    \begin{overpic}[width=0.49\linewidth]{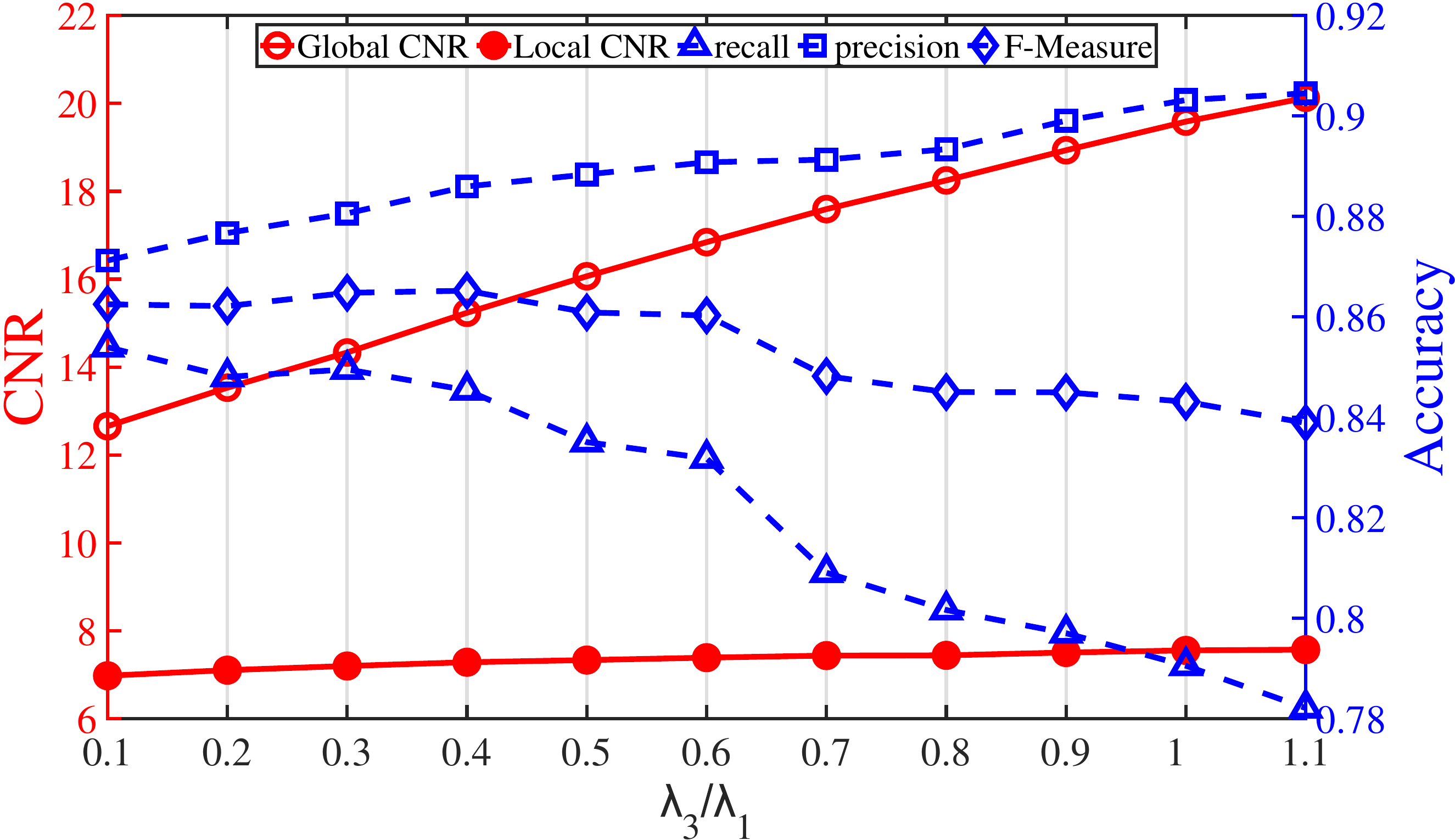}
      \put(9,52){\color{black}{(B)}}
    \end{overpic}
  }

  \label{fig5}
  \vspace{-0.8cm}
  \caption{Variations of five metrics in TV-TRPCA parameter optimization experiment. (A), (B) are the mean results as $\lambda_1$ and $\lambda_3$ vary, respectively.}
\end{figure*}

For quantitative analysis, experienced surgeons were invited to precisely segment the raw XCA images, and the ground truth (GT) vessel masks were obtained, as shown in \hyperref[fig7]{Fig.7.(a1)-(a4)}, which are used as the gold standard of the foreground layer. Due to the blurred vessel edge, the artificially selected vessel foreground area will be widened. In the experiment of vessel segmentation, the GT vessel mask is processed with smooth edge contraction to make the binary vessel mask more accurate, as shown in \hyperref[fig8]{Fig.8.(a1)-(a4)} .

\subsection{Experiment Design}

The main Experiment is divided into two parts: foreground extraction and vessel segmentation. In the \textit{TV-TRPCA} algorithm, since the angiography sequence is isotropic, we set $\sigma_m=\sigma_n=\sigma_t=1$. Other main parameters are recorded in \hyperref[tab2]{Tab.2}. The specific parameter optimization experiment will be introduced in the subsequent sections. In the \textit{TSRG} algorithm, the codes of RLF filtering are from the third-party libraries of MATLAB\footnote{\url{https://ww2.mathworks.cn/matlabcentral/fileexchange/27886-radon-like-features}}, the parameters are adjusted to $sup=25,or=1,sc=10$. And 4-neighborhood sampling is used in region growing. For comparison, we investigated some open-source state-of-the-art approaches and tested their performance.

In foreground extraction part, three RPCA methods \textit{AccAltProj}\footnote{\url{https://github.com/caesarcai/AccAltProj_for_RPCA}}\cite{cai2019accelerated}, \textit{MoG-RPCA}\footnote{\url{https://github.com/timmyzhao/MoG-RPCA}}\cite{zhao2014robust}, \textit{MCR-RPCA}\footnote{\url{https://github.com/Binjie-Qin/MCR-RPCA}}\cite{jin2017extracting}, three TRPCA methods \textit{ETRPCA}\footnote{\url{https://github.com/xdweixia/TPAMI2020_ETRPCA}}\cite{gao2020enhanced}, \textit{KBR-RPCA}\footnote{\url{https://github.com/XieQi2015/KBR-TC-and-RPCA}}\cite{xie2017kronecker}, \textit{TNN-TRPCA}\footnote{\url{https://github.com/canyilu/Tensor-Robust-Principal-Component-Analysis-TRPCA}}\cite{lu2019tensor} and three non-PCA-based matrix decomposition methods \textit{DECOLOR}\footnote{\url{https://github.com/GreenTeaHua/DECOLOR-}}\cite{zhou2012moving}, \textit{GoDec}\footnote{\url{https://sites.google.com/site/godecomposition/code}}\cite{zhou2011godec}, \textit{PRMF}\footnote{\url{http://winsty.net/prmf.html}}\cite{wang2012probabilistic} were introduced as the comparison groups. To make it more adaptive to XCA vessel layer extraction, we optimized the parameters of these nine algorithms in detail, as recorded in \hyperref[tab2]{Tab.2}. Furthermore, we tested the improvement of three commonly used vessel segmentation methods \textit{Frangi filtering}\footnote{\url{https://ww2.mathworks.cn/matlabcentral/fileexchange/24409-hessian-based-frangi-vesselness-filter}}\cite{frangi1998multiscale}, \textit{DSA}\cite{meaney1980digital} and \textit{U-net}\footnote{\url{https://lmb.informatik.uni-freiburg.de/people/ronneber/u-net/}}\cite{ronneberger2015u} by replacing the raw XCA image input with vessel layer image extracted by \textit{TV-TRPCA}. 

In vessel segmentation part, two advanced end to end deep learning method \textit{VGN}\footnote{\url{https://github.com/syshin1014/VGN}}\cite{shin2019deep} and \textit{SVS-net}\footnote{\url{https://github.com/Binjie-Qin/SVS-net}}\cite{hao2020sequential} were introduced as the comparison groups. A shared calibrated dataset was used as the neural network training set\footnote{\url{https://github.com/Binjie-Qin/SVS-net/tree/master/data}}. The structure and properties of these two networks are also described specifically in the subsequent sections.

All the experiments were programmed in Matlab R2020a and Python 3.6.0, ran on a low-cost personal laptop with an Intel (R) Core (TM) i7-11700H 3.2GHz CPU and 16GB 3200MHz Memory.

\begin{table*}[!t]
  \renewcommand{\arraystretch}{1}
  \centering
  \begin{threeparttable}
    \topcaption{Main parameters of different layer separation methods\tnote{1}\tnote{2}}
    \label{tab2}
    \begin{footnotesize}
      \begin{tabular}{c|c|c|c|c|c|c|c|c|c}
        \toprule \toprule
        DECOLOR      & GoDec         & PRMF          & AAP                          & MCR-R                              & MoG-R & ETRPCA                           & KBR-R                          & TNN-T                              & TV-T                               \\
        \midrule
        $\lambda=1$  & $r=2$         & $r_k=2$       & $r=\mu=2$                    & $\lambda_1=\dfrac{0.5}{\sqrt{MN}}$ & $k=3$ & $p=0.9$                          & $\beta=\dfrac{2}{M}$           & $\lambda_1=\dfrac{0.3}{\sqrt{MT}}$ & $\lambda_1=\dfrac{0.3}{\sqrt{MT}}$ \\
        $K=\sqrt{T}$ & $\sigma=10^4$ & $\lambda_U=1$ & $\beta=\dfrac{1}{\sqrt{MN}}$ & $\lambda_2=0.4\lambda_1$           & $r=2$ & $\lambda=\dfrac{0.2}{\sqrt{MT}}$ & $\gamma=100\beta$              & $\mu=10^{-4}$                      & $\lambda_2=100\lambda_1$           \\
                     & $q=0$         & $\lambda_V=1$ & $\beta_{init}=4\beta$        &                                    &       &                                  & $\lambda=\dfrac{1}{\sqrt{MT}}$ & $\nu=10^{-4}$                      & $\lambda_3=0.6\lambda_1$           \\
                     &               &               & $\gamma=0.65$                &                                    &       &                                  & $\mu=10$                       &                                    & $\mu=\nu=10^{-4}$                  \\
        \bottomrule \bottomrule
      \end{tabular}
      \begin{tablenotes}
        \item[1] For all layer separation methods, the termination tolerance is $\epsilon=10^{-3}$, the increase ratio is $\rho=1.1$, and the maximal iteration is $imax=100$.
        \item[2] In order to facilitate the presentation, some methods adopt the abbreviated form.
      \end{tablenotes}
    \end{footnotesize}
  \end{threeparttable}
\end{table*}

\subsection{Evaluation Metrics}

\subsubsection{Foreground Extraction Visibility Evaluation}

\begin{figure}[ht]
  \centering
  \subfigure[]{
    \begin{overpic}[width=0.19\linewidth]{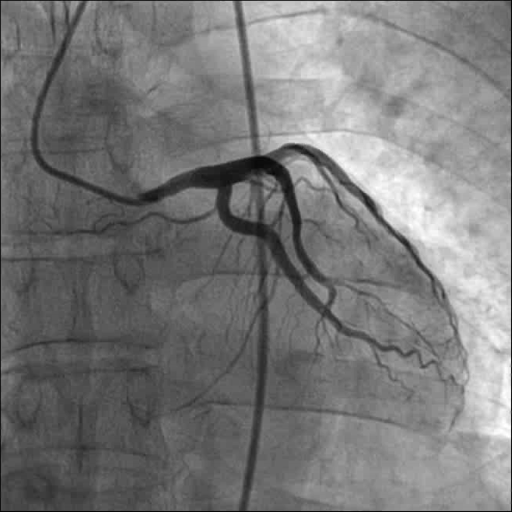}
      \put(3,5){\color{white}{(a)}}
    \end{overpic}
  }
  \subfigure[]{
    \hspace{-0.3cm}
    \begin{overpic}[width=0.19\linewidth]{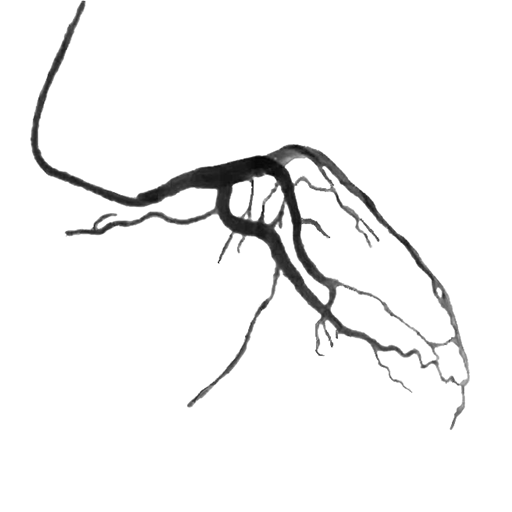}
      \put(3,5){\color{black}{(b)}}
    \end{overpic}
  }
  \subfigure[]{
    \hspace{-0.3cm}
    \begin{overpic}[width=0.19\linewidth]{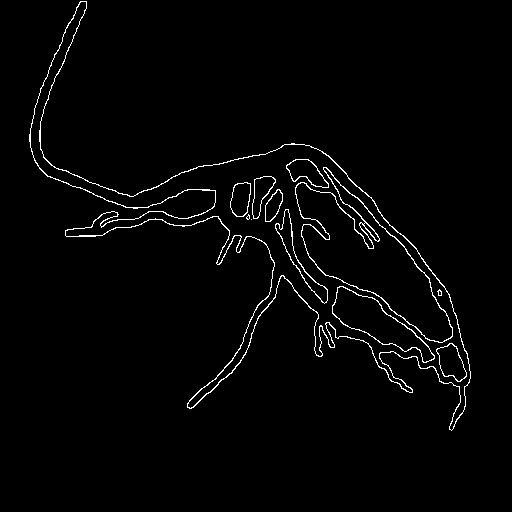}
      \put(3,5){\color{white}{(c)}}
    \end{overpic}
  }
  \subfigure[]{
    \hspace{-0.3cm}
    \begin{overpic}[width=0.19\linewidth]{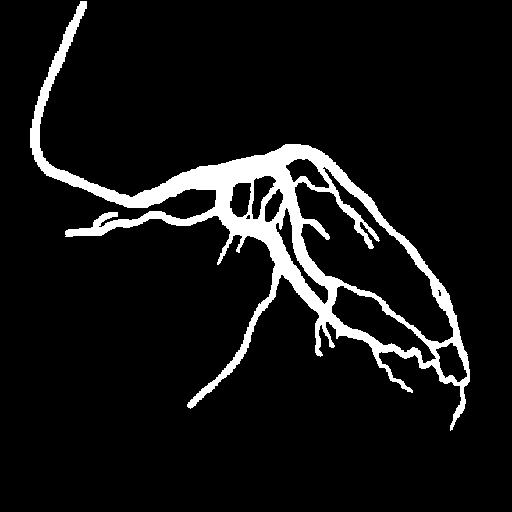}
      \put(3,5){\color{white}{(d)}}
    \end{overpic}
  }
  \subfigure[]{
    \hspace{-0.3cm}
    \begin{overpic}[width=0.19\linewidth]{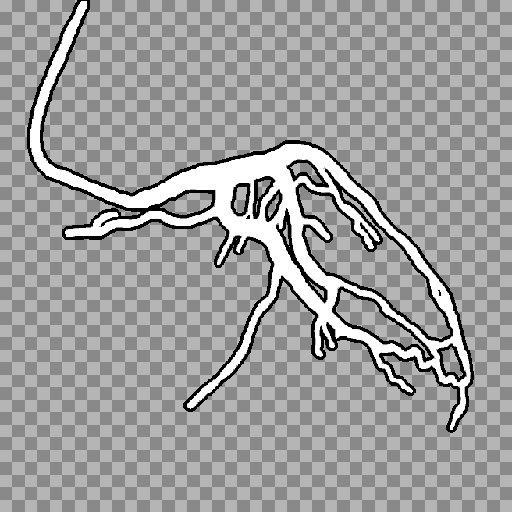}
      \put(3,5){\color{white}{(e)}}
    \end{overpic}
  }

  \label{fig6}
  \vspace{-0.8cm}
  \caption{Global CNR and Local CNR. (a) is the raw XCA image. (b) is the Ground truth vessel mask. (c) is the edge of vessel region. (d) is the demonstration of Global CNR. (e) is the demonstration of Local CNR. The evenly distributed squares represent transparent regions that are not included in the calculation.}
\end{figure}

In order to quantitatively evaluate and compare the visibility of the extracted foreground, the contrast-to-noise ratio (CNR)\cite{ma2017automatic} is introduced and defined as
\begin{equation}
  CNR=\frac{|\mu_v-\mu_b|}{\sigma_b}
  \label{eq35}
\end{equation}
where $\mu_v$ and $\mu_b$ are the mean gray values of vessel region (white area in \hyperref[fig6]{Fig.6.(d)(e)}) and background region (black area in \hyperref[fig6]{Fig.6.(d)(e)}) in foreground layer, $\sigma_b$ is the standard deviation of background region gray values. CNR reflects the gray value difference between vessel region and background region in foreground layer, the higher the CNR value, the higher the contrast of foreground layer and the better the visibility of vessel region.

Consider that there are a lot of artifacts along the edges of the vessels in foreground layer, to evaluate the global and local visibility respectively, we use two criteria to divide the background area. The first covers all the non-vessel region of foreground layer, as shown in \hyperref[fig6]{Fig.6.(d)}, which is called the Global CNR. The second is the 4-pixel wide neighborhood area surrounding the vessel region, as shown in \hyperref[fig6]{Fig.6.(e)}, which is called the Local CNR.

\subsubsection{Vessel Segmentation Accuracy Evaluation}

In order to quantitatively evaluate and compare the accuracy of the vessel segmentation results, the accuracy is expressed by means of recall, precision and their harmonic mean, the F-Measure, defined as\cite{brutzer2011evaluation}
\begin{equation}
  \begin{aligned}
    \text{recall}    & =\dfrac{\text{TP}}{\text{TP}+\text{FN}}                            \\
    \text{precision} & =\dfrac{\text{TP}}{\text{TP}+\text{FP}}                            \\
    \text{F-Measure} & =\dfrac{2*\text{recall}*\text{precision}}{\text{recall}+\text{precision}}
  \end{aligned}
  \label{eq36}
\end{equation}
where TP (true positive) is the part of correctly classified foreground pixels, FP (false positive) is the part of background pixels that are wrongly classified as foreground, and FN (false negative) is the part of foreground pixels that are wrongly classified as backgrounds. Thus, recall reflects the ability to identify foreground pixels, precision reflects the correctness of classification, and F-Measure reflects the comprehensive performance of the segmentation method.

\subsection{Foreground Extraction Experiment} \label{TVTRPCA}

\subsubsection{Parameter Optimization}

According to \hyperref[eq14]{Eq.14}, the main parameters that affect the performance of TV-TRPCA algorithm are $\lambda_1,\lambda_2,\lambda_3$. $\lambda_1$ is the equilibrium factor between the low-rank property of background layer and the sparsity property of foreground layer. $\lambda_2$ is the equilibrium factor of Gaussian noise. And $\lambda_3$ is the equilibrium factor of spatial-temporal continuity. Since the tested XCA image sequences have low noise, we set $\lambda_2=100\lambda_1$ to reduce the proportion of noise term in layer separation. In this section, the influence of $\lambda_1$ and $\lambda_3$ will be experimented on clinical XCA image sequences.

\begin{figure*}[!t]
  \centering
  \hspace{0.4cm}
  \begin{scriptsize}
    \textbf{GT} \ \ \ \ \ \ \ \ \ \ \textbf{DECOLOR} \ \ \ \ \ \ \ \ \textbf{GoDec} \ \ \ \ \ \ \ \ \ \ \ \textbf{PRMF} \ \ \ \ \ \ \ \textbf{AccAltProj} \ \ \ \textbf{MoG-RPCA} \ \textbf{MCR-RPCA} \ \ \ \ \textbf{ETRPCA} \ \ \ \ \textbf{KBR-RPCA} \ \textbf{TNN-TRPCA} \ \textbf{TV-TRPCA}
  \end{scriptsize}

  \subfigure[]{
    \begin{overpic}[width=0.089\linewidth]{Origin_0137.png}
      \begin{footnotesize}
        \put(3,5){\color{white}{(A1)}}
      \end{footnotesize}
    \end{overpic}
  }
  \subfigure[]{
    \hspace{-0.33cm}
    \begin{overpic}[width=0.089\linewidth]{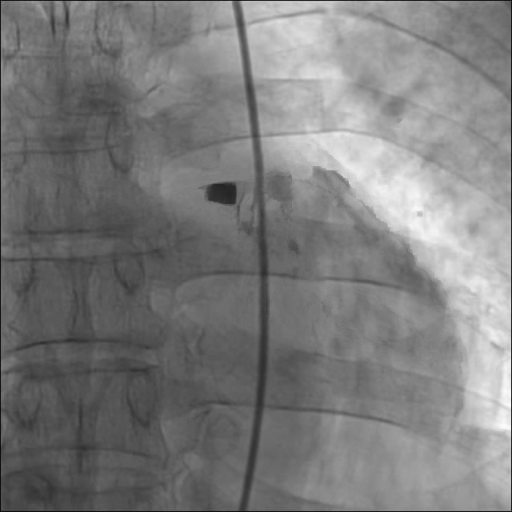}
      \begin{footnotesize}
        \put(3,5){\color{white}{(B1)}}
      \end{footnotesize}
    \end{overpic}
  }
  \subfigure[]{
    \hspace{-0.33cm}
    \begin{overpic}[width=0.089\linewidth]{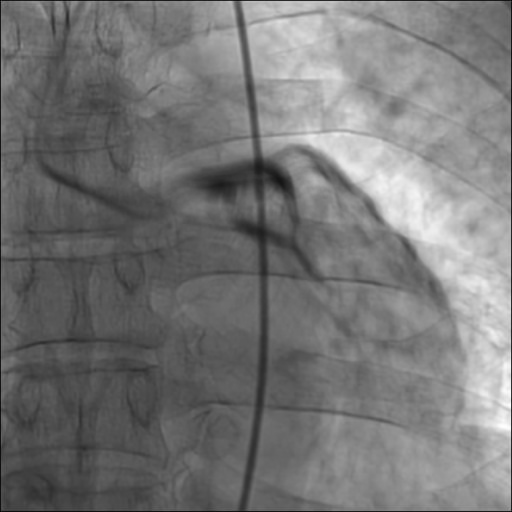}
      \begin{footnotesize}
        \put(3,5){\color{white}{(C1)}}
      \end{footnotesize}
    \end{overpic}
  }
  \subfigure[]{
    \hspace{-0.33cm}
    \begin{overpic}[width=0.089\linewidth]{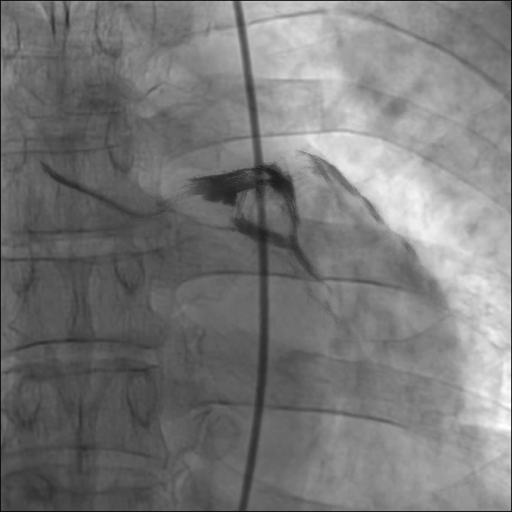}
      \begin{footnotesize}
        \put(3,5){\color{white}{(D1)}}
      \end{footnotesize}
    \end{overpic}
  }
  \subfigure[]{
    \hspace{-0.33cm}
    \begin{overpic}[width=0.089\linewidth]{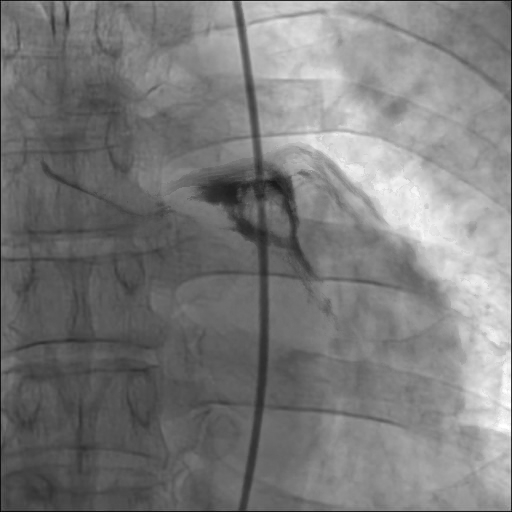}
      \begin{footnotesize}
        \put(3,5){\color{white}{(E1)}}
      \end{footnotesize}
    \end{overpic}
  }
  \subfigure[]{
    \hspace{-0.33cm}
    \begin{overpic}[width=0.089\linewidth]{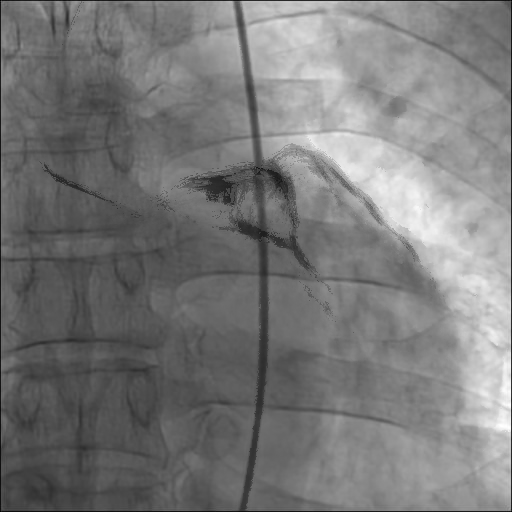}
      \begin{footnotesize}
        \put(3,5){\color{white}{(F1)}}
      \end{footnotesize}
    \end{overpic}
  }
  \subfigure[]{
    \hspace{-0.33cm}
    \begin{overpic}[width=0.089\linewidth]{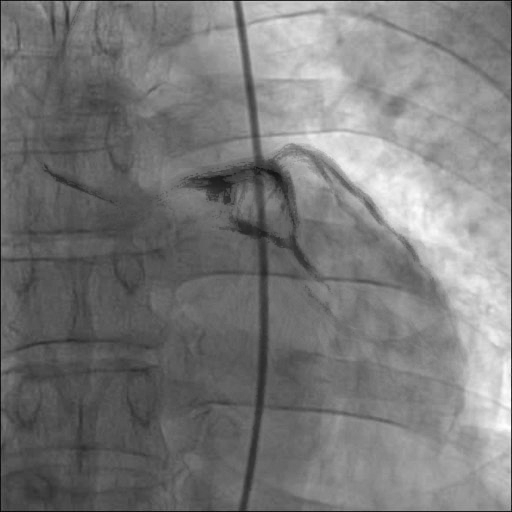}
      \begin{footnotesize}
        \put(3,5){\color{white}{(G1)}}
      \end{footnotesize}
    \end{overpic}
  }
  \subfigure[]{
    \hspace{-0.33cm}
    \begin{overpic}[width=0.089\linewidth]{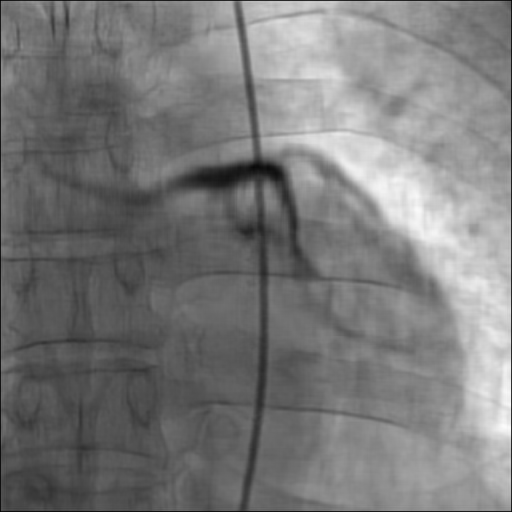}
      \begin{footnotesize}
        \put(3,5){\color{white}{(H1)}}
      \end{footnotesize}
    \end{overpic}
  }
  \subfigure[]{
    \hspace{-0.33cm}
    \begin{overpic}[width=0.089\linewidth]{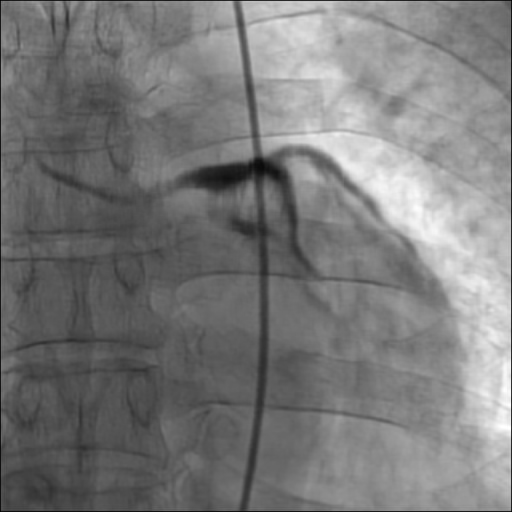}
      \begin{footnotesize}
        \put(3,5){\color{white}{(I1)}}
      \end{footnotesize}
    \end{overpic}
  }
  \subfigure[]{
    \hspace{-0.33cm}
    \begin{overpic}[width=0.089\linewidth]{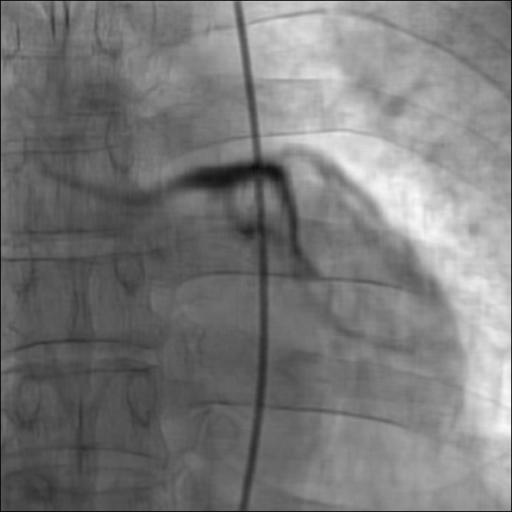}
      \begin{footnotesize}
        \put(3,5){\color{white}{(J1)}}
      \end{footnotesize}
    \end{overpic}
  }
  \subfigure[]{
    \hspace{-0.33cm}
    \begin{overpic}[width=0.089\linewidth]{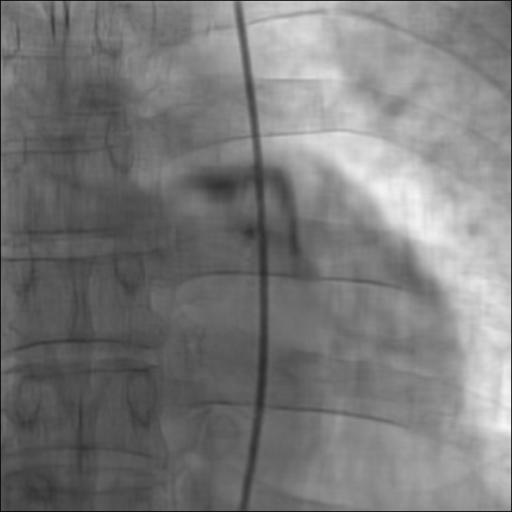}
      \begin{footnotesize}
        \put(3,5){\color{white}{(K1)}}
      \end{footnotesize}
    \end{overpic}
  }

  \vspace{-0.95cm}
  \subfigure[]{
    \begin{overpic}[width=0.089\linewidth]{GT_0137.png}
      \begin{footnotesize}
        \put(3,5){\color{black}{(a1)}}
      \end{footnotesize}
    \end{overpic}
  }
  \subfigure[]{
    \hspace{-0.33cm}
    \begin{overpic}[width=0.089\linewidth]{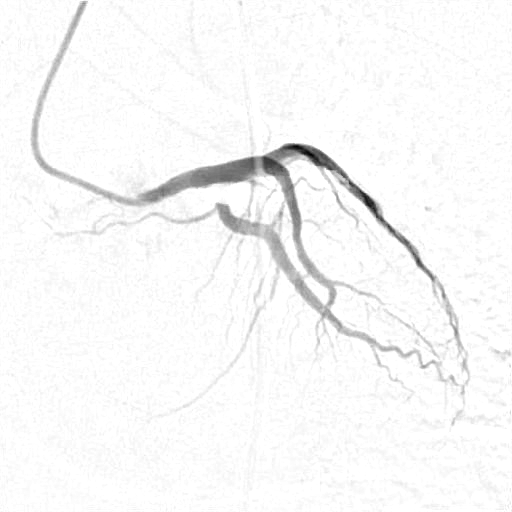}
      \begin{footnotesize}
        \put(3,5){\color{black}{(b1)}}
      \end{footnotesize}
    \end{overpic}
  }
  \subfigure[]{
    \hspace{-0.33cm}
    \begin{overpic}[width=0.089\linewidth]{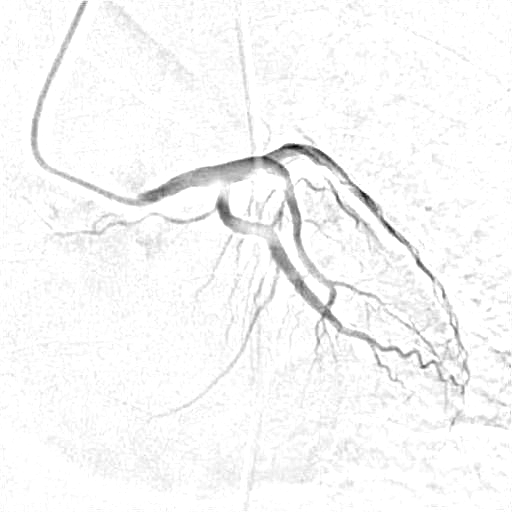}
      \begin{footnotesize}
        \put(3,5){\color{black}{(c1)}}
      \end{footnotesize}
    \end{overpic}
  }
  \subfigure[]{
    \hspace{-0.33cm}
    \begin{overpic}[width=0.089\linewidth]{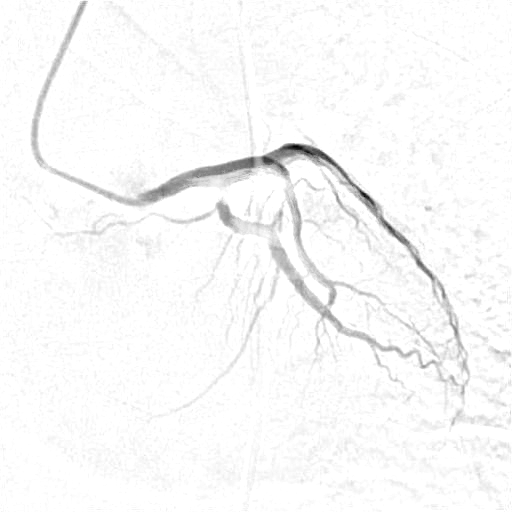}
      \begin{footnotesize}
        \put(3,5){\color{black}{(d1)}}
      \end{footnotesize}
    \end{overpic}
  }
  \subfigure[]{
    \hspace{-0.33cm}
    \begin{overpic}[width=0.089\linewidth]{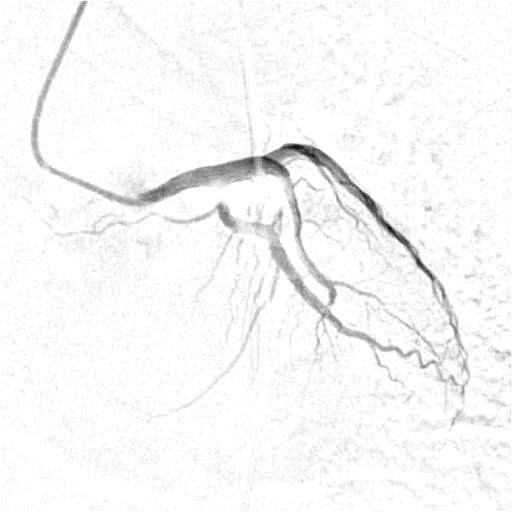}
      \begin{footnotesize}
        \put(3,5){\color{black}{(e1)}}
      \end{footnotesize}
    \end{overpic}
  }
  \subfigure[]{
    \hspace{-0.33cm}
    \begin{overpic}[width=0.089\linewidth]{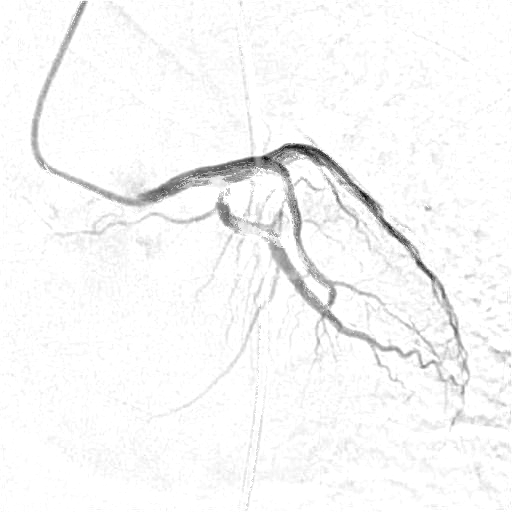}
      \begin{footnotesize}
        \put(3,5){\color{black}{(f1)}}
      \end{footnotesize}
    \end{overpic}
  }
  \subfigure[]{
    \hspace{-0.33cm}
    \begin{overpic}[width=0.089\linewidth]{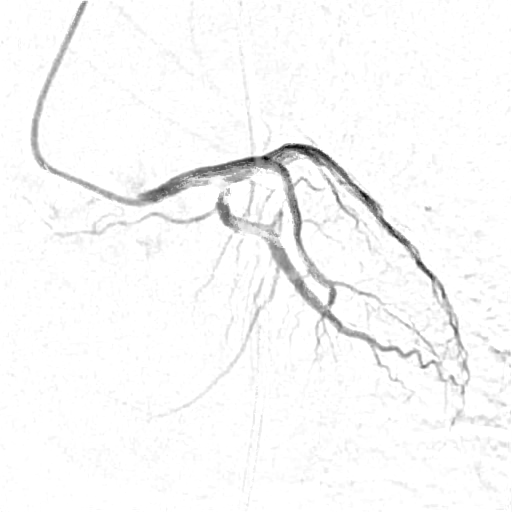}
      \begin{footnotesize}
        \put(3,5){\color{black}{(g1)}}
      \end{footnotesize}
    \end{overpic}
  }
  \subfigure[]{
    \hspace{-0.33cm}
    \begin{overpic}[width=0.089\linewidth]{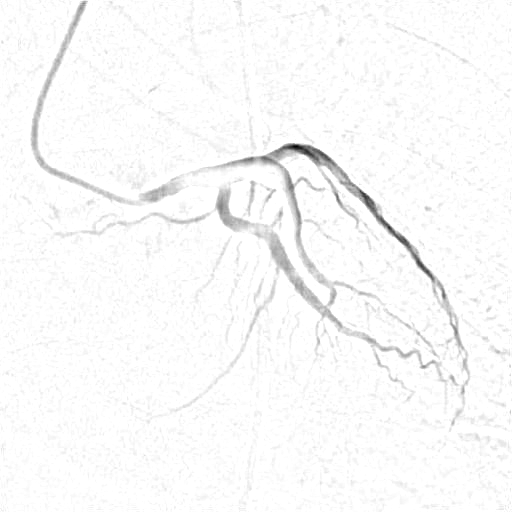}
      \begin{footnotesize}
        \put(3,5){\color{black}{(h1)}}
      \end{footnotesize}
    \end{overpic}
  }
  \subfigure[]{
    \hspace{-0.33cm}
    \begin{overpic}[width=0.089\linewidth]{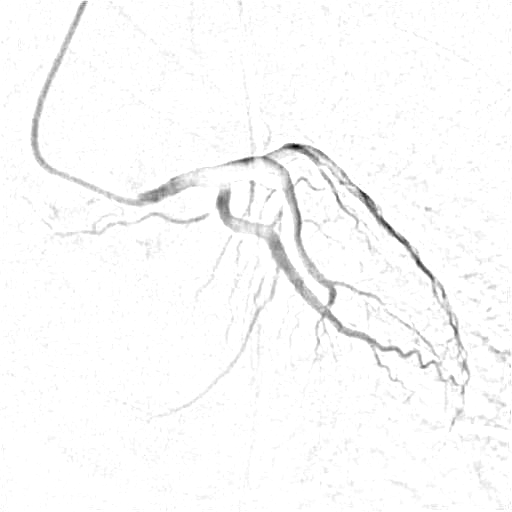}
      \begin{footnotesize}
        \put(3,5){\color{black}{(i1)}}
      \end{footnotesize}
    \end{overpic}
  }
  \subfigure[]{
    \hspace{-0.33cm}
    \begin{overpic}[width=0.089\linewidth]{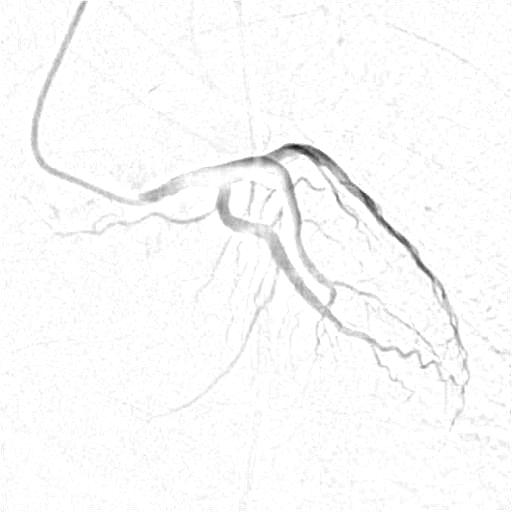}
      \begin{footnotesize}
        \put(3,5){\color{black}{(j1)}}
      \end{footnotesize}
    \end{overpic}
  }
  \subfigure[]{
    \hspace{-0.33cm}
    \begin{overpic}[width=0.089\linewidth]{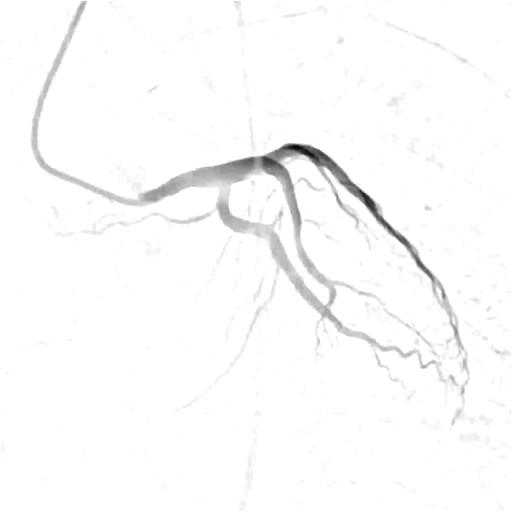}
      \begin{footnotesize}
        \put(3,5){\color{black}{(k1)}}
      \end{footnotesize}
    \end{overpic}
  }

  \vspace{-0.85cm}
  \subfigure[]{
    \begin{overpic}[width=0.089\linewidth]{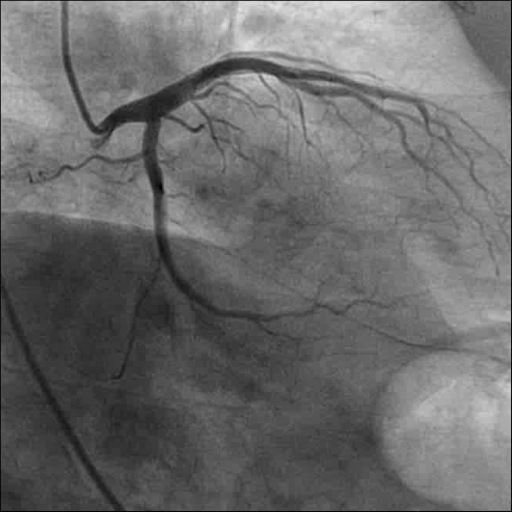}
      \begin{footnotesize}
        \put(3,5){\color{white}{(A2)}}
      \end{footnotesize}
    \end{overpic}
  }
  \subfigure[]{
    \hspace{-0.33cm}
    \begin{overpic}[width=0.089\linewidth]{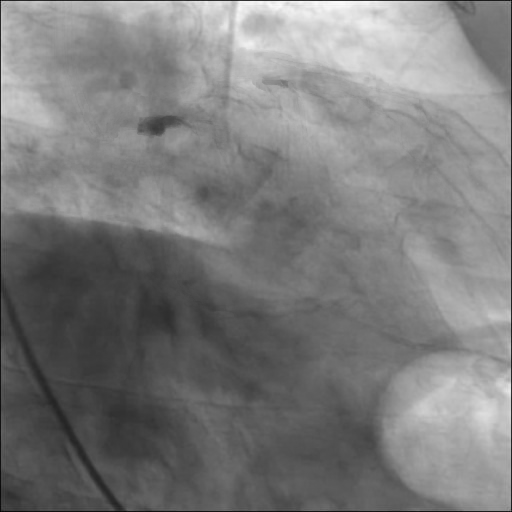}
      \begin{footnotesize}
        \put(3,5){\color{white}{(B2)}}
      \end{footnotesize}
    \end{overpic}
  }
  \subfigure[]{
    \hspace{-0.33cm}
    \begin{overpic}[width=0.089\linewidth]{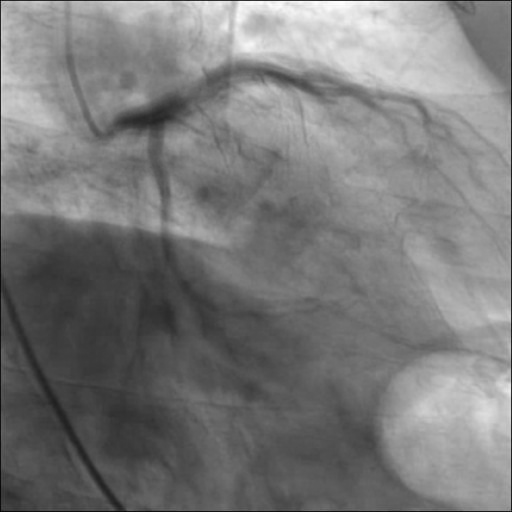}
      \begin{footnotesize}
        \put(3,5){\color{white}{(C2)}}
      \end{footnotesize}
    \end{overpic}
  }
  \subfigure[]{
    \hspace{-0.33cm}
    \begin{overpic}[width=0.089\linewidth]{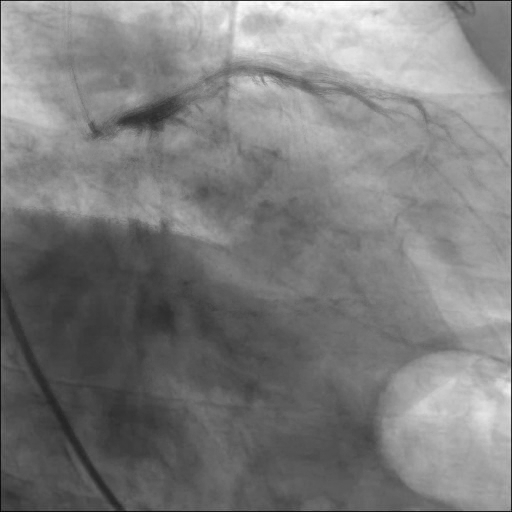}
      \begin{footnotesize}
        \put(3,5){\color{white}{(D2)}}
      \end{footnotesize}
    \end{overpic}
  }
  \subfigure[]{
    \hspace{-0.33cm}
    \begin{overpic}[width=0.089\linewidth]{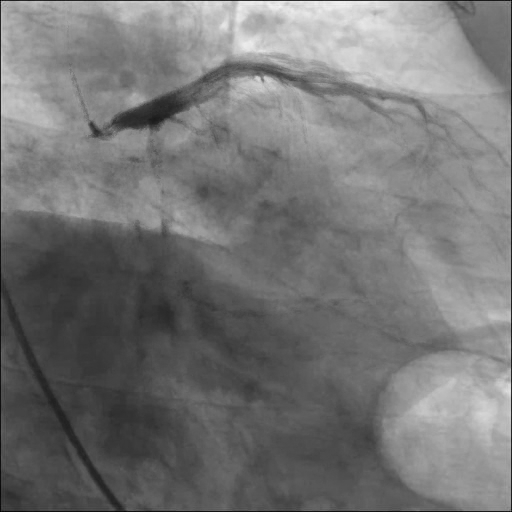}
      \begin{footnotesize}
        \put(3,5){\color{white}{(E2)}}
      \end{footnotesize}
    \end{overpic}
  }
  \subfigure[]{
    \hspace{-0.33cm}
    \begin{overpic}[width=0.089\linewidth]{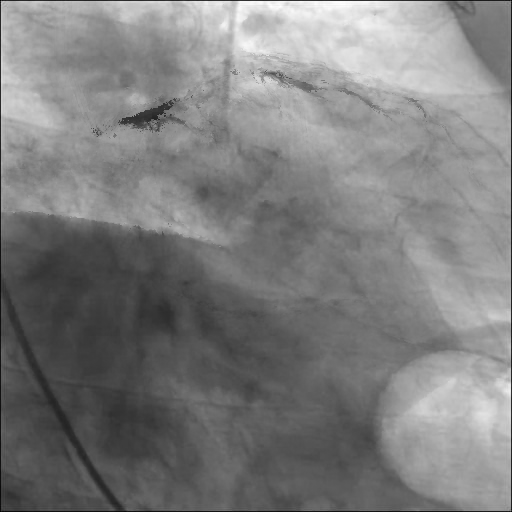}
      \begin{footnotesize}
        \put(3,5){\color{white}{(F2)}}
      \end{footnotesize}
    \end{overpic}
  }
  \subfigure[]{
    \hspace{-0.33cm}
    \begin{overpic}[width=0.089\linewidth]{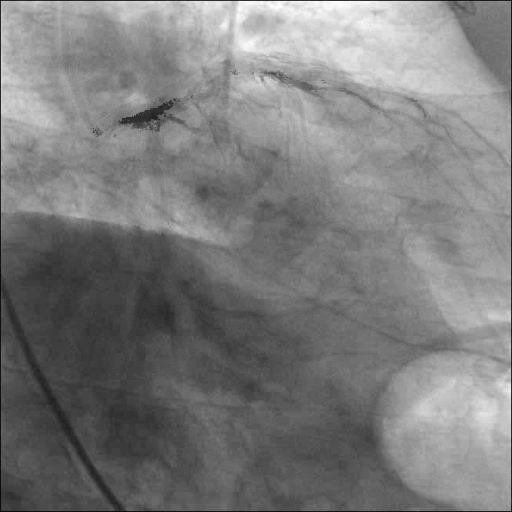}
      \begin{footnotesize}
        \put(3,5){\color{white}{(G2)}}
      \end{footnotesize}
    \end{overpic}
  }
  \subfigure[]{
    \hspace{-0.33cm}
    \begin{overpic}[width=0.089\linewidth]{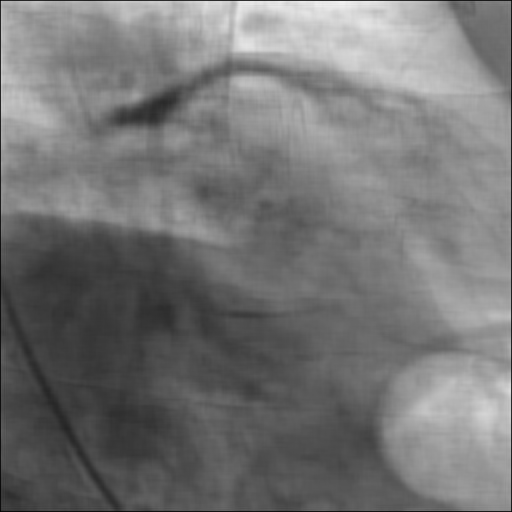}
      \begin{footnotesize}
        \put(3,5){\color{white}{(H2)}}
      \end{footnotesize}
    \end{overpic}
  }
  \subfigure[]{
    \hspace{-0.33cm}
    \begin{overpic}[width=0.089\linewidth]{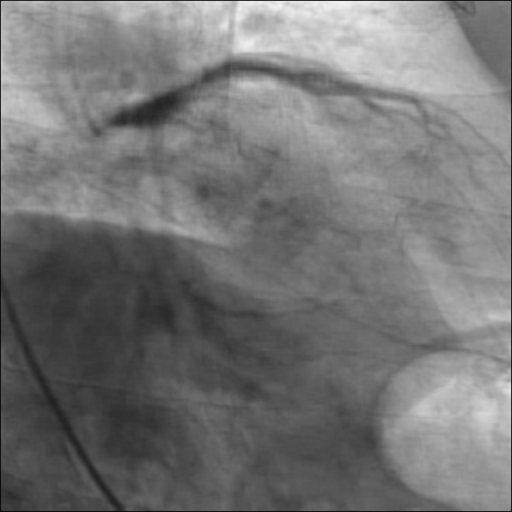}
      \begin{footnotesize}
        \put(3,5){\color{white}{(I2)}}
      \end{footnotesize}
    \end{overpic}
  }
  \subfigure[]{
    \hspace{-0.33cm}
    \begin{overpic}[width=0.089\linewidth]{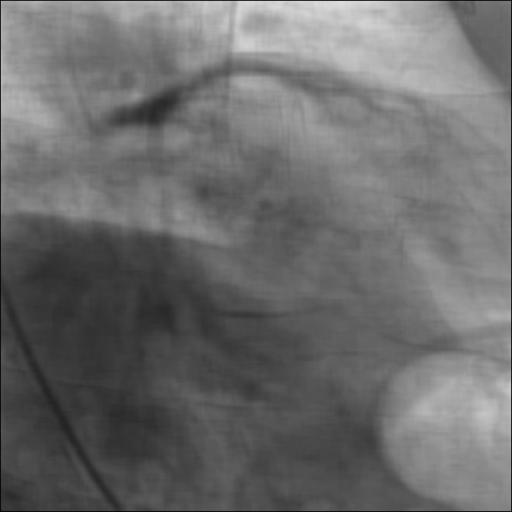}
      \begin{footnotesize}
        \put(3,5){\color{white}{(J2)}}
      \end{footnotesize}
    \end{overpic}
  }
  \subfigure[]{
    \hspace{-0.33cm}
    \begin{overpic}[width=0.089\linewidth]{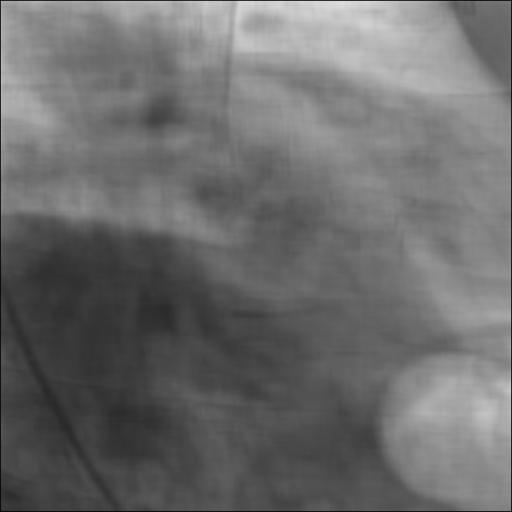}
      \begin{footnotesize}
        \put(3,5){\color{white}{(K2)}}
      \end{footnotesize}
    \end{overpic}
  }

  \vspace{-0.95cm}
  \subfigure[]{
    \begin{overpic}[width=0.089\linewidth]{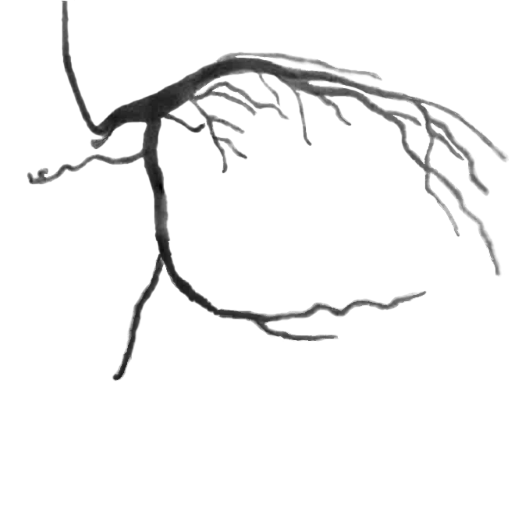}
      \begin{footnotesize}
        \put(3,5){\color{black}{(a2)}}
      \end{footnotesize}
    \end{overpic}
  }
  \subfigure[]{
    \hspace{-0.33cm}
    \begin{overpic}[width=0.089\linewidth]{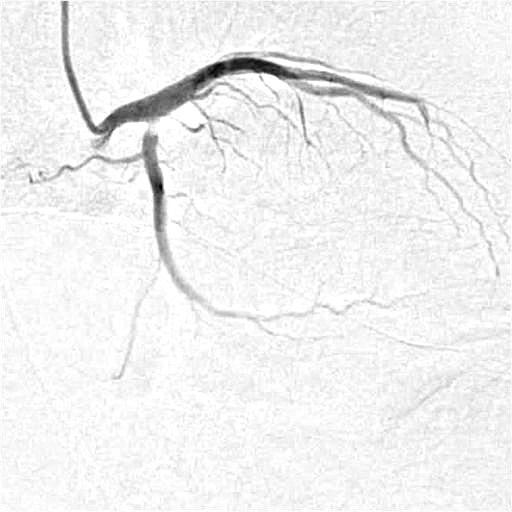}
      \begin{footnotesize}
        \put(3,5){\color{black}{(b2)}}
      \end{footnotesize}
    \end{overpic}
  }
  \subfigure[]{
    \hspace{-0.33cm}
    \begin{overpic}[width=0.089\linewidth]{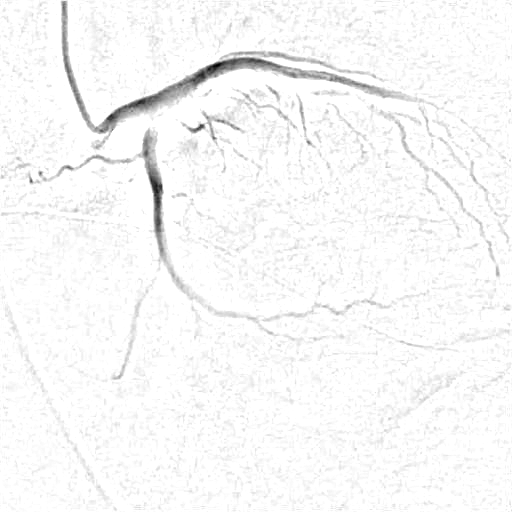}
      \begin{footnotesize}
        \put(3,5){\color{black}{(c2)}}
      \end{footnotesize}
    \end{overpic}
  }
  \subfigure[]{
    \hspace{-0.33cm}
    \begin{overpic}[width=0.089\linewidth]{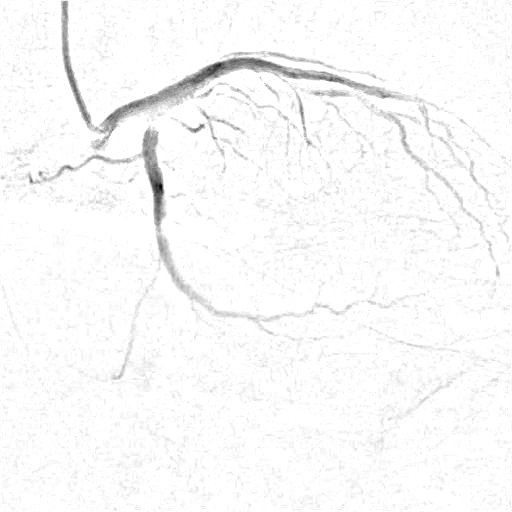}
      \begin{footnotesize}
        \put(3,5){\color{black}{(d2)}}
      \end{footnotesize}
    \end{overpic}
  }
  \subfigure[]{
    \hspace{-0.33cm}
    \begin{overpic}[width=0.089\linewidth]{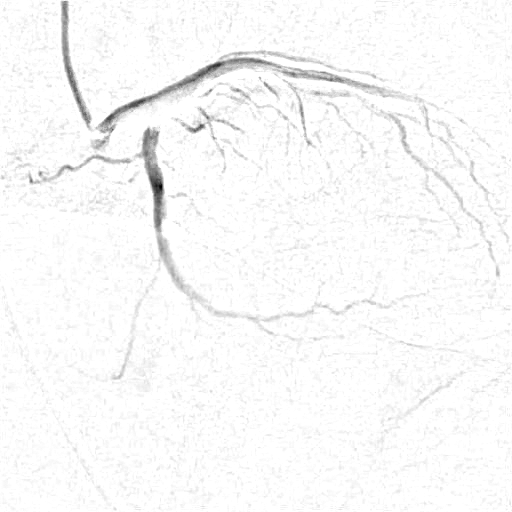}
      \begin{footnotesize}
        \put(3,5){\color{black}{(e2)}}
      \end{footnotesize}
    \end{overpic}
  }
  \subfigure[]{
    \hspace{-0.33cm}
    \begin{overpic}[width=0.089\linewidth]{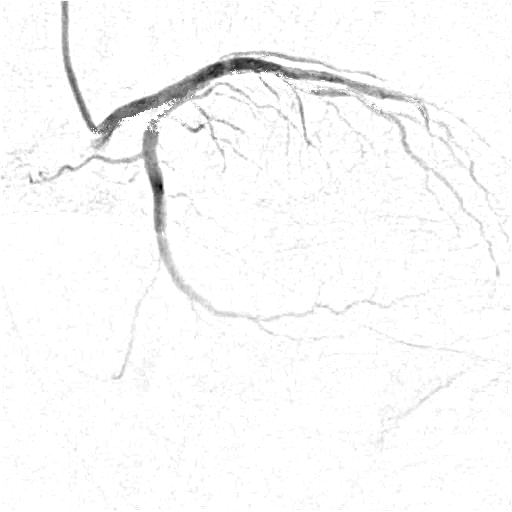}
      \begin{footnotesize}
        \put(3,5){\color{black}{(f2)}}
      \end{footnotesize}
    \end{overpic}
  }
  \subfigure[]{
    \hspace{-0.33cm}
    \begin{overpic}[width=0.089\linewidth]{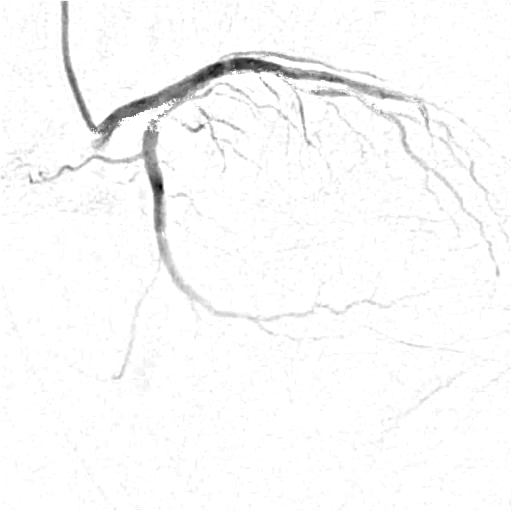}
      \begin{footnotesize}
        \put(3,5){\color{black}{(g2)}}
      \end{footnotesize}
    \end{overpic}
  }
  \subfigure[]{
    \hspace{-0.33cm}
    \begin{overpic}[width=0.089\linewidth]{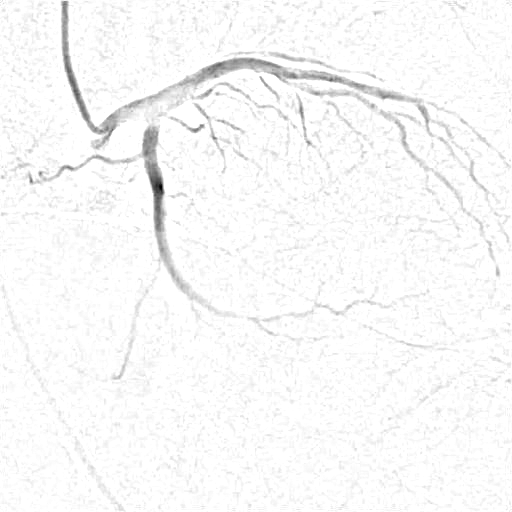}
      \begin{footnotesize}
        \put(3,5){\color{black}{(h2)}}
      \end{footnotesize}
    \end{overpic}
  }
  \subfigure[]{
    \hspace{-0.33cm}
    \begin{overpic}[width=0.089\linewidth]{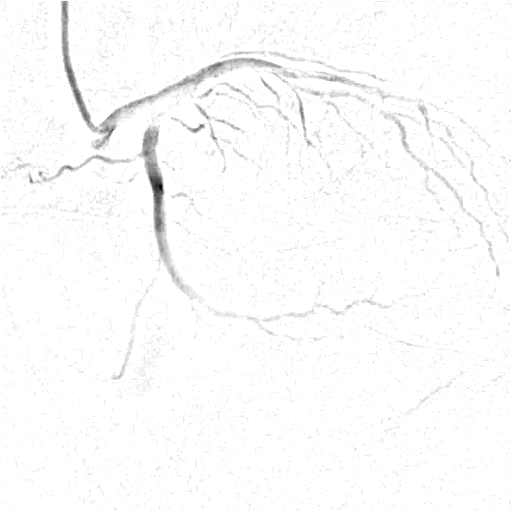}
      \begin{footnotesize}
        \put(3,5){\color{black}{(i2)}}
      \end{footnotesize}
    \end{overpic}
  }
  \subfigure[]{
    \hspace{-0.33cm}
    \begin{overpic}[width=0.089\linewidth]{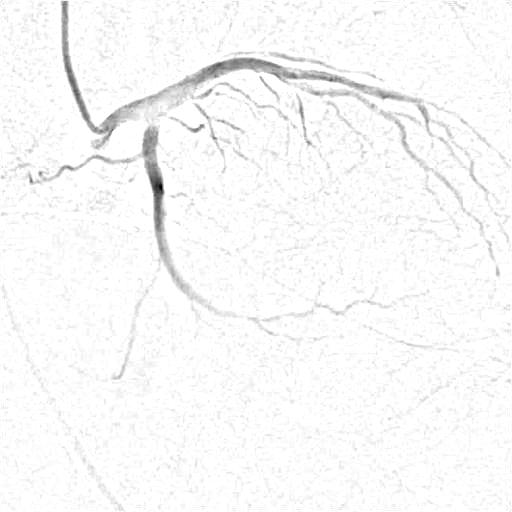}
      \begin{footnotesize}
        \put(3,5){\color{black}{(j2)}}
      \end{footnotesize}
    \end{overpic}
  }
  \subfigure[]{
    \hspace{-0.33cm}
    \begin{overpic}[width=0.089\linewidth]{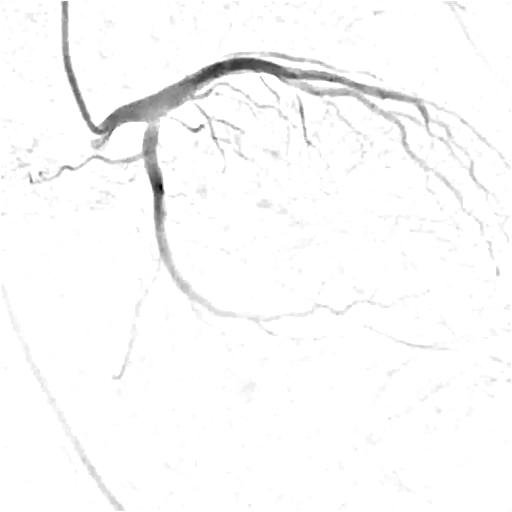}
      \begin{footnotesize}
        \put(3,5){\color{black}{(k2)}}
      \end{footnotesize}
    \end{overpic}
  }

  \vspace{-0.85cm}
  \subfigure[]{
    \begin{overpic}[width=0.089\linewidth]{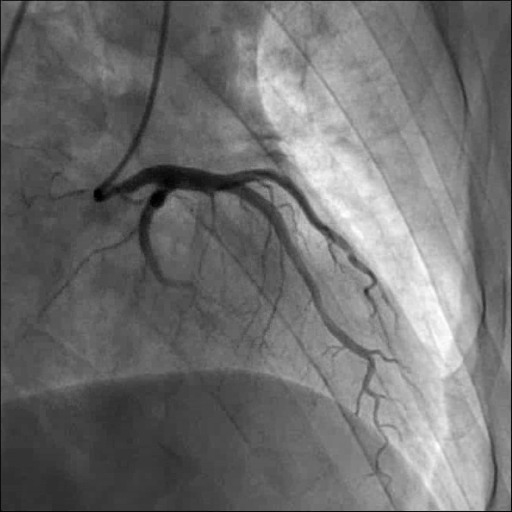}
      \begin{footnotesize}
        \put(3,5){\color{white}{(A3)}}
      \end{footnotesize}
    \end{overpic}
  }
  \subfigure[]{
    \hspace{-0.33cm}
    \begin{overpic}[width=0.089\linewidth]{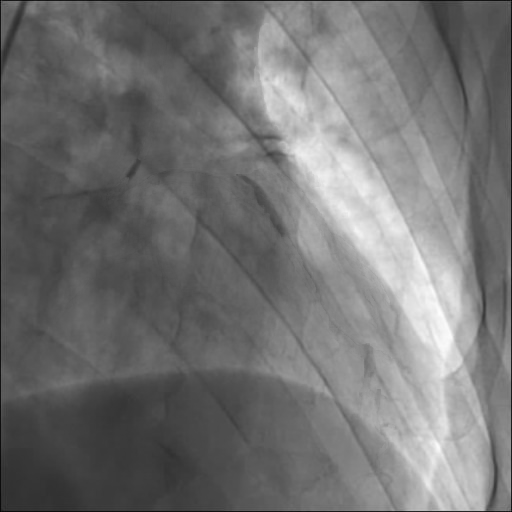}
      \begin{footnotesize}
        \put(3,5){\color{white}{(B3)}}
      \end{footnotesize}
    \end{overpic}
  }
  \subfigure[]{
    \hspace{-0.33cm}
    \begin{overpic}[width=0.089\linewidth]{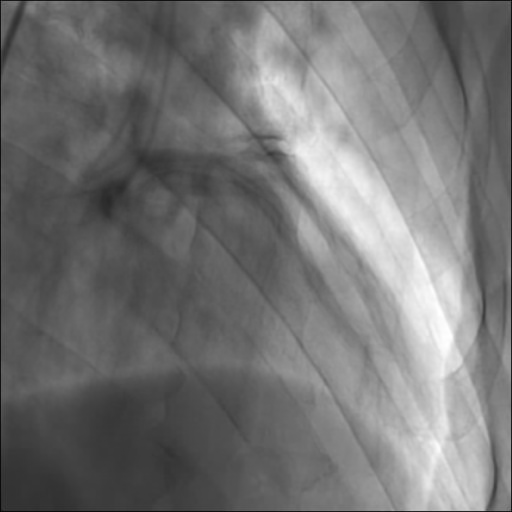}
      \begin{footnotesize}
        \put(3,5){\color{white}{(C3)}}
      \end{footnotesize}
    \end{overpic}
  }
  \subfigure[]{
    \hspace{-0.33cm}
    \begin{overpic}[width=0.089\linewidth]{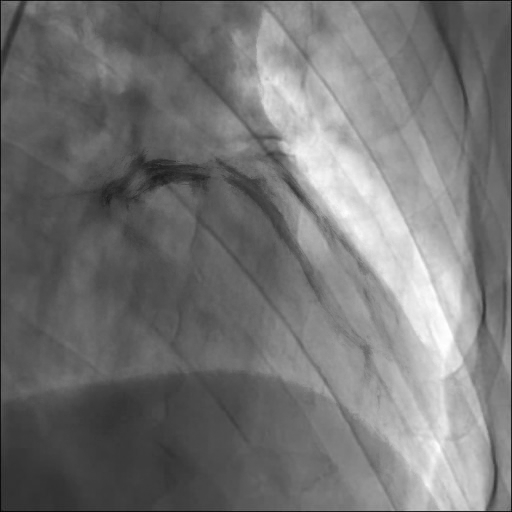}
      \begin{footnotesize}
        \put(3,5){\color{white}{(D3)}}
      \end{footnotesize}
    \end{overpic}
  }
  \subfigure[]{
    \hspace{-0.33cm}
    \begin{overpic}[width=0.089\linewidth]{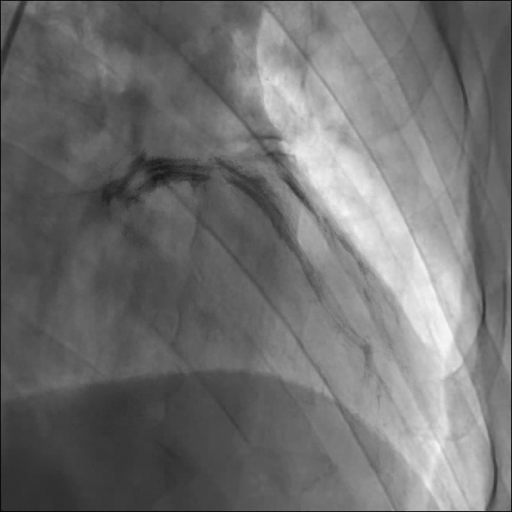}
      \begin{footnotesize}
        \put(3,5){\color{white}{(E3)}}
      \end{footnotesize}
    \end{overpic}
  }
  \subfigure[]{
    \hspace{-0.33cm}
    \begin{overpic}[width=0.089\linewidth]{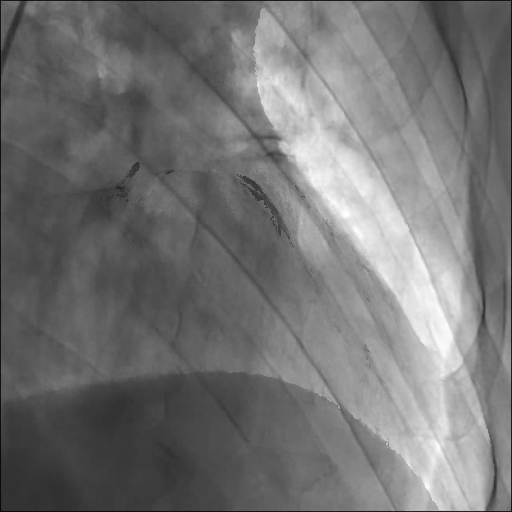}
      \begin{footnotesize}
        \put(3,5){\color{white}{(F3)}}
      \end{footnotesize}
    \end{overpic}
  }
  \subfigure[]{
    \hspace{-0.33cm}
    \begin{overpic}[width=0.089\linewidth]{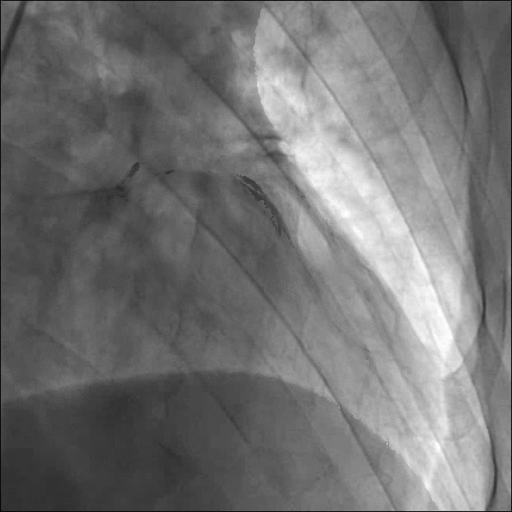}
      \begin{footnotesize}
        \put(3,5){\color{white}{(G3)}}
      \end{footnotesize}
    \end{overpic}
  }
  \subfigure[]{
    \hspace{-0.33cm}
    \begin{overpic}[width=0.089\linewidth]{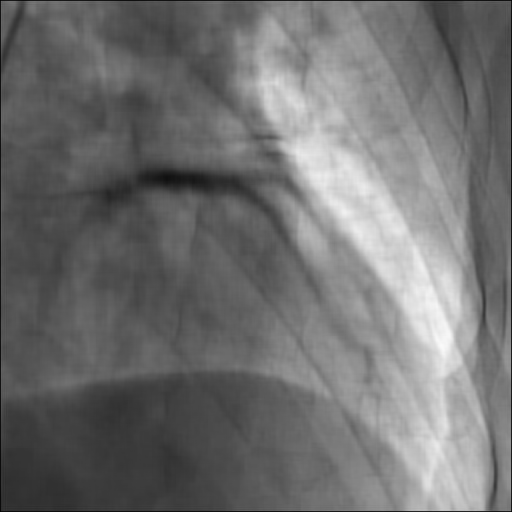}
      \begin{footnotesize}
        \put(3,5){\color{white}{(H3)}}
      \end{footnotesize}
    \end{overpic}
  }
  \subfigure[]{
    \hspace{-0.33cm}
    \begin{overpic}[width=0.089\linewidth]{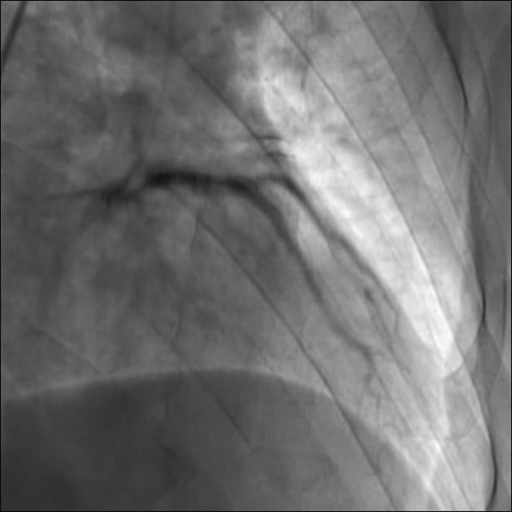}
      \begin{footnotesize}
        \put(3,5){\color{white}{(I3)}}
      \end{footnotesize}
    \end{overpic}
  }
  \subfigure[]{
    \hspace{-0.33cm}
    \begin{overpic}[width=0.089\linewidth]{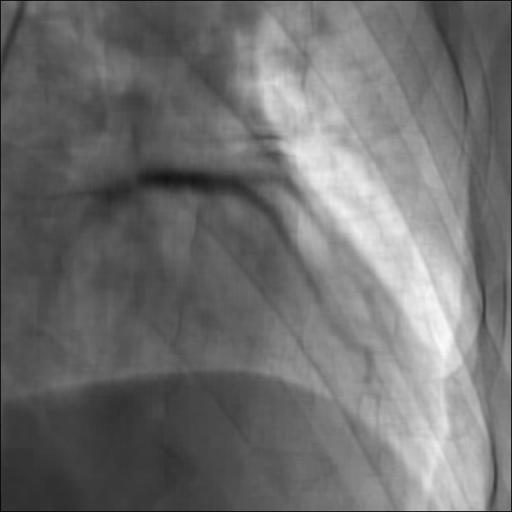}
      \begin{footnotesize}
        \put(3,5){\color{white}{(J3)}}
      \end{footnotesize}
    \end{overpic}
  }
  \subfigure[]{
    \hspace{-0.33cm}
    \begin{overpic}[width=0.089\linewidth]{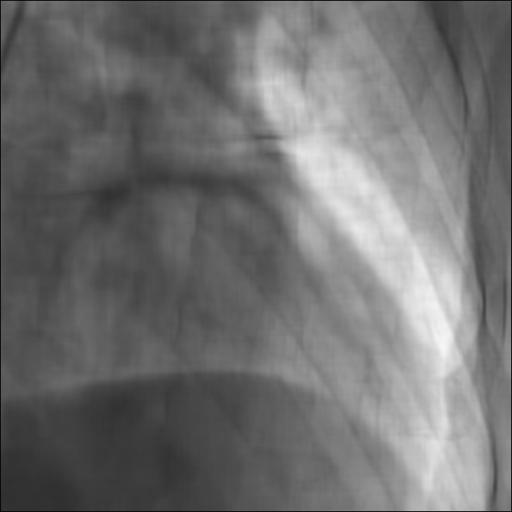}
      \begin{footnotesize}
        \put(3,5){\color{white}{(K3)}}
      \end{footnotesize}
    \end{overpic}
  }

  \vspace{-0.95cm}
  \subfigure[]{
    \begin{overpic}[width=0.089\linewidth]{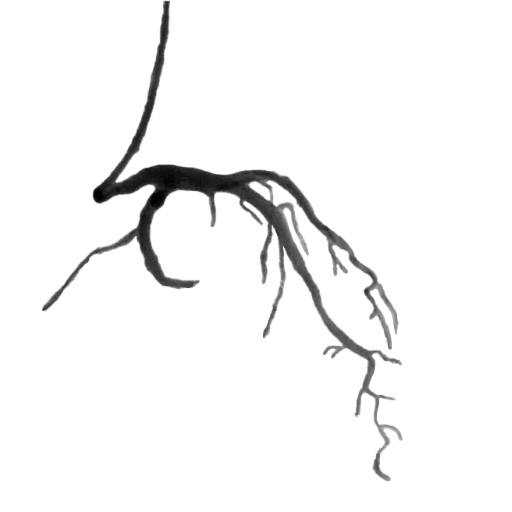}
      \begin{footnotesize}
        \put(3,5){\color{black}{(a3)}}
      \end{footnotesize}
    \end{overpic}
  }
  \subfigure[]{
    \hspace{-0.33cm}
    \begin{overpic}[width=0.089\linewidth]{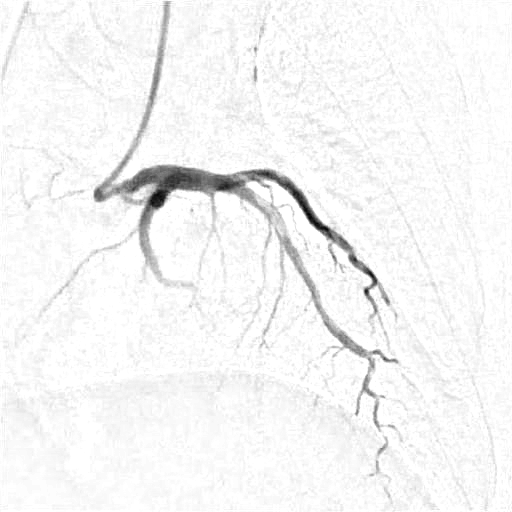}
      \begin{footnotesize}
        \put(3,5){\color{black}{(b3)}}
      \end{footnotesize}
    \end{overpic}
  }
  \subfigure[]{
    \hspace{-0.33cm}
    \begin{overpic}[width=0.089\linewidth]{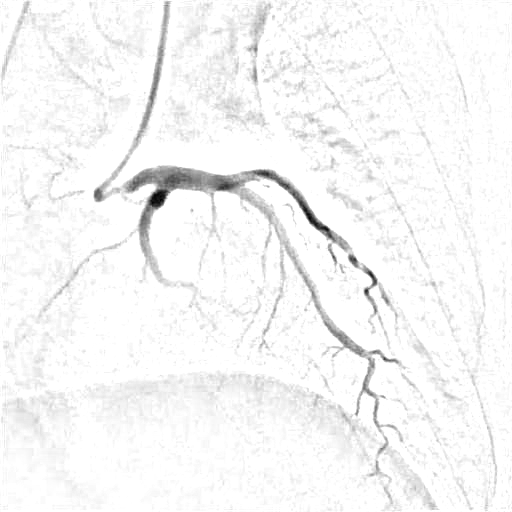}
      \begin{footnotesize}
        \put(3,5){\color{black}{(c3)}}
      \end{footnotesize}
    \end{overpic}
  }
  \subfigure[]{
    \hspace{-0.33cm}
    \begin{overpic}[width=0.089\linewidth]{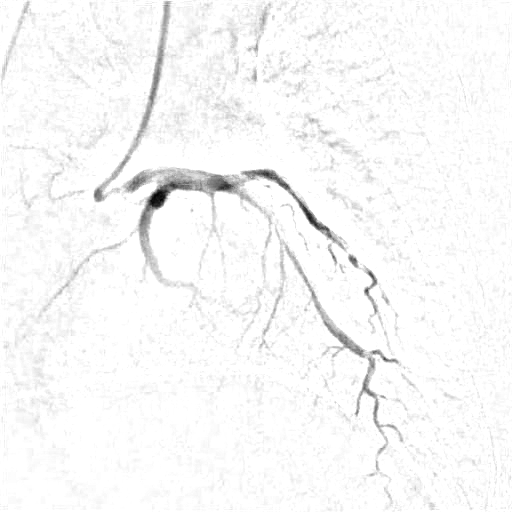}
      \begin{footnotesize}
        \put(3,5){\color{black}{(d3)}}
      \end{footnotesize}
    \end{overpic}
  }
  \subfigure[]{
    \hspace{-0.33cm}
    \begin{overpic}[width=0.089\linewidth]{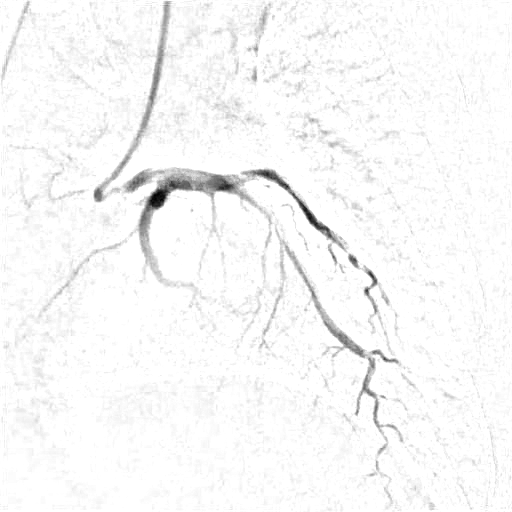}
      \begin{footnotesize}
        \put(3,5){\color{black}{(e3)}}
      \end{footnotesize}
    \end{overpic}
  }
  \subfigure[]{
    \hspace{-0.33cm}
    \begin{overpic}[width=0.089\linewidth]{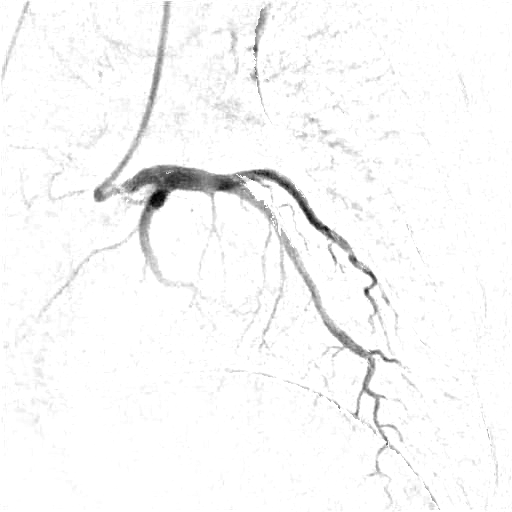}
      \begin{footnotesize}
        \put(3,5){\color{black}{(f3)}}
      \end{footnotesize}
    \end{overpic}
  }
  \subfigure[]{
    \hspace{-0.33cm}
    \begin{overpic}[width=0.089\linewidth]{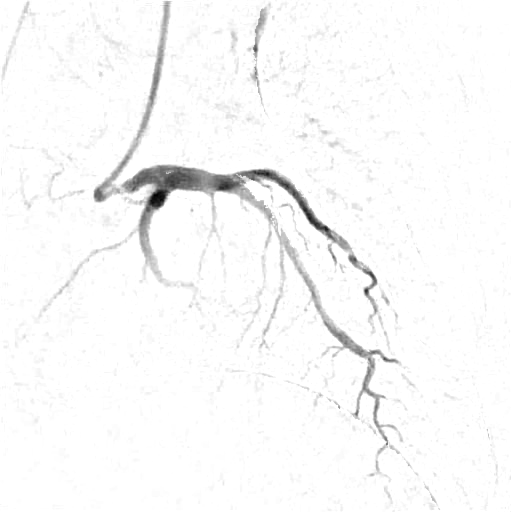}
      \begin{footnotesize}
        \put(3,5){\color{black}{(g3)}}
      \end{footnotesize}
    \end{overpic}
  }
  \subfigure[]{
    \hspace{-0.33cm}
    \begin{overpic}[width=0.089\linewidth]{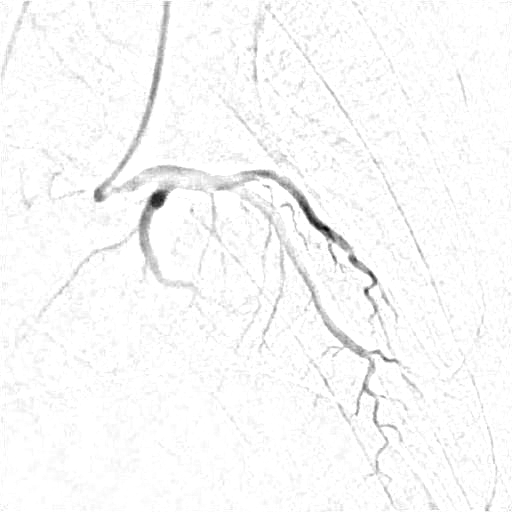}
      \begin{footnotesize}
        \put(3,5){\color{black}{(h3)}}
      \end{footnotesize}
    \end{overpic}
  }
  \subfigure[]{
    \hspace{-0.33cm}
    \begin{overpic}[width=0.089\linewidth]{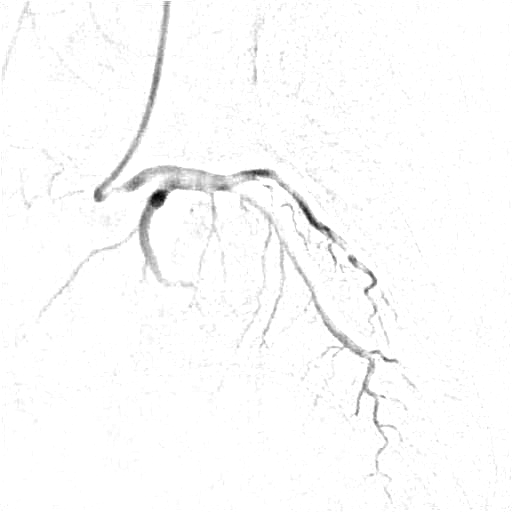}
      \begin{footnotesize}
        \put(3,5){\color{black}{(i3)}}
      \end{footnotesize}
    \end{overpic}
  }
  \subfigure[]{
    \hspace{-0.33cm}
    \begin{overpic}[width=0.089\linewidth]{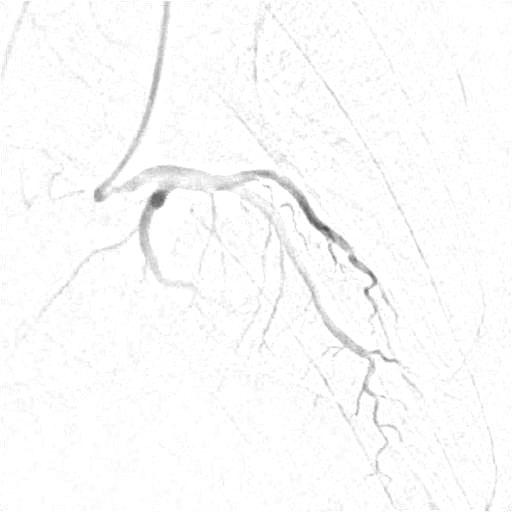}
      \begin{footnotesize}
        \put(3,5){\color{black}{(j3)}}
      \end{footnotesize}
    \end{overpic}
  }
  \subfigure[]{
    \hspace{-0.33cm}
    \begin{overpic}[width=0.089\linewidth]{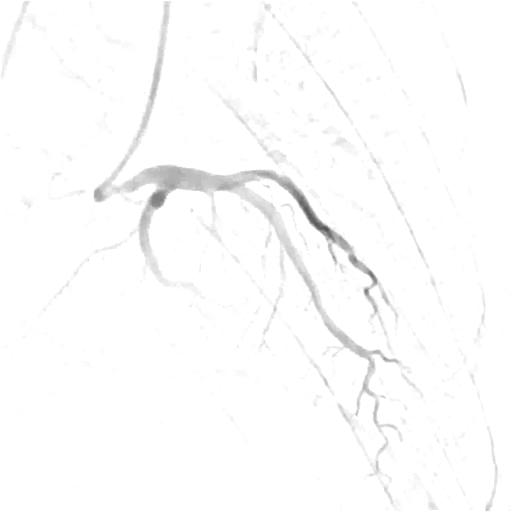}
      \begin{footnotesize}
        \put(3,5){\color{black}{(k3)}}
      \end{footnotesize}
    \end{overpic}
  }

  \vspace{-0.85cm}
  \subfigure[]{
    \begin{overpic}[width=0.089\linewidth]{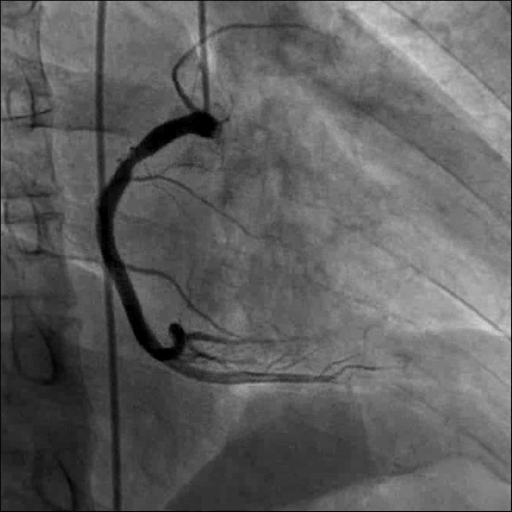}
      \begin{footnotesize}
        \put(3,5){\color{white}{(A4)}}
      \end{footnotesize}
    \end{overpic}
  }
  \subfigure[]{
    \hspace{-0.33cm}
    \begin{overpic}[width=0.089\linewidth]{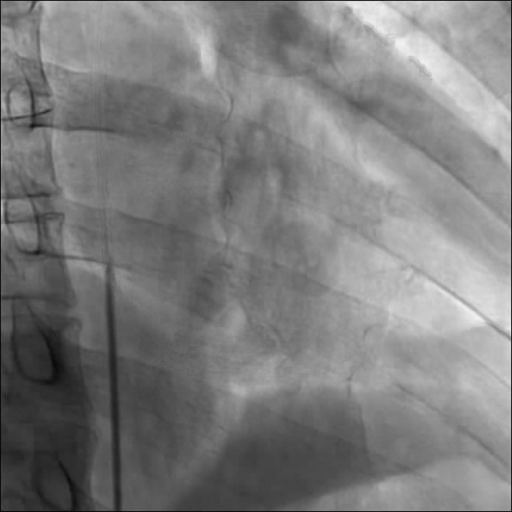}
      \begin{footnotesize}
        \put(3,5){\color{white}{(B4)}}
      \end{footnotesize}
    \end{overpic}
  }
  \subfigure[]{
    \hspace{-0.33cm}
    \begin{overpic}[width=0.089\linewidth]{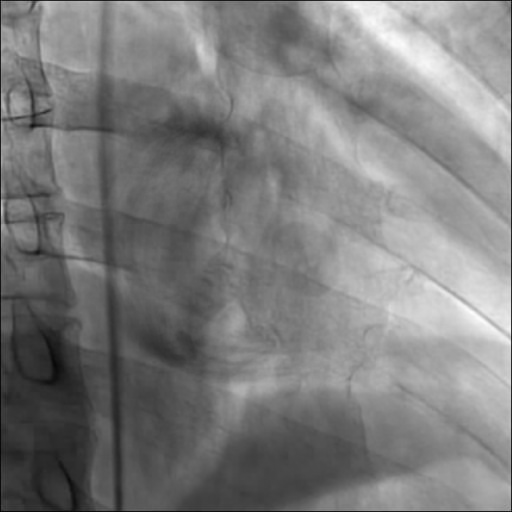}
      \begin{footnotesize}
        \put(3,5){\color{white}{(C4)}}
      \end{footnotesize}
    \end{overpic}
  }
  \subfigure[]{
    \hspace{-0.33cm}
    \begin{overpic}[width=0.089\linewidth]{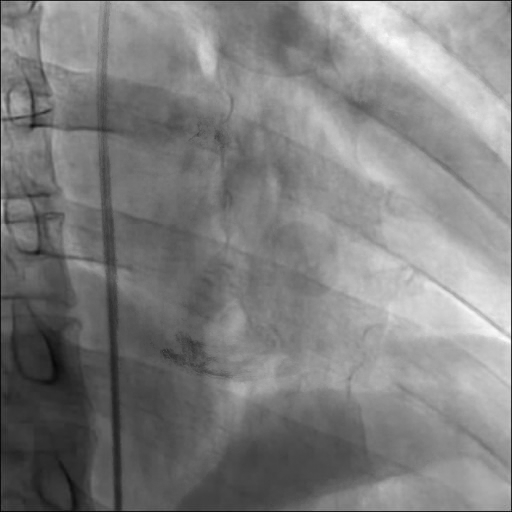}
      \begin{footnotesize}
        \put(3,5){\color{white}{(D4)}}
      \end{footnotesize}
    \end{overpic}
  }
  \subfigure[]{
    \hspace{-0.33cm}
    \begin{overpic}[width=0.089\linewidth]{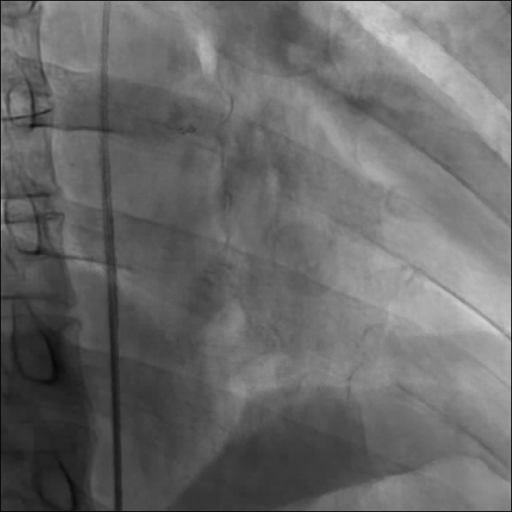}
      \begin{footnotesize}
        \put(3,5){\color{white}{(E4)}}
      \end{footnotesize}
    \end{overpic}
  }
  \subfigure[]{
    \hspace{-0.33cm}
    \begin{overpic}[width=0.089\linewidth]{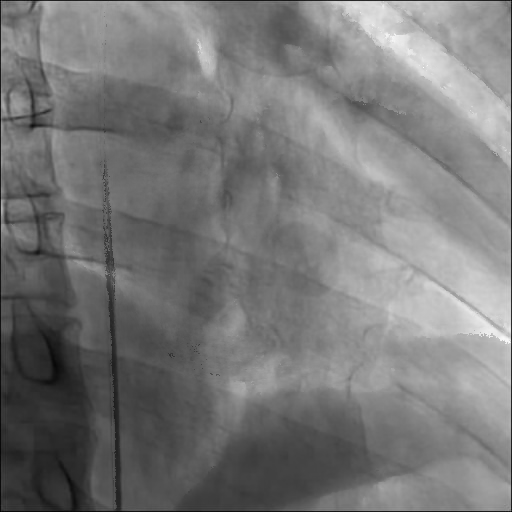}
      \begin{footnotesize}
        \put(3,5){\color{white}{(F4)}}
      \end{footnotesize}
    \end{overpic}
  }
  \subfigure[]{
    \hspace{-0.33cm}
    \begin{overpic}[width=0.089\linewidth]{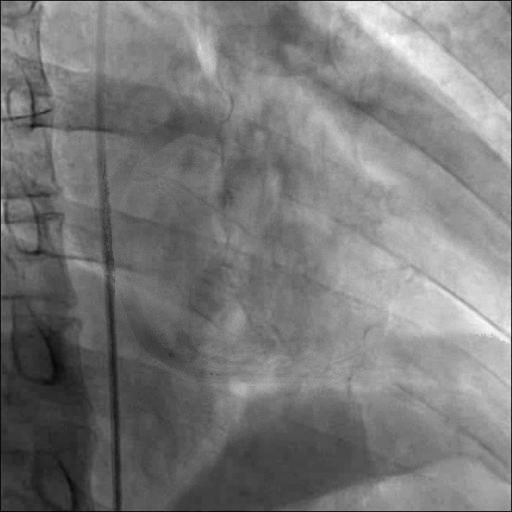}
      \begin{footnotesize}
        \put(3,5){\color{white}{(G4)}}
      \end{footnotesize}
    \end{overpic}
  }
  \subfigure[]{
    \hspace{-0.33cm}
    \begin{overpic}[width=0.089\linewidth]{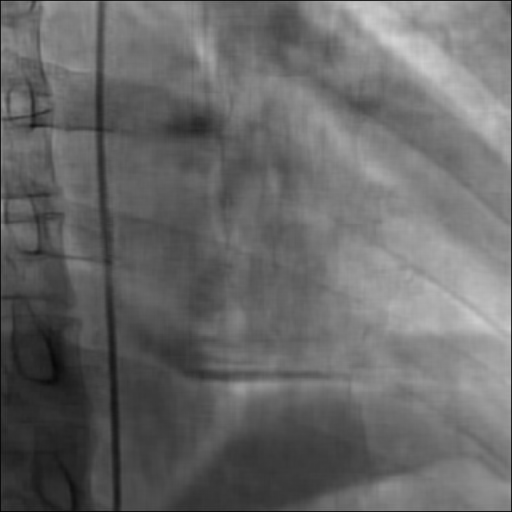}
      \begin{footnotesize}
        \put(3,5){\color{white}{(H4)}}
      \end{footnotesize}
    \end{overpic}
  }
  \subfigure[]{
    \hspace{-0.33cm}
    \begin{overpic}[width=0.089\linewidth]{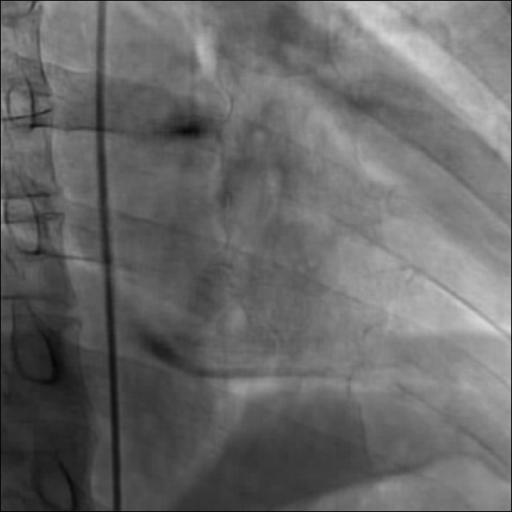}
      \begin{footnotesize}
        \put(3,5){\color{white}{(I4)}}
      \end{footnotesize}
    \end{overpic}
  }
  \subfigure[]{
    \hspace{-0.33cm}
    \begin{overpic}[width=0.089\linewidth]{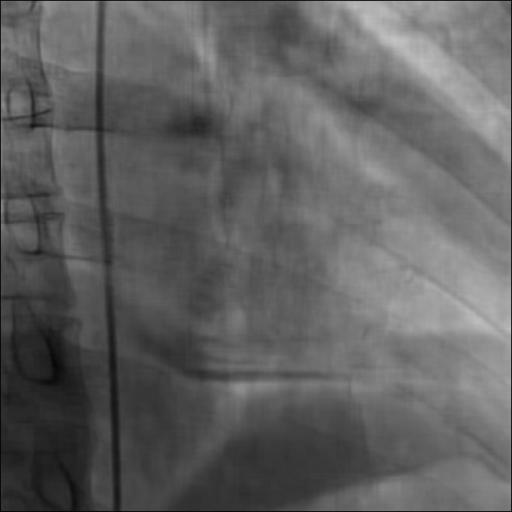}
      \begin{footnotesize}
        \put(3,5){\color{white}{(J4)}}
      \end{footnotesize}
    \end{overpic}
  }
  \subfigure[]{
    \hspace{-0.33cm}
    \begin{overpic}[width=0.089\linewidth]{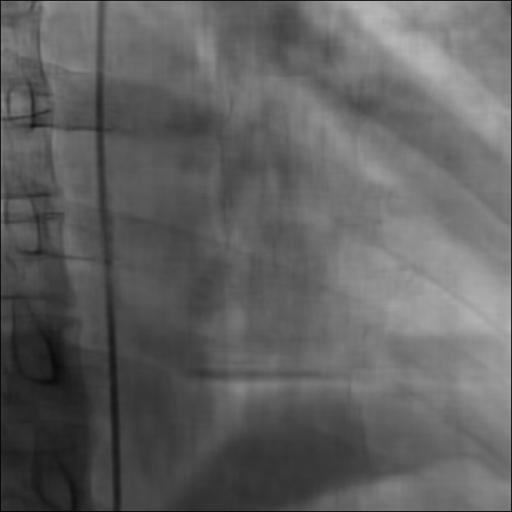}
      \begin{footnotesize}
        \put(3,5){\color{white}{(K4)}}
      \end{footnotesize}
    \end{overpic}
  }

  \vspace{-0.95cm}
  \subfigure[]{
    \begin{overpic}[width=0.089\linewidth]{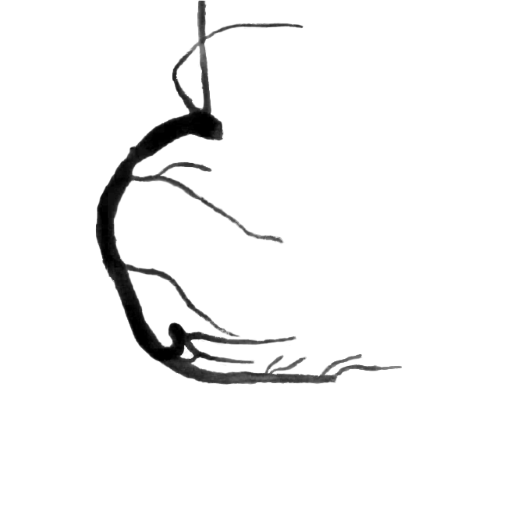}
      \begin{footnotesize}
        \put(3,5){\color{black}{(a4)}}
      \end{footnotesize}
    \end{overpic}
  }
  \subfigure[]{
    \hspace{-0.33cm}
    \begin{overpic}[width=0.089\linewidth]{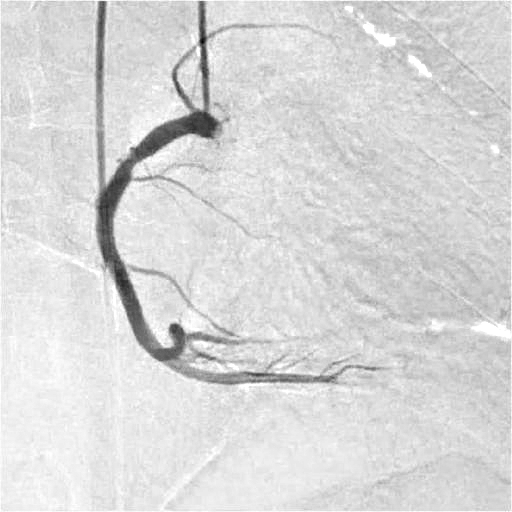}
      \begin{footnotesize}
        \put(3,5){\color{black}{(b4)}}
      \end{footnotesize}
    \end{overpic}
  }
  \subfigure[]{
    \hspace{-0.33cm}
    \begin{overpic}[width=0.089\linewidth]{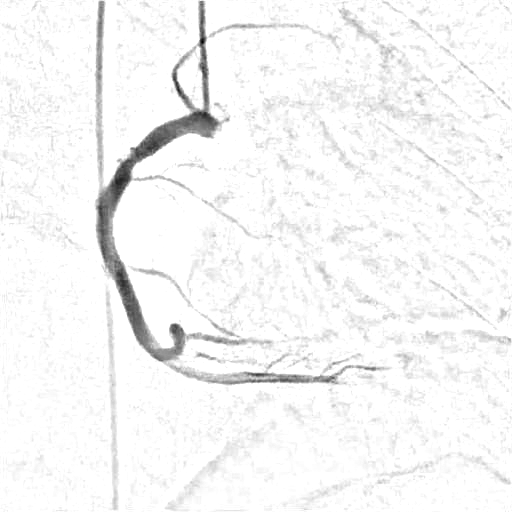}
      \begin{footnotesize}
        \put(3,5){\color{black}{(c4)}}
      \end{footnotesize}
    \end{overpic}
  }
  \subfigure[]{
    \hspace{-0.33cm}
    \begin{overpic}[width=0.089\linewidth]{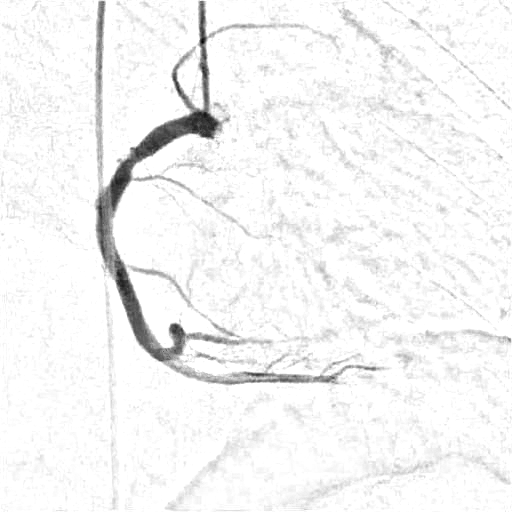}
      \begin{footnotesize}
        \put(3,5){\color{black}{(d4)}}
      \end{footnotesize}
    \end{overpic}
  }
  \subfigure[]{
    \hspace{-0.33cm}
    \begin{overpic}[width=0.089\linewidth]{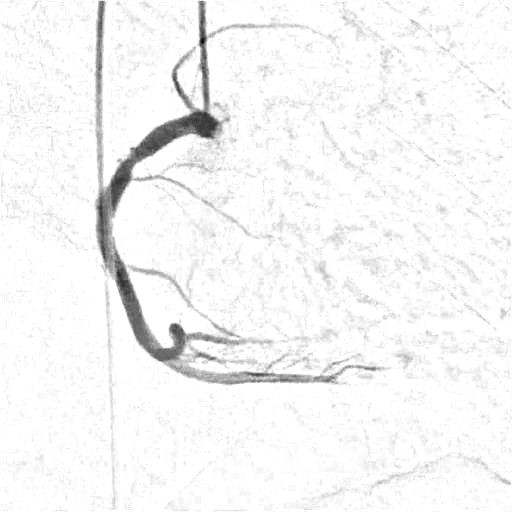}
      \begin{footnotesize}
        \put(3,5){\color{black}{(e4)}}
      \end{footnotesize}
    \end{overpic}
  }
  \subfigure[]{
    \hspace{-0.33cm}
    \begin{overpic}[width=0.089\linewidth]{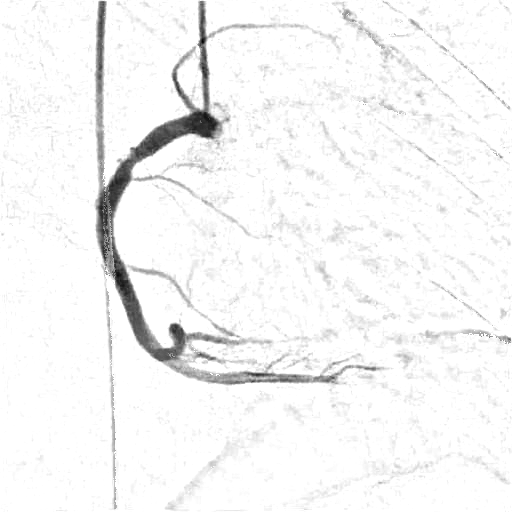}
      \begin{footnotesize}
        \put(3,5){\color{black}{(f4)}}
      \end{footnotesize}
    \end{overpic}
  }
  \subfigure[]{
    \hspace{-0.33cm}
    \begin{overpic}[width=0.089\linewidth]{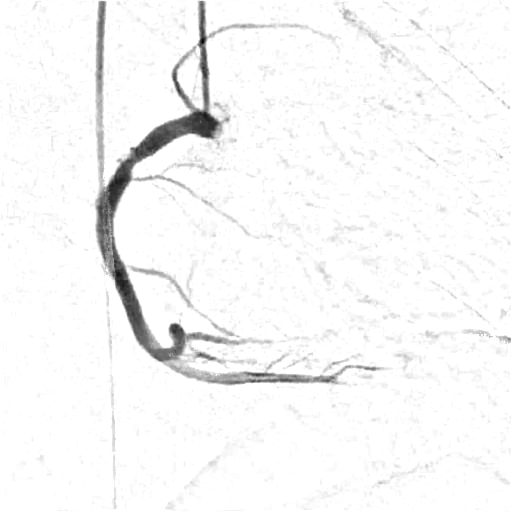}
      \begin{footnotesize}
        \put(3,5){\color{black}{(g4)}}
      \end{footnotesize}
    \end{overpic}
  }
  \subfigure[]{
    \hspace{-0.33cm}
    \begin{overpic}[width=0.089\linewidth]{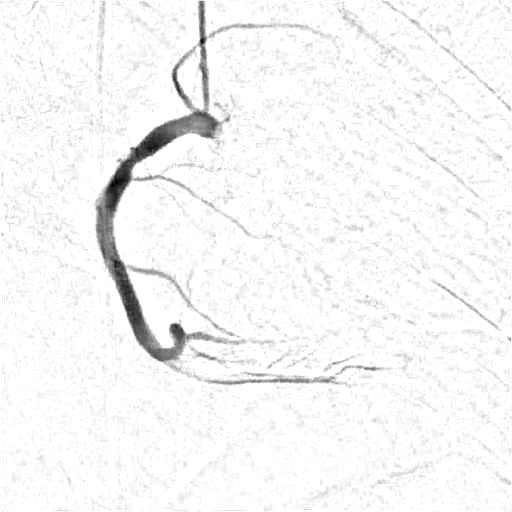}
      \begin{footnotesize}
        \put(3,5){\color{black}{(h4)}}
      \end{footnotesize}
    \end{overpic}
  }
  \subfigure[]{
    \hspace{-0.33cm}
    \begin{overpic}[width=0.089\linewidth]{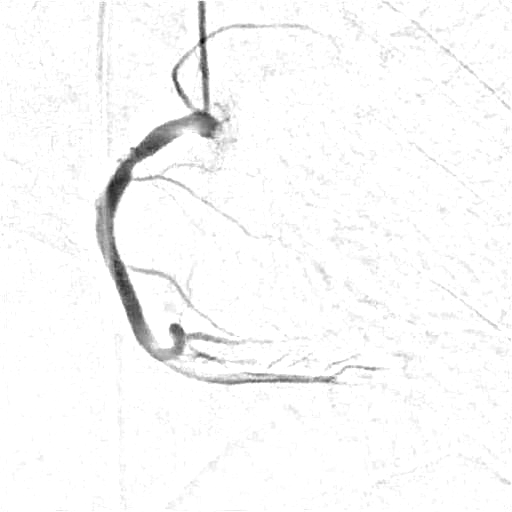}
      \begin{footnotesize}
        \put(3,5){\color{black}{(i4)}}
      \end{footnotesize}
    \end{overpic}
  }
  \subfigure[]{
    \hspace{-0.33cm}
    \begin{overpic}[width=0.089\linewidth]{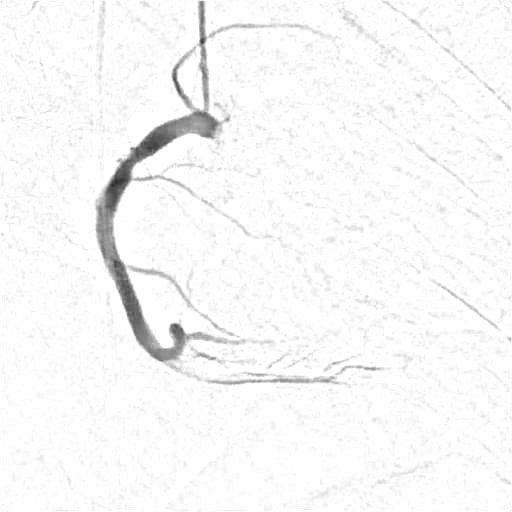}
      \begin{footnotesize}
        \put(3,5){\color{black}{(j4)}}
      \end{footnotesize}
    \end{overpic}
  }
  \subfigure[]{
    \hspace{-0.33cm}
    \begin{overpic}[width=0.089\linewidth]{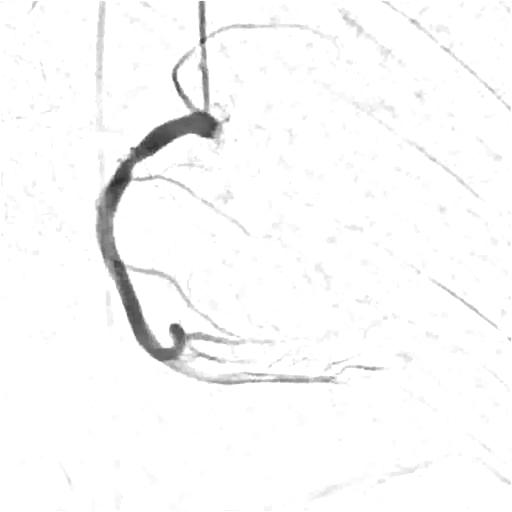}
      \begin{footnotesize}
        \put(3,5){\color{black}{(k4)}}
      \end{footnotesize}
    \end{overpic}
  }

  \label{fig7}
  \vspace{-0.8cm}
  \caption{Foreground extraction results of 4 XCA image sequences by 10 layer separation approaches. Each group of results contains a background layer image labeled by capital letters (B-K) and a vessel foreground layer image labeled by lowercase letters (b-k). (A1)-(A4) and (a1)-(a4) are the raw XCA images and their ground truth vessel masks selected by the surgeon. (B,b) DECOLOR. (C,c) GoDec. (D,d) PRMF. (E,e) AccAltProj. (F,f) Mog-RPCA. (G,g) MCR-RPCA. (H,h) ETRPCA. (I,i) KBR-RPCA. (J,j) TNN-TRPCA. (K,k) TV-TRPCA.}
\end{figure*}

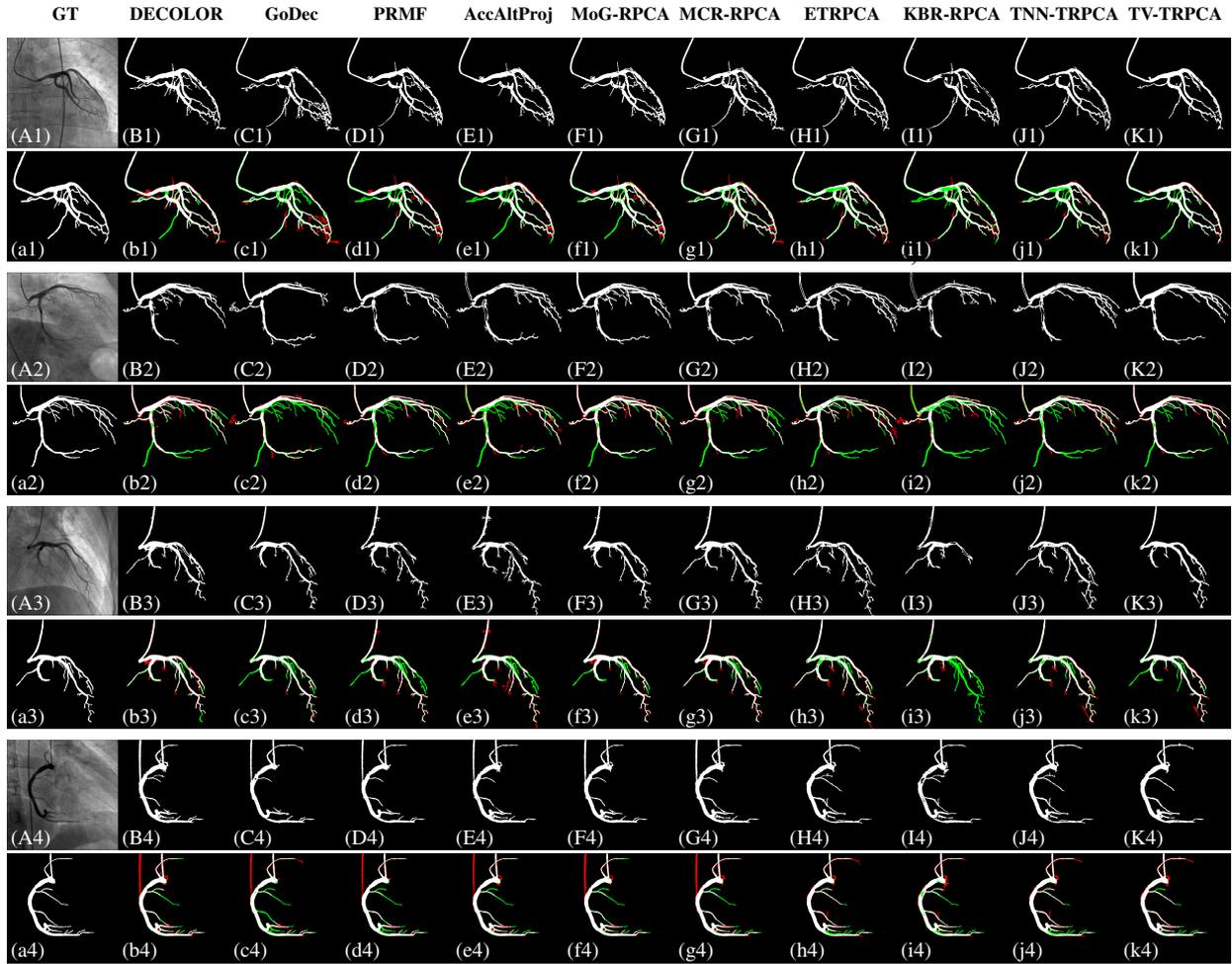
\begin{figure*}[!t]
  \centering
  \hspace{0.4cm}
  \begin{scriptsize}
    \textbf{GT} \ \ \ \ \ \ \ \ \ \ \textbf{DECOLOR} \ \ \ \ \ \ \ \ \textbf{GoDec} \ \ \ \ \ \ \ \ \ \ \ \textbf{PRMF} \ \ \ \ \ \ \ \textbf{AccAltProj} \ \ \ \textbf{MoG-RPCA} \ \textbf{MCR-RPCA} \ \ \ \ \textbf{ETRPCA} \ \ \ \ \textbf{KBR-RPCA} \ \textbf{TNN-TRPCA} \ \textbf{TV-TRPCA}
  \end{scriptsize}

  \subfigure[]{
    \begin{overpic}[width=0.089\linewidth]{Origin_0137.png}
      \begin{footnotesize}
        \put(3,5){\color{white}{(A1)}}
      \end{footnotesize}
    \end{overpic}
  }
  \subfigure[]{
    \hspace{-0.33cm}
    \begin{overpic}[width=0.089\linewidth]{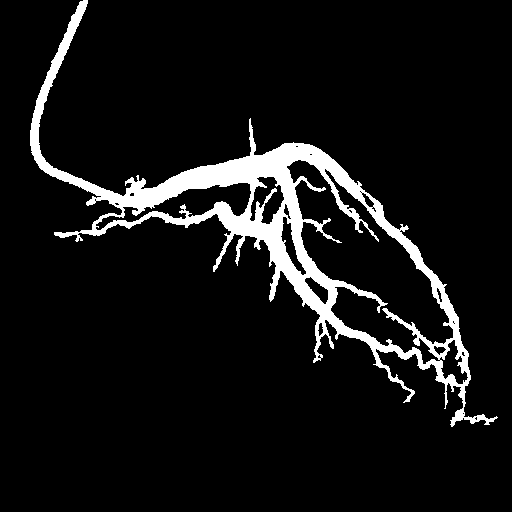}
      \begin{footnotesize}
        \put(3,5){\color{white}{(B1)}}
      \end{footnotesize}
    \end{overpic}
  }
  \subfigure[]{
    \hspace{-0.33cm}
    \begin{overpic}[width=0.089\linewidth]{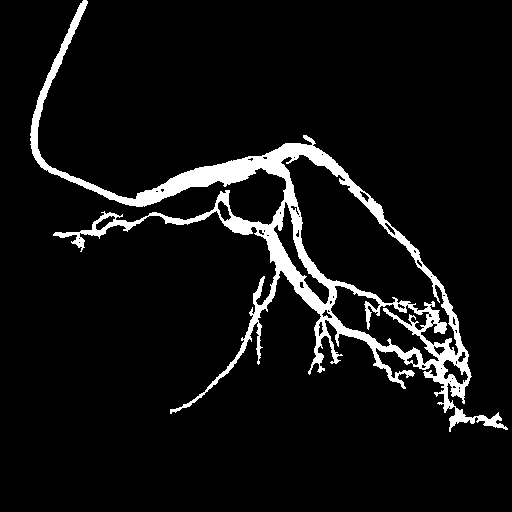}
      \begin{footnotesize}
        \put(3,5){\color{white}{(C1)}}
      \end{footnotesize}
    \end{overpic}
  }
  \subfigure[]{
    \hspace{-0.33cm}
    \begin{overpic}[width=0.089\linewidth]{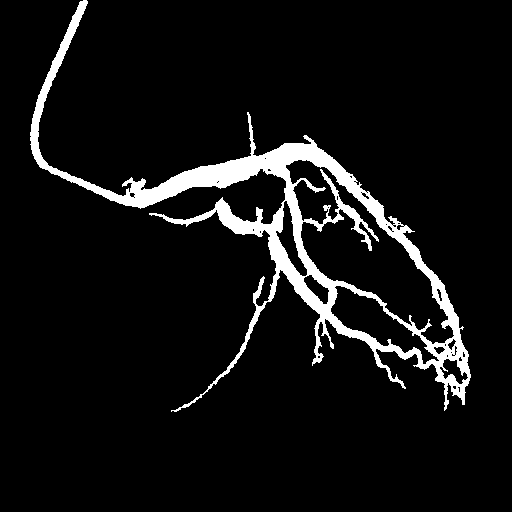}
      \begin{footnotesize}
        \put(3,5){\color{white}{(D1)}}
      \end{footnotesize}
    \end{overpic}
  }
  \subfigure[]{
    \hspace{-0.33cm}
    \begin{overpic}[width=0.089\linewidth]{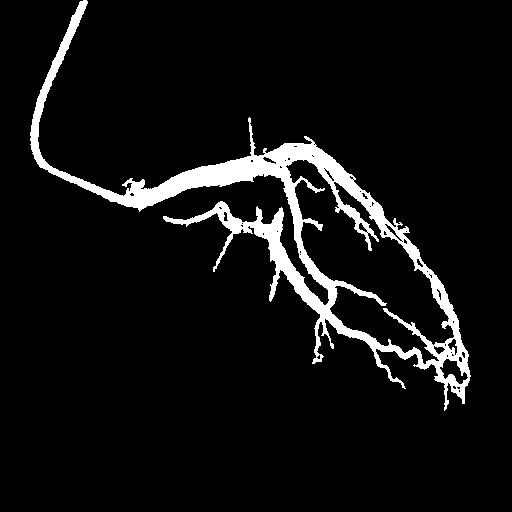}
      \begin{footnotesize}
        \put(3,5){\color{white}{(E1)}}
      \end{footnotesize}
    \end{overpic}
  }
  \subfigure[]{
    \hspace{-0.33cm}
    \begin{overpic}[width=0.089\linewidth]{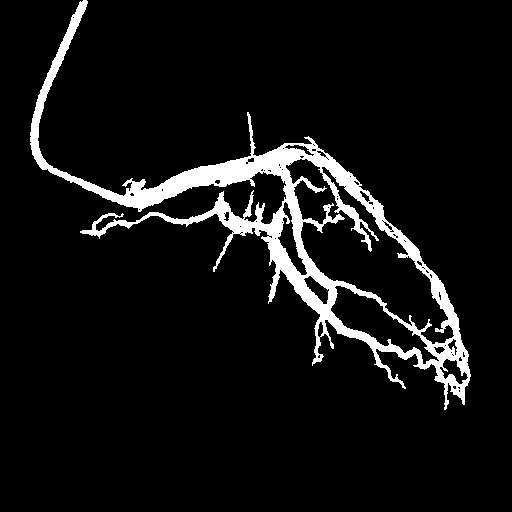}
      \begin{footnotesize}
        \put(3,5){\color{white}{(F1)}}
      \end{footnotesize}
    \end{overpic}
  }
  \subfigure[]{
    \hspace{-0.33cm}
    \begin{overpic}[width=0.089\linewidth]{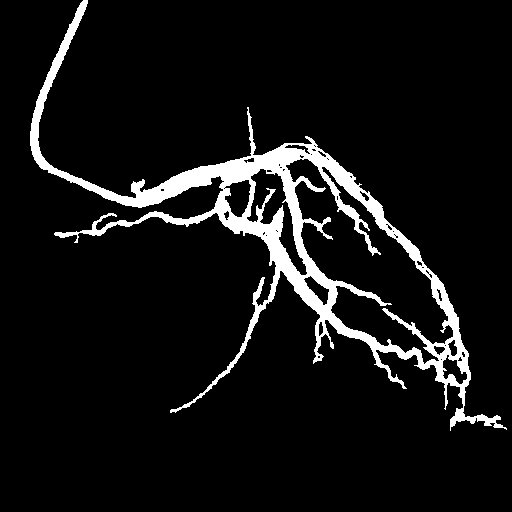}
      \begin{footnotesize}
        \put(3,5){\color{white}{(G1)}}
      \end{footnotesize}
    \end{overpic}
  }
  \subfigure[]{
    \hspace{-0.33cm}
    \begin{overpic}[width=0.089\linewidth]{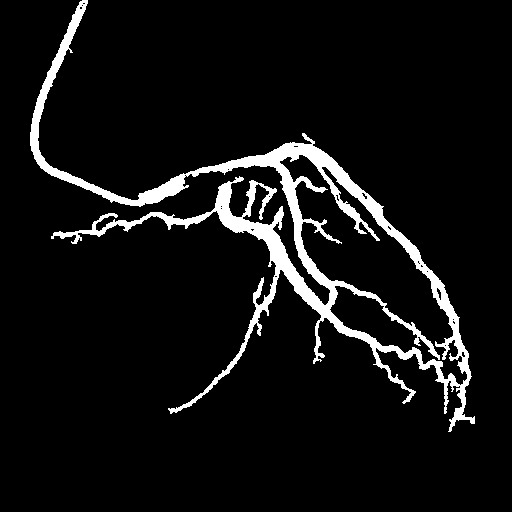}
      \begin{footnotesize}
        \put(3,5){\color{white}{(H1)}}
      \end{footnotesize}
    \end{overpic}
  }
  \subfigure[]{
    \hspace{-0.33cm}
    \begin{overpic}[width=0.089\linewidth]{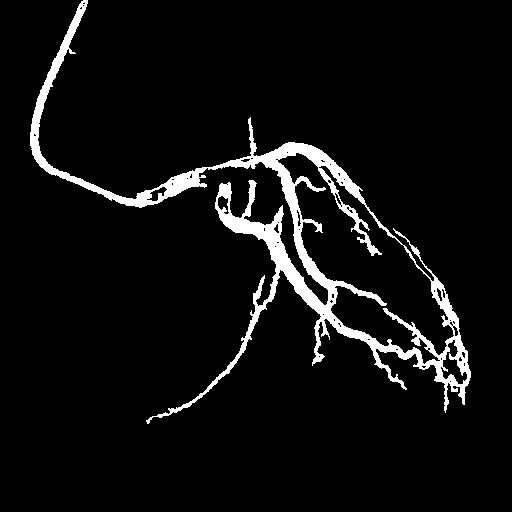}
      \begin{footnotesize}
        \put(3,5){\color{white}{(I1)}}
      \end{footnotesize}
    \end{overpic}
  }
  \subfigure[]{
    \hspace{-0.33cm}
    \begin{overpic}[width=0.089\linewidth]{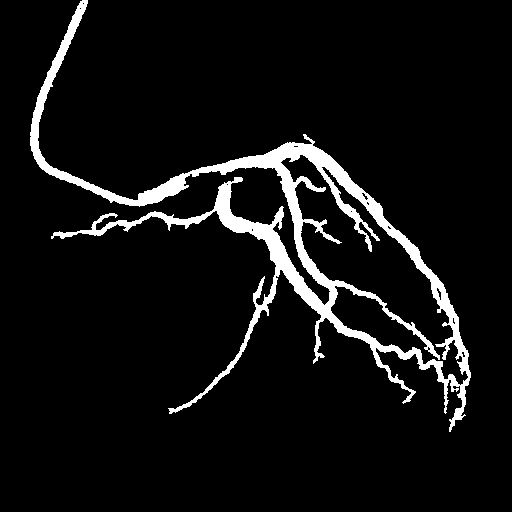}
      \begin{footnotesize}
        \put(3,5){\color{white}{(J1)}}
      \end{footnotesize}
    \end{overpic}
  }
  \subfigure[]{
    \hspace{-0.33cm}
    \begin{overpic}[width=0.089\linewidth]{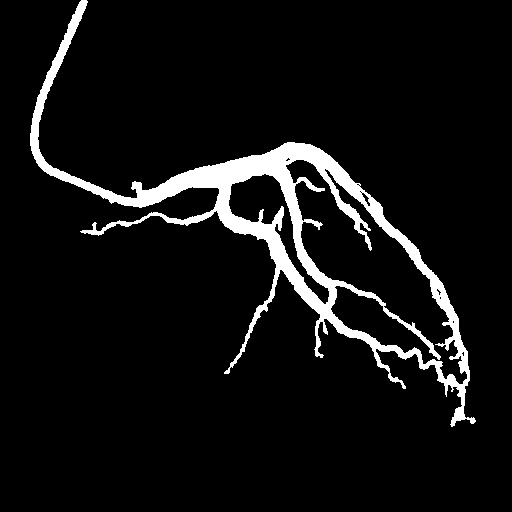}
      \begin{footnotesize}
        \put(3,5){\color{white}{(K1)}}
      \end{footnotesize}
    \end{overpic}
  }

  \vspace{-0.95cm}
  \subfigure[]{
    \begin{overpic}[width=0.089\linewidth]{GTBW_0137.png}
      \begin{footnotesize}
        \put(3,5){\color{white}{(a1)}}
      \end{footnotesize}
    \end{overpic}
  }
  \subfigure[]{
    \hspace{-0.33cm}
    \begin{overpic}[width=0.089\linewidth]{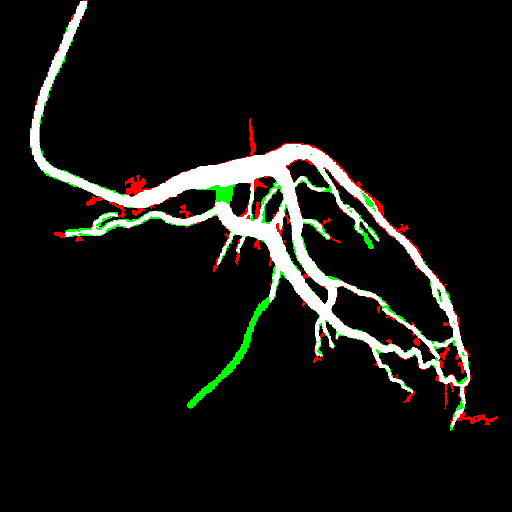}
      \begin{footnotesize}
        \put(3,5){\color{white}{(b1)}}
      \end{footnotesize}
    \end{overpic}
  }
  \subfigure[]{
    \hspace{-0.33cm}
    \begin{overpic}[width=0.089\linewidth]{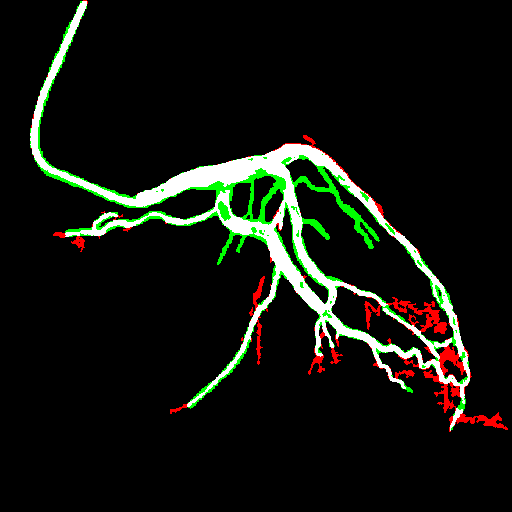}
      \begin{footnotesize}
        \put(3,5){\color{white}{(c1)}}
      \end{footnotesize}
    \end{overpic}
  }
  \subfigure[]{
    \hspace{-0.33cm}
    \begin{overpic}[width=0.089\linewidth]{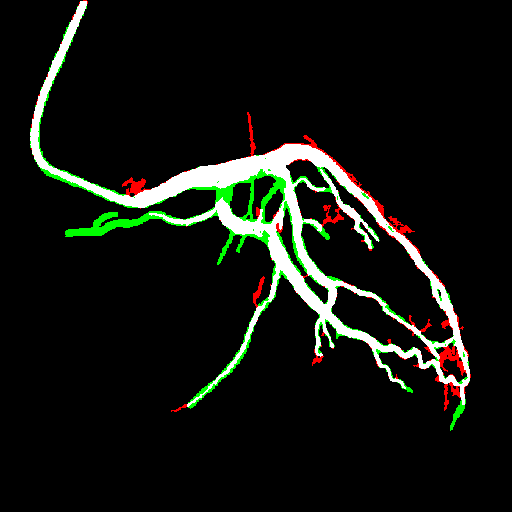}
      \begin{footnotesize}
        \put(3,5){\color{white}{(d1)}}
      \end{footnotesize}
    \end{overpic}
  }
  \subfigure[]{
    \hspace{-0.33cm}
    \begin{overpic}[width=0.089\linewidth]{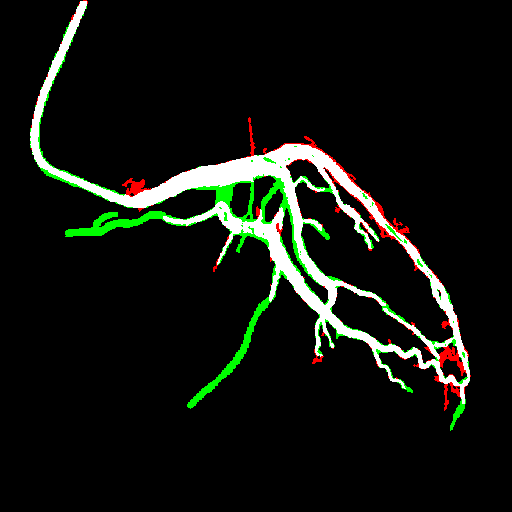}
      \begin{footnotesize}
        \put(3,5){\color{white}{(e1)}}
      \end{footnotesize}
    \end{overpic}
  }
  \subfigure[]{
    \hspace{-0.33cm}
    \begin{overpic}[width=0.089\linewidth]{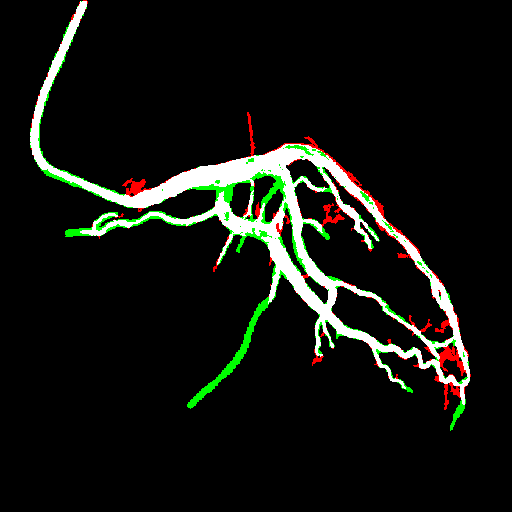}
      \begin{footnotesize}
        \put(3,5){\color{white}{(f1)}}
      \end{footnotesize}
    \end{overpic}
  }
  \subfigure[]{
    \hspace{-0.33cm}
    \begin{overpic}[width=0.089\linewidth]{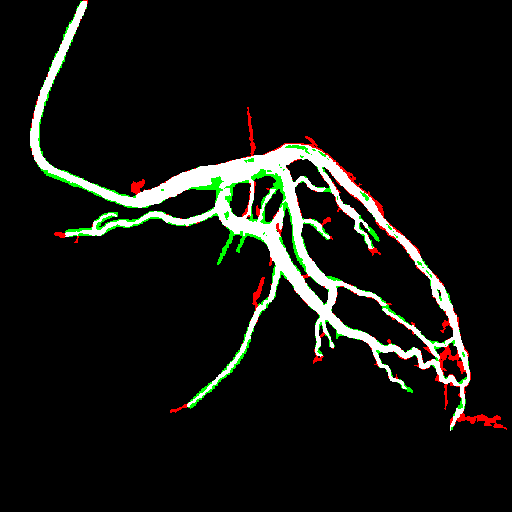}
      \begin{footnotesize}
        \put(3,5){\color{white}{(g1)}}
      \end{footnotesize}
    \end{overpic}
  }
  \subfigure[]{
    \hspace{-0.33cm}
    \begin{overpic}[width=0.089\linewidth]{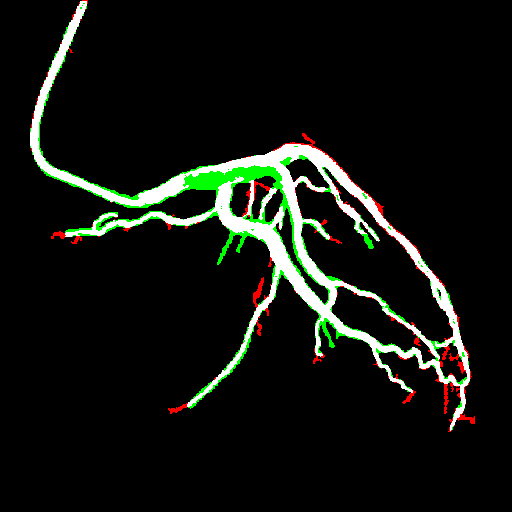}
      \begin{footnotesize}
        \put(3,5){\color{white}{(h1)}}
      \end{footnotesize}
    \end{overpic}
  }
  \subfigure[]{
    \hspace{-0.33cm}
    \begin{overpic}[width=0.089\linewidth]{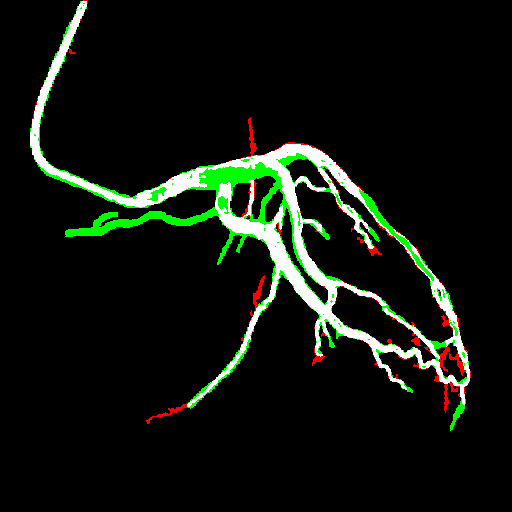}
      \begin{footnotesize}
        \put(3,5){\color{white}{(i1)}}
      \end{footnotesize}2)
    \end{overpic}
  }
  \subfigure[]{
    \hspace{-0.33cm}
    \begin{overpic}[width=0.089\linewidth]{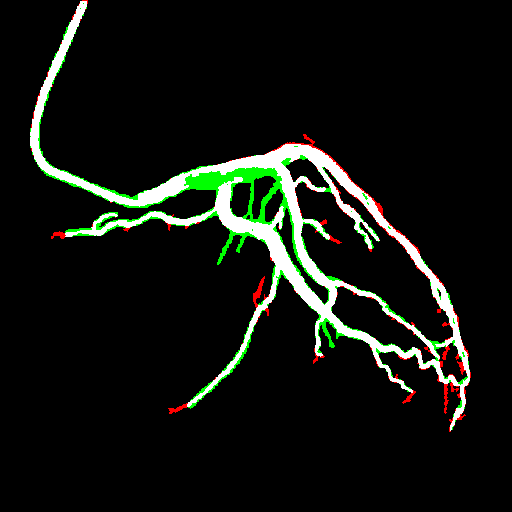}
      \begin{footnotesize}
        \put(3,5){\color{white}{(j1)}}
      \end{footnotesize}
    \end{overpic}
  }
  \subfigure[]{
    \hspace{-0.33cm}
    \begin{overpic}[width=0.089\linewidth]{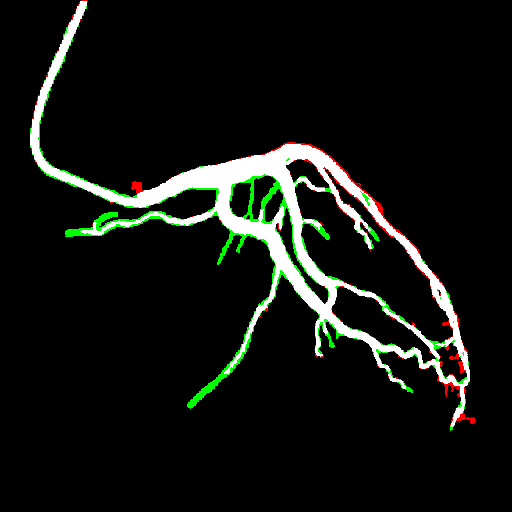}
      \begin{footnotesize}
        \put(3,5){\color{white}{(k1)}}
      \end{footnotesize}
    \end{overpic}
  }

  \vspace{-0.85cm}
  \subfigure[]{
    \begin{overpic}[width=0.089\linewidth]{Origin_0449.png}
      \begin{footnotesize}
        \put(3,5){\color{white}{(A2)}}
      \end{footnotesize}
    \end{overpic}
  }
  \subfigure[]{
    \hspace{-0.33cm}
    \begin{overpic}[width=0.089\linewidth]{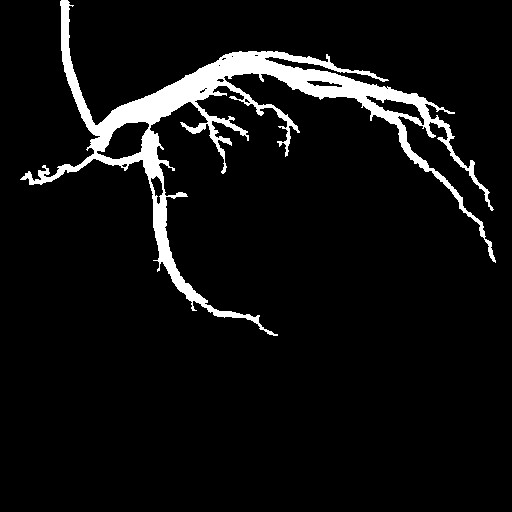}
      \begin{footnotesize}
        \put(3,5){\color{white}{(B2)}}
      \end{footnotesize}
    \end{overpic}
  }
  \subfigure[]{
    \hspace{-0.33cm}
    \begin{overpic}[width=0.089\linewidth]{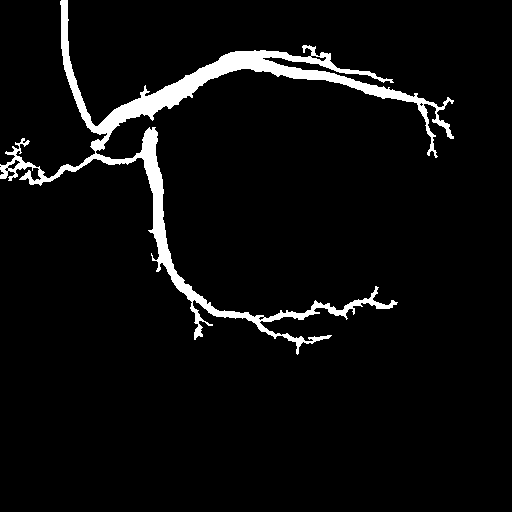}
      \begin{footnotesize}
        \put(3,5){\color{white}{(C2)}}
      \end{footnotesize}
    \end{overpic}
  }
  \subfigure[]{
    \hspace{-0.33cm}
    \begin{overpic}[width=0.089\linewidth]{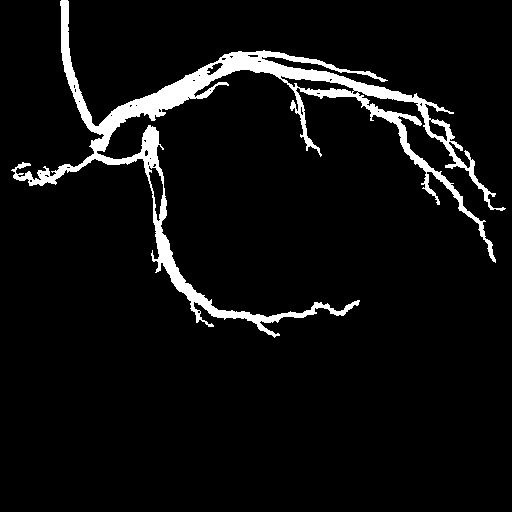}
      \begin{footnotesize}
        \put(3,5){\color{white}{(D2)}}
      \end{footnotesize}
    \end{overpic}
  }
  \subfigure[]{
    \hspace{-0.33cm}
    \begin{overpic}[width=0.089\linewidth]{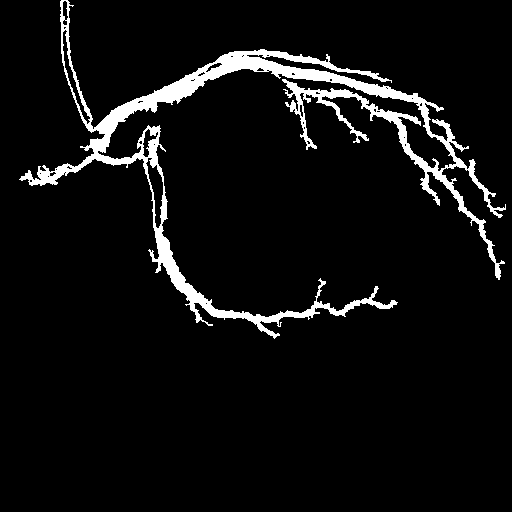}
      \begin{footnotesize}
        \put(3,5){\color{white}{(E2)}}
      \end{footnotesize}
    \end{overpic}
  }
  \subfigure[]{
    \hspace{-0.33cm}
    \begin{overpic}[width=0.089\linewidth]{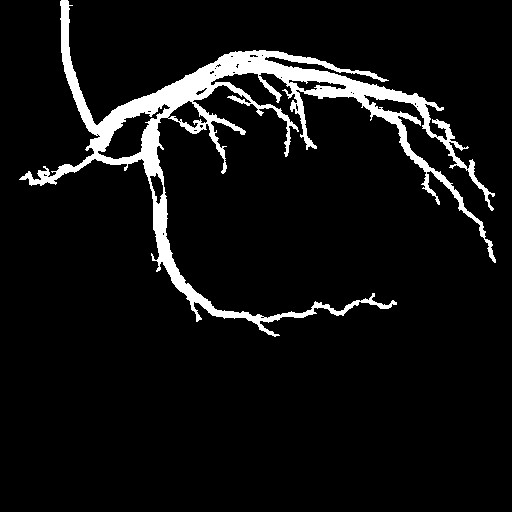}
      \begin{footnotesize}
        \put(3,5){\color{white}{(F2)}}
      \end{footnotesize}
    \end{overpic}
  }
  \subfigure[]{
    \hspace{-0.33cm}
    \begin{overpic}[width=0.089\linewidth]{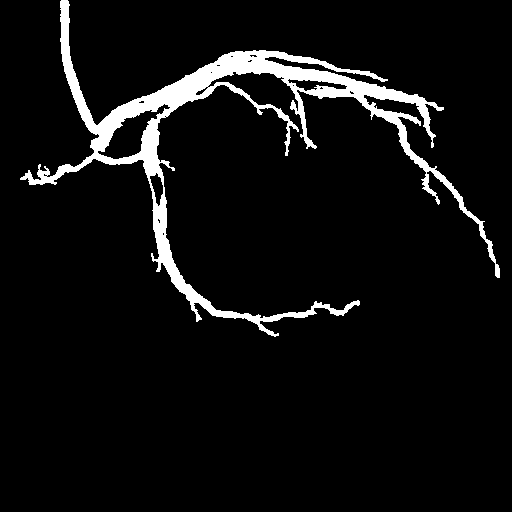}
      \begin{footnotesize}
        \put(3,5){\color{white}{(G2)}}
      \end{footnotesize}
    \end{overpic}
  }
  \subfigure[]{
    \hspace{-0.33cm}
    \begin{overpic}[width=0.089\linewidth]{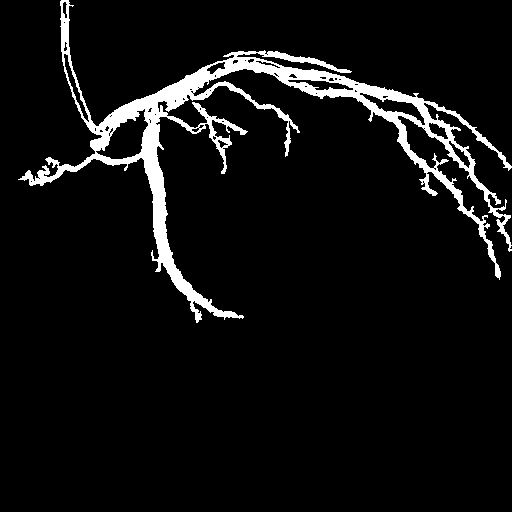}
      \begin{footnotesize}
        \put(3,5){\color{white}{(H2)}}
      \end{footnotesize}
    \end{overpic}
  }
  \subfigure[]{
    \hspace{-0.33cm}
    \begin{overpic}[width=0.089\linewidth]{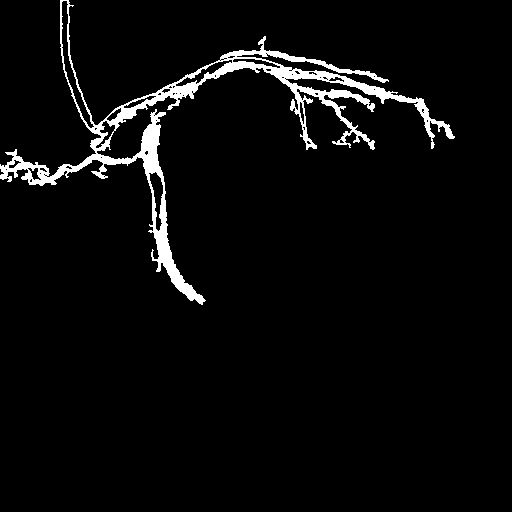}
      \begin{footnotesize}
        \put(3,5){\color{white}{(I2)}}
      \end{footnotesize}
    \end{overpic}
  }
  \subfigure[]{
    \hspace{-0.33cm}
    \begin{overpic}[width=0.089\linewidth]{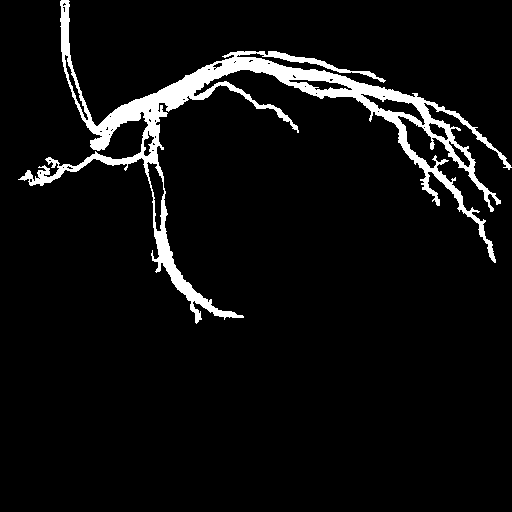}
      \begin{footnotesize}
        \put(3,5){\color{white}{(J2)}}
      \end{footnotesize}
    \end{overpic}
  }
  \subfigure[]{
    \hspace{-0.33cm}
    \begin{overpic}[width=0.089\linewidth]{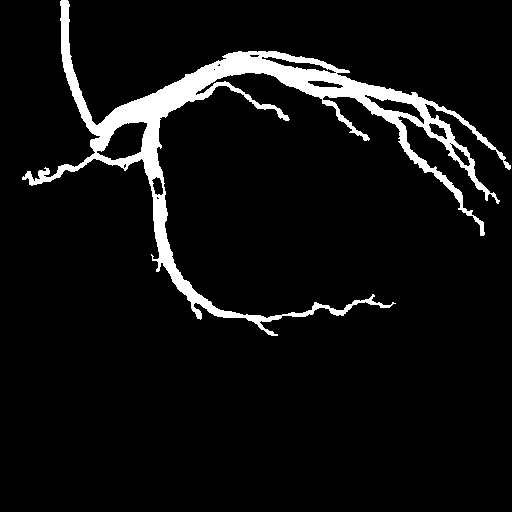}
      \begin{footnotesize}
        \put(3,5){\color{white}{(K2)}}
      \end{footnotesize}
    \end{overpic}
  }

  \vspace{-0.95cm}
  \subfigure[]{
    \begin{overpic}[width=0.089\linewidth]{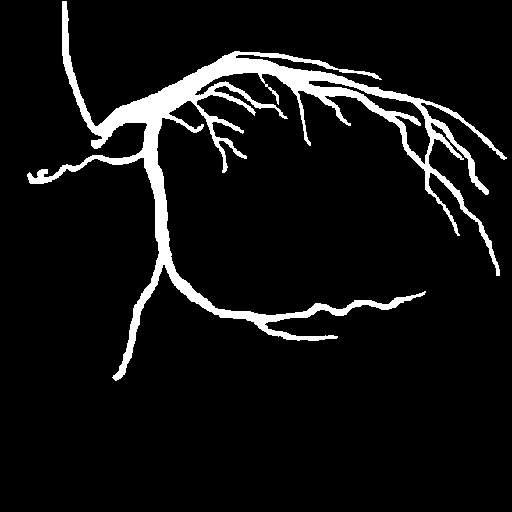}
      \begin{footnotesize}
        \put(3,5){\color{white}{(a2)}}
      \end{footnotesize}
    \end{overpic}
  }
  \subfigure[]{
    \hspace{-0.33cm}
    \begin{overpic}[width=0.089\linewidth]{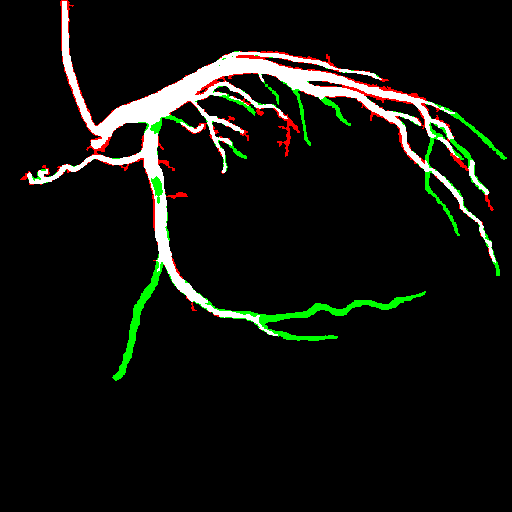}
      \begin{footnotesize}
        \put(3,5){\color{white}{(b2)}}
      \end{footnotesize}
    \end{overpic}
  }
  \subfigure[]{
    \hspace{-0.33cm}
    \begin{overpic}[width=0.089\linewidth]{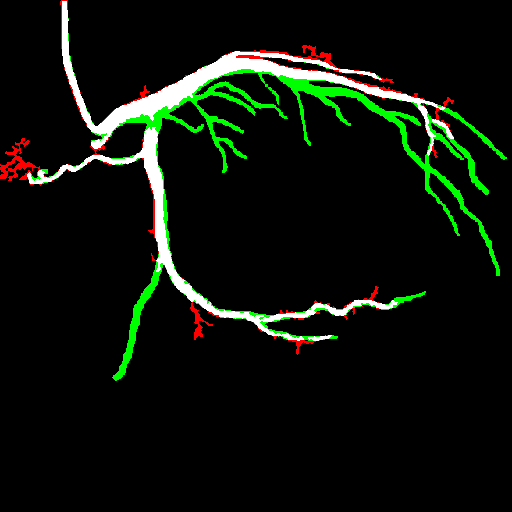}
      \begin{footnotesize}
        \put(3,5){\color{white}{(c2)}}
      \end{footnotesize}
    \end{overpic}
  }
  \subfigure[]{
    \hspace{-0.33cm}
    \begin{overpic}[width=0.089\linewidth]{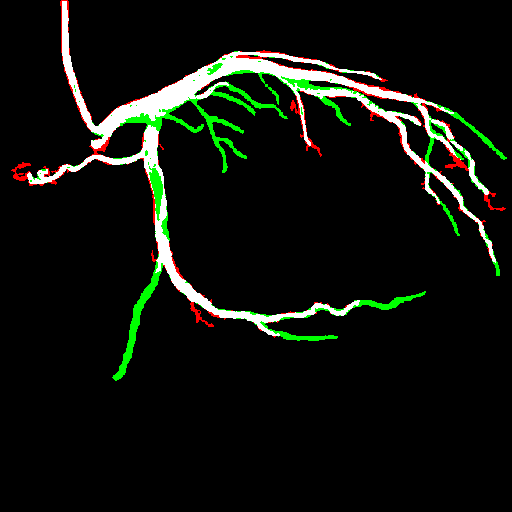}
      \begin{footnotesize}
        \put(3,5){\color{white}{(d2)}}
      \end{footnotesize}
    \end{overpic}
  }
  \subfigure[]{
    \hspace{-0.33cm}
    \begin{overpic}[width=0.089\linewidth]{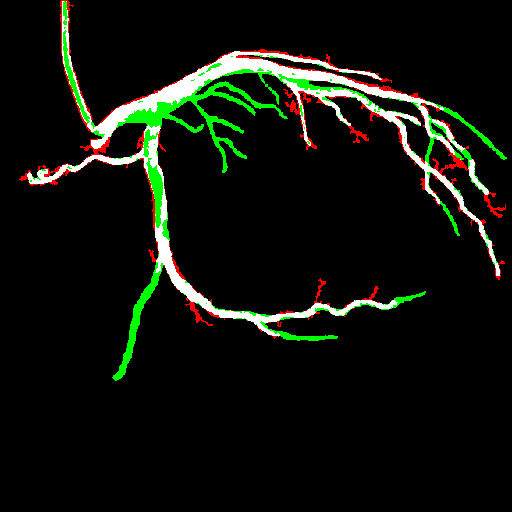}
      \begin{footnotesize}
        \put(3,5){\color{white}{(e2)}}
      \end{footnotesize}
    \end{overpic}
  }
  \subfigure[]{
    \hspace{-0.33cm}
    \begin{overpic}[width=0.089\linewidth]{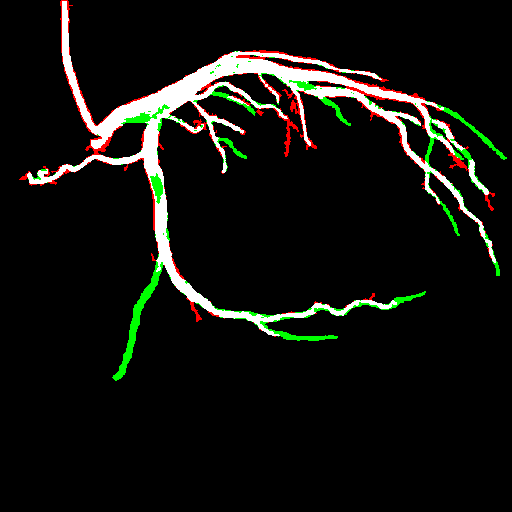}
      \begin{footnotesize}
        \put(3,5){\color{white}{(f2)}}
      \end{footnotesize}
    \end{overpic}
  }
  \subfigure[]{
    \hspace{-0.33cm}
    \begin{overpic}[width=0.089\linewidth]{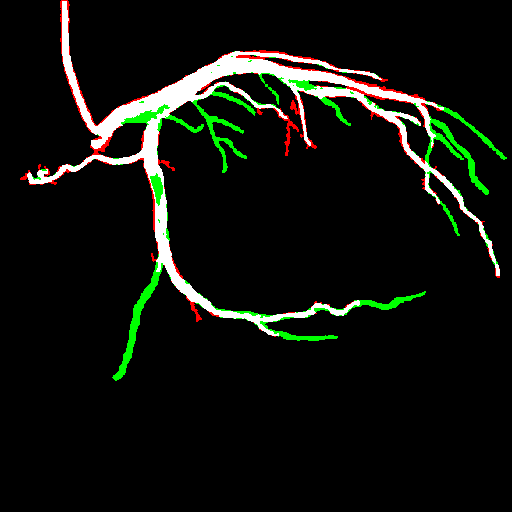}
      \begin{footnotesize}
        \put(3,5){\color{white}{(g2)}}
      \end{footnotesize}
    \end{overpic}
  }
  \subfigure[]{
    \hspace{-0.33cm}
    \begin{overpic}[width=0.089\linewidth]{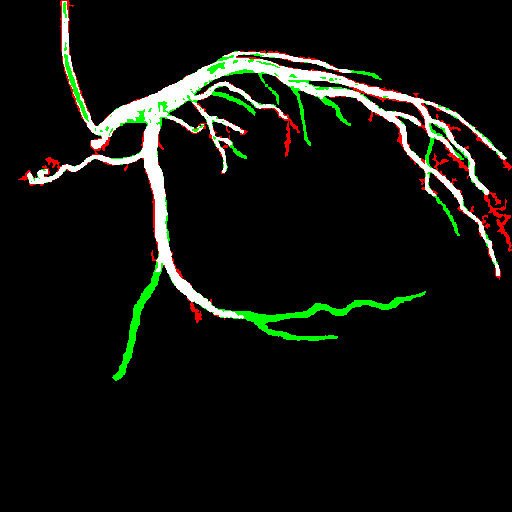}
      \begin{footnotesize}
        \put(3,5){\color{white}{(h2)}}
      \end{footnotesize}
    \end{overpic}
  }
  \subfigure[]{
    \hspace{-0.33cm}
    \begin{overpic}[width=0.089\linewidth]{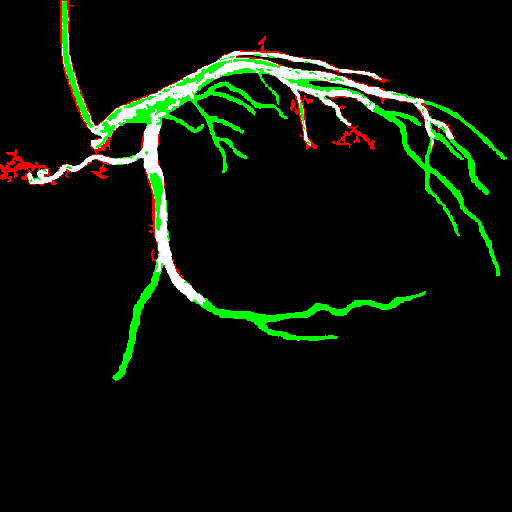}
      \begin{footnotesize}
        \put(3,5){\color{white}{(i2)}}
      \end{footnotesize}
    \end{overpic}
  }
  \subfigure[]{
    \hspace{-0.33cm}
    \begin{overpic}[width=0.089\linewidth]{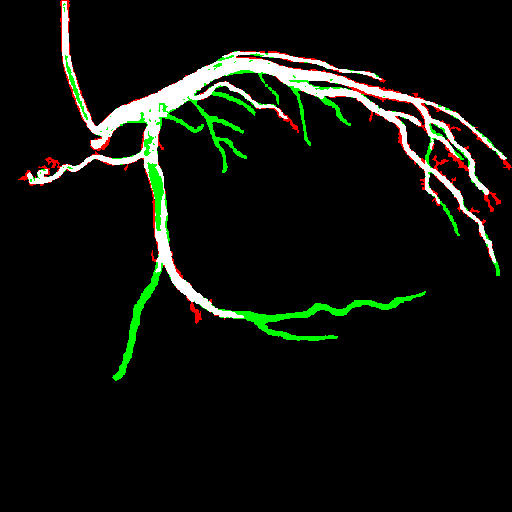}
      \begin{footnotesize}
        \put(3,5){\color{white}{(j2)}}
      \end{footnotesize}
    \end{overpic}
  }
  \subfigure[]{
    \hspace{-0.33cm}
    \begin{overpic}[width=0.089\linewidth]{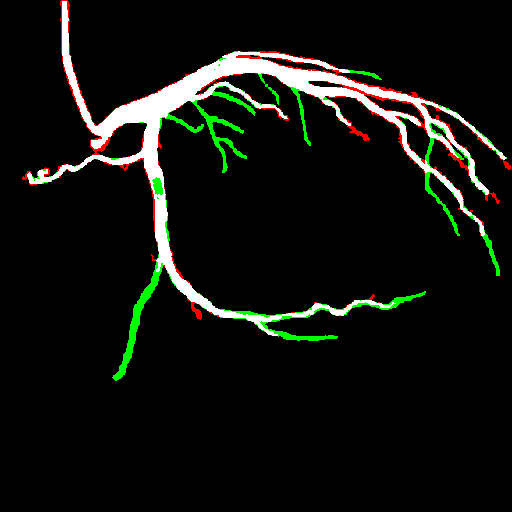}
      \begin{footnotesize}
        \put(3,5){\color{white}{(k2)}}
      \end{footnotesize}
    \end{overpic}
  }

  \vspace{-0.85cm}
  \subfigure[]{
    \begin{overpic}[width=0.089\linewidth]{Origin_0544.png}
      \begin{footnotesize}
        \put(3,5){\color{white}{(A3)}}
      \end{footnotesize}
    \end{overpic}
  }
  \subfigure[]{
    \hspace{-0.33cm}
    \begin{overpic}[width=0.089\linewidth]{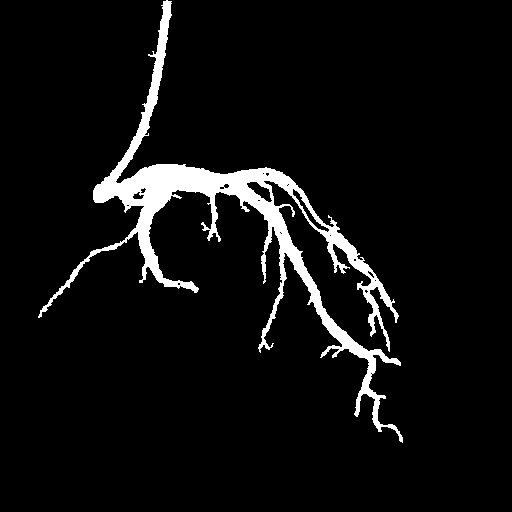}
      \begin{footnotesize}
        \put(3,5){\color{white}{(B3)}}
      \end{footnotesize}
    \end{overpic}
  }
  \subfigure[]{
    \hspace{-0.33cm}
    \begin{overpic}[width=0.089\linewidth]{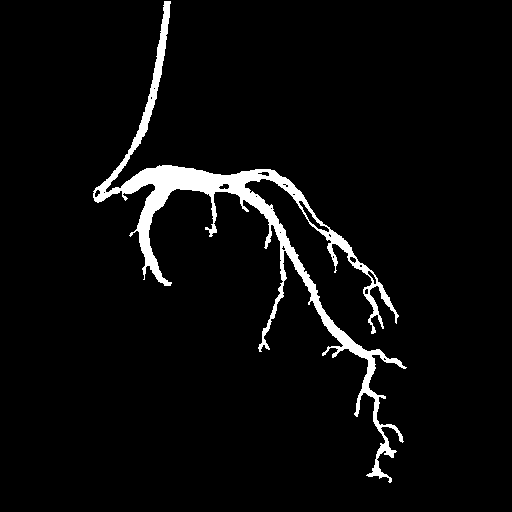}
      \begin{footnotesize}
        \put(3,5){\color{white}{(C3)}}
      \end{footnotesize}
    \end{overpic}
  }
  \subfigure[]{
    \hspace{-0.33cm}
    \begin{overpic}[width=0.089\linewidth]{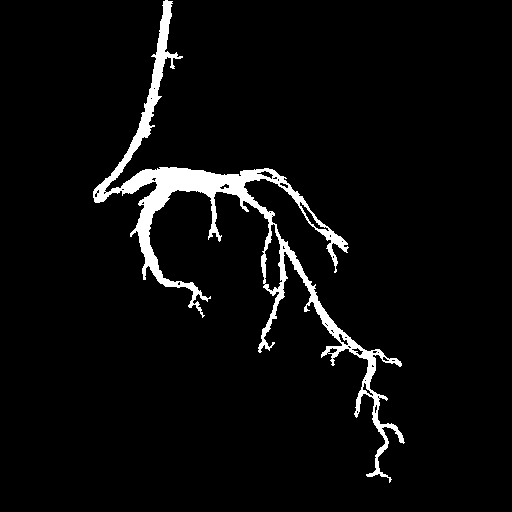}
      \begin{footnotesize}
        \put(3,5){\color{white}{(D3)}}
      \end{footnotesize}
    \end{overpic}
  }
  \subfigure[]{
    \hspace{-0.33cm}
    \begin{overpic}[width=0.089\linewidth]{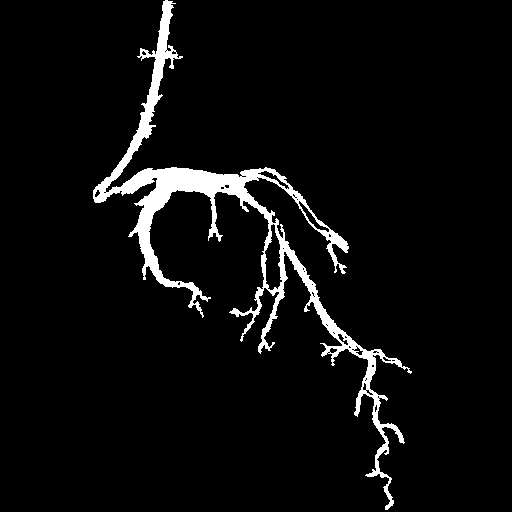}
      \begin{footnotesize}
        \put(3,5){\color{white}{(E3)}}
      \end{footnotesize}
    \end{overpic}
  }
  \subfigure[]{
    \hspace{-0.33cm}
    \begin{overpic}[width=0.089\linewidth]{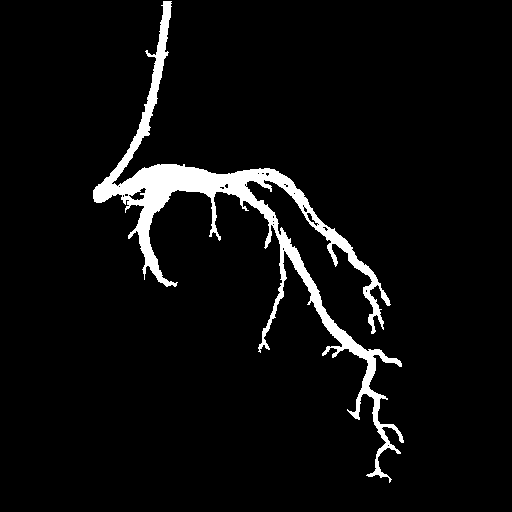}
      \begin{footnotesize}
        \put(3,5){\color{white}{(F3)}}
      \end{footnotesize}
    \end{overpic}
  }
  \subfigure[]{
    \hspace{-0.33cm}
    \begin{overpic}[width=0.089\linewidth]{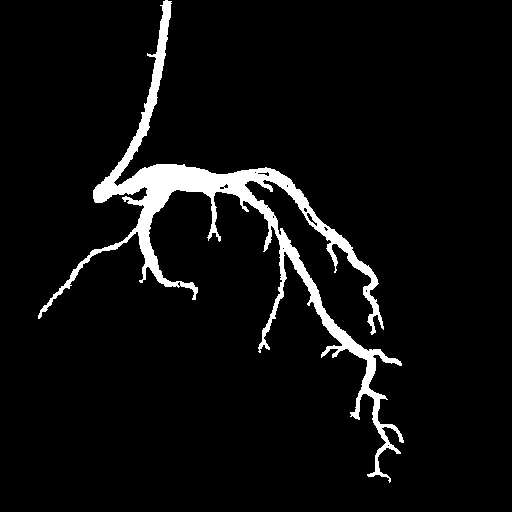}
      \begin{footnotesize}
        \put(3,5){\color{white}{(G3)}}
      \end{footnotesize}
    \end{overpic}
  }
  \subfigure[]{
    \hspace{-0.33cm}
    \begin{overpic}[width=0.089\linewidth]{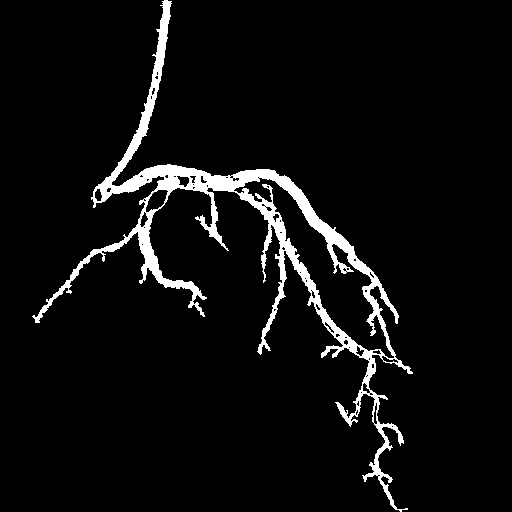}
      \begin{footnotesize}
        \put(3,5){\color{white}{(H3)}}
      \end{footnotesize}
    \end{overpic}
  }
  \subfigure[]{
    \hspace{-0.33cm}
    \begin{overpic}[width=0.089\linewidth]{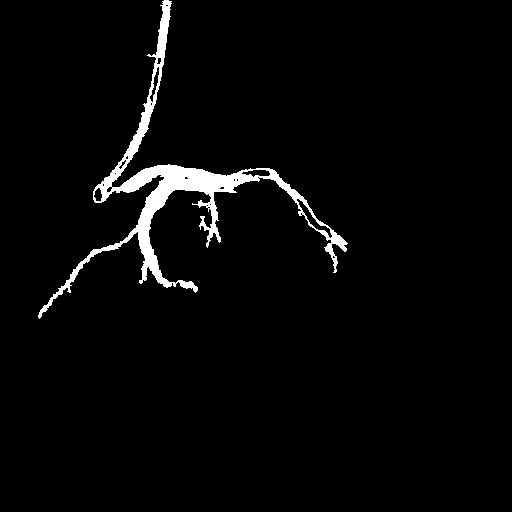}
      \begin{footnotesize}
        \put(3,5){\color{white}{(I3)}}
      \end{footnotesize}
    \end{overpic}
  }
  \subfigure[]{
    \hspace{-0.33cm}
    \begin{overpic}[width=0.089\linewidth]{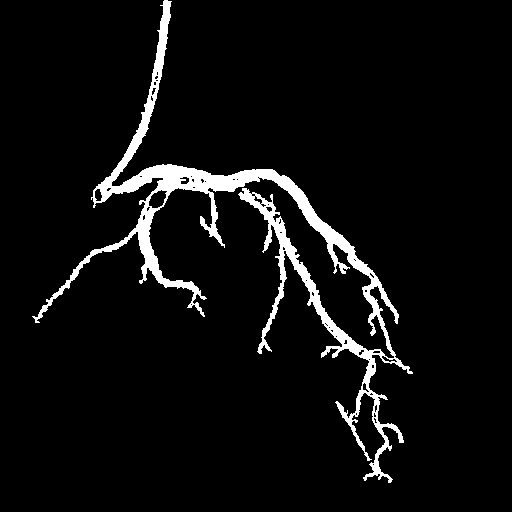}
      \begin{footnotesize}
        \put(3,5){\color{white}{(J3)}}
      \end{footnotesize}
    \end{overpic}
  }
  \subfigure[]{
    \hspace{-0.33cm}
    \begin{overpic}[width=0.089\linewidth]{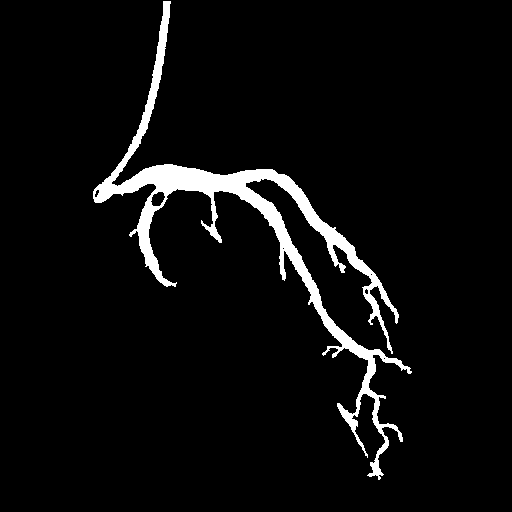}
      \begin{footnotesize}
        \put(3,5){\color{white}{(K3)}}
      \end{footnotesize}
    \end{overpic}
  }

  \vspace{-0.95cm}
  \subfigure[]{
    \begin{overpic}[width=0.089\linewidth]{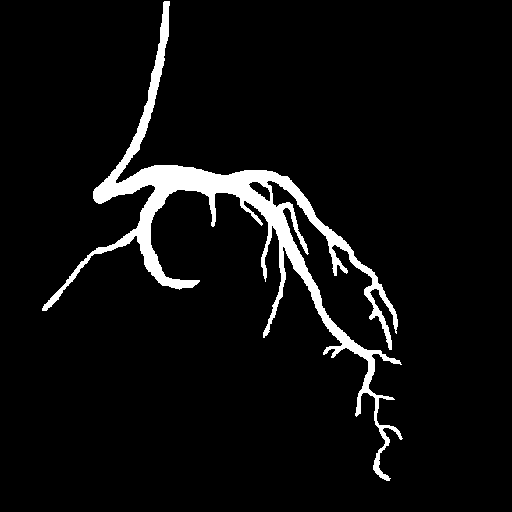}
      \begin{footnotesize}
        \put(3,5){\color{white}{(a3)}}
      \end{footnotesize}
    \end{overpic}
  }
  \subfigure[]{
    \hspace{-0.33cm}
    \begin{overpic}[width=0.089\linewidth]{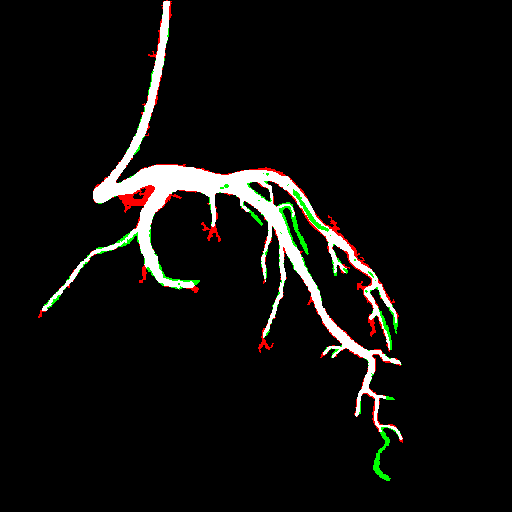}
      \begin{footnotesize}
        \put(3,5){\color{white}{(b3)}}
      \end{footnotesize}
    \end{overpic}
  }
  \subfigure[]{
    \hspace{-0.33cm}
    \begin{overpic}[width=0.089\linewidth]{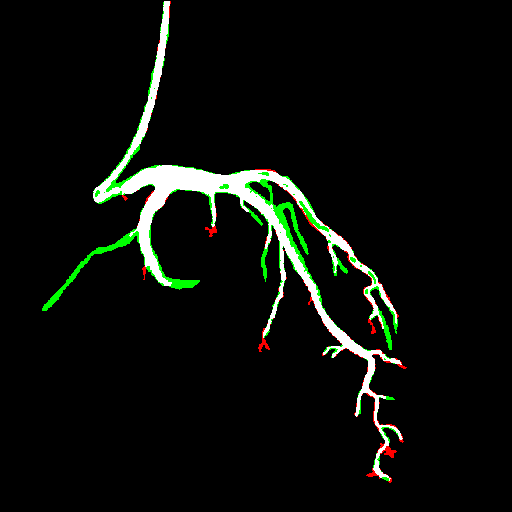}
      \begin{footnotesize}
        \put(3,5){\color{white}{(c3)}}
      \end{footnotesize}
    \end{overpic}
  }
  \subfigure[]{
    \hspace{-0.33cm}
    \begin{overpic}[width=0.089\linewidth]{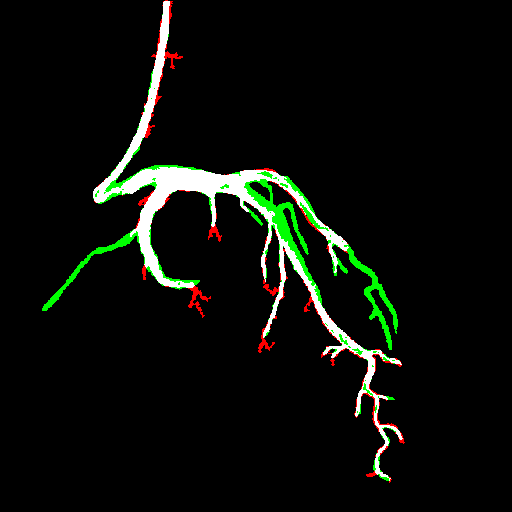}
      \begin{footnotesize}
        \put(3,5){\color{white}{(d3)}}
      \end{footnotesize}
    \end{overpic}
  }
  \subfigure[]{
    \hspace{-0.33cm}
    \begin{overpic}[width=0.089\linewidth]{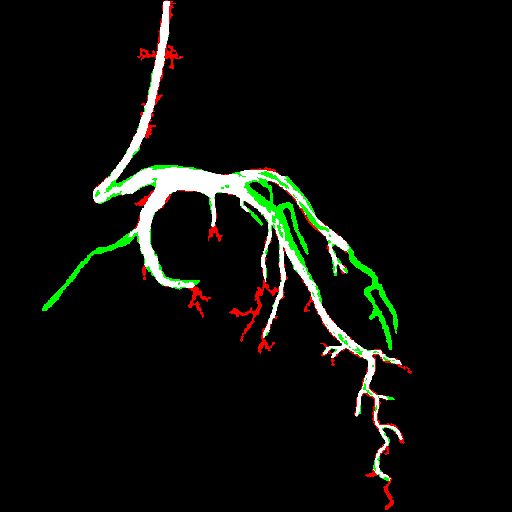}
      \begin{footnotesize}
        \put(3,5){\color{white}{(e3)}}
      \end{footnotesize}
    \end{overpic}
  }
  \subfigure[]{
    \hspace{-0.33cm}
    \begin{overpic}[width=0.089\linewidth]{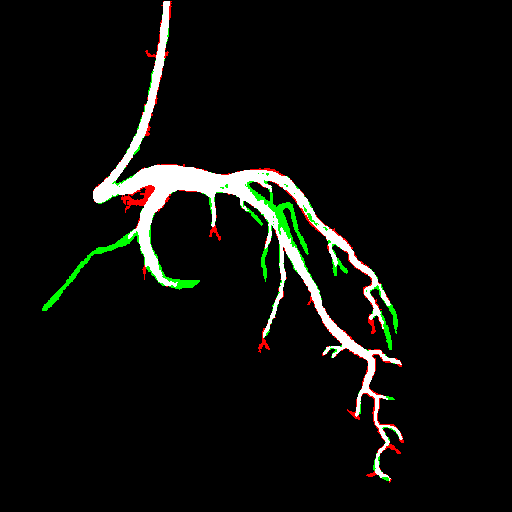}
      \begin{footnotesize}
        \put(3,5){\color{white}{(f3)}}
      \end{footnotesize}
    \end{overpic}
  }
  \subfigure[]{
    \hspace{-0.33cm}
    \begin{overpic}[width=0.089\linewidth]{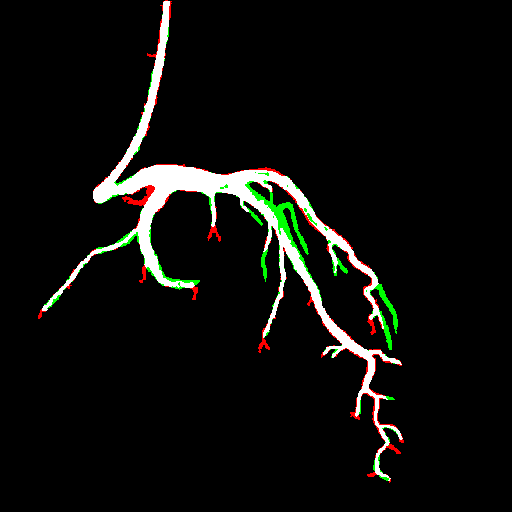}
      \begin{footnotesize}
        \put(3,5){\color{white}{(g3)}}
      \end{footnotesize}
    \end{overpic}
  }
  \subfigure[]{
    \hspace{-0.33cm}
    \begin{overpic}[width=0.089\linewidth]{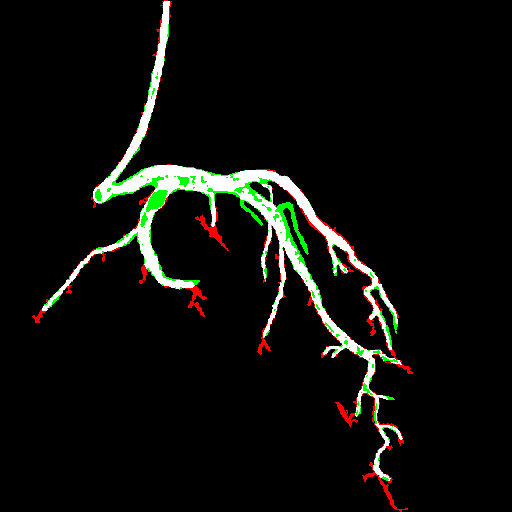}
      \begin{footnotesize}
        \put(3,5){\color{white}{(h3)}}
      \end{footnotesize}
    \end{overpic}
  }
  \subfigure[]{
    \hspace{-0.33cm}
    \begin{overpic}[width=0.089\linewidth]{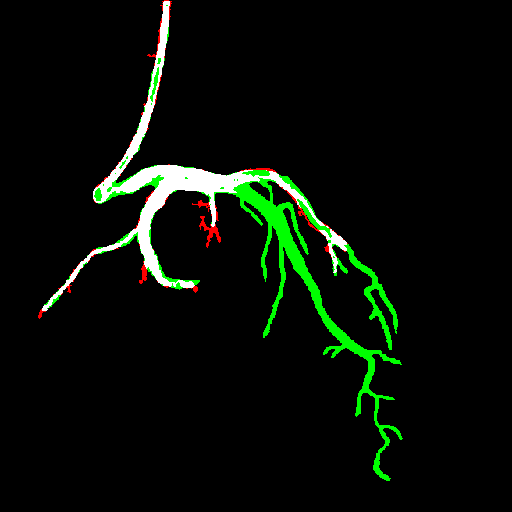}
      \begin{footnotesize}
        \put(3,5){\color{white}{(i3)}}
      \end{footnotesize}
    \end{overpic}
  }
  \subfigure[]{
    \hspace{-0.33cm}
    \begin{overpic}[width=0.089\linewidth]{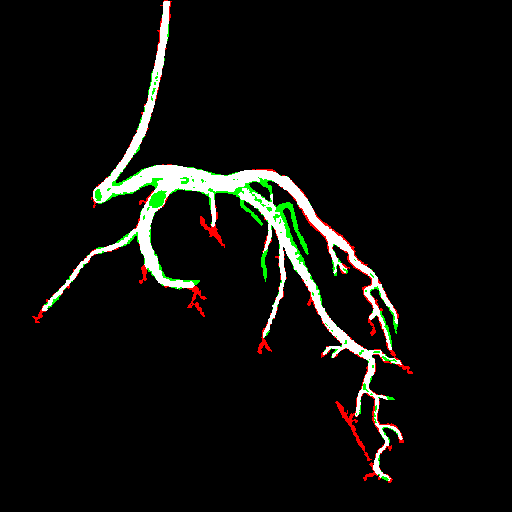}
      \begin{footnotesize}
        \put(3,5){\color{white}{(j3)}}
      \end{footnotesize}
    \end{overpic}
  }
  \subfigure[]{
    \hspace{-0.33cm}
    \begin{overpic}[width=0.089\linewidth]{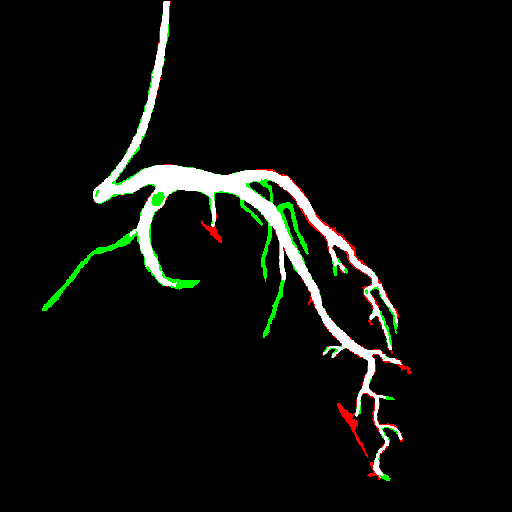}
      \begin{footnotesize}
        \put(3,5){\color{white}{(k3)}}
      \end{footnotesize}
    \end{overpic}
  }

  \vspace{-0.85cm}
  \subfigure[]{
    \begin{overpic}[width=0.089\linewidth]{Origin_0747.png}
      \begin{footnotesize}
        \put(3,5){\color{white}{(A4)}}
      \end{footnotesize}
    \end{overpic}
  }
  \subfigure[]{
    \hspace{-0.33cm}
    \begin{overpic}[width=0.089\linewidth]{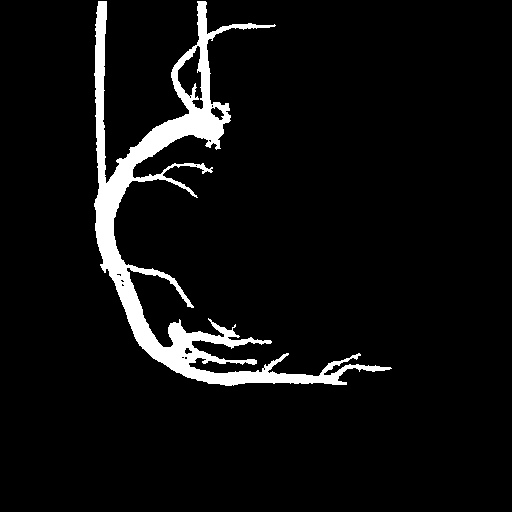}
      \begin{footnotesize}
        \put(3,5){\color{white}{(B4)}}
      \end{footnotesize}
    \end{overpic}
  }
  \subfigure[]{
    \hspace{-0.33cm}
    \begin{overpic}[width=0.089\linewidth]{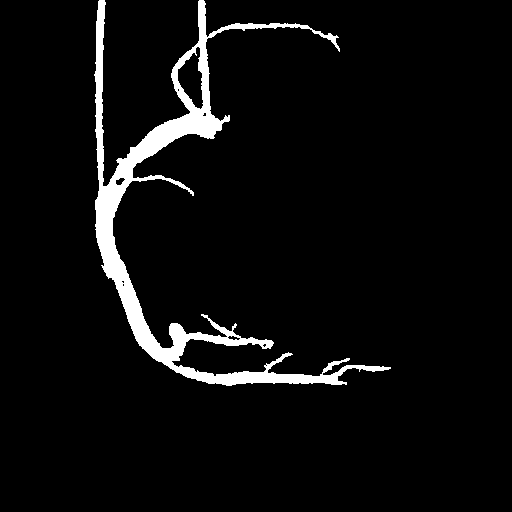}
      \begin{footnotesize}
        \put(3,5){\color{white}{(C4)}}
      \end{footnotesize}
    \end{overpic}
  }
  \subfigure[]{
    \hspace{-0.33cm}
    \begin{overpic}[width=0.089\linewidth]{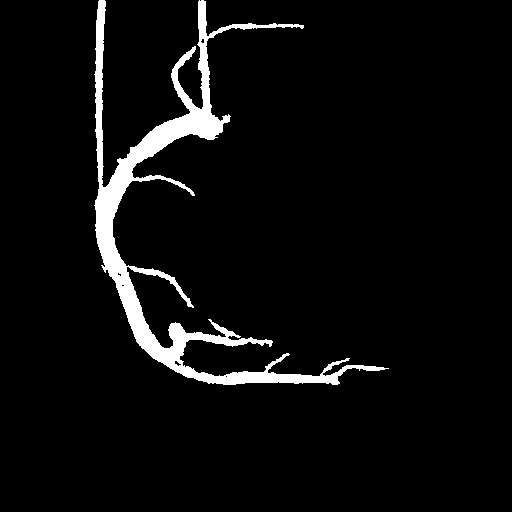}
      \begin{footnotesize}
        \put(3,5){\color{white}{(D4)}}
      \end{footnotesize}
    \end{overpic}
  }
  \subfigure[]{
    \hspace{-0.33cm}
    \begin{overpic}[width=0.089\linewidth]{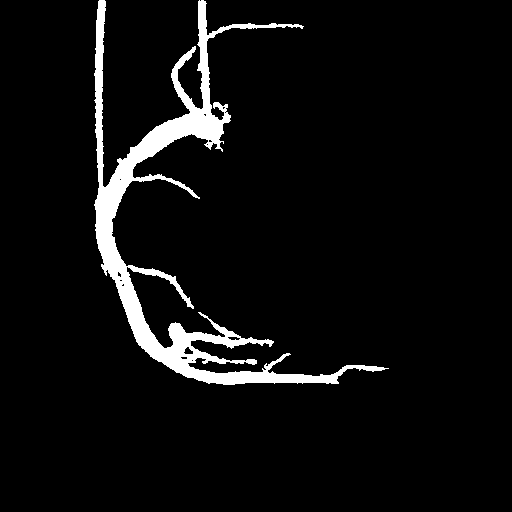}
      \begin{footnotesize}
        \put(3,5){\color{white}{(E4)}}
      \end{footnotesize}
    \end{overpic}
  }
  \subfigure[]{
    \hspace{-0.33cm}
    \begin{overpic}[width=0.089\linewidth]{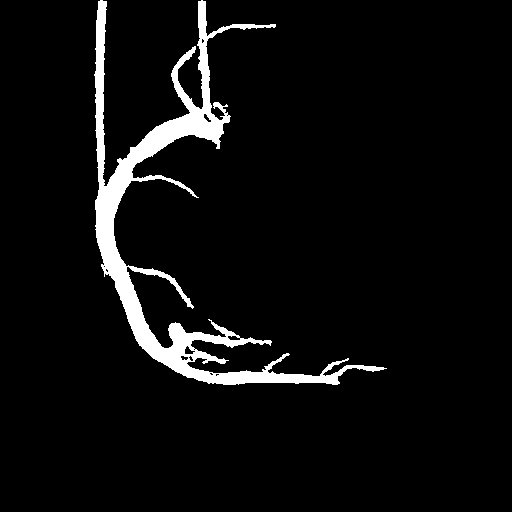}
      \begin{footnotesize}
        \put(3,5){\color{white}{(F4)}}
      \end{footnotesize}
    \end{overpic}
  }
  \subfigure[]{
    \hspace{-0.33cm}
    \begin{overpic}[width=0.089\linewidth]{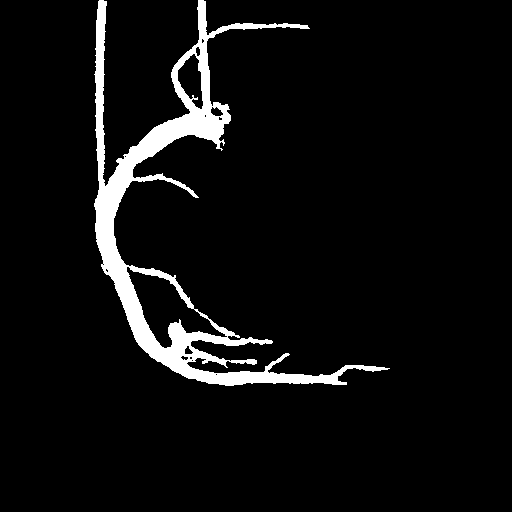}
      \begin{footnotesize}
        \put(3,5){\color{white}{(G4)}}
      \end{footnotesize}
    \end{overpic}
  }
  \subfigure[]{
    \hspace{-0.33cm}
    \begin{overpic}[width=0.089\linewidth]{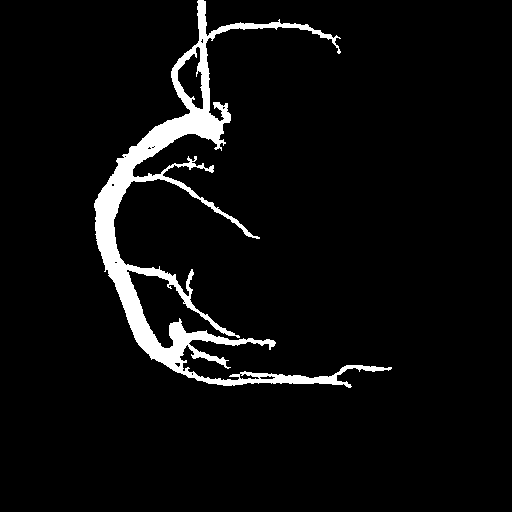}
      \begin{footnotesize}
        \put(3,5){\color{white}{(H4)}}
      \end{footnotesize}
    \end{overpic}
  }
  \subfigure[]{
    \hspace{-0.33cm}
    \begin{overpic}[width=0.089\linewidth]{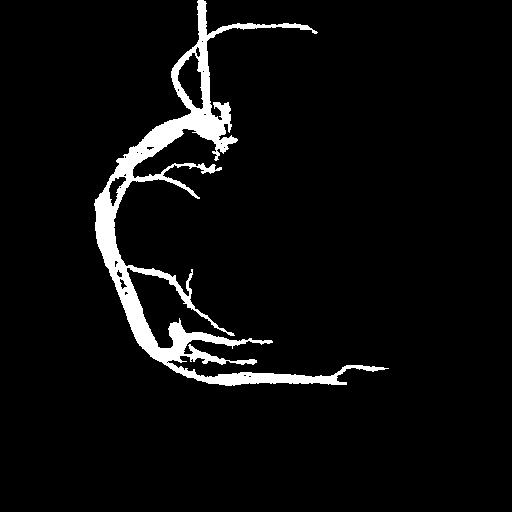}
      \begin{footnotesize}
        \put(3,5){\color{white}{(I4)}}
      \end{footnotesize}
    \end{overpic}
  }
  \subfigure[]{
    \hspace{-0.33cm}
    \begin{overpic}[width=0.089\linewidth]{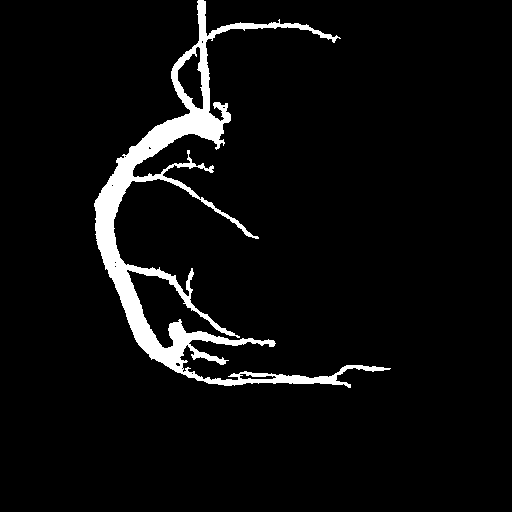}
      \begin{footnotesize}
        \put(3,5){\color{white}{(J4)}}
      \end{footnotesize}
    \end{overpic}
  }
  \subfigure[]{
    \hspace{-0.33cm}
    \begin{overpic}[width=0.089\linewidth]{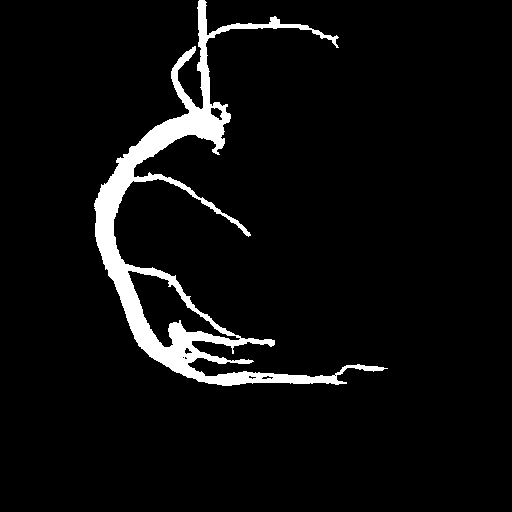}
      \begin{footnotesize}
        \put(3,5){\color{white}{(K4)}}
      \end{footnotesize}
    \end{overpic}
  }

  \vspace{-0.95cm}
  \subfigure[]{
    \begin{overpic}[width=0.089\linewidth]{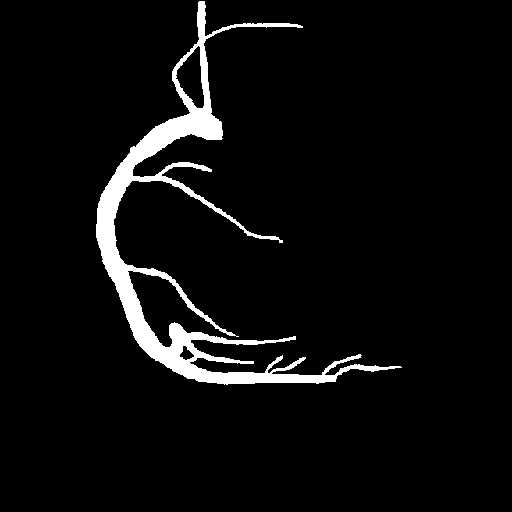}
      \begin{footnotesize}
        \put(3,5){\color{white}{(a4)}}
      \end{footnotesize}
    \end{overpic}
  }
  \subfigure[]{
    \hspace{-0.33cm}
    \begin{overpic}[width=0.089\linewidth]{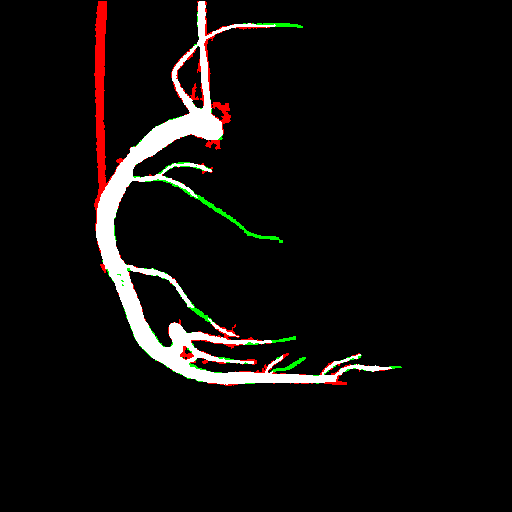}
      \begin{footnotesize}
        \put(3,5){\color{white}{(b4)}}
      \end{footnotesize}
    \end{overpic}
  }
  \subfigure[]{
    \hspace{-0.33cm}
    \begin{overpic}[width=0.089\linewidth]{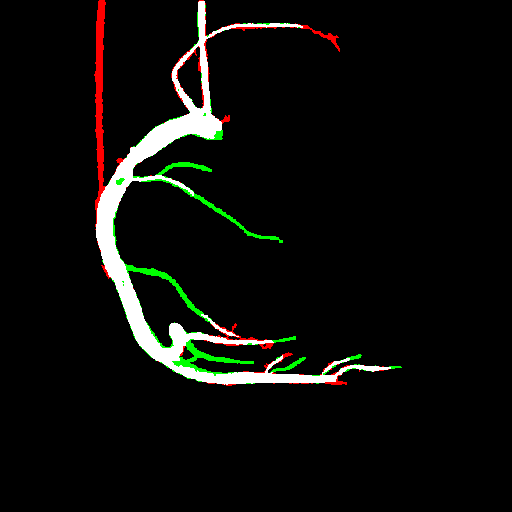}
      \begin{footnotesize}
        \put(3,5){\color{white}{(c4)}}
      \end{footnotesize}
    \end{overpic}
  }
  \subfigure[]{
    \hspace{-0.33cm}
    \begin{overpic}[width=0.089\linewidth]{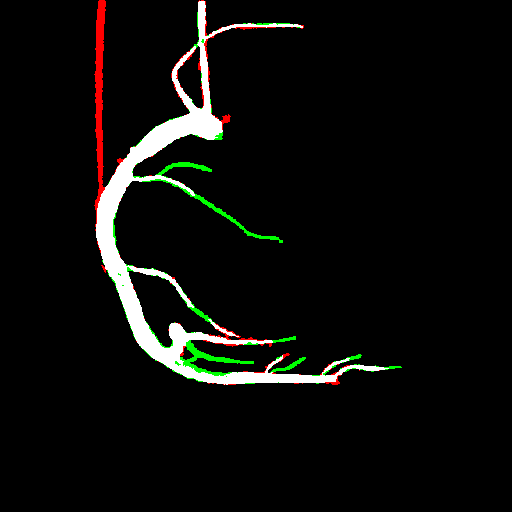}
      \begin{footnotesize}
        \put(3,5){\color{white}{(d4)}}
      \end{footnotesize}
    \end{overpic}
  }
  \subfigure[]{
    \hspace{-0.33cm}
    \begin{overpic}[width=0.089\linewidth]{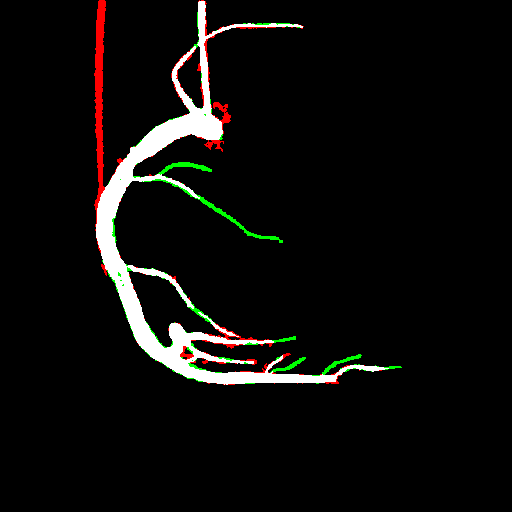}
      \begin{footnotesize}
        \put(3,5){\color{white}{(e4)}}
      \end{footnotesize}
    \end{overpic}
  }
  \subfigure[]{
    \hspace{-0.33cm}
    \begin{overpic}[width=0.089\linewidth]{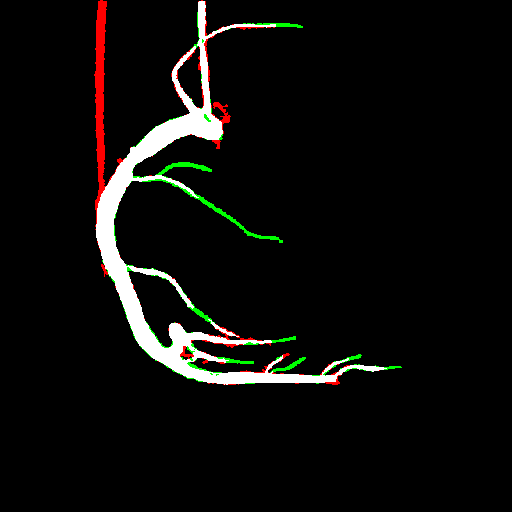}
      \begin{footnotesize}
        \put(3,5){\color{white}{(f4)}}
      \end{footnotesize}
    \end{overpic}
  }
  \subfigure[]{
    \hspace{-0.33cm}
    \begin{overpic}[width=0.089\linewidth]{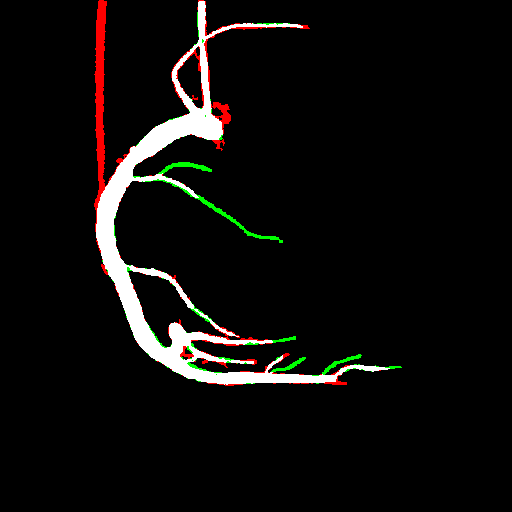}
      \begin{footnotesize}
        \put(3,5){\color{white}{(g4)}}
      \end{footnotesize}
    \end{overpic}
  }
  \subfigure[]{
    \hspace{-0.33cm}
    \begin{overpic}[width=0.089\linewidth]{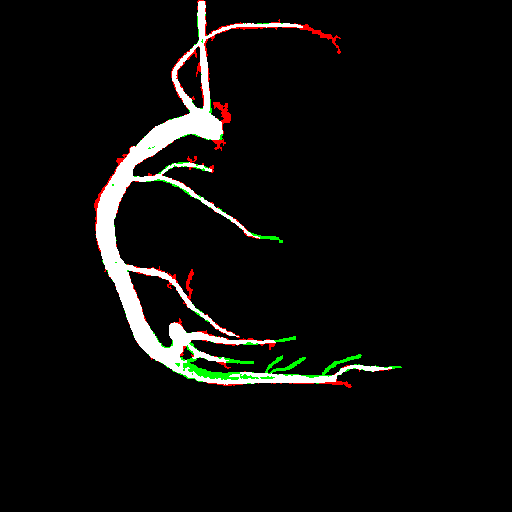}
      \begin{footnotesize}
        \put(3,5){\color{white}{(h4)}}
      \end{footnotesize}
    \end{overpic}
  }
  \subfigure[]{
    \hspace{-0.33cm}
    \begin{overpic}[width=0.089\linewidth]{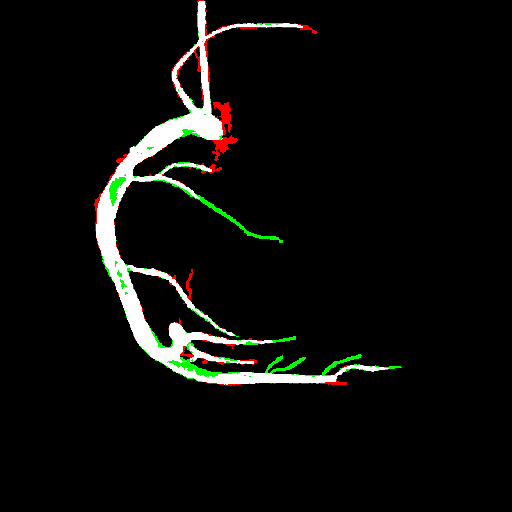}
      \begin{footnotesize}
        \put(3,5){\color{white}{(i4)}}
      \end{footnotesize}
    \end{overpic}
  }
  \subfigure[]{
    \hspace{-0.33cm}
    \begin{overpic}[width=0.089\linewidth]{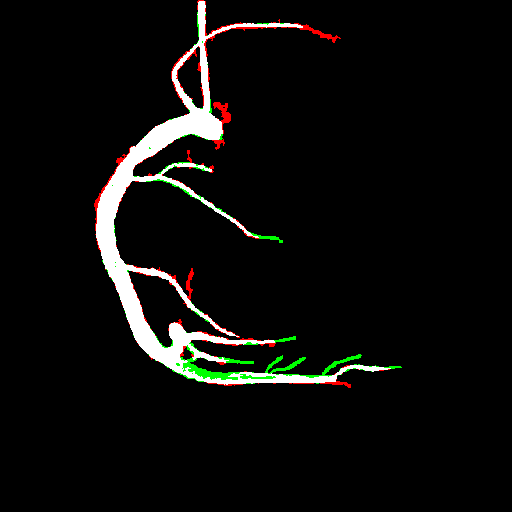}
      \begin{footnotesize}
        \put(3,5){\color{white}{(j4)}}
      \end{footnotesize}
    \end{overpic}
  }
  \subfigure[]{
    \hspace{-0.33cm}
    \begin{overpic}[width=0.089\linewidth]{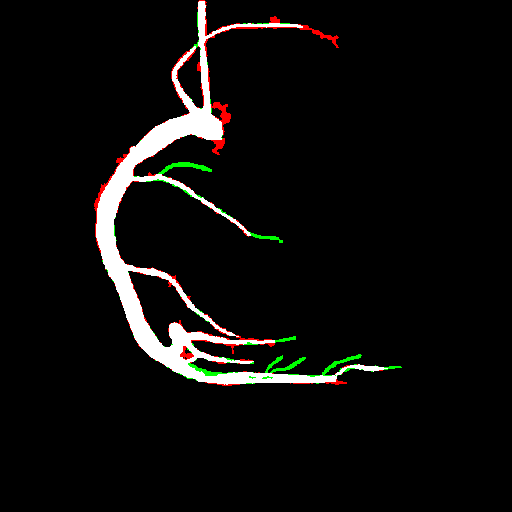}
      \begin{footnotesize}
        \put(3,5){\color{white}{(k4)}}
      \end{footnotesize}
    \end{overpic}
  }

  \label{fig8}
  \vspace{-0.8cm}
  \caption{Region growing segmentation results of 4 XCA image sequences on the vessel layer extracted by 10 layer separation approaches. Each group of results contains a segmentation image labeled by capital letters (B-K) and its contrast image with the ground truth labeled by lowercase letters (b-k). In (b-k), white area represents the true positive (TP) part, red area represents the false positive (FP) part, green area represents the false negative (FN) part. (A1)-(A4) and (a1)-(a4) are the raw XCA images and their ground truth binary vessel masks. (B,b) DECOLOR. (C,c) GoDec. (D,d) PRMF. (E,e) AccAltProj. (F,f) Mog-RPCA. (G,g) MCR-RPCA. (H,h) ETRPCA. (I,i) KBR-RPCA. (J,j) TNN-TRPCA. (K,k) TV-TRPCA.}
\end{figure*}

For $\lambda_1$, according to \cite{lu2019tensor} and \cite{candes2011robust}, define $\lambda_0=1/\sqrt{\max(M,N)*T}$, when $\lambda_1/\lambda_0\in [0,1]$, the decomposition could have better performance. In this experiment, we set $\lambda_1/\lambda_0\in [0.05,0.55],\lambda_3/\lambda_1=0.6$, and the foreground extraction results are presented in \hyperref[fig4]{Fig.4.(A1-K1)}. In addition, we applied TSRG on the vessel layer image and obtained binary vessel mask shown in \hyperref[fig4]{Fig.4.(a1-k1)}, where white area represents the true TP part, red area represents the FP part, and green area represents the FN part, compared with the ground truth. Quantitative analysis of foreground visibility and segmentation accuracy is shown in \hyperref[fig5]{Fig.5.(A)}, including five aforementioned metrics. Considering the variation trend of CNR value and segmentation accuracy, $\lambda_1/\lambda_0=0.3$ is selected as the optimal parameter.

For $\lambda_3$, we set $\lambda_3/\lambda_1\in [0.1,1.1],\lambda_1/\lambda_0=0.3$, the foreground extraction results are presented in \hyperref[fig4]{Fig.4.(A2-K2)}, and the vessel segmentation results are shown in \hyperref[fig4]{Fig.4.(a2-k2)}. Quantitative analysis of foreground visibility and segmentation accuracy is shown in \hyperref[fig5]{Fig.5.(B)}, and $\lambda_3/\lambda_1=0.6$ is selected as the optimal parameter.

\begin{figure*}[tbp]
  \centering
  \subfigure[]{\includegraphics[width=\linewidth]{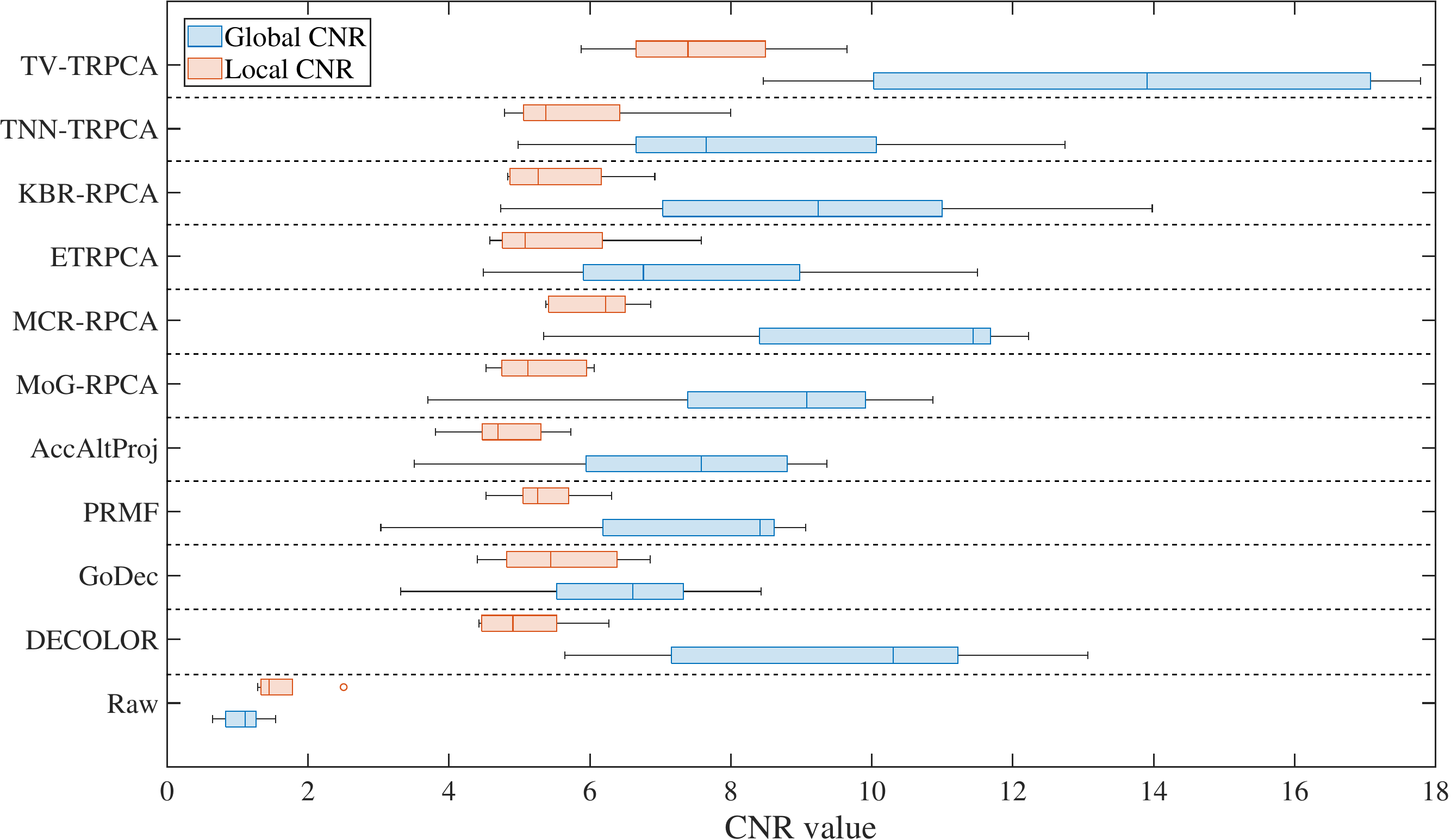}}

  \label{fig9}
  \vspace{-0.8cm}
  \caption{Distribution of Global CNR and Local CNR values of raw XCA image and vessel foreground extracted by different layer separation methods on clinical XCA image sequences.}
\end{figure*}

\begin{table*}[!t]
  \renewcommand{\arraystretch}{1}
  \centering
  \topcaption{The mean CNR, F-Measure values ($\pm$ standard deviation) and average time costs of different layer separation methods}
  \label{tab3}
  \begin{footnotesize}
    \begin{tabular}{l|ccr|c|ccr|c}
      \toprule \toprule
      \multirow{2}{*}{\textbf{Method}} & \multicolumn{4}{c|}{\textbf{Clinical}} & \multicolumn{4}{c}{\textbf{Dataset}}                                                                                                                                                          \\
      \cmidrule(l){2-9}
                                       & Global CNR                             & Local CNR                            & Time cost   & F-Measure                     & Global CNR                   & Local CNR                   & Time cost   & F-Measure                     \\
      \midrule
      Raw                         & 1.07 $\pm$ 0.33                        & 1.62 $\pm$ 0.50                      & $\diagdown$ & $\diagdown$                   & 1.25 $\pm$ 0.26              & 1.88 $\pm$ 0.39             & $\diagdown$ & $\diagdown$                   \\
      DECOLOR                          & 9.46 $\pm$ 2.86                        & 5.07 $\pm$ 0.76                      & 252.51s     & 0.817 $\pm$ 0.032             & 6.82 $\pm$ 2.35              & 3.48 $\pm$ 0.66             & 76.54s      & 0.748 $\pm$ 0.053             \\
      GoDec                            & 6.31 $\pm$ 1.87                        & 5.58 $\pm$ 0.98                      & \pmb{0.58s} & 0.749 $\pm$ 0.061             & 4.27 $\pm$ 1.23              & 3.70 $\pm$ 0.70             & \pmb{0.27s} & 0.621 $\pm$ 0.077             \\
      PRMF                             & 7.24 $\pm$ 2.44                        & 5.36 $\pm$ 0.64                      & 306.49s     & 0.763 $\pm$ 0.045             & 4.79 $\pm$ 1.72              & 3.72 $\pm$ 0.52             & 103.99s     & 0.655 $\pm$ 0.056             \\
      AccAltProj                       & 7.16 $\pm$ 2.27                        & 4.82 $\pm$ 0.71                      & 8.37s       & 0.749 $\pm$ 0.063             & 4.44 $\pm$ 1.69              & 3.21 $\pm$ 0.79             & 2.68s       & 0.619 $\pm$ 0.071             \\
      MoG-RPCA                         & 8.37 $\pm$ 2.74                        & 5.29 $\pm$ 0.67                      & 70.56s      & 0.801 $\pm$ 0.021             & 5.96 $\pm$ 2.28              & 4.47 $\pm$ 0.57             & 25.57s      & 0.722 $\pm$ 0.034             \\
      MCR-RPCA                         & 9.99 $\pm$ 2.80                        & 6.05 $\pm$ 0.64                      & 974.12s     & 0.807 $\pm$ 0.036             & 7.19 $\pm$ 2.66              & 4.93 $\pm$ 1.17             & 423.78s     & 0.765 $\pm$ 0.040             \\
      ETRPCA                           & 7.45 $\pm$ 2.61                        & 5.55 $\pm$ 1.21                      & 285.54s     & 0.781 $\pm$ 0.053             & 6.65 $\pm$ 2.71              & 5.45 $\pm$ 1.01             & 82.54s      & 0.707 $\pm$ 0.038             \\
      KBR-RPCA                         & 9.15 $\pm$ 3.37                        & 5.56 $\pm$ 0.87                      & 367.72s     & 0.675 $\pm$ 0.120             & 6.40 $\pm$ 2.59              & 5.08 $\pm$ 1.14             & 139.68s     & 0.533 $\pm$ 0.093             \\
      TNN-TRPCA                        & 8.35 $\pm$ 2.88                        & 5.84 $\pm$ 1.27                      & 157.77s     & 0.785 $\pm$ 0.059             & 6.98 $\pm$ 1.86              & 5.52 $\pm$ 1.27             & 73.70s      & 0.701 $\pm$ 0.042             \\
      TV-TRPCA                         & \pmb{13.51} $\pm$ \pmb{3.99}           & \pmb{7.59} $\pm$ \pmb{1.41}          & 321.56s     & \pmb{0.824} $\pm$ \pmb{0.024} & \pmb{11.87} $\pm$ \pmb{3.45} & \pmb{6.72} $\pm$ \pmb{1.31} & 90.14s      & \pmb{0.780} $\pm$ \pmb{0.023} \\
      \bottomrule \bottomrule
    \end{tabular}
  \end{footnotesize}
\end{table*}

\subsubsection{Comparison Test}

The vessel foreground extraction results of proposed TV-TRPCA method and 9 comparison methods are shown in \hyperref[fig7]{Fig.7}. Qualitatively, Three RPCA-based methods AccAltProj, MoG-RPCA and MCR-RPCA show better continuity of foreground vessel layer at the expense of smoothness of background layer. The background grayscale obtained by these three methods is extremely uneven, with a large number of tubular artifacts. This is because the high-dimensional information is destroyed in the operation of matricization. Non-RPCA-based methods GoDec and PRMF show similar properties with the former methods. DECOLOR performs well in the low-rank property of background layer, but remains a large number of artifacts in the foreground layer, destroying the sparsity. As a comparison, three TRPCA-based methods ETRPCA, KBR-RPCA and TNN-TRPCA have good gray uniformity of background layer. However, for the aorta moving slowly in the foreground, these three methods all classify it as the background part, resulting in the voids of the foreground vessel layer. Due to the high-order information recovery ability of TRPCA and the constraint of TV regularization on the foreground, the proposed method TV-TRPCA performs better in both the low-rank property of background and the continuity of foreground, the dynamic disturbance is also well suppressed.

Quantitatively, for clinical XCA image sequences and third-party dataset, we calculated Global CNR and Local CNR of different foreground extraction methods and recorded the time costs in \hyperref[tab3]{Tab.3}. And \hyperref[fig9]{Fig9} gives an intuitive distribution of CNR values for clinical images. To further test the influence of different methods on the segmentation accuracy, region growing was directly utilized on each vessel layer image, and the results are shown in \hyperref[fig8]{Fig8}, the F-Measure is recorded in \hyperref[tab3]{Tab.3}. It can be seen that the proposed TV-TRPCA method performs best both in CNR value and segmentation accuracy. But the segmentation by region growing without threshold method (TM) is not optimal, most of the binary vessel masks appear voids. This issue will be solved in the \hyperref[TSRG]{experiment} of TSRG. 

\subsubsection{Applications}

The vessel foreground layer image extracted by TV-TRPCA can serve as the input for commonly used vessel segmentation methods Frangi filtering, DSA and U-net, instead of the raw XCA image. The U-net simply contains a fully convolutional encoder and a deconvolution-based decoder. The feature extraction network (encoder) consists of four convolution and pooling layer, and the feature fusion network (decoder) consists of four convolution and upsampling layer. Between the encoder and decoder network is the operation of concatenating.

\begin{figure*}[htb]
  \centering
  \subfigure[]{
    \begin{overpic}[width=0.12\linewidth]{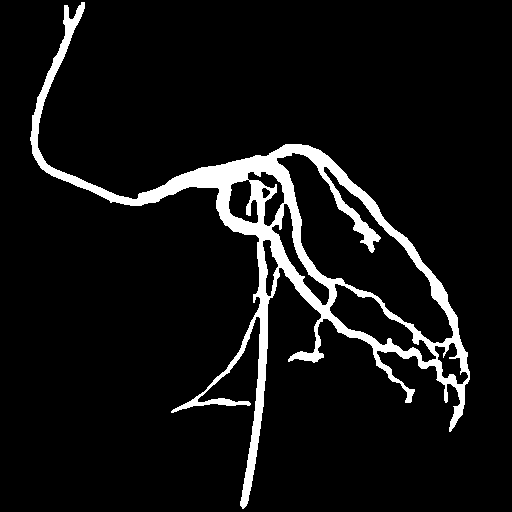}
      \begin{small}
        \put(3,5){\color{white}{(A1)}}
      \end{small}
    \end{overpic}
  }
  \subfigure[]{
    \hspace{-0.3cm}
    \begin{overpic}[width=0.12\linewidth]{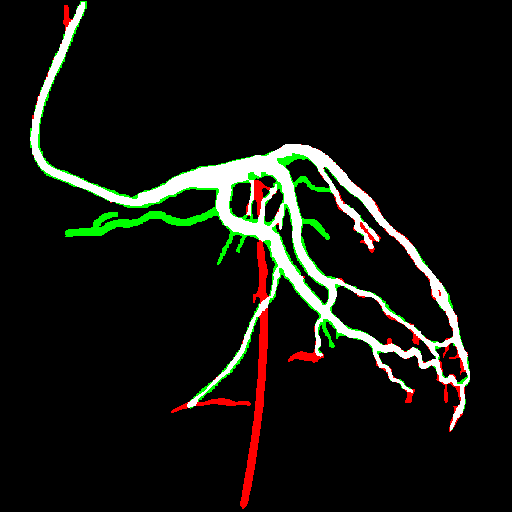}
      \begin{small}
        \put(3,5){\color{white}{(a1)}}
      \end{small}
    \end{overpic}
  }
  \subfigure[]{
    \hspace{-0.3cm}
    \begin{overpic}[width=0.12\linewidth]{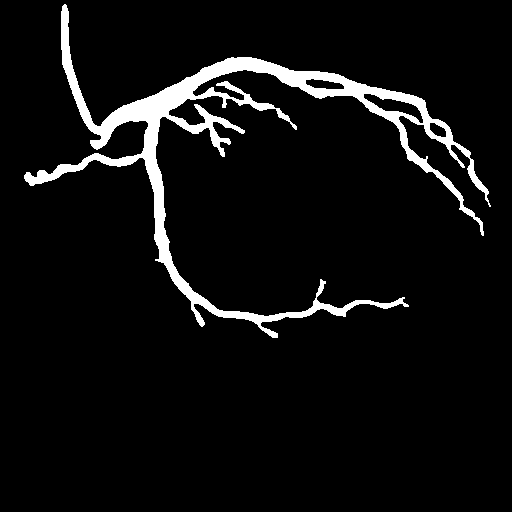}
      \begin{small}
        \put(67,5){\color{white}{(B1)}}
      \end{small}
    \end{overpic}
  }
  \subfigure[]{
    \hspace{-0.3cm}
    \begin{overpic}[width=0.12\linewidth]{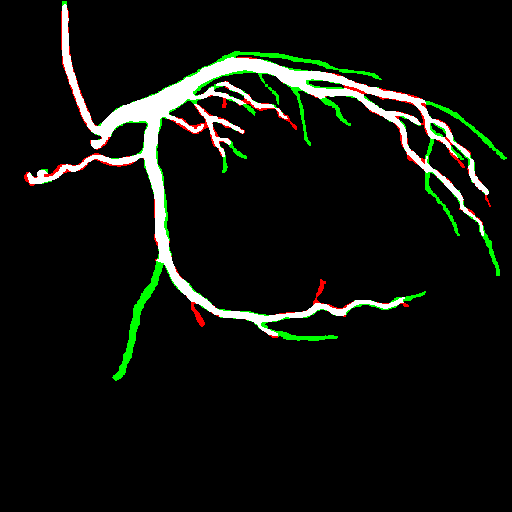}
      \begin{small}
        \put(67,5){\color{white}{(b1)}}
      \end{small}
    \end{overpic}
  }
  \subfigure[]{
    \hspace{-0.3cm}
    \begin{overpic}[width=0.12\linewidth]{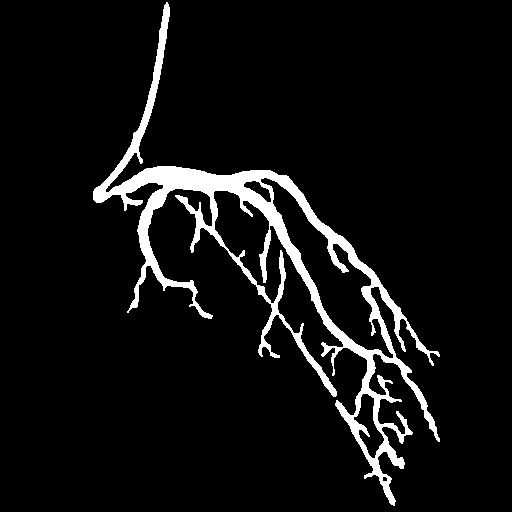}
      \begin{small}
        \put(3,5){\color{white}{(C1)}}
      \end{small}
    \end{overpic}
  }
  \subfigure[]{
    \hspace{-0.3cm}
    \begin{overpic}[width=0.12\linewidth]{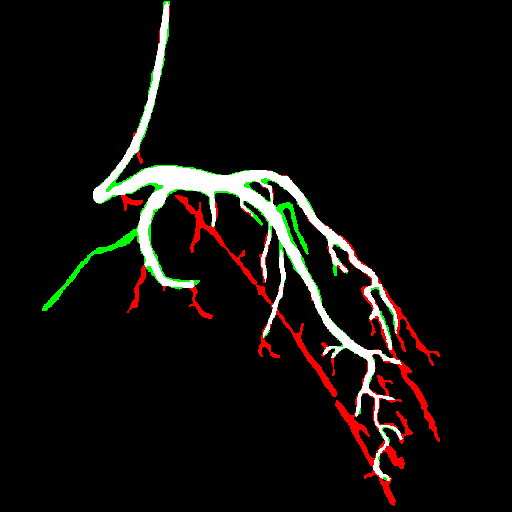}
      \begin{small}
        \put(3,5){\color{white}{(c1)}}
      \end{small}
    \end{overpic}
  }
  \subfigure[]{
    \hspace{-0.3cm}
    \begin{overpic}[width=0.12\linewidth]{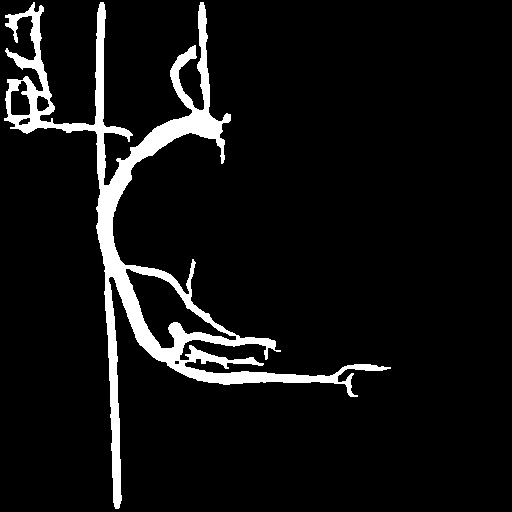}
      \begin{small}
        \put(67,5){\color{white}{(D1)}}
      \end{small}
    \end{overpic}
  }
  \subfigure[]{
    \hspace{-0.3cm}
    \begin{overpic}[width=0.12\linewidth]{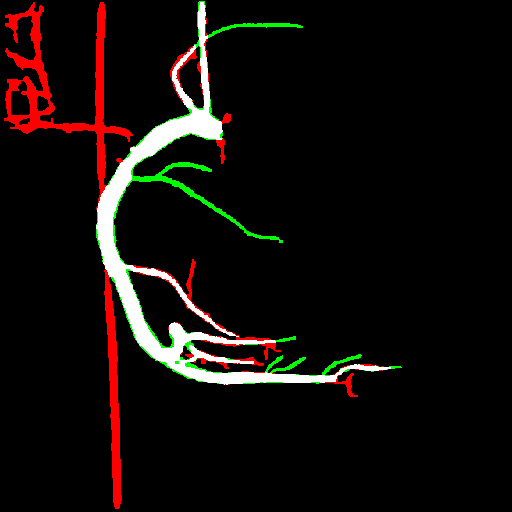}
      \begin{small}
        \put(67,5){\color{white}{(d1)}}
      \end{small}
    \end{overpic}
  }

  \vspace{-0.95cm}
  \subfigure[]{
    \begin{overpic}[width=0.12\linewidth]{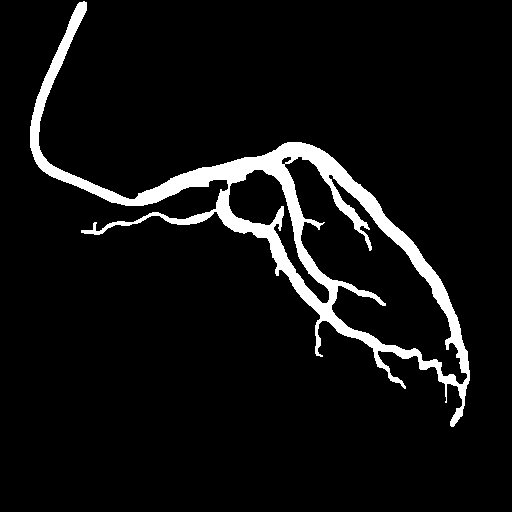}
      \begin{small}
        \put(3,5){\color{white}{(A2)}}
      \end{small}
    \end{overpic}
  }
  \subfigure[]{
    \hspace{-0.3cm}
    \begin{overpic}[width=0.12\linewidth]{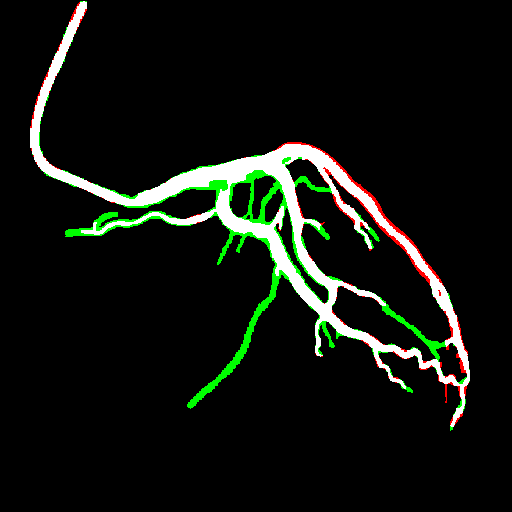}
      \begin{small}
        \put(3,5){\color{white}{(a2)}}
      \end{small}
    \end{overpic}
  }
  \subfigure[]{
    \hspace{-0.3cm}
    \begin{overpic}[width=0.12\linewidth]{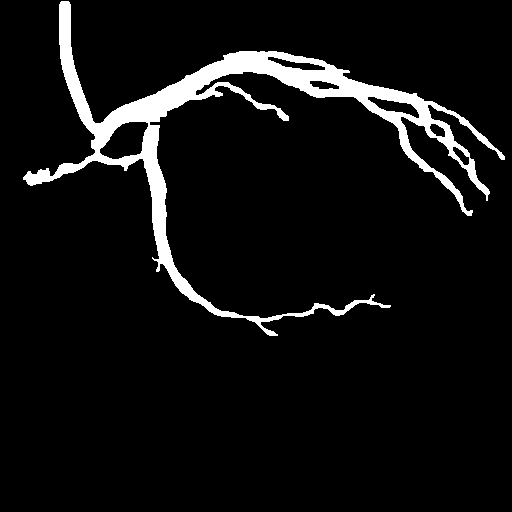}
      \begin{small}
        \put(67,5){\color{white}{(B2)}}
      \end{small}
    \end{overpic}
  }
  \subfigure[]{
    \hspace{-0.3cm}
    \begin{overpic}[width=0.12\linewidth]{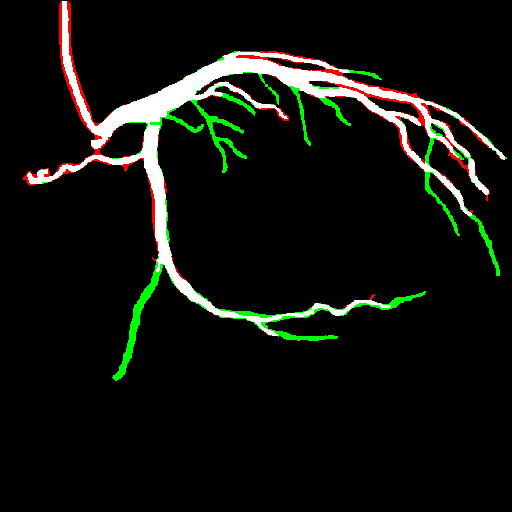}
      \begin{small}
        \put(67,5){\color{white}{(b2)}}
      \end{small}
    \end{overpic}
  }
  \subfigure[]{
    \hspace{-0.3cm}
    \begin{overpic}[width=0.12\linewidth]{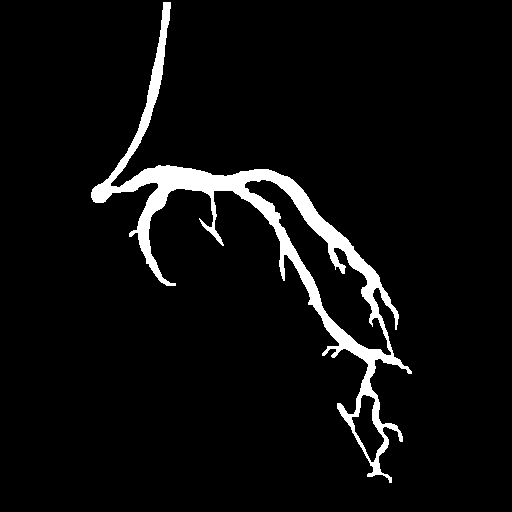}
      \begin{small}
        \put(3,5){\color{white}{(C2)}}
      \end{small}
    \end{overpic}
  }
  \subfigure[]{
    \hspace{-0.3cm}
    \begin{overpic}[width=0.12\linewidth]{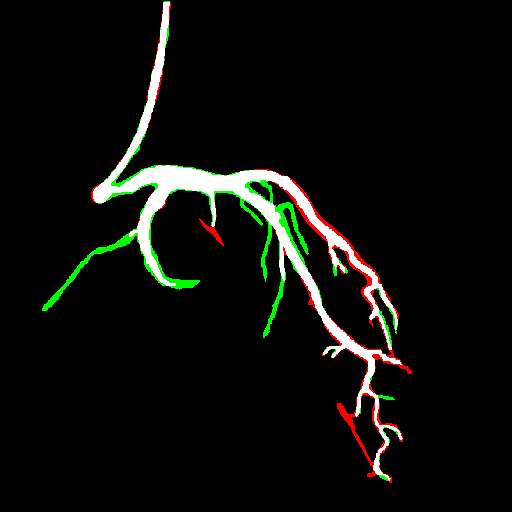}
      \begin{small}
        \put(3,5){\color{white}{(c2)}}
      \end{small}
    \end{overpic}
  }
  \subfigure[]{
    \hspace{-0.3cm}
    \begin{overpic}[width=0.12\linewidth]{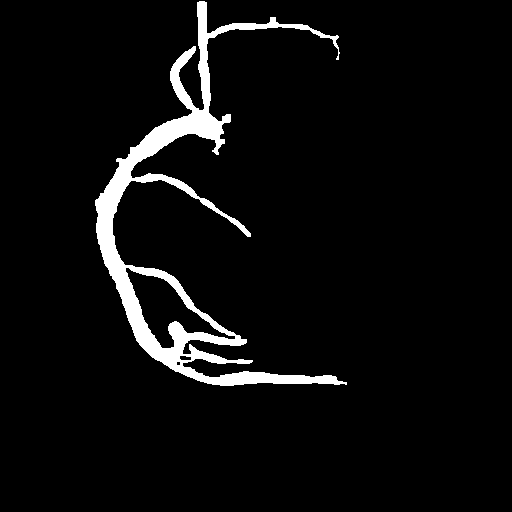}
      \begin{small}
        \put(67,5){\color{white}{(D2)}}
      \end{small}
    \end{overpic}
  }
  \subfigure[]{
    \hspace{-0.3cm}
    \begin{overpic}[width=0.12\linewidth]{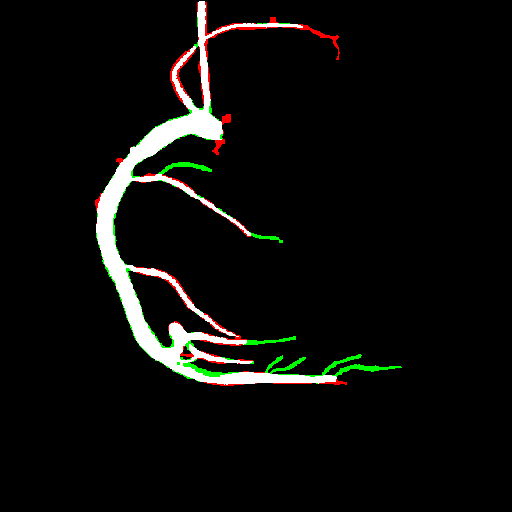}
      \begin{small}
        \put(67,5){\color{white}{(d2)}}
      \end{small}
    \end{overpic}
  }

  \vspace{-0.85cm}
  \subfigure[]{
    \begin{overpic}[width=0.12\linewidth]{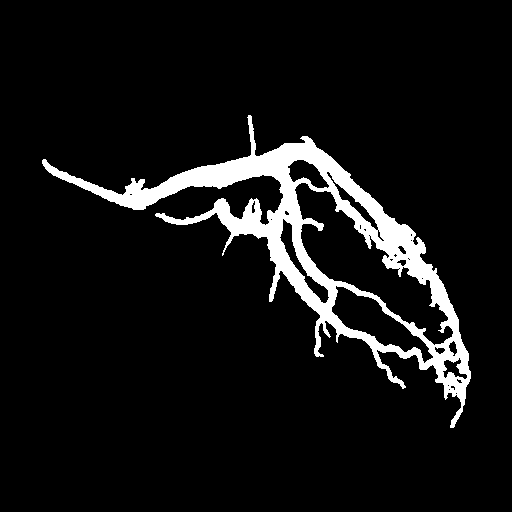}
      \begin{small}
        \put(3,5){\color{white}{(A3)}}
      \end{small}
    \end{overpic}
  }
  \subfigure[]{
    \hspace{-0.3cm}
    \begin{overpic}[width=0.12\linewidth]{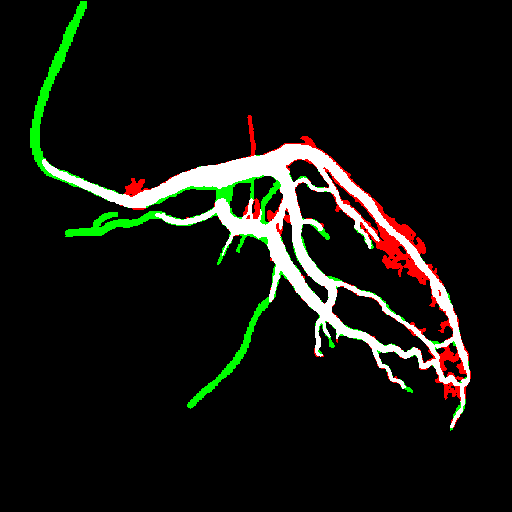}
      \begin{small}
        \put(3,5){\color{white}{(a3)}}
      \end{small}
    \end{overpic}
  }
  \subfigure[]{
    \hspace{-0.3cm}
    \begin{overpic}[width=0.12\linewidth]{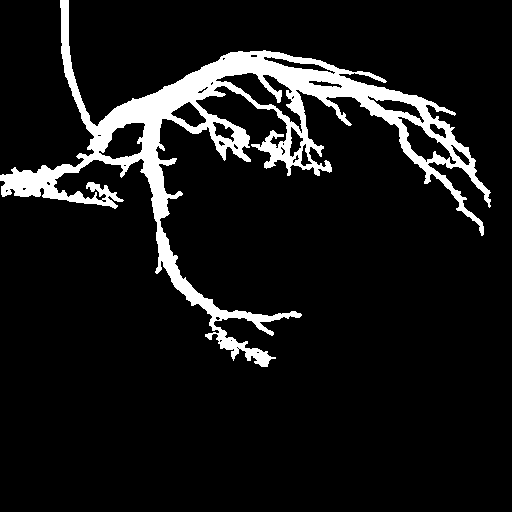}
      \begin{small}
        \put(67,5){\color{white}{(B3)}}
      \end{small}
    \end{overpic}
  }
  \subfigure[]{
    \hspace{-0.3cm}
    \begin{overpic}[width=0.12\linewidth]{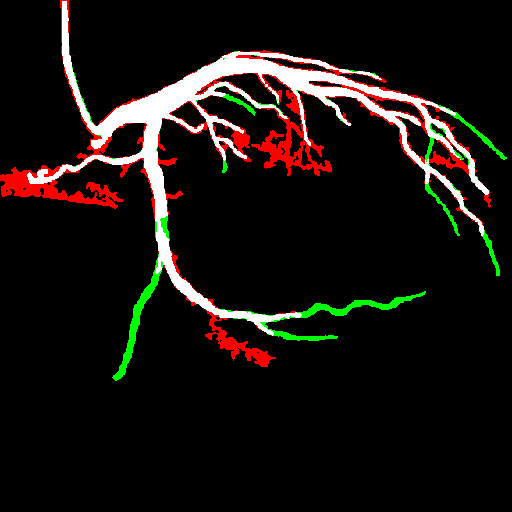}
      \begin{small}
        \put(67,5){\color{white}{(b3)}}
      \end{small}
    \end{overpic}
  }
  \subfigure[]{
    \hspace{-0.3cm}
    \begin{overpic}[width=0.12\linewidth]{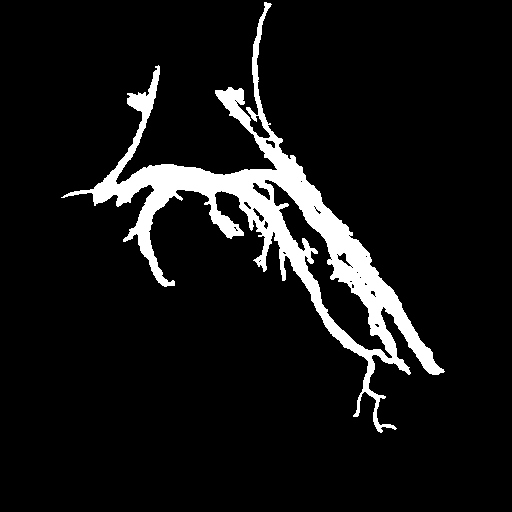}
      \begin{small}
        \put(3,5){\color{white}{(C3)}}
      \end{small}
    \end{overpic}
  }
  \subfigure[]{
    \hspace{-0.3cm}
    \begin{overpic}[width=0.12\linewidth]{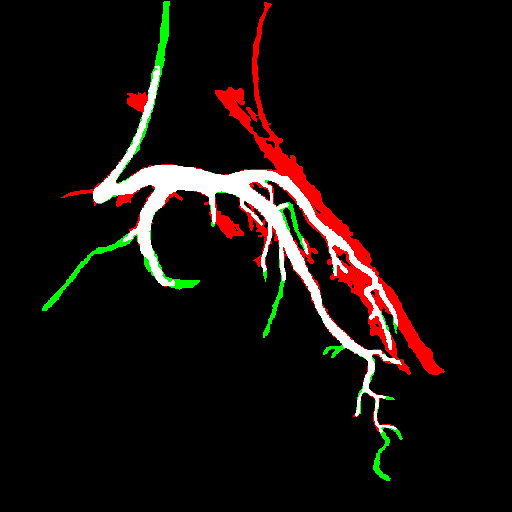}
      \begin{small}
        \put(3,5){\color{white}{(c3)}}
      \end{small}
    \end{overpic}
  }
  \subfigure[]{
    \hspace{-0.3cm}
    \begin{overpic}[width=0.12\linewidth]{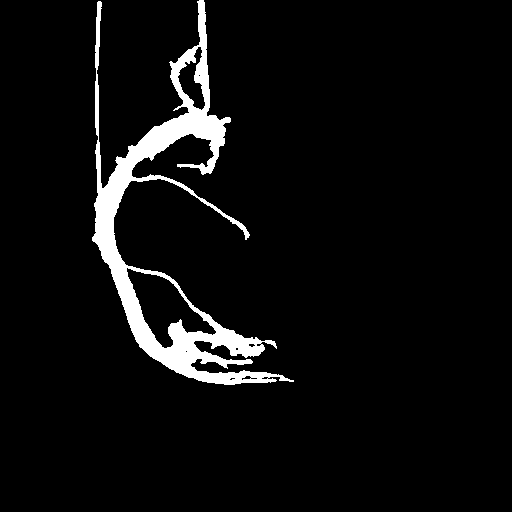}
      \begin{small}
        \put(67,5){\color{white}{(D3)}}
      \end{small}
    \end{overpic}
  }
  \subfigure[]{
    \hspace{-0.3cm}
    \begin{overpic}[width=0.12\linewidth]{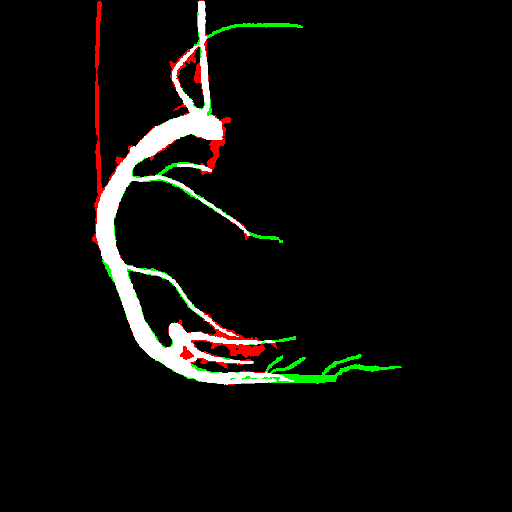}
      \begin{small}
        \put(67,5){\color{white}{(d3)}}
      \end{small}
    \end{overpic}
  }

  \vspace{-0.95cm}
  \subfigure[]{
    \begin{overpic}[width=0.12\linewidth]{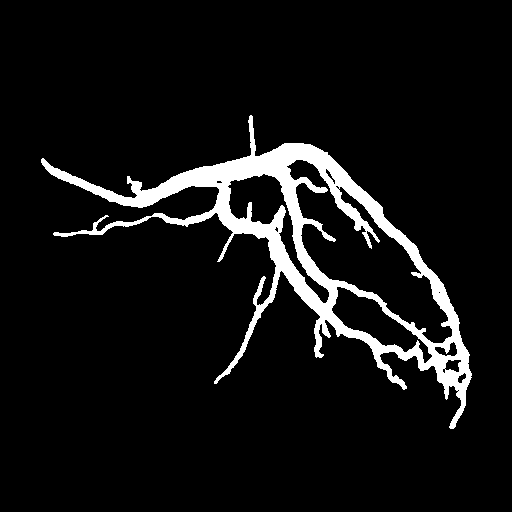}
      \begin{small}
        \put(3,5){\color{white}{(A4)}}
      \end{small}
    \end{overpic}
  }
  \subfigure[]{
    \hspace{-0.3cm}
    \begin{overpic}[width=0.12\linewidth]{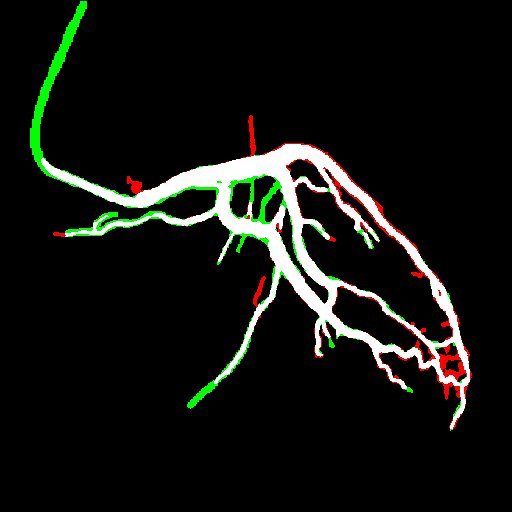}
      \begin{small}
        \put(3,5){\color{white}{(a4)}}
      \end{small}
    \end{overpic}
  }
  \subfigure[]{
    \hspace{-0.3cm}
    \begin{overpic}[width=0.12\linewidth]{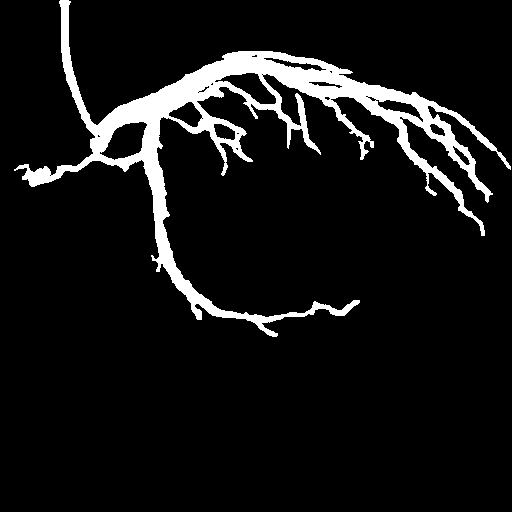}
      \begin{small}
        \put(67,5){\color{white}{(B4)}}
      \end{small}
    \end{overpic}
  }
  \subfigure[]{
    \hspace{-0.3cm}
    \begin{overpic}[width=0.12\linewidth]{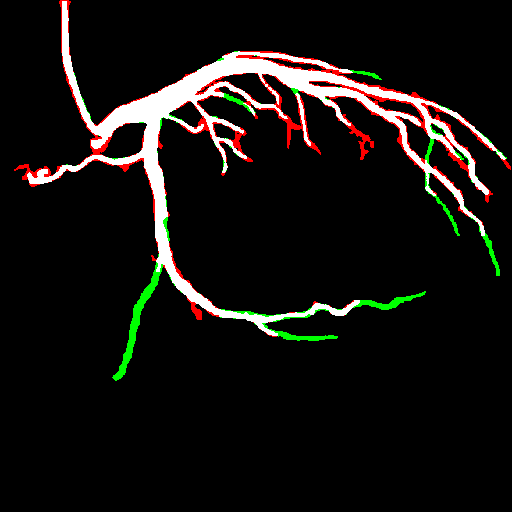}
      \begin{small}
        \put(67,5){\color{white}{(b4)}}
      \end{small}
    \end{overpic}
  }
  \subfigure[]{
    \hspace{-0.3cm}
    \begin{overpic}[width=0.12\linewidth]{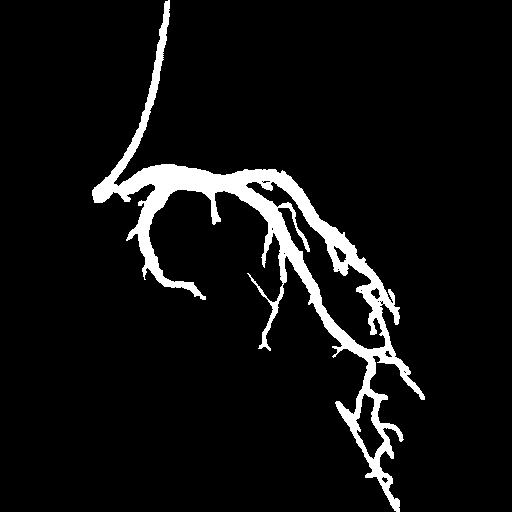}
      \begin{small}
        \put(3,5){\color{white}{(C4)}}
      \end{small}
    \end{overpic}
  }
  \subfigure[]{
    \hspace{-0.3cm}
    \begin{overpic}[width=0.12\linewidth]{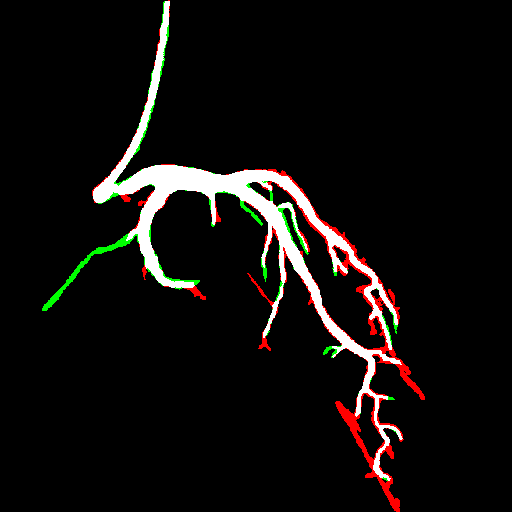}
      \begin{small}
        \put(3,5){\color{white}{(c4)}}
      \end{small}
    \end{overpic}
  }
  \subfigure[]{
    \hspace{-0.3cm}
    \begin{overpic}[width=0.12\linewidth]{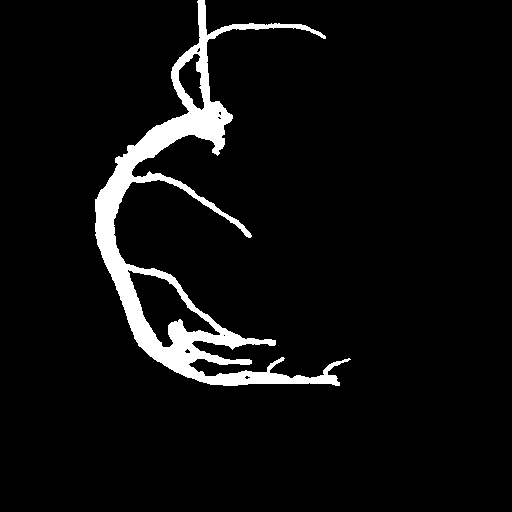}
      \begin{small}
        \put(67,5){\color{white}{(D4)}}
      \end{small}
    \end{overpic}
  }
  \subfigure[]{
    \hspace{-0.3cm}
    \begin{overpic}[width=0.12\linewidth]{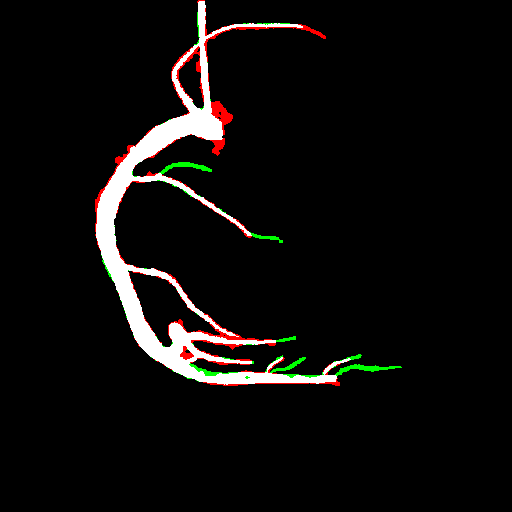}
      \begin{small}
        \put(67,5){\color{white}{(d4)}}
      \end{small}
    \end{overpic}
  }

  \vspace{-0.85cm}
  \subfigure[]{
    \begin{overpic}[width=0.12\linewidth]{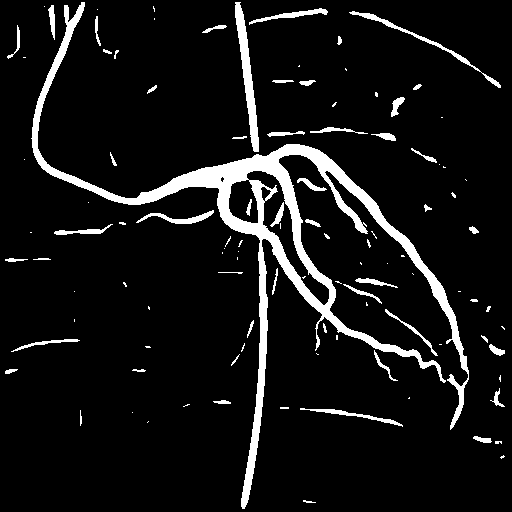}
      \begin{small}
        \put(3,5){\color{white}{(A5)}}
      \end{small}
    \end{overpic}
  }
  \subfigure[]{
    \hspace{-0.3cm}
    \begin{overpic}[width=0.12\linewidth]{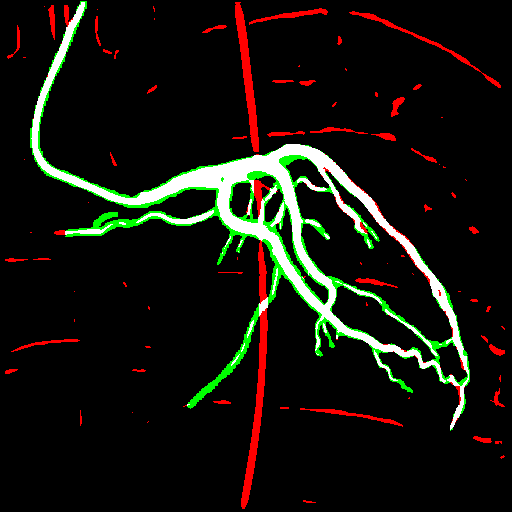}
      \begin{small}
        \put(3,5){\color{white}{(a5)}}
      \end{small}
    \end{overpic}
  }
  \subfigure[]{
    \hspace{-0.3cm}
    \begin{overpic}[width=0.12\linewidth]{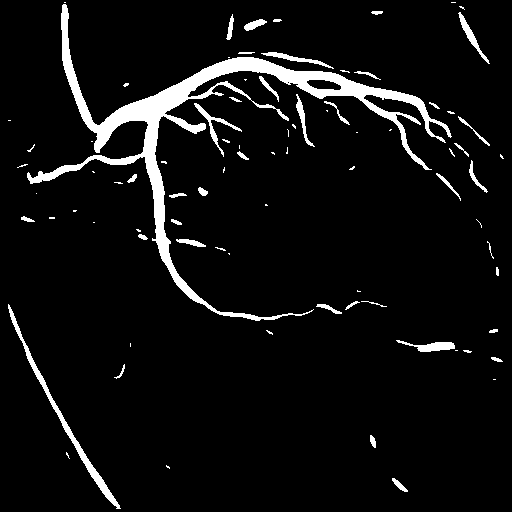}
      \begin{small}
        \put(67,5){\color{white}{(B5)}}
      \end{small}
    \end{overpic}
  }
  \subfigure[]{
    \hspace{-0.3cm}
    \begin{overpic}[width=0.12\linewidth]{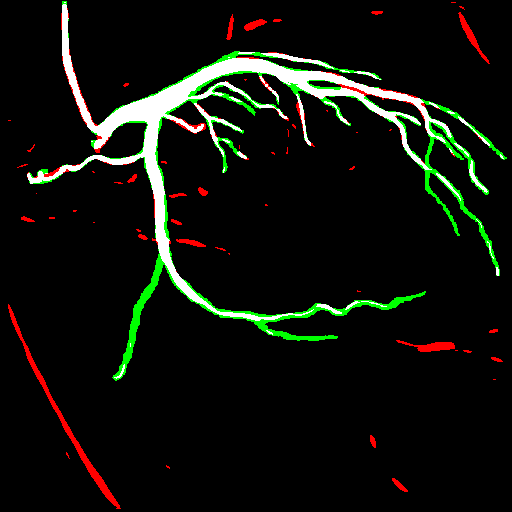}
      \begin{small}
        \put(67,5){\color{white}{(b5)}}
      \end{small}
    \end{overpic}
  }
  \subfigure[]{
    \hspace{-0.3cm}
    \begin{overpic}[width=0.12\linewidth]{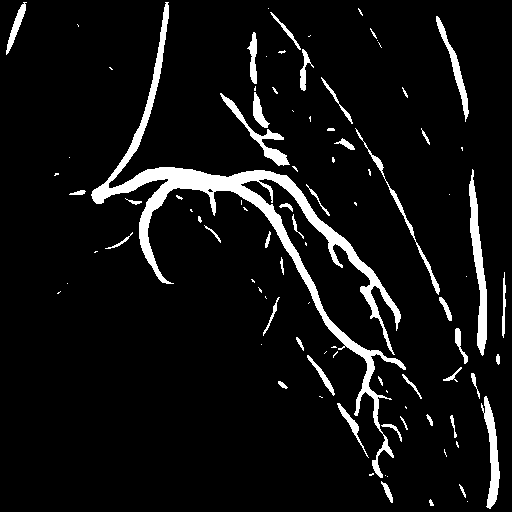}
      \begin{small}
        \put(3,5){\color{white}{(C5)}}
      \end{small}
    \end{overpic}
  }
  \subfigure[]{
    \hspace{-0.3cm}
    \begin{overpic}[width=0.12\linewidth]{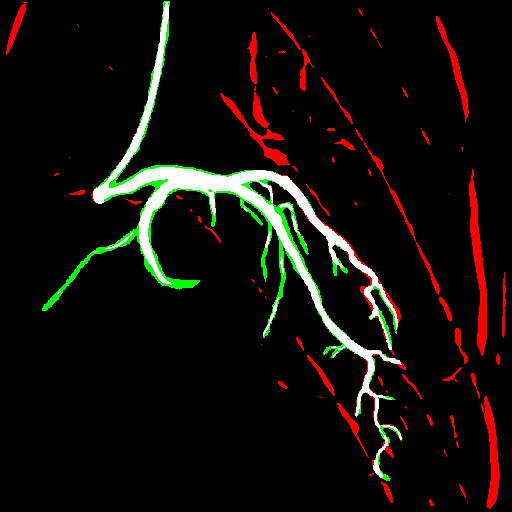}
      \begin{small}
        \put(3,5){\color{white}{(c5)}}
      \end{small}
    \end{overpic}
  }
  \subfigure[]{
    \hspace{-0.3cm}
    \begin{overpic}[width=0.12\linewidth]{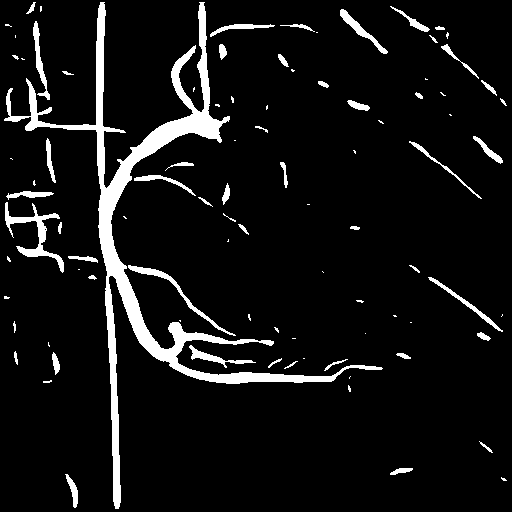}
      \begin{small}
        \put(67,5){\color{white}{(D5)}}
      \end{small}
    \end{overpic}
  }
  \subfigure[]{
    \hspace{-0.3cm}
    \begin{overpic}[width=0.12\linewidth]{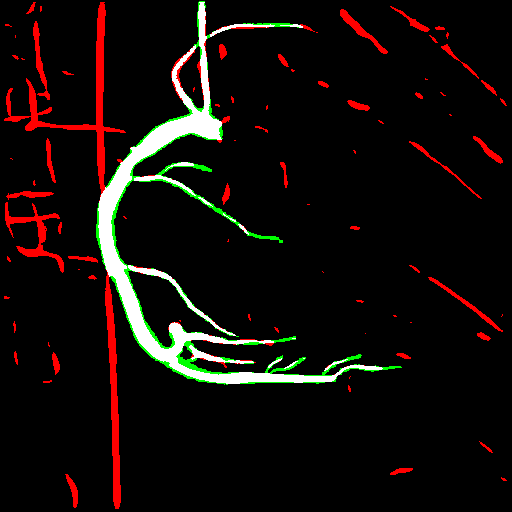}
      \begin{small}
        \put(67,5){\color{white}{(d5)}}
      \end{small}
    \end{overpic}
  }

  \vspace{-0.95cm}
  \subfigure[]{
    \begin{overpic}[width=0.12\linewidth]{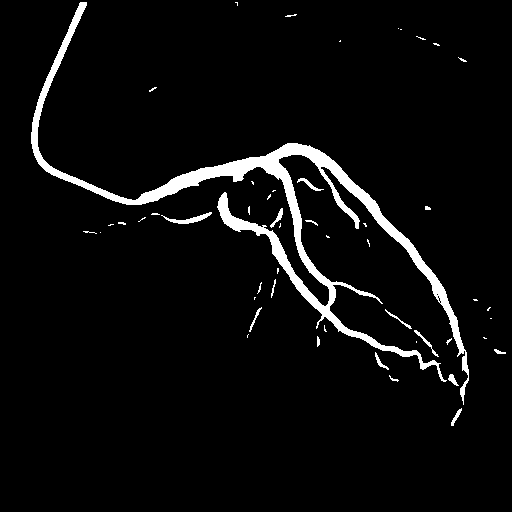}
      \begin{small}
        \put(3,5){\color{white}{(A6)}}
      \end{small}
    \end{overpic}
  }
  \subfigure[]{
    \hspace{-0.3cm}
    \begin{overpic}[width=0.12\linewidth]{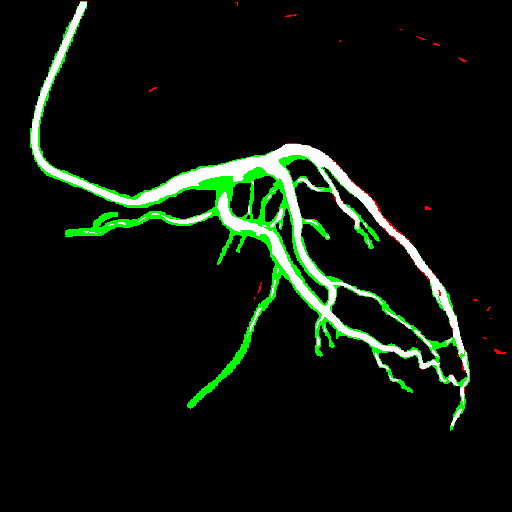}
      \begin{small}
        \put(3,5){\color{white}{(a6)}}
      \end{small}
    \end{overpic}
  }
  \subfigure[]{
    \hspace{-0.3cm}
    \begin{overpic}[width=0.12\linewidth]{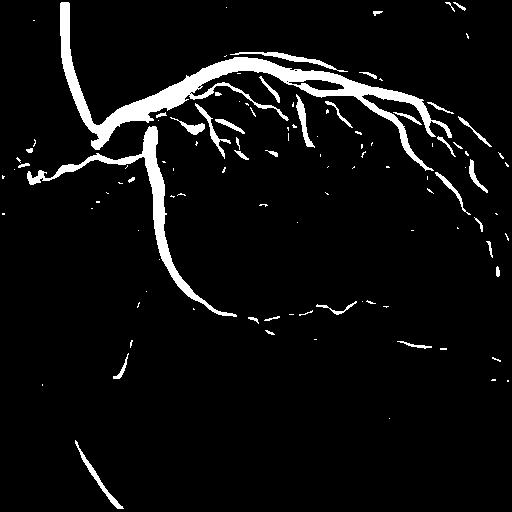}
      \begin{small}
        \put(67,5){\color{white}{(B6)}}
      \end{small}
    \end{overpic}
  }
  \subfigure[]{
    \hspace{-0.3cm}
    \begin{overpic}[width=0.12\linewidth]{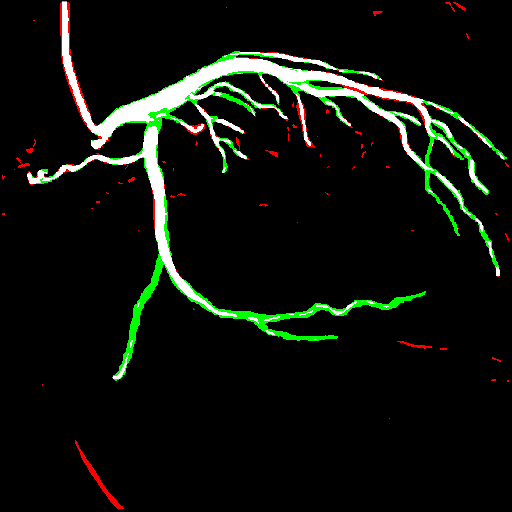}
      \begin{small}
        \put(67,5){\color{white}{(b6)}}
      \end{small}
    \end{overpic}
  }
  \subfigure[]{
    \hspace{-0.3cm}
    \begin{overpic}[width=0.12\linewidth]{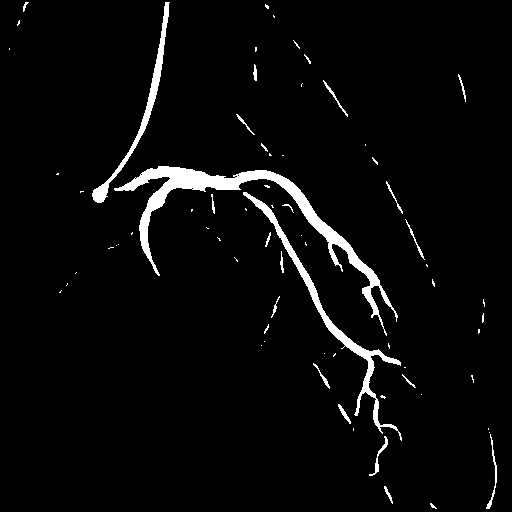}
      \begin{small}
        \put(3,5){\color{white}{(C6)}}
      \end{small}
    \end{overpic}
  }
  \subfigure[]{
    \hspace{-0.3cm}
    \begin{overpic}[width=0.12\linewidth]{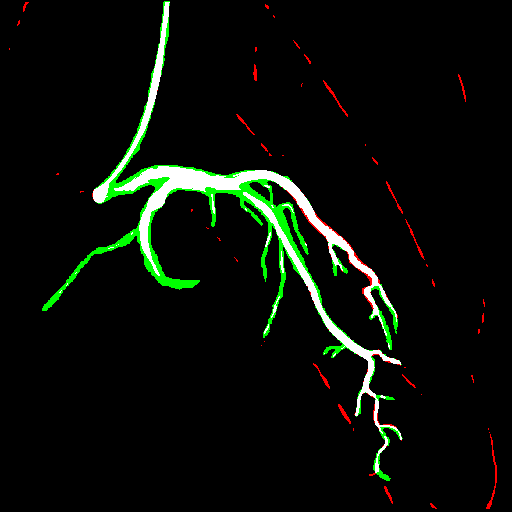}
      \begin{small}
        \put(3,5){\color{white}{(c6)}}
      \end{small}
    \end{overpic}
  }
  \subfigure[]{
    \hspace{-0.3cm}
    \begin{overpic}[width=0.12\linewidth]{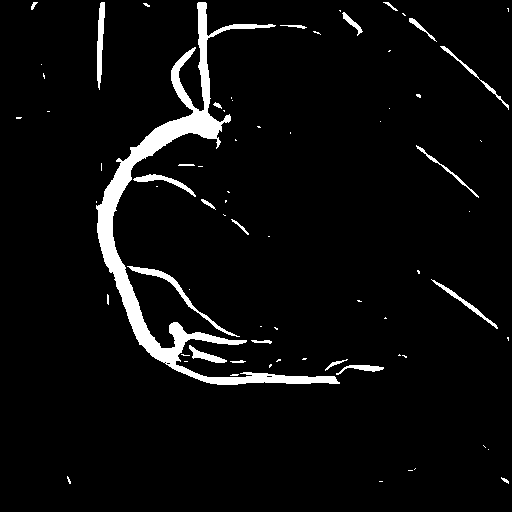}
      \begin{small}
        \put(67,5){\color{white}{(D6)}}
      \end{small}
    \end{overpic}
  }
  \subfigure[]{
    \hspace{-0.3cm}
    \begin{overpic}[width=0.12\linewidth]{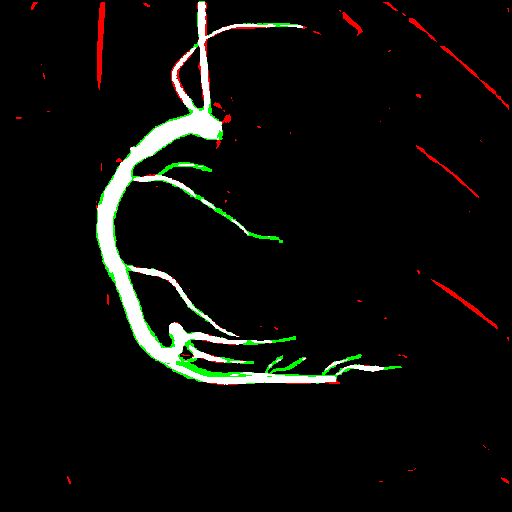}
      \begin{small}
        \put(67,5){\color{white}{(d6)}}
      \end{small}
    \end{overpic}
  }

  \label{fig10}
  \vspace{-0.8cm}
  \caption{Vessel segmentation results on raw XCA images ($1^{st},3^{rd},5^{th}$ row) and vessel layer images extracted by TV-TRPCA ($2^{nd},4^{th},6^{th}$ row) by 3 commonly used methods. Each group of results contains a segmentation image labeled by capital letters (A-D) and its contrast image with the ground truth labeled by lowercase letters (a-d). (1,2) Frangi filtering. (3,4) DSA. (5,6) U-net.}
\end{figure*}

The segmentation results on clinical XCA images of these three methods with different input are shown in \hyperref[fig10]{Fig.10}. And the accuracy metrics are recorded in \hyperref[tab4]{Tab.4}. With the raw XCA image as input, all methods suffer from the problem of irrelevant structures and tissue residue, some noise is also amplified. Especially, for U-net method, it identifies most of the tubular objects as segments of blood vessels, which will seriously reduce the segmentation precision. After changing the vessel layer image as input, the problem of irrelevant residue has been greatly improved, and the tubular structure of coronary is clearer. However, some noise interference is still difficult to eliminate completely, which will affect the observation of the whole coronary structure.

\subsection{Vessel Segmentation Experiment} \label{TSRG}

\subsubsection{Parameter Optimization}

In the first stage of TSRG vessel segmentation algorithm, the threshold method (TM) is applied to delete residual low gray level noise. To determine the optimal threshold method and parameter, the contrast test was designed as follows. In the first group, $91\%-99\%$ of the maximum gray intensity was used as the global threshold. The second group included global thresholding algorithm OTSU and local adaptive thresholding algorithm based on mean value and Gaussian window\footnote{\url{https://docs.opencv.org/5.x/d7/d4d/tutorial_py_thresholding.html}}. The parameters of mean value-based method is optimized to $blockSize=31, c=4$, and the parameters of Gaussian window-based method is optimized to $blockSize=51, c=4$.

\begin{table*}[!t]
  \renewcommand{\arraystretch}{1}
  \centering
  \topcaption{The mean recall, precision and F-Measure values ($\pm$ standard deviation) of 3 vessel segmentation methods on raw image and vessel layer}
  \label{tab4}
  \begin{small}
    \begin{tabular}{l|cc|cc|cc}
      \toprule \toprule
      \multirow{2}{*}{\textbf{Metrics}} & \multicolumn{2}{c|}{\textbf{Frangi filtering}} & \multicolumn{2}{c|}{\textbf{DSA}} & \multicolumn{2}{c}{\textbf{U-net}}                                                                                                 \\
      \cmidrule(l){2-7}
                                        & Raw image                                      & Vessel layer                      & Raw image                         & Vessel layer                  & Raw image                     & Vessel layer                  \\
      \midrule
      recall                            & 0.730 $\pm$ 0.062                              & \pmb{0.753} $\pm$ \pmb{0.047}     & 0.786 $\pm$ 0.033                 & \pmb{0.837} $\pm$ \pmb{0.051} & \pmb{0.665} $\pm$ \pmb{0.055} & 0.613 $\pm$ 0.096             \\
      precision                         & 0.747 $\pm$ 0.133                              & \pmb{0.863} $\pm$ \pmb{0.055}     & 0.688 $\pm$ 0.096                 & \pmb{0.819} $\pm$ \pmb{0.078} & 0.574 $\pm$ 0.128             & \pmb{0.847} $\pm$ \pmb{0.069} \\
      F-Measure                         & 0.729 $\pm$ 0.052                              & \pmb{0.802} $\pm$ \pmb{0.025}     & 0.730 $\pm$ 0.053                 & \pmb{0.824} $\pm$ \pmb{0.029} & 0.609 $\pm$ 0.073             & \pmb{0.706} $\pm$ \pmb{0.058} \\
      \bottomrule \bottomrule
    \end{tabular}
  \end{small}
\end{table*}

\begin{figure*}[!t]
  \centering
  \subfigure[]{
    \begin{overpic}[width=0.12\linewidth]{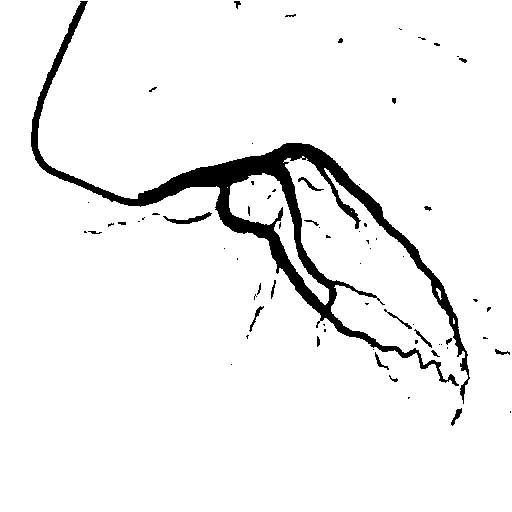}
      \begin{small}
        \put(3,5){\color{black}{(a1)}}
        \put(70,85){\color{blue}{91\%}}
      \end{small}
    \end{overpic}
  }
  \subfigure[]{
    \hspace{-0.3cm}
    \begin{overpic}[width=0.12\linewidth]{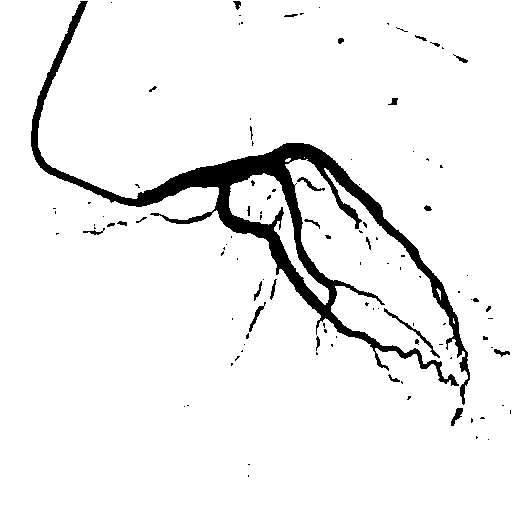}
      \begin{small}
        \put(3,5){\color{black}{(b1)}}
        \put(70,85){\color{blue}{93\%}}
      \end{small}
    \end{overpic}
  }
  \subfigure[]{
    \hspace{-0.3cm}
    \begin{overpic}[width=0.12\linewidth]{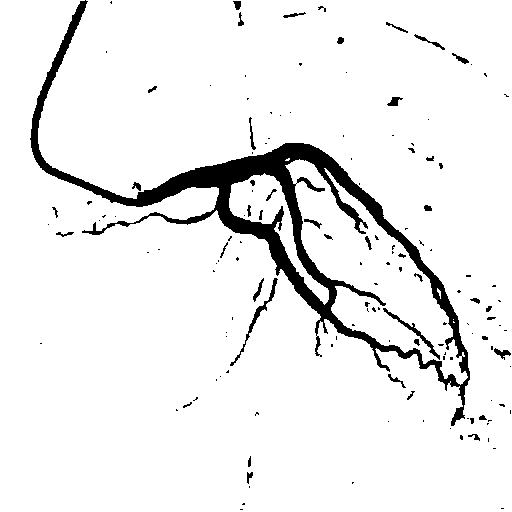}
      \begin{small}
        \put(3,5){\color{black}{(c1)}}
        \put(70,85){\color{blue}{95\%}}
      \end{small}
    \end{overpic}
  }
  \subfigure[]{
    \hspace{-0.3cm}
    \begin{overpic}[width=0.12\linewidth]{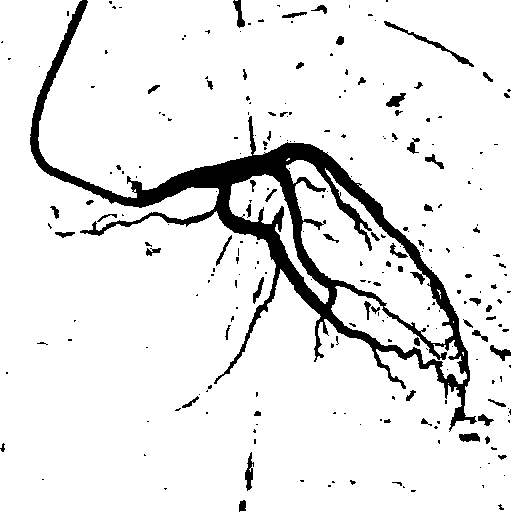}
      \begin{small}
        \put(3,5){\color{black}{(d1)}}
        \put(70,85){\color{blue}{97\%}}
      \end{small}
    \end{overpic}
  }
  \subfigure[]{
    \hspace{-0.3cm}
    \begin{overpic}[width=0.12\linewidth]{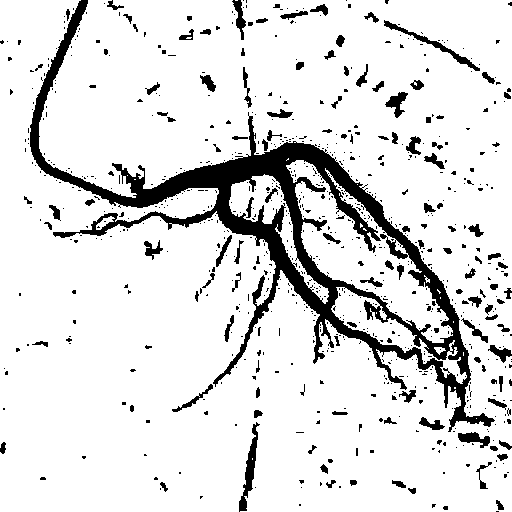}
      \begin{small}
        \put(3,5){\color{black}{(e1)}}
        \put(70,85){\color{blue}{99\%}}
      \end{small}
    \end{overpic}
  }
  \subfigure[]{
    \hspace{-0.3cm}
    \begin{overpic}[width=0.12\linewidth]{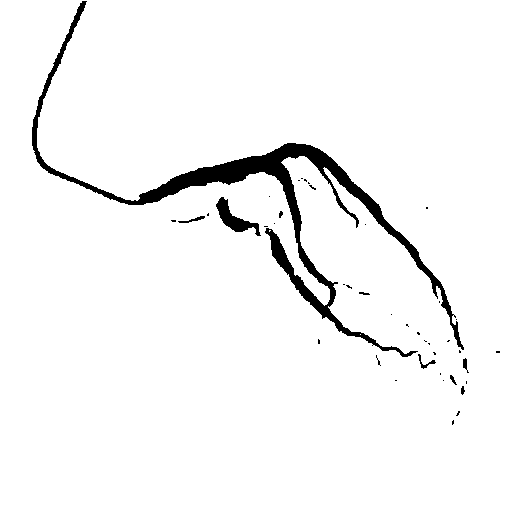}
      \begin{small}
        \put(3,5){\color{black}{(f1)}}
        \put(56,85){\color{blue}{OTSU}}
      \end{small}
    \end{overpic}
  }
  \subfigure[]{
    \hspace{-0.3cm}
    \begin{overpic}[width=0.12\linewidth]{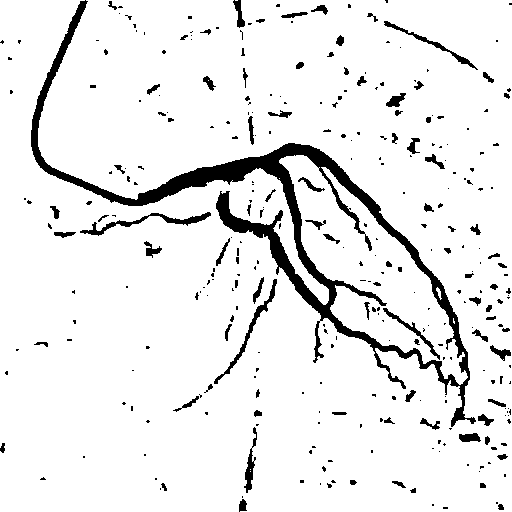}
      \begin{small}
        \put(3,5){\color{black}{(g1)}}
        \put(62,85){\color{blue}{Mean}}
      \end{small}
    \end{overpic}
  }
  \subfigure[]{
    \hspace{-0.3cm}
    \begin{overpic}[width=0.12\linewidth]{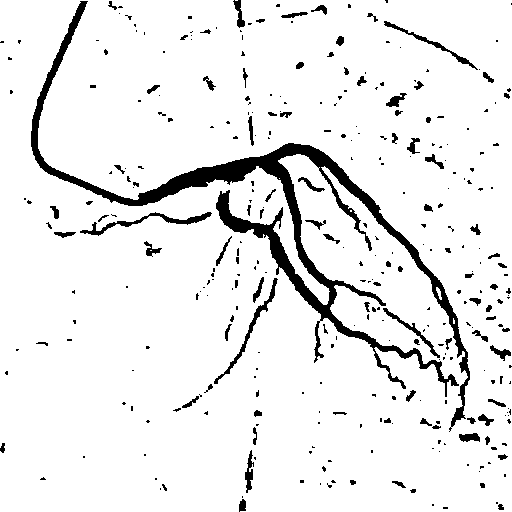}
      \begin{small}
        \put(3,5){\color{black}{(h1)}}
        \put(39,85){\color{blue}{Gaussian}}
      \end{small}
    \end{overpic}
  }

  \vspace{-0.95cm}
  \subfigure[]{
    \begin{overpic}[width=0.12\linewidth]{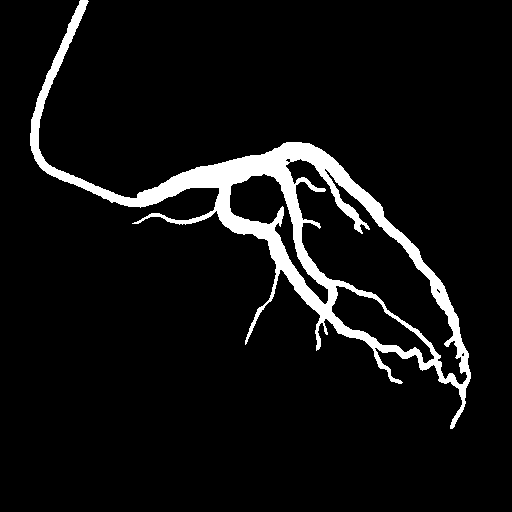}
      \begin{small}
        \put(3,5){\color{white}{(a2)}}
        \put(70,85){\color{yellow}{91\%}}
      \end{small}
    \end{overpic}
  }
  \subfigure[]{
    \hspace{-0.3cm}
    \begin{overpic}[width=0.12\linewidth]{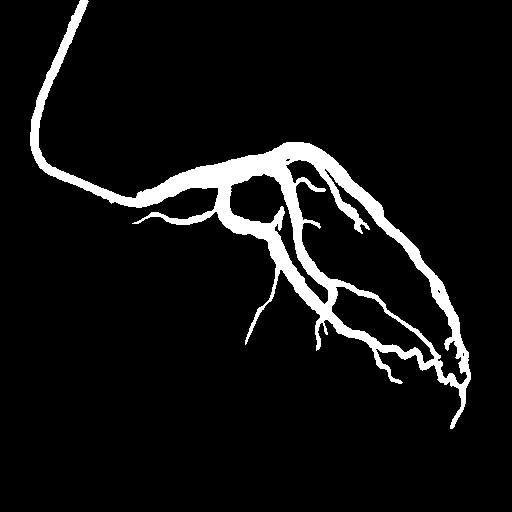}
      \begin{small}
        \put(3,5){\color{white}{(b2)}}
        \put(70,85){\color{yellow}{93\%}}
      \end{small}
    \end{overpic}
  }
  \subfigure[]{
    \hspace{-0.3cm}
    \begin{overpic}[width=0.12\linewidth]{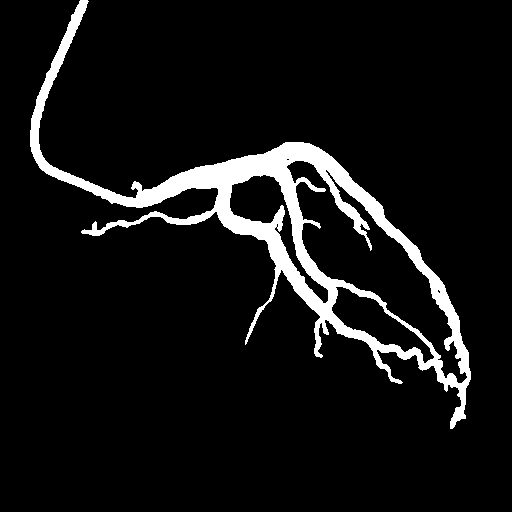}
      \begin{small}
        \put(3,5){\color{white}{(c2)}}
        \put(70,85){\color{yellow}{95\%}}
      \end{small}
    \end{overpic}
  }
  \subfigure[]{
    \hspace{-0.3cm}
    \begin{overpic}[width=0.12\linewidth]{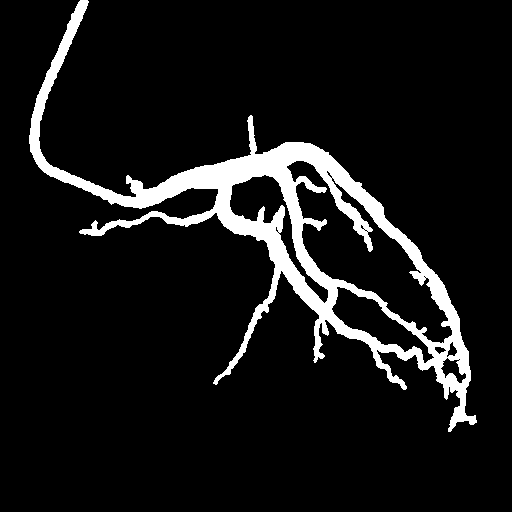}
      \begin{small}
        \put(3,5){\color{white}{(d2)}}
        \put(70,85){\color{yellow}{97\%}}
      \end{small}
    \end{overpic}
  }
  \subfigure[]{
    \hspace{-0.3cm}
    \begin{overpic}[width=0.12\linewidth]{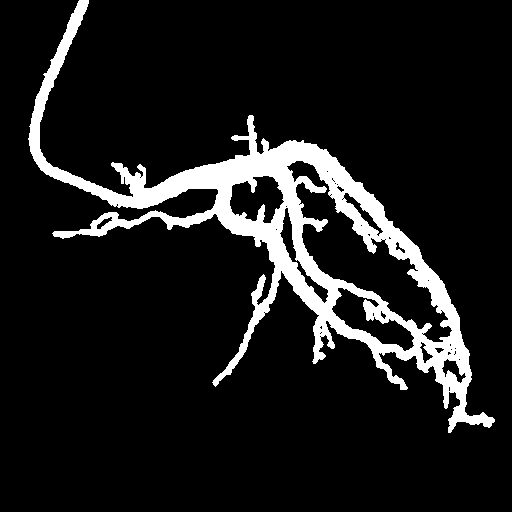}
      \begin{small}
        \put(3,5){\color{white}{(e2)}}
        \put(70,85){\color{yellow}{99\%}}
      \end{small}
    \end{overpic}
  }
  \subfigure[]{
    \hspace{-0.3cm}
    \begin{overpic}[width=0.12\linewidth]{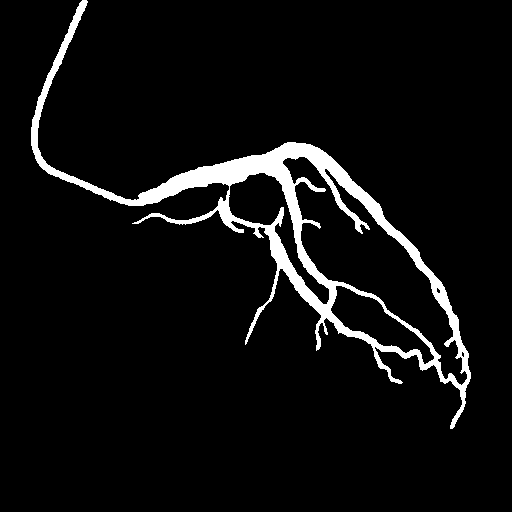}
      \begin{small}
        \put(3,5){\color{white}{(f2)}}
        \put(56,85){\color{yellow}{OTSU}}
      \end{small}
    \end{overpic}
  }
  \subfigure[]{
    \hspace{-0.3cm}
    \begin{overpic}[width=0.12\linewidth]{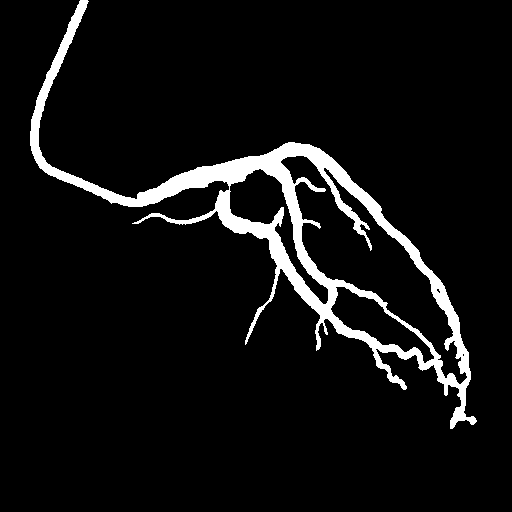}
      \begin{small}
        \put(3,5){\color{white}{(g2)}}
        \put(62,85){\color{yellow}{Mean}}
      \end{small}
    \end{overpic}
  }
  \subfigure[]{
    \hspace{-0.3cm}
    \begin{overpic}[width=0.12\linewidth]{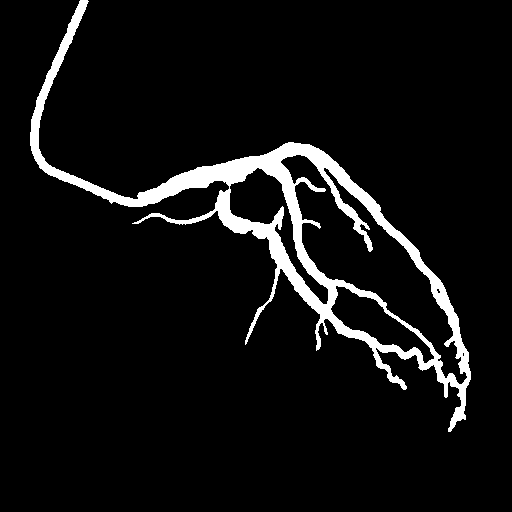}
      \begin{small}
        \put(3,5){\color{white}{(h2)}}
        \put(39,85){\color{yellow}{Gaussian}}
      \end{small}
    \end{overpic}
  }

  \vspace{-0.95cm}
  \subfigure[]{
    \begin{overpic}[width=0.12\linewidth]{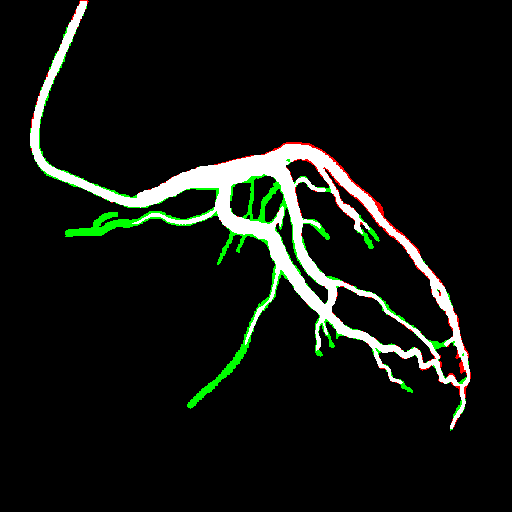}
      \begin{small}
        \put(3,5){\color{white}{(a3)}}
        \put(70,85){\color{yellow}{91\%}}
      \end{small}
    \end{overpic}
  }
  \subfigure[]{
    \hspace{-0.3cm}
    \begin{overpic}[width=0.12\linewidth]{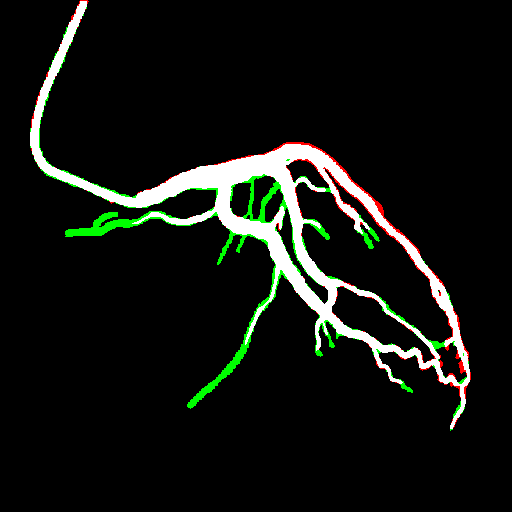}
      \begin{small}
        \put(3,5){\color{white}{(b3)}}
        \put(70,85){\color{yellow}{93\%}}
      \end{small}
    \end{overpic}
  }
  \subfigure[]{
    \hspace{-0.3cm}
    \begin{overpic}[width=0.12\linewidth]{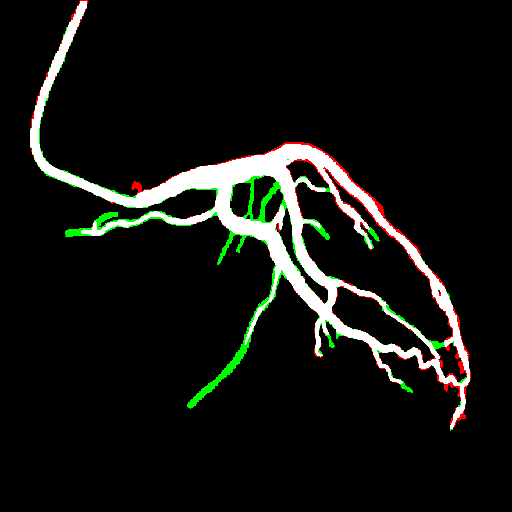}
      \begin{small}
        \put(3,5){\color{white}{(c3)}}
        \put(70,85){\color{yellow}{95\%}}
      \end{small}
    \end{overpic}
  }
  \subfigure[]{
    \hspace{-0.3cm}
    \begin{overpic}[width=0.12\linewidth]{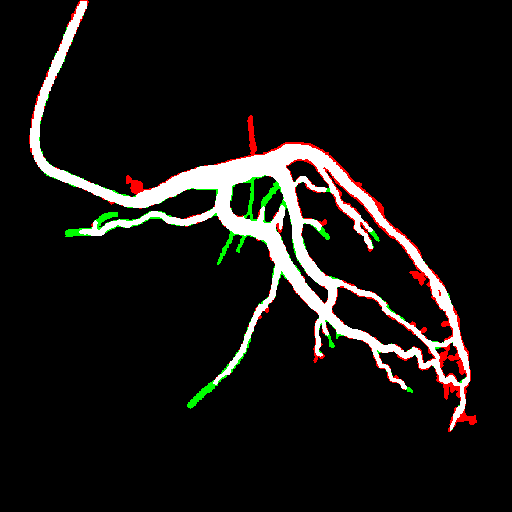}
      \begin{small}
        \put(3,5){\color{white}{(d3)}}
        \put(70,85){\color{yellow}{97\%}}
      \end{small}
    \end{overpic}
  }
  \subfigure[]{
    \hspace{-0.3cm}
    \begin{overpic}[width=0.12\linewidth]{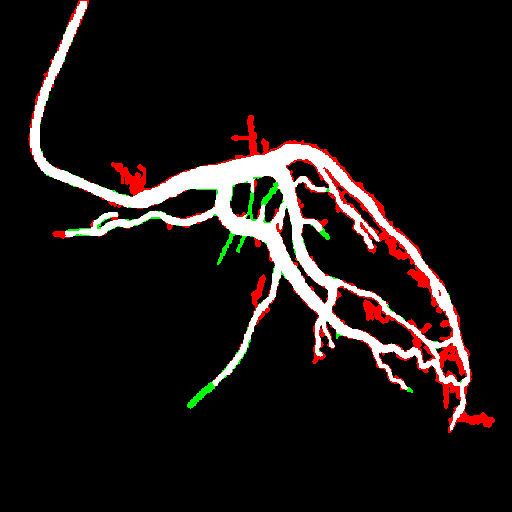}
      \begin{small}
        \put(3,5){\color{white}{(e3)}}
        \put(70,85){\color{yellow}{99\%}}
      \end{small}
    \end{overpic}
  }
  \subfigure[]{
    \hspace{-0.3cm}
    \begin{overpic}[width=0.12\linewidth]{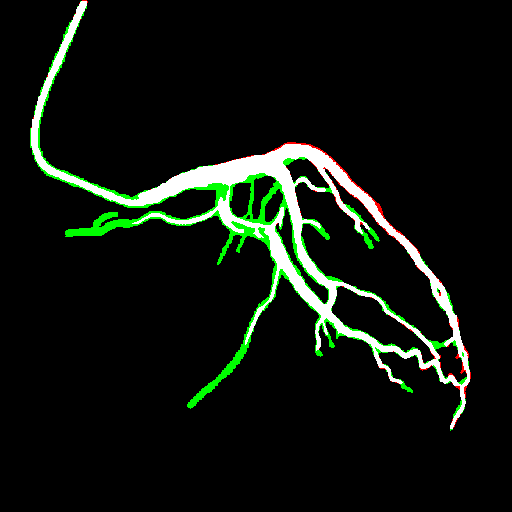}
      \begin{small}
        \put(3,5){\color{white}{(f3)}}
        \put(56,85){\color{yellow}{OTSU}}
      \end{small}
    \end{overpic}
  }
  \subfigure[]{
    \hspace{-0.3cm}
    \begin{overpic}[width=0.12\linewidth]{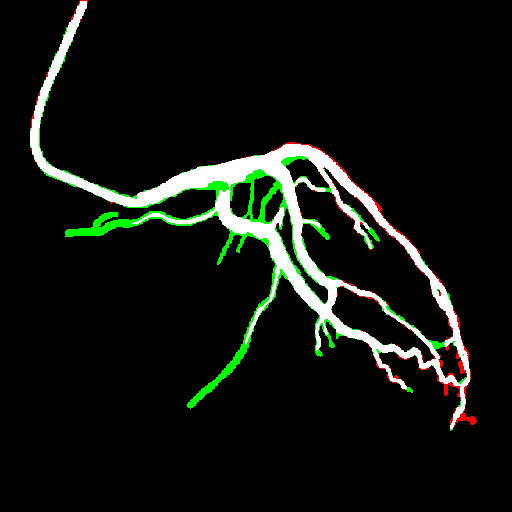}
      \begin{small}
        \put(3,5){\color{white}{(g3)}}
        \put(62,85){\color{yellow}{Mean}}
      \end{small}
    \end{overpic}
  }
  \subfigure[]{
    \hspace{-0.3cm}
    \begin{overpic}[width=0.12\linewidth]{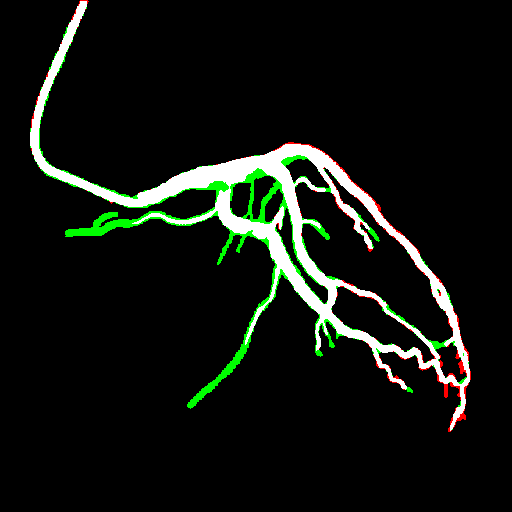}
      \begin{small}
        \put(3,5){\color{white}{(h3)}}
        \put(39,85){\color{yellow}{Gaussian}}
      \end{small}
    \end{overpic}
  }

  \label{fig11}
  \vspace{-0.8cm}
  \caption{Parameter optimization experiment for TSRG. (a1)-(e1) are the initial segmentation results as threshold varies from 91\% - 99\% of maximum gray intensity. (f1-h1) are three other commonly used threshold method OTSU, mean value-based and Gaussian window-based. (a2)-(h2) and (a3)-(h3) are the final segmentation results by TSRG and their contrast images with the ground truth.}
\end{figure*}

\begin{figure*}[!t]
  \centering
  \subfigure[]{\includegraphics[width=\linewidth]{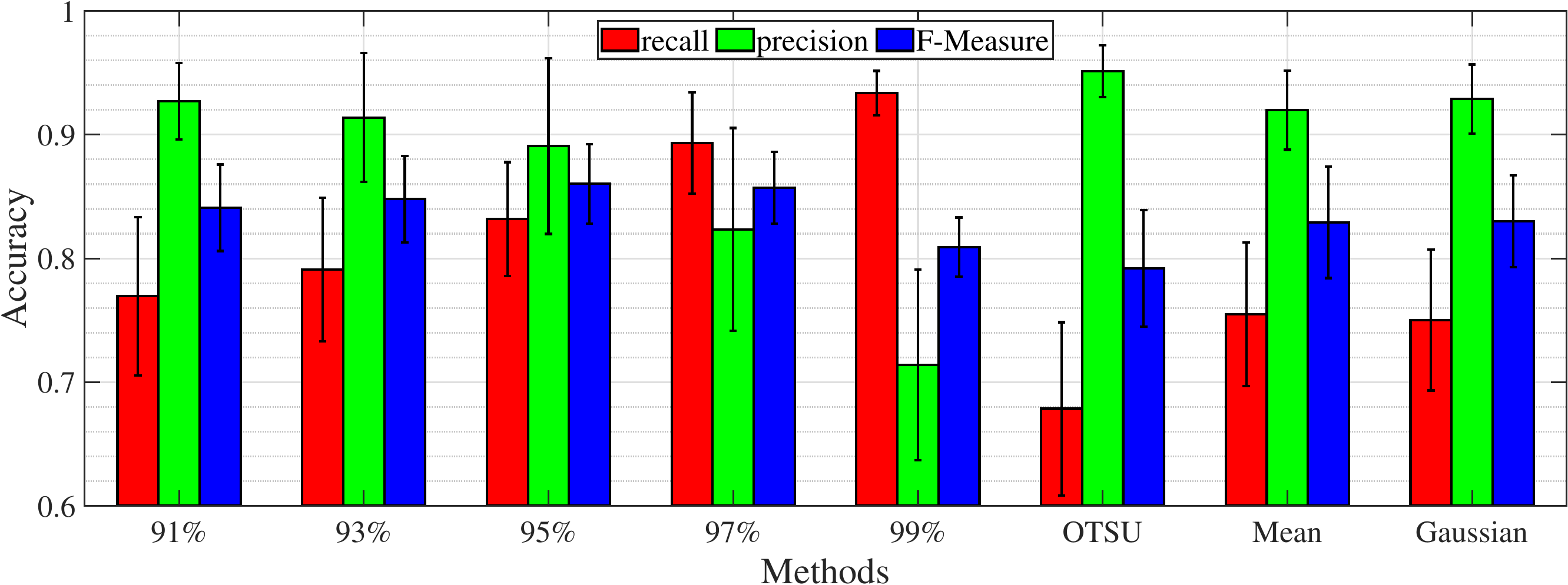}}

  \label{fig12}
  \vspace{-0.8cm}
  \caption{Variations of recall, precision and F-Measure values in the TSRG parameter optimization experiment.}
\end{figure*}

The initial and final segmentation results on clinical XCA image sequences are shown in \hyperref[fig11]{Fig.11.(a)-(h)}. Quantitative analysis of segmentation accuracy is shown in \hyperref[fig12]{Fig.12}. Comprehensively considering the topological correctness of final binary vessel mask and the variation trend of each metrics, global method with 95\% of the maximum gray intensity as the threshold is selected as the preprocessing of first stage region growing. 

\begin{figure*}[htbp]
  \centering
  \vspace{-2cm}
  \hspace{-0.1cm}
  \textbf{Ground Truth} \ \ \ \ \ \ \ \ \ \ \ \ \ \textbf{VGN} \ \ \ \ \ \ \ \ \ \ \ \ \ \ \ \ \ \ \textbf{SVS-net} \ \ \ \ \ \ \ \ \ \ \ \ \ \ \textbf{TV-T+RG} \ \ \ \ \ \ \ \textbf{TV-T+TM+RG} \ \ \ \ \ \textbf{TV-T+TSRG}

  \subfigure[]{
    \begin{overpic}[width=0.16\linewidth]{Origin_0137.png}
      \put(3,5){\color{white}{(A1)}}
    \end{overpic}
  }
  \subfigure[]{
    \hspace{-0.28cm}
    \begin{overpic}[width=0.16\linewidth]{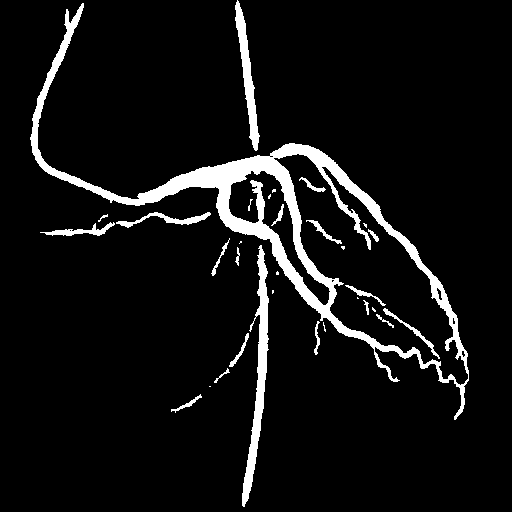}
      \put(3,5){\color{white}{(B1)}}
    \end{overpic}
  }
  \subfigure[]{
    \hspace{-0.28cm}
    \begin{overpic}[width=0.16\linewidth]{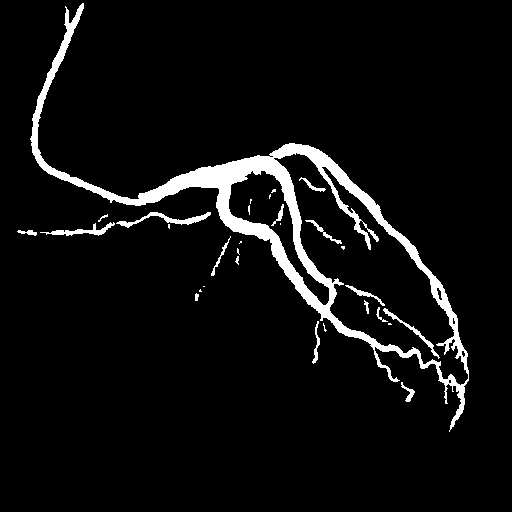}
      \put(3,5){\color{white}{(C1)}}
    \end{overpic}
  }
  \subfigure[]{
    \hspace{-0.28cm}
    \begin{overpic}[width=0.16\linewidth]{tvtrpca_0137_bw.png}
      \put(3,5){\color{white}{(D1)}}
    \end{overpic}
  }
  \subfigure[]{
    \hspace{-0.28cm}
    \begin{overpic}[width=0.16\linewidth]{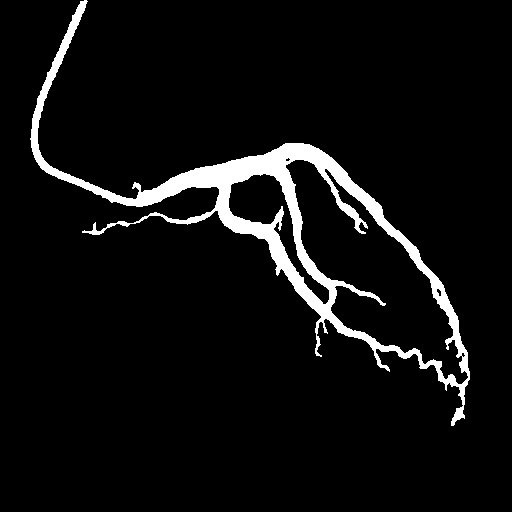}
      \put(3,5){\color{white}{(E1)}}
    \end{overpic}
  }
  \subfigure[]{
    \hspace{-0.28cm}
    \begin{overpic}[width=0.16\linewidth]{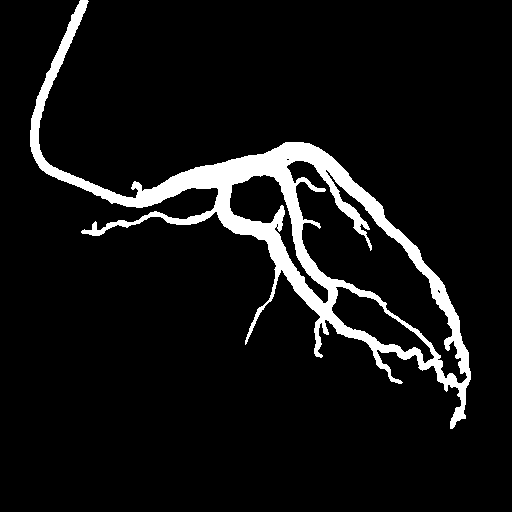}
      \put(3,5){\color{white}{(F1)}}
    \end{overpic}
  }

  \vspace{-0.95cm}
  \subfigure[]{
    \begin{overpic}[width=0.16\linewidth]{GTBW_0137.png}
      \put(3,5){\color{white}{(a1)}}
    \end{overpic}
  }
  \subfigure[]{
    \hspace{-0.28cm}
    \begin{overpic}[width=0.16\linewidth]{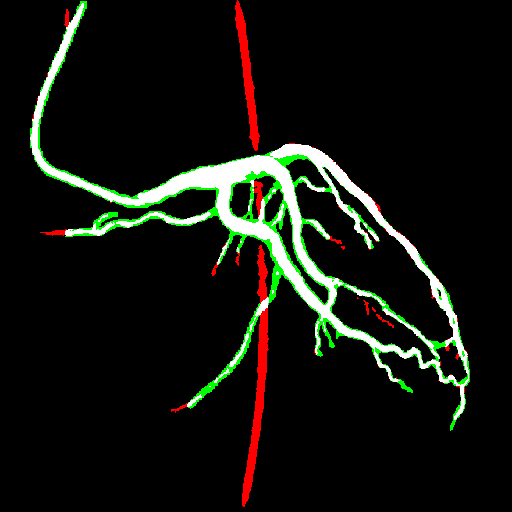}
      \put(3,5){\color{white}{(b1)}}
    \end{overpic}
  }
  \subfigure[]{
    \hspace{-0.28cm}
    \begin{overpic}[width=0.16\linewidth]{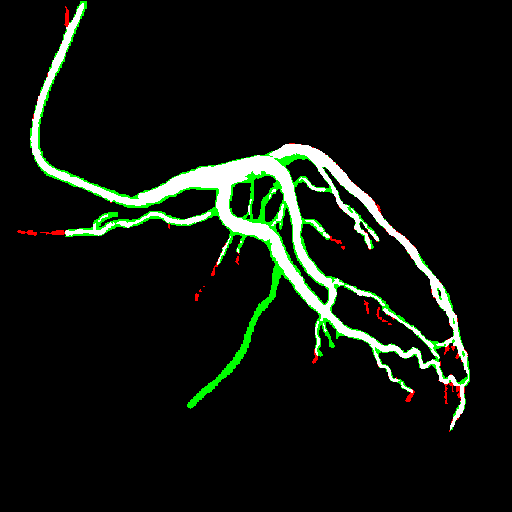}
      \put(3,5){\color{white}{(c1)}}
    \end{overpic}
  }
  \subfigure[]{
    \hspace{-0.28cm}
    \begin{overpic}[width=0.16\linewidth]{tvtrpca_0137_gt.png}
      \put(3,5){\color{white}{(d1)}}
    \end{overpic}
  }
  \subfigure[]{
    \hspace{-0.28cm}
    \begin{overpic}[width=0.16\linewidth]{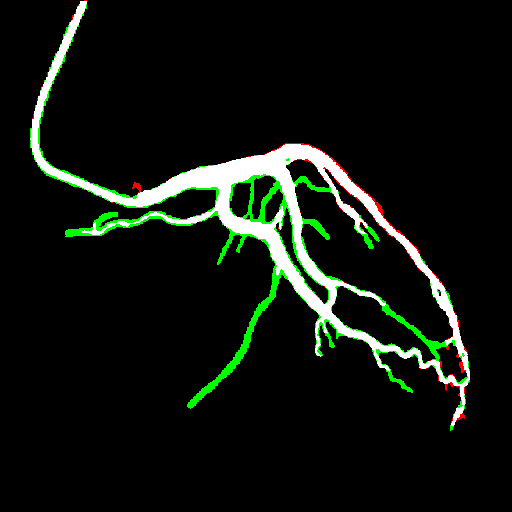}
      \put(3,5){\color{white}{(e1)}}
    \end{overpic}
  }
  \subfigure[]{
    \hspace{-0.28cm}
    \begin{overpic}[width=0.16\linewidth]{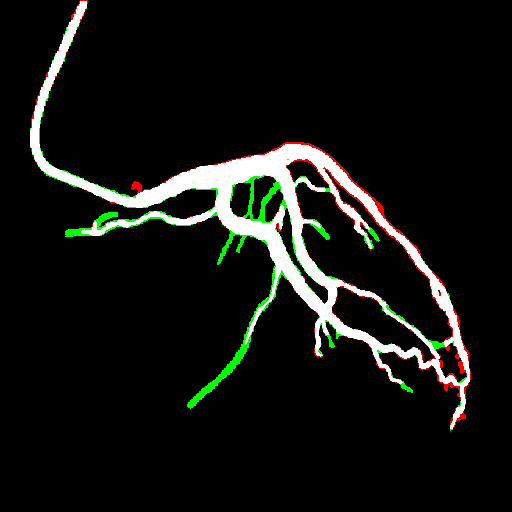}
      \put(3,5){\color{white}{(f1)}}
    \end{overpic}
  }

  \vspace{-0.85cm}
  \subfigure[]{
    \begin{overpic}[width=0.16\linewidth]{Origin_0449.png}
      \put(71,5){\color{white}{(A2)}}
    \end{overpic}
  }
  \subfigure[]{
    \hspace{-0.28cm}
    \begin{overpic}[width=0.16\linewidth]{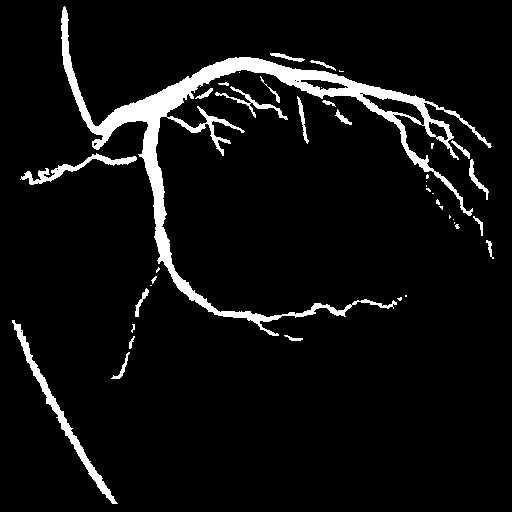}
      \put(71,5){\color{white}{(B2)}}
    \end{overpic}
  }
  \subfigure[]{
    \hspace{-0.28cm}
    \begin{overpic}[width=0.16\linewidth]{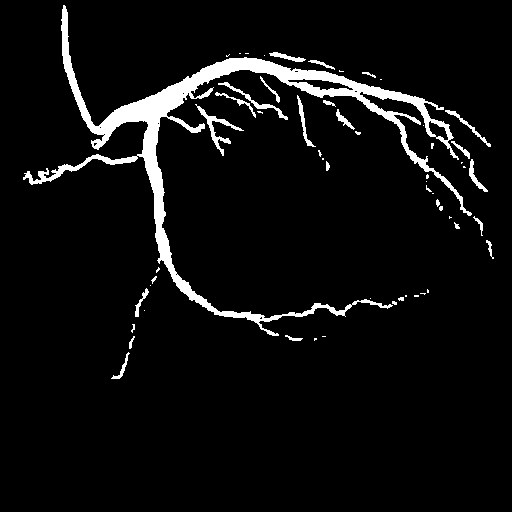}
      \put(71,5){\color{white}{(C2)}}
    \end{overpic}
  }
  \subfigure[]{
    \hspace{-0.28cm}
    \begin{overpic}[width=0.16\linewidth]{tvtrpca_0449_bw.png}
      \put(71,5){\color{white}{(D2)}}
    \end{overpic}
  }
  \subfigure[]{
    \hspace{-0.28cm}
    \begin{overpic}[width=0.16\linewidth]{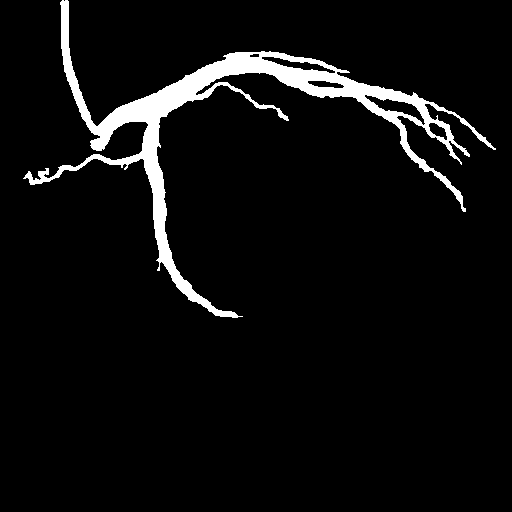}
      \put(71,5){\color{white}{(E2)}}
    \end{overpic}
  }
  \subfigure[]{
    \hspace{-0.28cm}
    \begin{overpic}[width=0.16\linewidth]{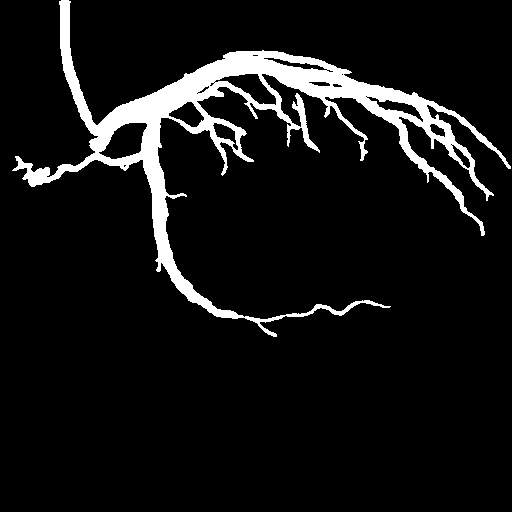}
      \put(71,5){\color{white}{(F2)}}
    \end{overpic}
  }

  \vspace{-0.95cm}
  \subfigure[]{
    \begin{overpic}[width=0.16\linewidth]{GTBW_0449.png}
      \put(71,5){\color{white}{(a2)}}
    \end{overpic}
  }
  \subfigure[]{
    \hspace{-0.28cm}
    \begin{overpic}[width=0.16\linewidth]{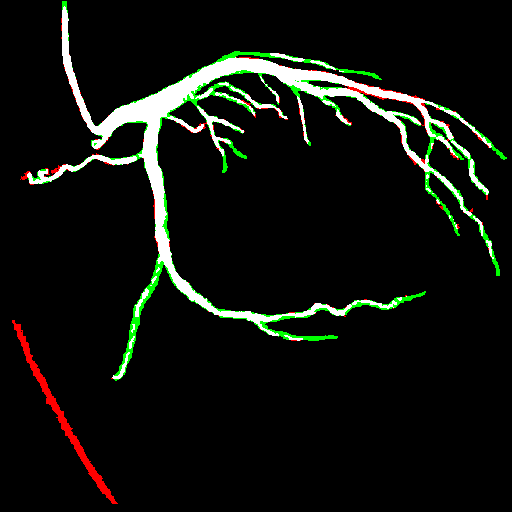}
      \put(71,5){\color{white}{(b2)}}
    \end{overpic}
  }
  \subfigure[]{
    \hspace{-0.28cm}
    \begin{overpic}[width=0.16\linewidth]{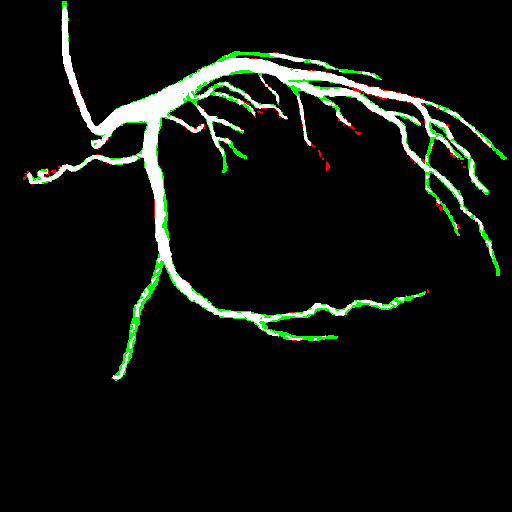}
      \put(71,5){\color{white}{(c2)}}
    \end{overpic}
  }
  \subfigure[]{
    \hspace{-0.28cm}
    \begin{overpic}[width=0.16\linewidth]{tvtrpca_0449_gt.png}
      \put(71,5){\color{white}{(d2)}}
    \end{overpic}
  }
  \subfigure[]{
    \hspace{-0.28cm}
    \begin{overpic}[width=0.16\linewidth]{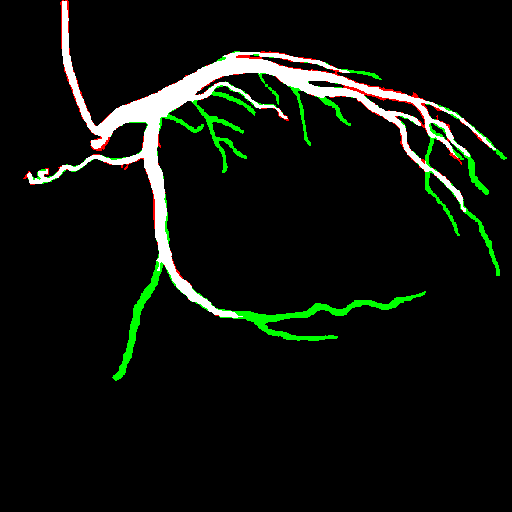}
      \put(71,5){\color{white}{(e2)}}
    \end{overpic}
  }
  \subfigure[]{
    \hspace{-0.28cm}
    \begin{overpic}[width=0.16\linewidth]{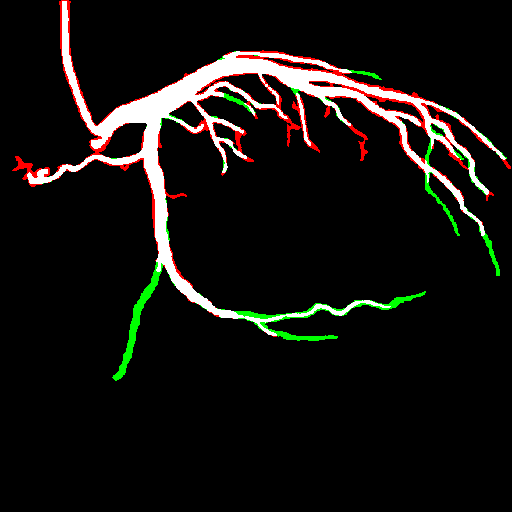}
      \put(71,5){\color{white}{(f2)}}
    \end{overpic}
  }

  \vspace{-0.85cm}
  \subfigure[]{
    \begin{overpic}[width=0.16\linewidth]{Origin_0544.png}
      \put(3,5){\color{white}{(A3)}}
    \end{overpic}
  }
  \subfigure[]{
    \hspace{-0.28cm}
    \begin{overpic}[width=0.16\linewidth]{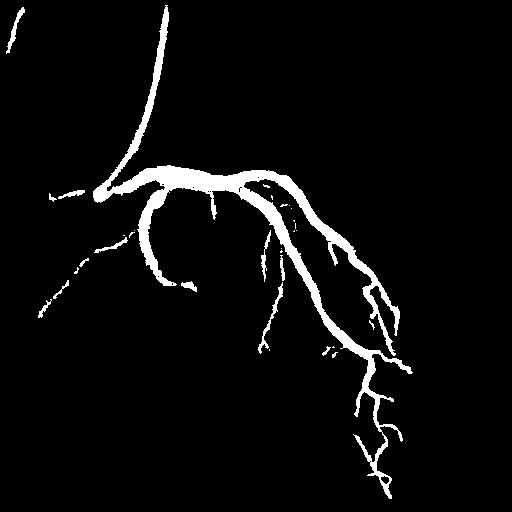}
      \put(3,5){\color{white}{(B3)}}
    \end{overpic}
  }
  \subfigure[]{
    \hspace{-0.28cm}
    \begin{overpic}[width=0.16\linewidth]{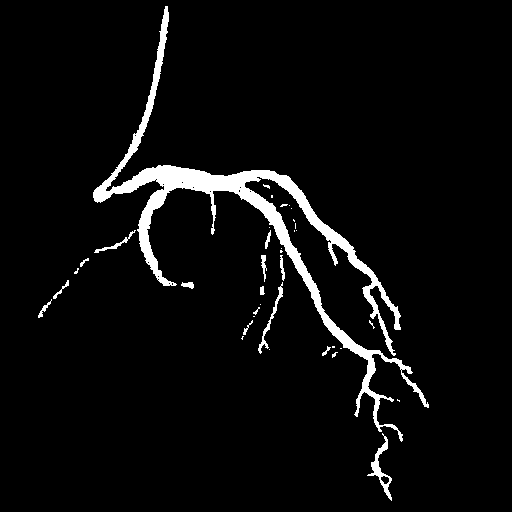}
      \put(3,5){\color{white}{(C3)}}
    \end{overpic}
  }
  \subfigure[]{
    \hspace{-0.28cm}
    \begin{overpic}[width=0.16\linewidth]{tvtrpca_0544_bw.png}
      \put(3,5){\color{white}{(D3)}}
    \end{overpic}
  }
  \subfigure[]{
    \hspace{-0.28cm}
    \begin{overpic}[width=0.16\linewidth]{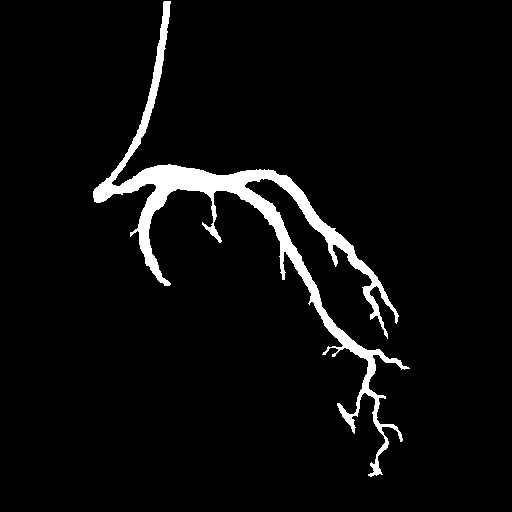}
      \put(3,5){\color{white}{(E3)}}
    \end{overpic}
  }
  \subfigure[]{
    \hspace{-0.28cm}
    \begin{overpic}[width=0.16\linewidth]{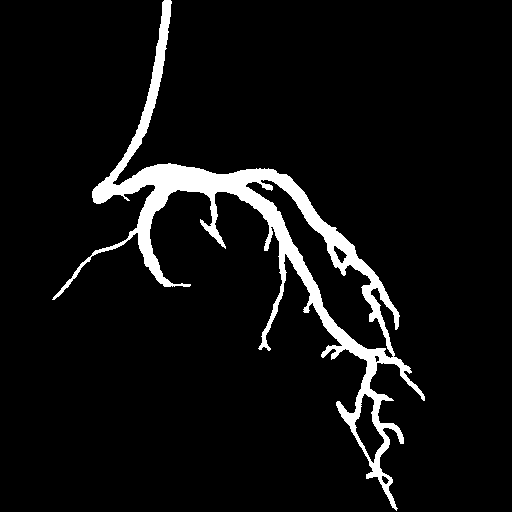}
      \put(3,5){\color{white}{(F3)}}
    \end{overpic}
  }

  \vspace{-0.95cm}
  \subfigure[]{
    \begin{overpic}[width=0.16\linewidth]{GTBW_0544.png}
      \put(3,5){\color{white}{(a3)}}
    \end{overpic}
  }
  \subfigure[]{
    \hspace{-0.28cm}
    \begin{overpic}[width=0.16\linewidth]{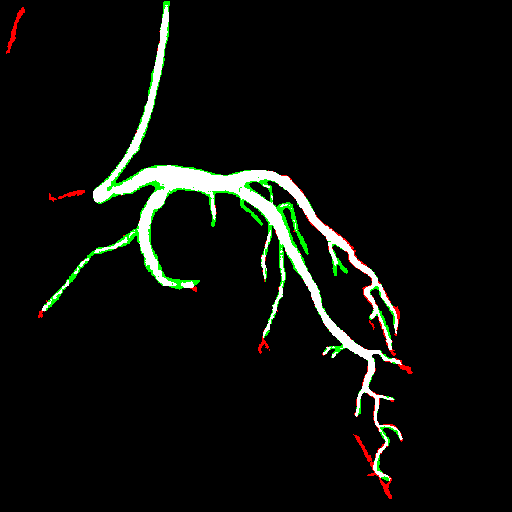}
      \put(3,5){\color{white}{(b3)}}
    \end{overpic}
  }
  \subfigure[]{
    \hspace{-0.28cm}
    \begin{overpic}[width=0.16\linewidth]{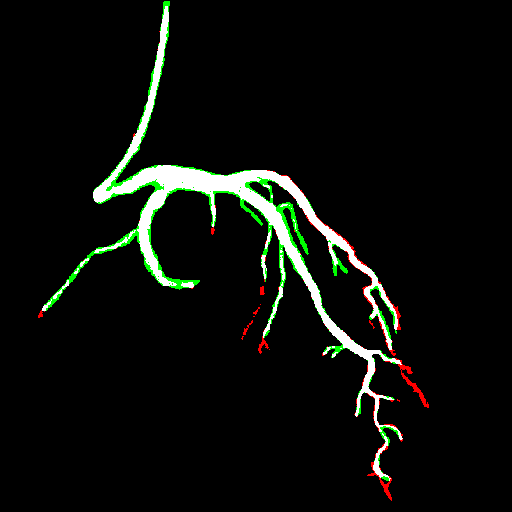}
      \put(3,5){\color{white}{(c3)}}
    \end{overpic}
  }
  \subfigure[]{
    \hspace{-0.28cm}
    \begin{overpic}[width=0.16\linewidth]{tvtrpca_0544_gt.png}
      \put(3,5){\color{white}{(d3)}}
    \end{overpic}
  }
  \subfigure[]{
    \hspace{-0.28cm}
    \begin{overpic}[width=0.16\linewidth]{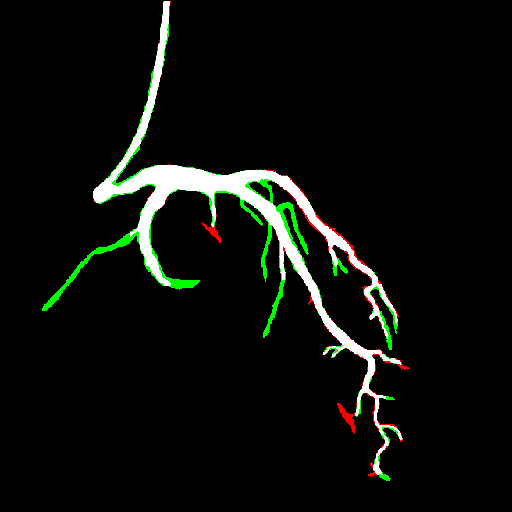}
      \put(3,5){\color{white}{(e3)}}
    \end{overpic}
  }
  \subfigure[]{
    \hspace{-0.28cm}
    \begin{overpic}[width=0.16\linewidth]{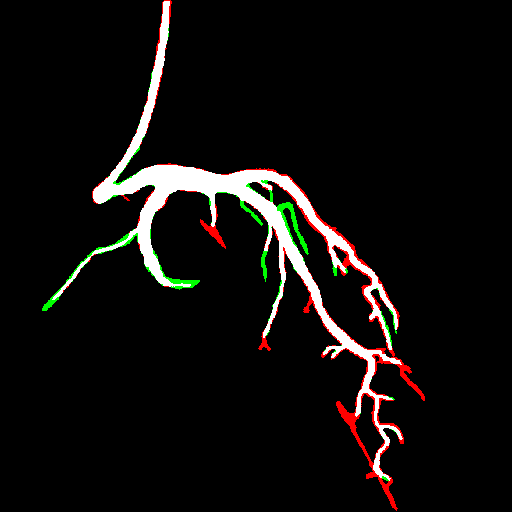}
      \put(3,5){\color{white}{(f3)}}
    \end{overpic}
  }

  \vspace{-0.85cm}
  \subfigure[]{
    \begin{overpic}[width=0.16\linewidth]{Origin_0747.png}
      \put(71,5){\color{white}{(A4)}}
    \end{overpic}
  }
  \subfigure[]{
    \hspace{-0.28cm}
    \begin{overpic}[width=0.16\linewidth]{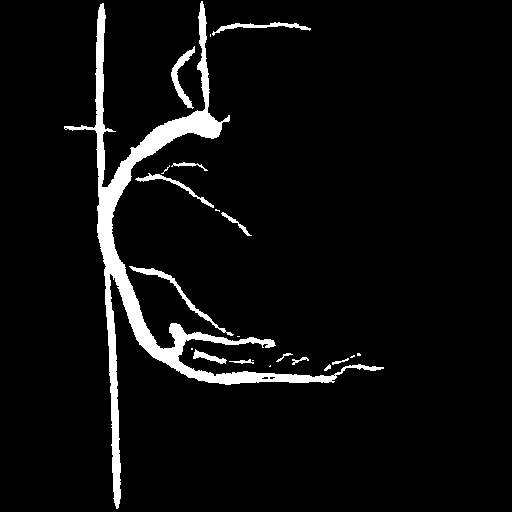}
      \put(71,5){\color{white}{(B4)}}
    \end{overpic}
  }
  \subfigure[]{
    \hspace{-0.28cm}
    \begin{overpic}[width=0.16\linewidth]{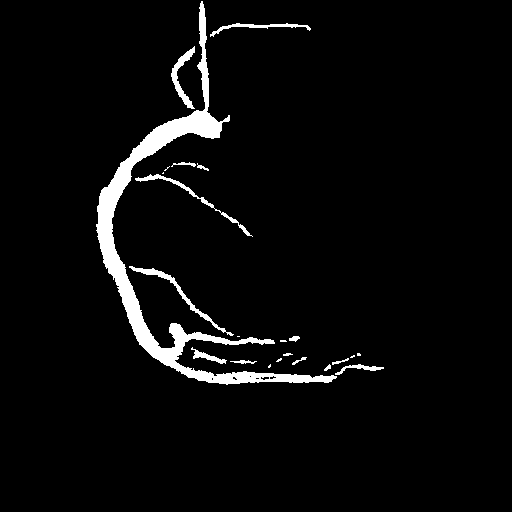}
      \put(71,5){\color{white}{(C4)}}
    \end{overpic}
  }
  \subfigure[]{
    \hspace{-0.28cm}
    \begin{overpic}[width=0.16\linewidth]{tvtrpca_0747_bw.png}
      \put(71,5){\color{white}{(D4)}}
    \end{overpic}
  }
  \subfigure[]{
    \hspace{-0.28cm}
    \begin{overpic}[width=0.16\linewidth]{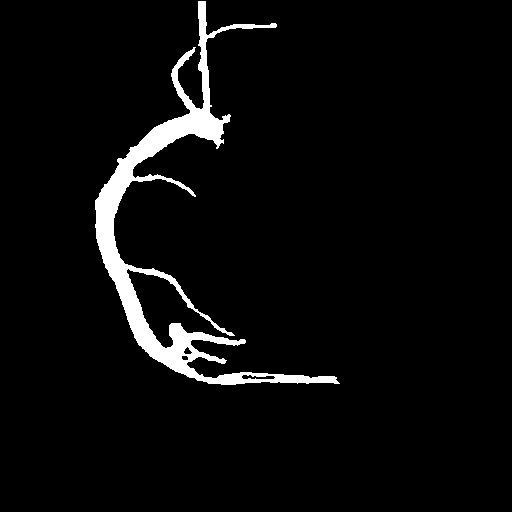}
      \put(71,5){\color{white}{(E4)}}
    \end{overpic}
  }
  \subfigure[]{
    \hspace{-0.28cm}
    \begin{overpic}[width=0.16\linewidth]{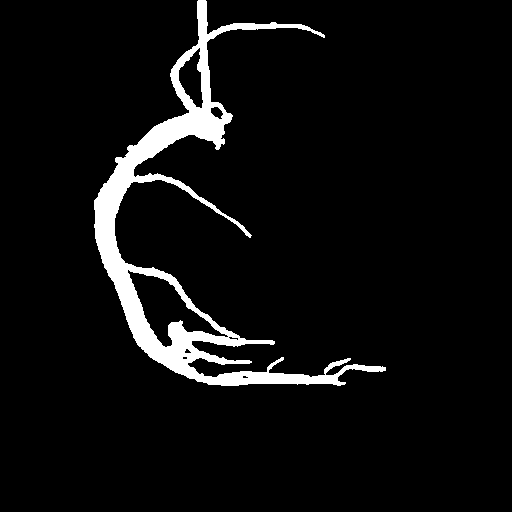}
      \put(71,5){\color{white}{(F4)}}
    \end{overpic}
  }

  \vspace{-0.95cm}
  \subfigure[]{
    \begin{overpic}[width=0.16\linewidth]{GTBW_0747.png}
      \put(71,5){\color{white}{(a4)}}
    \end{overpic}
  }
  \subfigure[]{
    \hspace{-0.28cm}
    \begin{overpic}[width=0.16\linewidth]{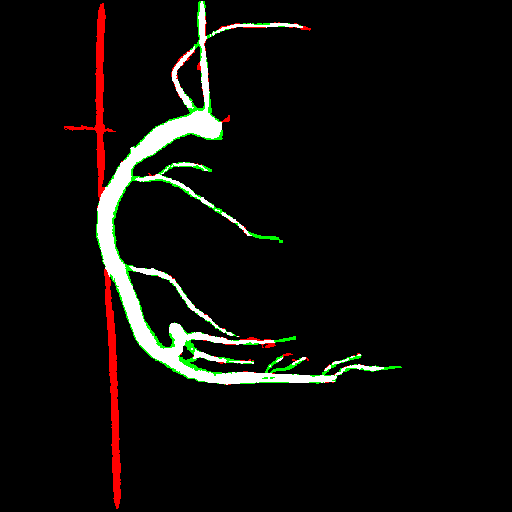}
      \put(71,5){\color{white}{(b4)}}
    \end{overpic}
  }
  \subfigure[]{
    \hspace{-0.28cm}
    \begin{overpic}[width=0.16\linewidth]{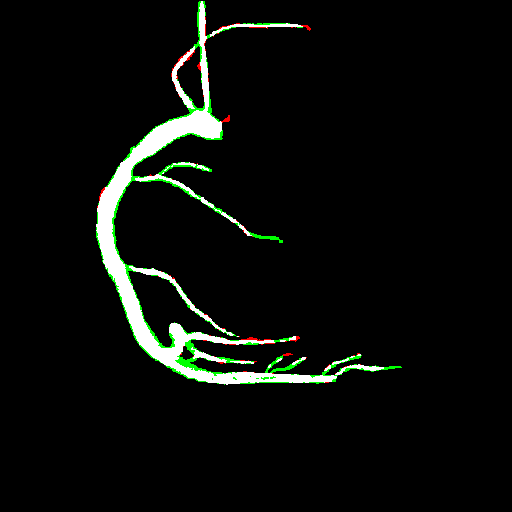}
      \put(71,5){\color{white}{(c4)}}
    \end{overpic}
  }
  \subfigure[]{
    \hspace{-0.28cm}
    \begin{overpic}[width=0.16\linewidth]{tvtrpca_0747_gt.png}
      \put(71,5){\color{white}{(d4)}}
    \end{overpic}
  }
  \subfigure[]{
    \hspace{-0.28cm}
    \begin{overpic}[width=0.16\linewidth]{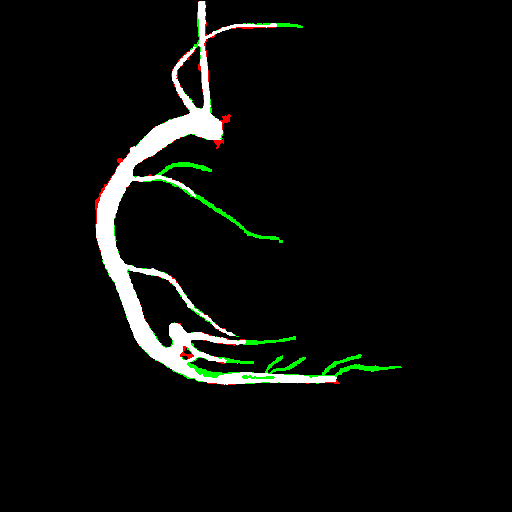}
      \put(71,5){\color{white}{(e4)}}
    \end{overpic}
  }
  \subfigure[]{
    \hspace{-0.28cm}
    \begin{overpic}[width=0.16\linewidth]{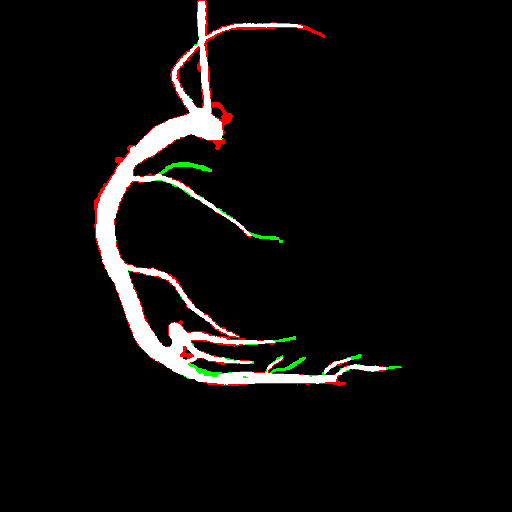}
      \put(71,5){\color{white}{(f4)}}
    \end{overpic}
  }

  \label{fig13}
  \vspace{-0.9cm}
  \caption{Vessel segmentation results of 4 XCA image sequences by 5 methods. Each group of results contains a segmentation image labeled by capital letters (B-F) and its contrast image labeled by lowercase letters (b-f). (A1)-(A4) and (a1)-(a4) are the raw XCA images and their ground truth binary vessel masks. (B,b) VGN. (C,c) SVS-net. (D,d) TV-TRPCA+RG. (E,e) TV-TRPCA+TM+RG (First stage). (F,f) TV-TRPCA+TSRG.}
\end{figure*}

\subsubsection{Comparison Test}

Though 4 vessel segmentation methods region growing, Frangi filtering, DSA and U-net have been tested in the \hyperref[TVTRPCA]{experiment} of TV-TRPCA. However, these simple segmentation methods still have many limitations and cannot achieve ideal accuracy. Advanced deep learning method VGN and SVS-net were proposed recent years. The former contains a CNN module which generates pixelwise features corresponding to vessel probabilities, a GNN module which generates features reflecting the vascular connectivity, and an inference module which complementarily combines two features to produce the final vessel probability map\cite{shin2019deep}. Based on U-net, the latter contains an encoder network extracting 3D feature from the input sequence and an decoder network learning the salient feature via upsampling and operation of channel attention block (CAB), between the encoder and decoder network exists the skip connection layers with feature fusion operation (FFO)\cite{hao2020sequential}.

\begin{figure*}[tbp]
  \centering
  \subfigure[]{\includegraphics[width=\linewidth]{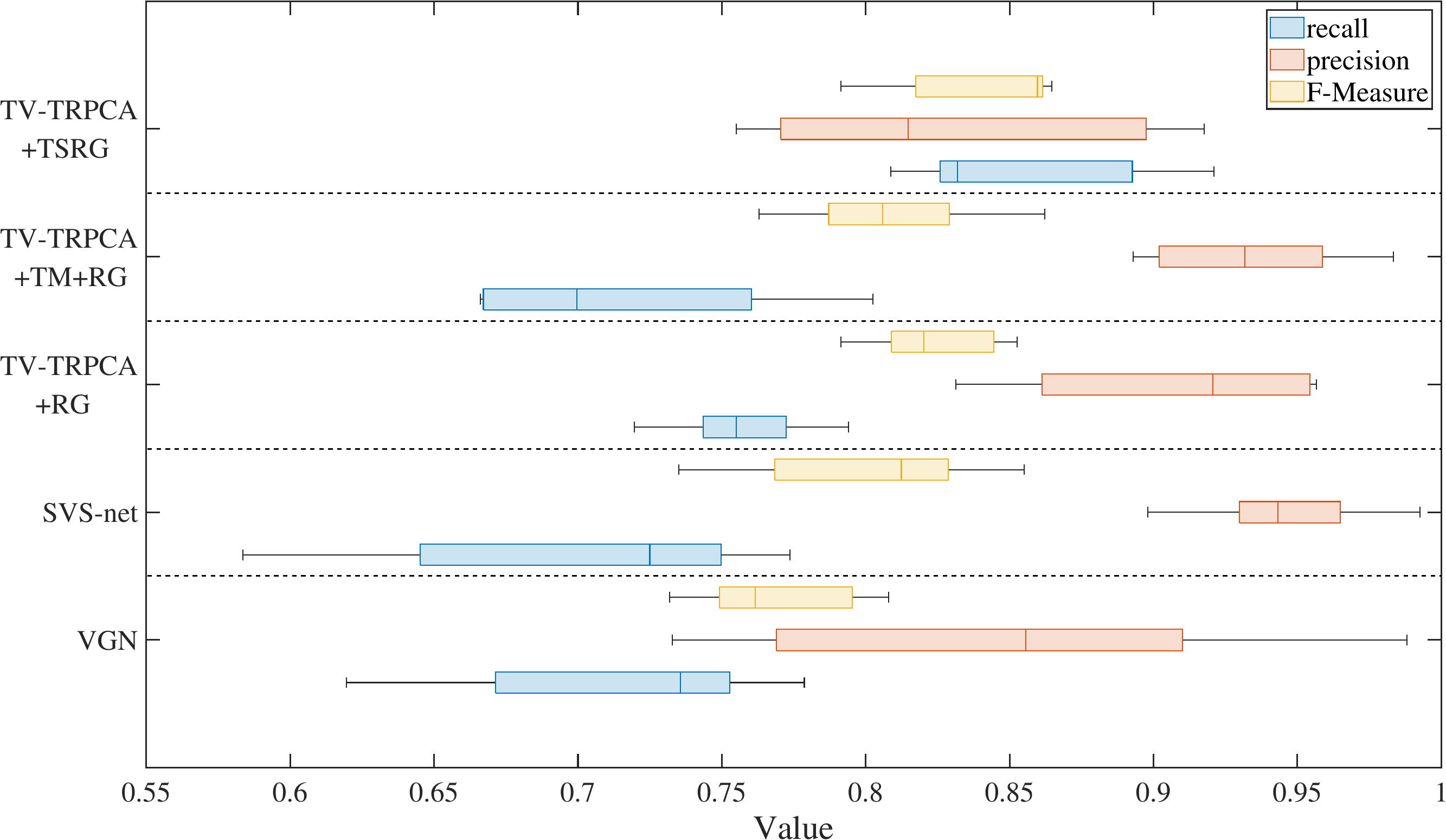}}

  \label{fig14}
  \vspace{-0.8cm}
  \caption{Distribution of recall, precision and F-Measure values of different vessel segmentation methods on clinical XCA images.}
\end{figure*}

\begin{table*}[!t]
  \renewcommand{\arraystretch}{1}
  \centering
  \topcaption{The mean recall, precision and F-Measure values ($\pm$ standard deviation) of different vessel segmentation methods}
  \label{tab5}
  \begin{footnotesize}
    \begin{tabular}{l|ccc|ccc}
      \toprule \toprule
      \multirow{2}{*}{\textbf{Method}} & \multicolumn{3}{c|}{\textbf{Clinical}} & \multicolumn{3}{c}{\textbf{Dataset}}                                                                                                                                 \\
      \cmidrule(l){2-7}
                                       & recall                                 & precision                            & F-Measure                     & recall                        & precision                     & F-Measure                     \\
      \midrule
      VGN                              & 0.713 $\pm$ 0.062                      & 0.848 $\pm$ 0.098                    & 0.770 $\pm$ 0.030             & 0.733 $\pm$ 0.054             & 0.812 $\pm$ 0.077             & 0.768 $\pm$ 0.039             \\
      SVS-net                          & 0.698 $\pm$ 0.075                      & \pmb{0.946} $\pm$ \pmb{0.034}        & 0.800 $\pm$ 0.045             & 0.679 $\pm$ 0.060             & \pmb{0.931} $\pm$ \pmb{0.023} & 0.795 $\pm$ 0.021             \\
      TV-TRPCA+RG                      & 0.757 $\pm$ 0.027                      & 0.907 $\pm$ 0.054                    & 0.824 $\pm$ 0.024             & 0.708 $\pm$ 0.041             & 0.870 $\pm$ 0.031             & 0.777 $\pm$ 0.051             \\
      TV-TRPCA+TM+RG                   & 0.716 $\pm$ 0.058                      & 0.933 $\pm$ 0.036                    & 0.809 $\pm$ 0.036             & 0.659 $\pm$ 0.037             & 0.921 $\pm$ 0.026             & 0.752 $\pm$ 0.033             \\
      TV-TRPCA+TSRG                    & \pmb{0.855} $\pm$ \pmb{0.046}          & 0.831 $\pm$ 0.071                    & \pmb{0.840} $\pm$ \pmb{0.032} & \pmb{0.798} $\pm$ \pmb{0.043} & 0.805 $\pm$ 0.069             & \pmb{0.801} $\pm$ \pmb{0.032} \\
      \bottomrule \bottomrule
    \end{tabular}
  \end{footnotesize}
\end{table*}

We compared the vessel segmentation results of three TV-TRPCA-based methods and two deep learning-based methods experimentally, as shown in \hyperref[fig13]{Fig.13}. Qualitatively, for vessel graph network (VGN), the CNN module extracts main branch and GNN module enhances vessel segments with weak contrast. This method completely preserves the structure and distribution of vessels in the raw image. However, it suffers from the same problem of misidentification of other tubular tissues as U-net, which reduces the precision. For sequential vessel segmentation deep network (SVS-net), since the input of encoder module is four consecutive frames, some high-dimensional information of the image sequence can be preserved, and irrelevant tissues can also be suppressed in the CAB module. As can be seen in \hyperref[fig13]{Fig.13.(C,c)}, the segmentation precision of this method is very high, most of the noise and interference are filtered. However, the strong inhibition effect could affect the width and continuity of the vessel region to some extent. For TV-TRPCA-based methods, as mentioned in previous section, due to the uneven gray distribution of foreground layer, directly utilizing region growing on vessel layer image (TV-TRPCA+RG) will cause voids in the binary vessel masks. For proposed TSRG method, in the first stage with threshold method and region growing (TM+RG), the continuous vessel area is well segmented, but minor segments with local weak contrast are ignored. In the second stage, region growing is performed on the image merged by the edge information enhanced by RLF filtering and the segmentation result in the first stage, as shown in \hyperref[fig3]{Fig.3} and \hyperref[fig13]{Fig.13.(F,f)}, the vessel area is extended to better display the whole coronary artery tree.

Quantitatively, for clinical XCA image sequences and third-party dataset, we calculated recall, precision and F-Measure of different vessel segmentation methods and recorded them in \hyperref[tab5]{Tab.5}. And \hyperref[fig14]{Fig14} gives an intuitive distribution of the results for clinical images. On the whole, these methods have good comprehensive effects, among which SVS-net has the highest precision, while proposed TV-TRPCA+TSRG method has more advantages in recognition rate and overall accuracy.

\subsection{Overall Discussion}

In the previous experiments of TV-TRPCA and TSRG methods, we demonstrated the superiority of this study by the visibility of foreground extraction and the accuracy of vessel segmentation. On the one hand, compared with RPCA-based and other matrix decomposition layer separation methods, due to the high-order information recovery ability of TRPCA and the constraint of TV regularization on the foreground, the proposed TV-TRPCA method performs better in both the low-rank property of background and the spatial-temporal continuity of foreground with higher Global and Local CNR values, which can effectively improve the performance of commonly used vessel segmentation methods. On the other hand, the proposed TSRG method segments the vessel layer in two stages. This improves the ability to present details while ensuring accuracy, and the continuity and integrity of vessel is preserved. Finally, binary coronary artery mask images with unique tubular structure are extracted completely.

The method proposed in this paper also has some limitations and deficiencies. In foreground extraction, the optimization of $TV/\ell_1$ norm including multiple 3D-DFT is complex and time-consuming, higher CPU processor performance is needed to achieve better real-time performance in clinic. Meanwhile, as can be seen in \hyperref[tab3]{Tab.3} and \hyperref[tab5]{Tab5}, the effect of foreground extraction and vessel segmentation will be reduced for datasets with fewer frames. Thus, the proposed TV-TRPCA method is more suitable for a complete XCA image sequence. In vessel segmentation, since the RLF filter enhances the edge information, the proposed TSRG method will widen the vessel area, resulting in reduced precision. Aiming at these deficiencies, in our future works, we are prepared to research the weakly supervised deep learning algorithm based on the theory of TV-TRPCA to further improve the real-time performance of foreground extraction and the accuracy of vessel segmentation.

\section{Conclusion} \label{Conclusion}

A robust implementation of foreground extraction and vessel segmentation for X-ray coronary angiography image sequence is proposed in this paper. In clinical medicine, raw XCA images with irrelevant tissues, organ overlaps and high-frequency noise are difficult to use directly for pre-operative diagnosis. The existing vessel extraction methods such as Hessian-based filtering, digital subtraction angiography (DSA), robust principal component analysis (RPCA) and deep neural networks have many shortcomings, such as remaining irrelevant tissues, amplifying local noise and destroying high-dimensional information. To address these issues, in this study, the vessel extraction step is divided into two steps: foreground extraction and vessel segmentation.

In foreground extraction, a tensor nuclear norm (TNN) minimization-based tensor robust principal component analysis (TRPCA) method is proposed and derived. Compared with matrix decomposition methods like RPCA, TRPCA operates directly on the raw data form, which preserves more high-dimensional information. Total variation (TV) regularization is introduced to further enhance the spatial-temporal continuity of the foreground layer. And the whole model is optimized by ADMM strategy. In vessel segmentation, a two-stage region growing (TSRG) method is presented. In the first stage, a global threshold segmentation is used as preprocessing to obtain the main branch and remove the residual low-grayscale noise. In the second stage, the Radon-like features (RLF) filter is utilized to enhance minor vessel segments, and the two intermediate results are reconnected by region growing to get final binary coronary artery mask image. Experiments on clinical XCA image sequences and third-party dataset qualitatively and quantitatively demonstrate the high visibility of the vessel foreground layers extracted by TV-TRPCA and the high accuracy of the binary vessel masks segmented by TSRG. In addition, commonly used vessel extraction methods can also benefit from the TV-TRPCA layer separation method for providing a high contrast template.

The results of this study can serve as a theoretical basis for other quantitative analyses of X-ray coronary angiography images. For example, vessel centerline extraction and calibration. It is more accurate and efficient to extract the centerline from the binary vessel mask than directly from the raw grayscale XCA image. And the centerline can further provide a prerequisite for percutaneous coronary intervention (PCI) path planning and guidewire navigation. Another important application of this study is the 2D-3D registration of intra-operative 2D X-ray coronary angiography (XCA) images with the pre-operative 3D computed tomography angiography (CTA) model. Our method improves the signal-to-noise ratio (SNR) of raw XCA images and significantly enhances the visibility of vessel areas, thus improving the robustness and accuracy of 2D-3D registration.

In future works, weakly supervised deep neural networks based on the theory of TV-TRPCA is going to be researched to further improve the real-time performance of foreground extraction and the accuracy of vessel segmentation, especially for the XCA image sequences with few frames. Furthermore, based on the results of this study, we are prepared to track and extract the coronary centerline by combining image thinning algorithms and deep reinforcement learning (DRL) algorithms.


%

\appendix
\section{Proof of Eq.29 in $\mathcal{H}$ sub-problem}  \label{APPENDIX}
\setcounter{equation}{0}
\renewcommand{\theequation}{A.\arabic{equation}}

\hyperref[eq27]{Eq.27} can be rewritten as
\begin{equation}
  \pmb{A}f(\mathcal{H}_{k+1})= f(\mathcal{K})
  \label{eqa1}
\end{equation}
where
\begin{equation}
  \pmb{A}=\left(\mu_k\pmb{I}+\nu_k\left(\sigma_m^2\pmb{D}_m^T\pmb{D}_m+\sigma_m^2\pmb{D}_n^T\pmb{D}_n+\sigma_m^2\pmb{D}_t^T\pmb{D}_t\right)\right)
  \label{eqa2}
\end{equation}
$\pmb{D}_m,\pmb{D}_n,\pmb{D}_t\in\mathbb{R}^{MNT\times MNT}$ are block circulant matrices and $f(\mathcal{V}_{k+1}),f(\mathcal{K})\in\mathbb{R}^{MNT}$ are tensor vectorization of $\mathcal{V}_{k+1}$ and $\mathcal{K}$.

Consider the following equation
\begin{equation}
  (\pmb{I}+\pmb{B}^T\pmb{B})\pmb{x}= \pmb{y}
  \label{eqa3}
\end{equation}
where $\pmb{B}$ is a block circulant matrix, $\pmb{x}$ and $\pmb{y}$ are column vectors. It can be regarded as a simplified version of \hyperref[eqa1]{Eq.A.1}.

According to matrix transformation theories\cite{golub2013matrix}, $\pmb{B}$ can be diagonalized by the DFT matrix $\pmb{F}$, i.e.,
\begin{equation}
  \pmb{B}=\pmb{F}^H\pmb{\varLambda}\pmb{F}
  \label{eqa4}
\end{equation}
where $\pmb{F}^H$ is the Hermitian conjugate matrix of $\pmb{F}$ and $\pmb{\varLambda}=diagm(\pmb{fft}(\pmb{B}(:,1)))$ forms a diagonal matrix storing the eigenvalues of $\pmb{B}$. Notice that
$\pmb{F}^H\pmb{F}=\pmb{F}\pmb{F}^H=\pmb{I}$, \hyperref[eqa3]{Eq.A.3} can be rewritten as
\begin{equation}
  \pmb{F}^H(\pmb{I}+\pmb{\varLambda }^T\pmb{\varLambda })\pmb{F}\pmb{x}= \pmb{y}
  \label{eqa5}
\end{equation}
Multiplying both sides of above equation by $\pmb{F}$, it can be further rewritten as
\begin{equation}
  \begin{aligned}
    (\pmb{I}+\pmb{\varLambda }^T\pmb{\varLambda })\pmb{F}\pmb{x} & = \pmb{F}\pmb{y}     \\
    \Rightarrow (\pmb{I}+|\pmb{\varLambda}|^2)\pmb{fft}(\pmb{x}) & = \pmb{fft}(\pmb{y})
    \label{eqa6}
  \end{aligned}
\end{equation}
Since $\pmb{I}+|\pmb{\varLambda}|^2$ is a diagonal matrix, \hyperref[eqa3]{Eq.A.3} has the following solution
\begin{equation}
  \pmb{x}=\pmb{ifft}\left(\frac{\pmb{fft}(\pmb{y})}{diagv(\pmb{I}+|\pmb{\varLambda}|^2)}\right)
  \label{eqa7}
\end{equation}
where $diagv(\cdot)$ forms a column vector from the elements of the diagonal of a matrix, $|\cdot|^2$ is the element-wise square, and the division is also performed element-wisely.

It is not hard to find that \hyperref[eqa1]{Eq.A.1} has a similar form with \hyperref[eqa3]{Eq.A.3}, its solution \hyperref[eq29]{Eq.29} can be derived by analogy.

\section*{Acknowledgment}
This work was greatly supported by the National Natural Science Foundation of China (Grant No. 61973210) and the Medical-engineering Cross Projects of SJTU (Grant No. YG2019ZDA17, ZH2018QNB23). The authors would like to thank the doctors of Ruijin Hospital affiliated to Shanghai Jiao Tong University for their helps, and all the authors of open-source codes and dataset used in this study.





\bibliographystyle{elsarticle-num.bst}\biboptions{numbers}
\bibliography{refs}

\begin{thebibliography}{10}
\expandafter\ifx\csname url\endcsname\relax
  \def\url#1{\texttt{#1}}\fi
\expandafter\ifx\csname urlprefix\endcsname\relax\def\urlprefix{URL }\fi
\expandafter\ifx\csname href\endcsname\relax
  \def\href#1#2{#2} \def\path#1{#1}\fi

\bibitem{world2021world}
W.~H. Organization, et~al., World health statistics 2021, Tech. rep., World
  Health Organization (2021).

\bibitem{kiricsli2013standardized}
H.~Kiri{\c{s}}li, M.~Schaap, C.~Metz, A.~Dharampal, W.~B. Meijboom, S.-L.
  Papadopoulou, A.~Dedic, K.~Nieman, M.~A. de~Graaf, M.~Meijs, et~al.,
  Standardized evaluation framework for evaluating coronary artery stenosis
  detection, stenosis quantification and lumen segmentation algorithms in
  computed tomography angiography, Medical image analysis 17~(8) (2013)
  859--876.

\bibitem{jin2017extracting}
M.~Jin, R.~Li, J.~Jiang, B.~Qin, Extracting contrast-filled vessels in x-ray
  angiography by graduated rpca with motion coherency constraint, Pattern
  Recognition 63 (2017) 653--666.

\bibitem{qin2019accurate}
B.~Qin, M.~Jin, D.~Hao, Y.~Lv, Q.~Liu, Y.~Zhu, S.~Ding, J.~Zhao, B.~Fei,
  Accurate vessel extraction via tensor completion of background layer in x-ray
  coronary angiograms, Pattern recognition 87 (2019) 38--54.

\bibitem{moccia2018blood}
S.~Moccia, E.~De~Momi, S.~El~Hadji, L.~S. Mattos, Blood vessel segmentation
  algorithms—review of methods, datasets and evaluation metrics, Computer
  methods and programs in biomedicine 158 (2018) 71--91.

\bibitem{frangi1998multiscale}
A.~F. Frangi, W.~J. Niessen, K.~L. Vincken, M.~A. Viergever, Multiscale vessel
  enhancement filtering, in: International conference on medical image
  computing and computer-assisted intervention, Springer, 1998, pp. 130--137.

\bibitem{kumar2010radon}
R.~Kumar, A.~V{\'a}zquez-Reina, H.~Pfister, Radon-like features and their
  application to connectomics, in: 2010 IEEE computer society conference on
  computer vision and pattern recognition-workshops, IEEE, 2010, pp. 186--193.

\bibitem{syeda2012finding}
T.~Syeda-Mahmood, F.~Wang, R.~Kumar, D.~Beymer, Y.~Zhang, R.~Lundstrom,
  E.~McNulty, Finding similar 2d x-ray coronary angiograms, in: International
  Conference on Medical Image Computing and Computer-Assisted Intervention,
  Springer, 2012, pp. 501--508.

\bibitem{otsu1979threshold}
N.~Otsu, A threshold selection method from gray-level histograms, IEEE
  transactions on systems, man, and cybernetics 9~(1) (1979) 62--66.

\bibitem{kerkeni2016coronary}
A.~Kerkeni, A.~Benabdallah, A.~Manzanera, M.~H. Bedoui, A coronary artery
  segmentation method based on multiscale analysis and region growing,
  Computerized Medical Imaging and Graphics 48 (2016) 49--61.

\bibitem{nasr2018segmentation}
E.~Nasr-Esfahani, N.~Karimi, M.~H. Jafari, S.~M.~R. Soroushmehr, S.~Samavi,
  B.~Nallamothu, K.~Najarian, Segmentation of vessels in angiograms using
  convolutional neural networks, Biomedical Signal Processing and Control 40
  (2018) 240--251.

\bibitem{wan2021automatic}
T.~Wan, J.~Chen, Z.~Zhang, D.~Li, Z.~Qin, Automatic vessel segmentation in
  x-ray angiogram using spatio-temporal fully-convolutional neural network,
  Biomedical Signal Processing and Control 68 (2021) 102646.

\bibitem{ronneberger2015u}
O.~Ronneberger, P.~Fischer, T.~Brox, U-net: Convolutional networks for
  biomedical image segmentation, in: International Conference on Medical image
  computing and computer-assisted intervention, Springer, 2015, pp. 234--241.

\bibitem{shin2019deep}
S.~Y. Shin, S.~Lee, I.~D. Yun, K.~M. Lee, Deep vessel segmentation by learning
  graphical connectivity, Medical image analysis 58 (2019) 101556.

\bibitem{hao2020sequential}
D.~Hao, S.~Ding, L.~Qiu, Y.~Lv, B.~Fei, Y.~Zhu, B.~Qin, Sequential vessel
  segmentation via deep channel attention network, Neural Networks 128 (2020)
  172--187.

\bibitem{zhang2020weakly}
J.~Zhang, G.~Wang, H.~Xie, S.~Zhang, N.~Huang, S.~Zhang, L.~Gu, Weakly
  supervised vessel segmentation in x-ray angiograms by self-paced learning
  from noisy labels with suggestive annotation, Neurocomputing 417 (2020)
  114--127.

\bibitem{meaney1980digital}
T.~F. Meaney, M.~Weinstein, E.~Buonocore, W.~Pavlicek, G.~P. Borkowski, J.~H.
  Gallagher, B.~Sufka, W.~J. Maclntyre, Digital subtraction angiography of the
  human cardiovascular system, in: Application of Optical Instrumentation in
  Medicine VIII, Vol. 233, SPIE, 1980, pp. 272--278.

\bibitem{bentoutou2002invariant}
Y.~Bentoutou, N.~Taleb, M.~C. El~Mezouar, M.~Taleb, L.~Jetto, An invariant
  approach for image registration in digital subtraction angiography, Pattern
  Recognition 35~(12) (2002) 2853--2865.

\bibitem{song2019patch}
S.~Song, J.~Yang, D.~Ai, C.~Du, Y.~Huang, H.~Song, L.~Zhang, Y.~Han, Y.~Wang,
  A.~F. Frangi, Patch-based adaptive background subtraction for vascular
  enhancement in x-ray cineangiograms, IEEE journal of biomedical and health
  informatics 23~(6) (2019) 2563--2575.

\bibitem{tang2012application}
S.~Tang, Y.~Wang, Y.-W. Chen, Application of ica to x-ray coronary digital
  subtraction angiography, Neurocomputing 79 (2012) 168--172.

\bibitem{xia2019vessel}
S.~Xia, H.~Zhu, X.~Liu, M.~Gong, X.~Huang, L.~Xu, H.~Zhang, J.~Guo, Vessel
  segmentation of x-ray coronary angiographic image sequence, IEEE Transactions
  on Biomedical Engineering 67~(5) (2019) 1338--1348.

\bibitem{candes2011robust}
E.~J. Cand{\`e}s, X.~Li, Y.~Ma, J.~Wright, Robust principal component
  analysis?, Journal of the ACM (JACM) 58~(3) (2011) 1--37.

\bibitem{ma2017automatic}
H.~Ma, A.~Hoogendoorn, E.~Regar, W.~J. Niessen, T.~van Walsum, Automatic online
  layer separation for vessel enhancement in x-ray angiograms for percutaneous
  coronary interventions, Medical image analysis 39 (2017) 145--161.

\bibitem{zhang2018vesselness}
J.~Zhang, G.~Wang, H.~Xie, S.~Zhang, Z.~Shi, L.~Gu, Vesselness-constrained
  robust pca for vessel enhancement in x-ray coronary angiograms, Physics in
  Medicine \& Biology 63~(15) (2018) 155019.

\bibitem{song2019spatio}
S.~Song, C.~Du, D.~Ai, Y.~Huang, H.~Song, Y.~Wang, J.~Yang, Spatio-temporal
  constrained online layer separation for vascular enhancement in x-ray
  angiographic image sequence, IEEE Transactions on Circuits and Systems for
  Video Technology 30~(10) (2019) 3558--3570.

\bibitem{kolda2009tensor}
T.~G. Kolda, B.~W. Bader, Tensor decompositions and applications, SIAM review
  51~(3) (2009) 455--500.

\bibitem{lu2019tensor}
C.~Lu, J.~Feng, Y.~Chen, W.~Liu, Z.~Lin, S.~Yan, Tensor robust principal
  component analysis with a new tensor nuclear norm, IEEE transactions on
  pattern analysis and machine intelligence 42~(4) (2019) 925--938.

\bibitem{goldfarb2014robust}
D.~Goldfarb, Z.~Qin, Robust low-rank tensor recovery: Models and algorithms,
  SIAM Journal on Matrix Analysis and Applications 35~(1) (2014) 225--253.

\bibitem{xie2017kronecker}
Q.~Xie, Q.~Zhao, D.~Meng, Z.~Xu, Kronecker-basis-representation based tensor
  sparsity and its applications to tensor recovery, IEEE transactions on
  pattern analysis and machine intelligence 40~(8) (2017) 1888--1902.

\bibitem{liu2018improved}
Y.~Liu, L.~Chen, C.~Zhu, Improved robust tensor principal component analysis
  via low-rank core matrix, IEEE Journal of Selected Topics in Signal
  Processing 12~(6) (2018) 1378--1389.

\bibitem{lu2016tensor}
C.~Lu, J.~Feng, Y.~Chen, W.~Liu, Z.~Lin, S.~Yan, Tensor robust principal
  component analysis: Exact recovery of corrupted low-rank tensors via convex
  optimization, in: Proceedings of the IEEE conference on computer vision and
  pattern recognition, 2016, pp. 5249--5257.

\bibitem{gao2020enhanced}
Q.~Gao, P.~Zhang, W.~Xia, D.~Xie, X.~Gao, D.~Tao, Enhanced tensor rpca and its
  application, IEEE Transactions on Pattern Analysis and Machine Intelligence
  43~(6) (2020) 2133--2140.

\bibitem{xie2016weighted}
Y.~Xie, S.~Gu, Y.~Liu, W.~Zuo, W.~Zhang, L.~Zhang, Weighted schatten $ p $-norm
  minimization for image denoising and background subtraction, IEEE
  transactions on image processing 25~(10) (2016) 4842--4857.

\bibitem{chan2011augmented}
S.~H. Chan, R.~Khoshabeh, K.~B. Gibson, P.~E. Gill, T.~Q. Nguyen, An augmented
  lagrangian method for total variation video restoration, IEEE Transactions on
  Image Processing 20~(11) (2011) 3097--3111.

\bibitem{cao2016total}
W.~Cao, Y.~Wang, J.~Sun, D.~Meng, C.~Yang, A.~Cichocki, Z.~Xu, Total variation
  regularized tensor rpca for background subtraction from compressive
  measurements, IEEE Transactions on Image Processing 25~(9) (2016) 4075--4090.

\bibitem{kilmer2011factorization}
M.~E. Kilmer, C.~D. Martin, Factorization strategies for third-order tensors,
  Linear Algebra and its Applications 435~(3) (2011) 641--658.

\bibitem{zhao2014robust}
Q.~Zhao, D.~Meng, Z.~Xu, W.~Zuo, L.~Zhang, Robust principal component analysis
  with complex noise, in: International conference on machine learning, PMLR,
  2014, pp. 55--63.

\bibitem{lin2010augmented}
Z.~Lin, M.~Chen, Y.~Ma, The augmented lagrange multiplier method for exact
  recovery of corrupted low-rank matrices, arXiv preprint arXiv:1009.5055
  (2010).

\bibitem{golub2013matrix}
G.~H. Golub, C.~F. Van~Loan, Matrix computations, JHU press, 2013.

\bibitem{tremeau1997region}
A.~Tremeau, N.~Borel, A region growing and merging algorithm to color
  segmentation, Pattern recognition 30~(7) (1997) 1191--1203.

\bibitem{cai2019accelerated}
H.~Cai, J.-F. Cai, K.~Wei, Accelerated alternating projections for robust
  principal component analysis, The Journal of Machine Learning Research 20~(1)
  (2019) 685--717.

\bibitem{zhou2012moving}
X.~Zhou, C.~Yang, W.~Yu, Moving object detection by detecting contiguous
  outliers in the low-rank representation, IEEE transactions on pattern
  analysis and machine intelligence 35~(3) (2012) 597--610.

\bibitem{zhou2011godec}
T.~Zhou, D.~Tao, Godec: Randomized low-rank \& sparse matrix decomposition in
  noisy case, in: Proceedings of the 28th International Conference on Machine
  Learning, ICML 2011, 2011, p.~1.

\bibitem{wang2012probabilistic}
N.~Wang, T.~Yao, J.~Wang, D.-Y. Yeung, A probabilistic approach to robust
  matrix factorization, in: Computer Vision--ECCV 2012: 12th European
  Conference on Computer Vision, Florence, Italy, October 7-13, 2012,
  Proceedings, Part VII 12, Springer, 2012, pp. 126--139.

\bibitem{brutzer2011evaluation}
S.~Brutzer, B.~H{\"o}ferlin, G.~Heidemann, Evaluation of background subtraction
  techniques for video surveillance, in: CVPR 2011, IEEE, 2011, pp. 1937--1944.

\end{thebibliography}

\end{document}